\documentclass[manuscript,screen,nonacm]{acmart}

\usepackage{array}
\usepackage{textcomp}
\usepackage{stfloats}
\usepackage{verbatim}
\usepackage{graphicx}

\usepackage{mathtools}
\usepackage{amsfonts}
\usepackage{bbm}
\usepackage{bm}
\usepackage{arydshln}
\usepackage{booktabs}
\usepackage{pifont}
\usepackage{microtype}
\usepackage{latexsym}
\usepackage{times}
\usepackage{bbding}
\usepackage{pdflscape}
\usepackage{colortbl}    

\usepackage{tcolorbox}
\usepackage{enumitem}
\usepackage{longtable}
\usepackage{booktabs}
\usepackage{adjustbox}  
\usepackage{multirow}
\usepackage{xcolor}
\usepackage{wrapfig}

\usepackage[edges]{forest}
\definecolor{black}{RGB}{0, 0, 0}
\definecolor{light-green}{RGB}{240, 247, 236}
\definecolor{light-blue}{RGB}{237, 242, 248}
\definecolor{light-gray}{RGB}{247, 247, 247}
\definecolor{light-yellow}{RGB}{253, 246, 238}
\definecolor{paired-dark-blue}{RGB}{49, 130, 188}
\definecolor{paired-light-orange}{RGB}{251, 208, 162}
\definecolor{paired-dark-orange}{RGB}{230, 85, 12}
\definecolor{paired-light-green}{RGB}{199, 233, 193}
\definecolor{paired-dark-green}{RGB}{49, 163, 83}
\definecolor{paired-light-purple}{RGB}{218, 218, 235}
\definecolor{paired-dark-purple}{RGB}{117, 107, 176}
\definecolor{paired-light-gray}{RGB}{217, 217, 217}
\definecolor{paired-dark-gray}{RGB}{99, 99, 99}
\definecolor{paired-light-pink}{RGB}{222, 158, 214}
\definecolor{paired-dark-pink}{RGB}{123, 65, 115}
\definecolor{paired-light-red}{RGB}{231, 150, 156}
\definecolor{paired-dark-red}{RGB}{131, 60, 56}
\definecolor{paired-light-yellow}{RGB}{231, 204, 149}
\definecolor{paired-dark-yellow}{RGB}{141, 109, 49}
\tikzset{%
    parent/.style = {align=center,text width=2cm,rounded corners=3pt, line width=0.2mm, fill=light-yellow,draw=black},
    child/.style = {align=center,text width=2.3cm,rounded corners=3pt, fill=blue!10,draw=blue!80,line width=0.3mm},
    grandchild/.style = {align=center,text width=2cm,rounded corners=3pt},
    greatgrandchild/.style = {align=center,text width=1.5cm,rounded corners=3pt},
    greatgrandchild2/.style = {align=center,text width=1.5cm,rounded corners=3pt},    
    referenceblock/.style = {align=center,text width=1.5cm,rounded corners=2pt},
    methods/.style = {align=center,text width=2.5cm,rounded corners=3pt, fill=light-green,draw=black,line width=0.2mm},   
    methods_work/.style = {align=center, text width=6.3cm, rounded corners=3pt, fill=light-green,draw=black,line width=0.2mm},  
    representation/.style = {align=center,text width=2.5cm,rounded corners=3pt, fill=light-blue,draw=black,line width=0.2mm},   
    representation_work/.style = {align=center,text width=6.3cm,rounded corners=3pt, fill=light-blue,draw=black,line width=0.2mm},    
    probing/.style = {align=center,text width=2.5cm,rounded corners=3pt, fill= light-gray,draw=black,line width=0.2mm},   
    probing_work/.style = {align=center,text width=6.3cm,rounded corners=3pt, fill= light-gray,draw= black,line width=0.2mm}    
}

\begin{document}

\title{Large Language Models for Planning: A Comprehensive and Systematic Survey}

\author{Pengfei Cao}
\affiliation{%
  \institution{The Key Laboratory of Cognition and Decision Intelligence for Complex Systems, CASIA, and School of Artificial Intelligence, University of Chinese Academy of Sciences}
  \country{China}
}
\email{pengfei.cao@nlpr.ia.ac.cn}

\author{Tianyi Men}
\affiliation{%
  \institution{The Key Laboratory of Cognition and Decision Intelligence for Complex Systems, CASIA, and School of Artificial Intelligence, University of Chinese Academy of Sciences}
  \country{China}
  }
\email{tianyi.men@nlpr.ia.ac.cn}

\author{Wencan Liu}
\affiliation{%
  \institution{Harbin Institute of Technology}
  \country{China}
  }

\author{Jingwen Zhang}
\affiliation{%
 \institution{Institute of Information Engineering, Chinese Academy of Sciences}
 \country{China}
 }
 \email{zhangjingwen2024@iie.ac.cn}

\author{Xuzhao Li}
\affiliation{%
 \institution{School of Automation, Beijing Institute of Technology}
 \country{China}
 }
 \email{xuzhaoli2024@bit.edu.cn}

\author{Xixun Lin}
\affiliation{%
 \institution{Institute of Information Engineering, Chinese Academy of Sciences}
 \country{China}
 }
 \email{linxixun@iie.ac.cn}

\author{Dianbo Sui}
\affiliation{%
  \institution{Harbin Institute of Technology}
  \country{China}
  }
  \email{suidianbo@hit.edu.cn}

\author{Yanan Cao}
\affiliation{%
 \institution{Institute of Information Engineering, Chinese Academy of Sciences}
 \country{China}
 }
 \email{caoyanan@iie.ac.cn}

\author{Kang Liu}
\affiliation{%
  \institution{The Key Laboratory of Cognition and Decision Intelligence for Complex Systems, CASIA, and School of Artificial Intelligence, University of Chinese Academy of Sciences}
  \country{China}
  }
\email{kliu@nlpr.ia.ac.cn}

\author{Jun Zhao}
\affiliation{%
  \institution{The Key Laboratory of Cognition and Decision Intelligence for Complex Systems, CASIA, and School of Artificial Intelligence, University of Chinese Academy of Sciences}
  \country{China}
  }
\email{jzhao@nlpr.ia.ac.cn}

\renewcommand{\shortauthors}{Cao et al.}

\begin{abstract}
Planning represents a fundamental capability of intelligent agents, requiring comprehensive environmental understanding, rigorous logical reasoning, and effective sequential decision-making. While Large Language Models (LLMs) have demonstrated remarkable performance on certain planning tasks, their broader application in this domain warrants systematic investigation. This paper presents a comprehensive review of LLM-based planning.  Specifically, this survey is structured as follows: First, we establish the theoretical foundations by introducing essential definitions and categories about automated planning. Next, we provide a detailed taxonomy and analysis of contemporary LLM-based planning methodologies, categorizing them into three principal approaches: 1) \textit{External Module Augmented Methods} that combine LLMs with additional components for planning, 2) \textit{Finetuning-based Methods} that involve using trajectory data and feedback signals to adjust LLMs in order to improve their planning abilities, and 3) \textit{Searching-based Methods} that break down complex tasks into simpler components, navigate the planning space, or enhance decoding strategies to find the best solutions. Subsequently, we systematically summarize existing evaluation frameworks, including benchmark datasets, evaluation metrics and performance comparisons between representative planning methods. Finally, we discuss the underlying mechanisms enabling LLM-based planning and outline promising research directions for this rapidly evolving field. We hope this survey will serve as a valuable resource to inspire innovation and drive progress in this field.
\end{abstract}

\settopmatter{printacmref=False}

\maketitle

\section{Introduction}
Large Language Models (LLMs) have demonstrated impressive capabilities across a broad spectrum of natural language processing tasks. Building on these advancements, recent research has investigated the use of LLMs as autonomous agents to address real-world planning problems, achieving promising results. Planning—an essential component of intelligent behavior—involves complex processes such as environmental comprehension, logical reasoning, and sequential decision-making \cite{ghallab2004automated}. It holds significant practical value and has been widely applied in domains including web navigation (across Android, desktop, and web platforms), embodied intelligence (e.g., autonomous driving, Minecraft, and virtual environments), and travel planning.

To tackle complex planning tasks, traditional approaches have primarily relied on symbolic planners, especially those based on the Planning Domain Definition Language (PDDL) framework \cite{aeronautiques1998pddl,haslum2019introduction}. However, these methods require translating flexible, natural language problem descriptions into rigid symbolic representations—an often labor-intensive process that depends heavily on human expertise. Moreover, symbolic approaches tend to lack robustness, as even minor modeling errors can result in complete planning failures \cite{russell2016artificial,georgievski2015htn}. Recently, the emergence and rapid development of LLMs have enabled a new wave of autonomous planning systems, which exhibit promising levels of planning capability \cite{zhu2024knowagent,wucan,zeng2023agenttuning,hutree}. LLM-based planning approaches can be broadly categorized into three main types: \textit{1) External Module Augmented Methods}, which integrate LLMs with external components for planning, including planners (i.e., planner enhanced methods) \cite{liu2023llm+,hao2024large,tang2024itinera} and memory (i.e., memory enhanced methods) \cite{zheng2023synapse,park2023generative,majumder2023clin,kang2023think}. \textit{2) Finetuning-based Methods}, which enhance planning abilities by finetuning LLMs using trajectory data (i.e., imitation learning-based methods) \cite{zhang2023you,chen2024agent,chen2023fireact} and feedback signals (i.e., feedback-based methods) \cite{yuan2024advancing,yang2024react,wang2024llms}. \textit{3) Searching-based Methods}, which aim to identify optimal solutions by decomposing complex tasks (i.e., decomposition-based methods) \cite{shen2023hugginggpt,zhou2022least,kim2024rada}, exploring the planning space (i.e., exploration-based methods) \cite{yao2023tree,xu2024search,hu2023avis}, or improving decoding strategies (i.e., decoding-based methods) \cite{xie2024self,wang2024chain,liu2024don}. In addition to proposing new planning techniques, several studies have introduced benchmark datasets across diverse scenarios to assess the planning capabilities of LLMs \cite{deng2023mind2web,jia2024langsuit,xie2024travelplanner}. Furthermore, recent research has begun to probe the underlying mechanisms of LLM-based planning, offering insights into how these models reason and make decisions \cite{men2024unlocking,stechly2024chain,lazarotype}.

\begin{figure*}[t]
\centering
  \includegraphics[width=0.99\textwidth,height=0.5\textheight,keepaspectratio]{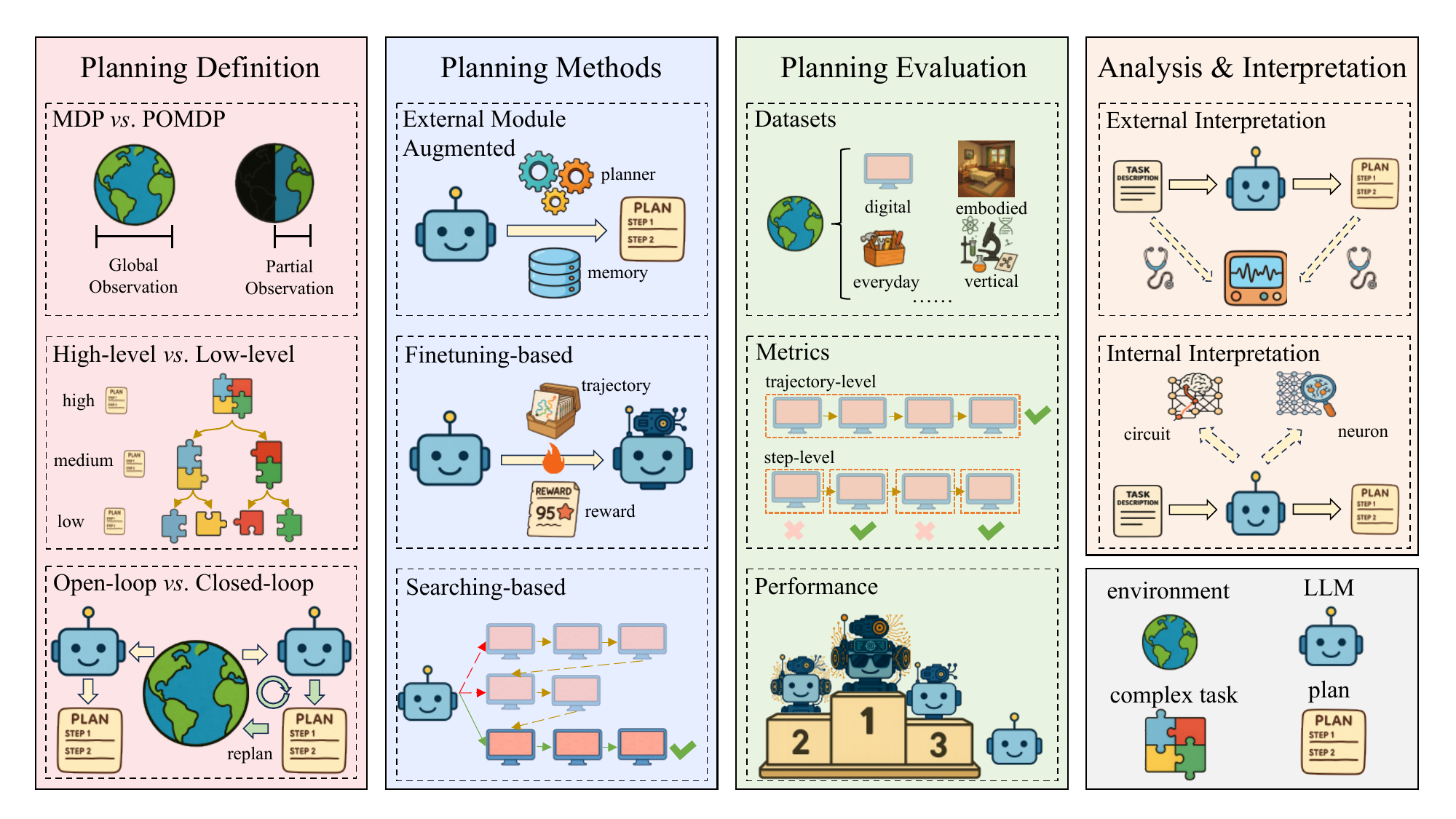}
  \caption{An overview of LLM-based agent planning, covering its definition, methods, evaluation approaches, and analysis and interpretation.}
  \label{overview}
\end{figure*}

Given the rapidly growing interest in LLM-based planning, we believe that an up-to-date and comprehensive survey can offer a valuable and holistic perspective for future research. Currently, there are only a few surveys exploring the planning of LLMs. For example, Huang et al. \cite{huang2024understanding} review some works on LLM-based agent planning, but their survey is not exhaustive, only includes literature up to February 2024, and lacks a summary of planning evaluation and interpretability aspects. Some surveys \cite{aghzal2025survey,tantakoun2025llms,wei2025plangenllms} review works but lack comprehensiveness or fine-grained categorization. Other surveys primarily focus on LLM-based agents more broadly, with planning discussed only in part, thus lacking a comprehensive view of the topic \cite{wang2024survey, luo2025large, li2024survey}. Therefore, we conduct a more comprehensive survey for LLM-based agent planning to enlighten and guide
researchers and practitioners. Figure \ref{overview} presents the overview of the LLM-based agent planning, which is organized into four closely related parts, involving planning definitions, planning methods, planning evaluation, as well as analysis and interpretation.

The key contributions of this paper are summarized as follows:

\begin{itemize}
	\item \textbf{Systematic Review}: We present a comprehensive review of LLM-based planning, including planning definitions, methodologies, evaluations, and mechanistic interpretations. This systematic analysis provides a structured framework for understanding current advancements in the field.

        \item \textbf{Insightful Analysis}: We analyze the strengths and limitations of existing planning methods, offering practical guidance for researchers in selecting appropriate baselines for their specific applications. We also comprehensively summarize the existing datasets, evaluation metrics, and analyze the performance of representative methods.

        \item \textbf{Future Directions}: We systematically outline current challenges in LLM-based planning and propose promising research directions. These insights are intended to inspire innovative solutions and drive future advances in the field.

        \item \textbf{Open Resources}: We curate and publicly release a comprehensive collection of resources, including over 300 planning-related papers, made accessible via GitHub\footnote{\url{https://github.com/Quester-one/Awesome-LLM-Planning}} to support community engagement and collaboration.
\end{itemize}

\textit{Organization of Survey}: The remainder of this survey is organized as follows. Section \ref{definition} introduces the definitions and categories of planning. Section \ref{PM} provides a detailed overview of LLM-based methodologies for planning. Section \ref{Eva} presents widely used datasets, evaluation metrics and performance of classical planning methods. Section \ref{AI} explores the underlying mechanisms of LLM-based planning. Section \ref{direction} discusses promising directions for future research. Finally, Section \ref{conclusion} concludes the survey.

\section{Definitions and Categories} \label{definition}
\subsection{Planning Definitions}
A planning problem $P$ is defined as a tuple $(S,A,T,s^{init},S^{goal})$ \cite{liu2023llm+,aghzal2025survey}, where $S$ denotes the set of the environment states, $A$ is the set of agent actions, and $T$ is the state transition function, defined as $T: S \times A \to S$. The initial state of the environment is denoted by $s^{init} \in S$, and the set of goal states is denoted by $S^{goal} \subseteq S$, which satisfy specific constraints aligned with user goals. A solution to the planning problem is an action sequence $\pi = (a_0, a_1, \dots, a_{n-1})$ that transitions the system from the initial state $s_0=s^{init}$ to $s_n \in S^{goal}$, following the transition function $s_{i+1}=T(s_{i}, a_{i})$, where $s_i$ and $a_i$ denote the state and action at time step $i \in [0, n)$. The planning can be formulated as 
\begin{equation}
\pi = (a_0, a_1, \dots, a_{n-1})={\rm Planning}(S,A,T,s^{init},S^{goal};\Theta),
\end{equation}
where $\Theta$ denote the parameters of the LLM.

\subsection{Planning Categories}
The planning problem can be further categorized based on three key factors: the type of environment, the type of agent, and the nature of interaction between the agent and the environment.

\subsubsection{MDP vs. POMDP}

Depending on whether the environment is fully observable, planning tasks can be formulated as either a \textit{Markov Decision Process} (MDP) \cite{bellman1957markovian} or a \textit{Partially Observable Markov Decision Process} (POMDP) \cite{aastrom1965optimal,kaelbling1998planning}.

An MDP is typically defined as a tuple $(S, A, T, R, \gamma)$, where $S$ represents the state space, $A$ denotes the action space, $T$ is the state transition function, defined as $T:S \times A \to S$, $R$ is the reward function, defined as $R: S \times A \to \mathbb{R}$, where $\mathbb{R}$ denotes the reward value and $\gamma$ represents the discount factor.

In contrast, a POMDP extends the MDP framework to account for partial observability and is defined as a tuple $(S, A, T, O, Z, R, \gamma)$, where $S$, $A$, $T$, $R$, and $\gamma$ retain the same definitions as in the MDP, $O$ denotes the observation space, and $Z$ is the observation function, defined as $Z: S \times A \to O$.

\subsubsection{High-level vs. Low-level}
According to hierarchical task planning frameworks \cite{barto2003recent,pateria2021hierarchical,aghzal2025survey}, planning can be divided into \textit{high-level task planning} and \textit{low-level task planning}.

In high-level task planning, the process is represented as
\begin{equation}
P_H \rightarrow [P_{M1}, P_{M2}, \dots, P_{Mn}],
\end{equation}
where $P_H$ denotes the high-level task, and $P_{Mi}$ is the $i$-th medium sub-task resulting from decomposition.

In low-level task planning, the process is given by
\begin{equation}
P_{Mi} \rightarrow [a_1^{(i)}, a_2^{(i)}, \dots, a_{m_i}^{(i)}],
\end{equation}
where $a_{j}^{(i)}$ is $j$-th low level action for the $i$-th medium sub-task $P_{Mi}$, with $m_{i}$ indicating the number of actions in the sequence.

The final action plan $\pi$ is a concatenation of these action sequences:
\begin{equation}
\pi = (a_1^{(1)}, \dots, a_{m_1}^{(1)}, a_1^{(2)}, \dots, a_{m_2}^{(2)}, \dots, a_1^{(n)}, \dots, a_{m_n}^{(n)}).
\end{equation}

\subsubsection{Open-loop vs. Closed-loop}
Based on whether planning relies on multiple observations during the interaction process \cite{hu2023tree,sun2023adaplanner}, it can be categorized into \textit{open-loop planning} and \textit{closed-loop planning}.

In open-loop planning, the process can be formulated as
\begin{equation}
\pi = (a_0, a_1, \dots, a_{n-1})=\arg\max_{a_{0:n-1}} p_\theta(a_{0:n-1}|s_0),
\end{equation}
where $a_{i:j}$ denotes the action sequence executed from time step $i$ to $j$, $p_\theta$ denote the model's predicted probability, $s_i$ is the state at time step $i$, and $n$ indicates the total length of the action sequence. 

In closed-loop planning, the process can be expressed as
\begin{equation}
\pi = (a_0, a_1, \dots, a_{n-1}), with \, a_{i}=\arg\max_{a_{i}} p_\theta(a_{i}|s_{0:i}),
\end{equation}
where $a_{i}$ is the action taken at time step $i$, $p_\theta$ is the predicted probability from the model, $s_{i:j}$ denotes the sequence of states from time step $i$ to $j$.

\begin{figure*}[t]
\centering
\footnotesize
\begin{forest}
    for tree={
        forked edges,
        grow'=0,
        draw,
        rounded corners,
        node options={align=center,},
        text width=7cm,
        s sep=6pt,
        calign=edge midpoint,
    }
    [Planning Methods (\S\ref{PM}), fill=gray!45, parent
        [External Module Augmented Methods (\S\ref{EMAM}), for tree={methods}
            [Planner Enhanced Methods (\S\ref{planner}), methods
                [LLM+P \cite{liu2023llm+}\text{,} LLM-DP \cite{dagan2023dynamic}\text{,} PDDLEGO \cite{zhang2024pddlego}\text{,} LLM+ASP \cite{yang2023coupling}\text{,} Ada \cite{wong2023learning}\text{,} Hao et al. \cite{hao2024large}\text{,} Gundawar et al. \cite{gundawar2024robust}\text{,} Xie et al. \cite{xie2023translating}\text{,} Zhang et al. \cite{zhang2024proc2pddl}\text{,} Kambhampati et al. \cite{kambhampatiposition}\text{,} Guan et al. \cite{guan2023leveraging}\text{,} etc., methods_work]
            ]
            [Memory Enhanced Methods (\S\ref{memory}), methods
                [MemoryBank \cite{zhong2024memorybank}\text{,} TiM \cite{liu2023think}\text{,} MemGPT \cite{packer2023memgpt}\text{,} MoT \cite{li2023mot}\text{,} CLIN \cite{majumder2023clin}\text{,} DT-Mem \cite{kang2023think}\text{,} Yin et al. \cite{yin2024explicit}\text{,} Zheng et al. \cite{zheng2023synapse}\text{,} Wang et al. \cite{wang2024jarvis}\text{,} Zhang et al. \cite{zhang2024large}\text{,} Zhu et al. \cite{zhu2023ghost}\text{,} etc., methods_work]
            ]
        ]
        [Finetuning-based Methods (\S\ref{FM}), for tree={representation}
            [Imitation Learning-based Methods (\S\ref{imitation}), representation
                [Zhang et al. \cite{zhang2023you}\text{,} Agent-FLAN \cite{chen2024agent}\text{,} FireAct \cite{chen2023fireact}\text{,} CoPlanner \cite{wang2024cooperative}\text{,} KnowAgent \cite{zhu2024knowagent}\text{,} AUTOACT \cite{qiao2024autoact}\text{,} Gandhi et al. \cite{gandhi2024stream}\text{,} Su et al. \cite{su2024dualformer}\text{,} Lehnert et al. \cite{lehnert2024beyond}\text{,} AGENTGEN \cite{hu2024agentgen}\text{,} KnowSelf \cite{qiao2025agentic}\text{,} etc., representation_work]
            ]
            [Feedback-based Methods (\S\ref{feedback}), representation
                [Yuan et al. \cite{yuan2024advancing}\text{,} A3T \cite{yang2024react}\text{,} Wang et al. \cite{wang2024llms}\text{,} S2RCQL \cite{deng2024can}\text{,} WEBRL \cite{qi2024webrl}\text{,} ReST-MCTS* \cite{zhang2024rest}\text{,} Jiao et al. \cite{jiao2024learning}\text{,} RefAug \cite{zhang2024learn}\text{,} SPPO \cite{wu2024self}\text{,} Wang et al. \cite{wang2024cpl}\text{,} \cite{wu2024thinking}\text{,} Step-DPO \cite{lai2024step}\text{,} ToolPlanner \cite{wu2024toolplanner}\text{,} etc., representation_work]
            ]
        ]
        [Searching-based Methods (\S\ref{SM}), for tree={probing}
            [Decomposition-based Methods (\S\ref{decomposition}), probing
                [HuggingGPT \cite{shen2023hugginggpt}\text{,} Zhou et al. \cite{zhou2022least}\text{,} RaDA \cite{kim2024rada}\text{,} Zhang et al. \cite{zhang2024meta}\text{,} Sun et al. \cite{sun2024retrieval}\text{,} UrbanLLM \cite{jiang2024urbanllm}\text{,}  Sprague et al. \cite{sprague2023deductive}\text{,} Li et al. \cite{li2024can}\text{,} ADAPT \cite{prasad2023adapt}\text{,} TRIP \cite{zhang2024strength}\text{,} HRV \cite{cornelio2025hierarchical}\text{,} Protrix \cite{wu2403protrix}\text{,} etc., probing_work]
            ]
            [Exploration-based Methods (\S\ref{exploration_method}), probing
                [ToT \cite{yao2023tree}\text{,} SearChain \cite{xu2024search}\text{,} AVIS \cite{hu2023avis}\text{,} AoT \cite{sel2023algorithm}\text{,} GoT \cite{besta2024graph}\text{,} Tree-Planner \cite{hu2023tree}\text{,} AGoT \cite{pandey2025adaptive}\text{,} LLM-MCTS \cite{zhao2023large}\text{,} MC-DML \cite{shimonte}\text{,}  WebPilot \cite{zhang2024webpilot}\text{,} DR-MCTS \cite{liu2025doubly}\text{,} LLM-A* \cite{meng2024llm}\text{,} RAP \cite{hao2023reasoning}\text{,} ReasonPlanner \cite{dinh2024reasonplanner}\text{,} QLASS \cite{lin2025qlass}\text{,} ARMAP \cite{chen2025scaling} etc., probing_work]
            ]
            [Decoding-based Methods (\S\ref{decoding}), probing
                [Xie et al. \cite{xie2024self}\text{,} Wang et al. \cite{wang2024chain}\text{,} Liu et al. \cite{liu2024don}\text{,} Roy et al. \cite{roy2024flap}\text{,} Predictive decoding \cite{ma2024non}\text{,} Contrastive decoding \cite{o2023contrastive}\text{,} Grounded decoding \cite{huang2024grounded}\text{,} etc., probing_work]
            ]
        ]
    ]
\end{forest}   
\caption{The typology of LLM-based planning methods, which includes external module augmented methods, finetuning-based methods, and searching-based methods.}
\label{fig:tax}
\end{figure*}
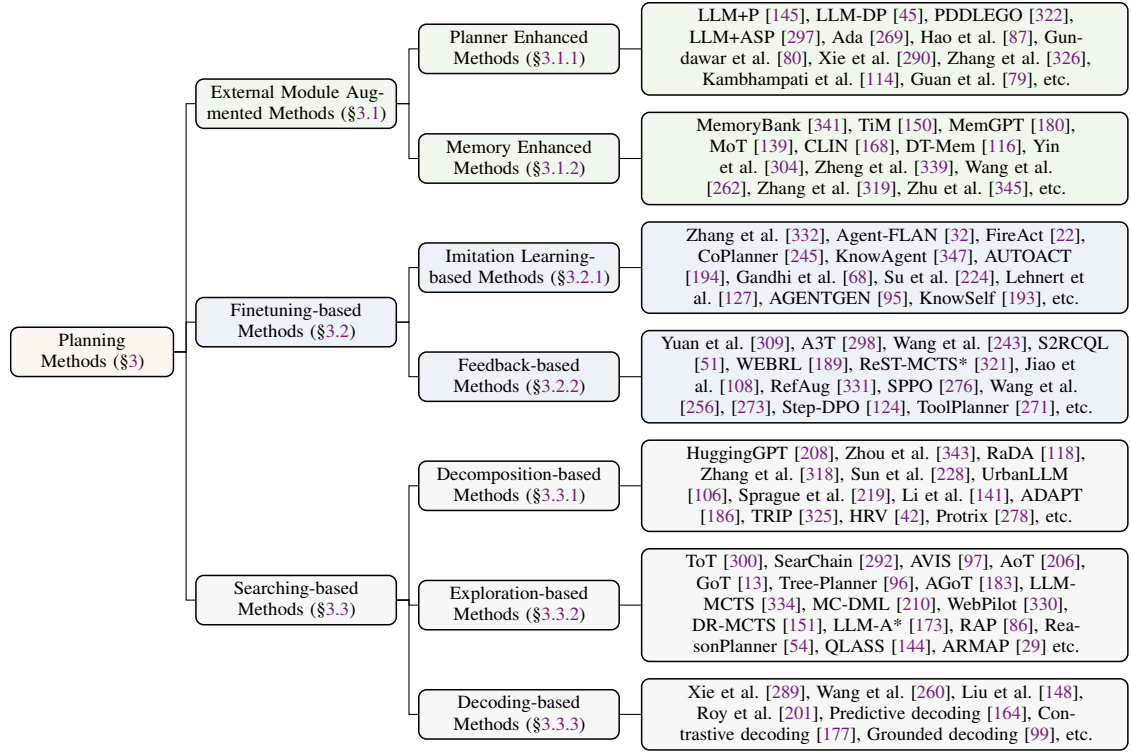

\section{Planning Methods} \label{PM}
This section presents a comprehensive overview for LLM-based planning methods. As shown in Figure \ref{fig:tax}, existing methods can be divided into External Module Augmented
Methods (\S\ref{EMAM}), Finetuning-based Methods (\S\ref{FM}), and Searching-based Methods (\S\ref{SM}). We will illustrate each type of method in detail.

\subsection{External Module Augmented Methods} \label{EMAM}
\begin{figure*}[t]
\centering
  \includegraphics[width=0.9\textwidth,height=0.5\textheight,keepaspectratio]{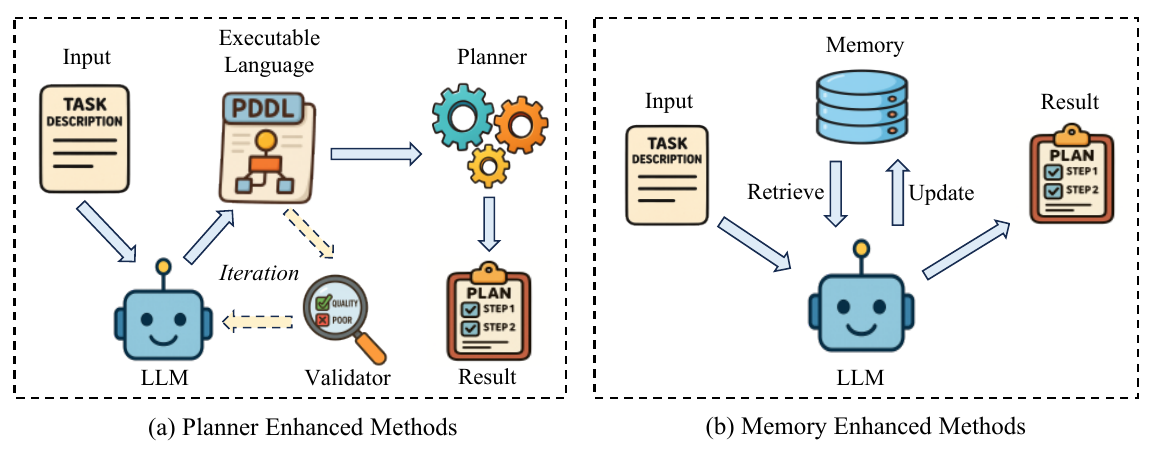}
  \caption{The illustration of external module augmented methods, which includes planner enhanced methods and memory enhanced methods.}
  \label{external}
\end{figure*}

LLMs demonstrate impressive reasoning capabilities and possess extensive knowledge, yet they face significant challenges when addressing complex planning tasks. To overcome these limitations, some research has explored augmenting LLMs with external components such as symbolic planners \cite{liu2023llm+,dagan2023dynamic,zhang2024pddlego,zhang2024proc2pddl,wu2024can,gundawar2024robust,chu2025llm+} and memory modules \cite{yin2024explicit,kang2023think,zheng2023synapse,park2023generative,zhong2024memorybank,wang2023ldm2,cornelio2025hierarchical}. These hybrid approaches aim to enhance the planning performance of LLMs by integrating structured reasoning and long-term memory capabilities. This type of methods can be further divided into \textbf{Planner Enhanced Methods} (\S\ref{planner}) and \textbf{Memory Enhanced Methods} (\S\ref{memory}), as shown in Figure \ref{external}.

\subsubsection{Planner Enhanced Methods} \label{planner}
External planners play an important role in the field of automated planning, which typically takes executable codes as input to generate reliable plans \cite{de2008z3,barrett2011cvc4}. Within this paradigm, the quality of the generated code becomes a critical bottleneck in effectively solving planning tasks. To this end, several studies explore the use of LLMs as translators that convert natural language descriptions of planning problems into executable code \cite{guan2023leveraging,zhang2024pddlego,wong2023learning}.

To be specific, LLM+P \cite{liu2023llm+} begins by transforming a natural language description of a planning problem into a Planning Domain Definition Language (PDDL) file, a standardized format for representing classical planning problems. Next, classical planners are employed to efficiently identify an optimal solution or plan based on the PDDL file. Finally, the generated solution is translated back into natural language via the LLM, ensuring it is accessible and understandable to humans. LLM-DP \cite{dagan2023dynamic} also introduces a similar neuro-symbolic framework, which uses the LLM to translate natural language instructions into PDDL goals and utilizes a symbolic planner to generate executable plans. Additionally, it monitors the world state to adapt to environmental changes and support informed decision-making. TRIP-PAL \cite{de2024trip} is also a hybrid method that first employs LLMs to format travel and user information, and then utilizes automated planning solver to generate travel plans. Although these methods of using LLMs as PDDL translators have made progress, the generated PDDL may contain errors that directly affect the subsequent planning. Therefore, some researchers have attempted to improve the quality of the generated PPDL. Guan et al. \cite{guan2023leveraging} propose a framework that prompts an LLM to construct initial PDDL models and then corrects errors based on external feedback sources, including PDDL model validation tools and human domain experts. The refined PDDL can then be passed to either a classical planner or an LLM-modulo planner to produce the final plan. PDDLEGO \cite{zhang2024pddlego} designs an iterative strategy to improve PDDL files. In detail, when the file lacks sufficient information to achieve the end goal, it recursively falls back to a provided sub-goal. This iterative process enables environmental exploration and progressive refinement until the goal is achieved. Similarly, Mahdavi et al. \cite{mahdavileveraging} also propose an iterative refinement method that generates multiple PDDL candidates and employs the exploration walk metric to provide feedback signals, guiding the selection and enhancement of the final PDDL file.

In addition to translating planning problems into PDDL, some researchers also reformulate them into other forms and employ specialized solvers for plan generation. LLM+ASP \cite{yang2023coupling} utilizes an LLM to convert a problem description into atomic facts. Then, an Answer Set Programming (ASP) solver along with background knowledge takes the facts as the input to predict the answer. Hao et al. \cite{hao2024large} formalize the complex multi-constraint planning problems as constrained satisfiability problems. Their approach first leverages LLMs to translate user queries into code, and subsequently applies a satisfiability modulo theories (SMT) solver to solve the problem. Ju et al. \cite{ju2024globe} propose a real-time planning system called TTG that employs LLMs to translate language inputs into symbolic forms, and utilizes Mixed Integer Linear Programming (MILP) solvers to generate solutions. Kambhampati et al. \cite{kambhampatiposition} propose that the LLM-Modulo framework offers an effective approach to integrate the approximate knowledge generation capabilities of LLMs with external verifiers, enabling robust and reliable planning. To further ensure the correctness of the output, Gundawar et al. \cite{gundawar2024robust} integrate LLMs with a comprehensive set of symbolic verifiers that validate the outputs, and trigger reprompting when validation fails.

Meanwhile, some researchers also attempt to evaluate the translation capability of of LLMs from natural languages to planning languages. In detail, Xie et al. \cite{xie2023translating} conduct an empirical study evaluating the effectiveness of LLMs in translating natural language goals into PDDL. Their findings reveal that LLMs are significantly more adept at translation tasks than at performing planning tasks. Similarly,  Zhang et al. \cite{zhang2024proc2pddl} construct a dataset called \textsc{Proc2Pddl} that contains procedural texts paired with expert-annotated PDDL representations. They use the dataset to evaluate whether LLMs can translate natural language texts to PDDL symbolic language in open-domain simulated environments. To accurately measure the quality of generated PDDL, Zuo et al. \cite{zuo2024planetarium} design a benchmark called Planetarium that contains different tasks with varying levels of difficulty. In addition, Stein et al. \cite{stein2023autoplanbench} propose a method called \textsc{AutoPlanBench} that converts PDDL representations of planning problems to natural language encoding for evaluating planning capabilities of LLMs.

\subsubsection{Memory Enhanced Methods} \label{memory}
The memory module is a critical component in the architecture of LLM-based agents, serving to store information obtained from historical states or the environment \cite{wang2024survey,xi2023rise}. Several studies leverage memory to enhance the planning capabilities of LLMs \cite{yin2024explicit,wang2024jarvis,park2023generative,wang2023ldm2,kang2023think}.

Specifically, Zhong et al. \cite{zhong2024memorybank} propose a memory mechanism called MemoryBank, which stores and recalls historical interactions to enhance performance in long-term interaction scenarios. Similarly, Liu et al. \cite{liu2023think} introduce the Think-in-Memory (TiM) framework, designed to retain and retrieve historical thoughts for improved reasoning. To further enhance retrieval efficiency, they develop a specialized hashing algorithm for long-term conversations. MemGPT \cite{packer2023memgpt}, inspired by operating system design, addresses context length limitations and demonstrates effectiveness in tasks such as document analysis and conversational agents. Zhang et al. \cite{zhang2024large} present an agent framework called \textsc{Rememberer}, tailored for decision-making and planning tasks. This framework equips LLMs with long-term experience memory and employs a reinforcement learning algorithm to update that memory over time. Li et al. \cite{li2023mot} propose a reasoning framework named MoT, which performs ``pre-thinking'' on unlabeled data to store high-confidence thoughts as external memory. During the testing phase, relevant thoughts are retrieved to support reasoning.

To generalize better to new environments, a memory-augmented language agent called CLIN is proposed for sequential decision-making tasks \cite{majumder2023clin}. It employs a continually evolving memory to store causal abstractions derived from reflections on past experiences, which in turn supports future planning. Similarly, Zhu et al. \cite{zhu2023ghost} design a text-based memory for LLM-based agent to store and retrieve gained experiential knowledge, leading to improved generalization in planning tasks. Yin et al. \cite{yin2024explicit} propose a memory learning framework called EM$^{2}$, which treats the explicit memory as a latent and utilizes the Expectation-Maximization (EM) algorithm to update it. In addition, storing past successful experiences has also been shown to improve generalization in planning \cite{wang2024jarvis,sarch2023open,zheng2023synapse}. For example, Wang et al. \cite{wang2024jarvis} introduce a new agent called JARVIS-1 for long-horizon planning tasks. To enhance the correctness and consistency of planning, they augment the agent with a multimodal memory, which stores past successful planning experiences. Zheng et al. \cite{zheng2023synapse} present \textsc{Synapse}, an LLM-based agent tailored for planning tasks that leverages an exemplar memory to store successful trajectories and retrieve them to improve generalization on novel tasks. Inspired by the concept of working memory \cite{graves2014neural,sukhbaatar2015end}, Kang et al. \cite{kang2023think} propose a new approach called DT-Mem for decision-making tasks. It stores and manipulates information in a parametric manner, demonstrating strong generalization and planning capabilities across multiple game-based tasks. 


Several works also focus on modeling memory structures to enhance planning capabilities \cite{yoo2024exploratory}. For example, Wang et al. \cite{wang2024crafting} regard user data generated on personal devices as memories and organize it in a graph form. They further introduce a retrieval-augmented generation technique to utilize the memory for downstream applications. Yoo et al. \cite{yoo2024exploratory} introduce an exploratory retrieval-augmented planning framework, which employs a graph-based context memory to enhance embodied reasoning capabilities of LLMs via the accumulation of past observations. For partially observable markov decision processes (POMDP) tasks, Yue et al. \cite{yue2024learning} introduce the concept of memory dependency pairs to recall past observations. Meanwhile, they also propose a method called AttentionTuner to incorporate the memory dependency pairs into self-attention modules.

\begin{tcolorbox}[colback=blue!5!white,colframe=black!75!white,
  title=\textbf{Summary of External Module Augmented Methods}, 
  fonttitle=\bfseries, coltitle=white, colbacktitle=black, rounded corners]
  \begin{itemize}[leftmargin=1.5em]
    \item \textit{Planner Enhanced Methods} first usually generate executable codes and then employ an external planner to output reliable plans. However, the effect is limited by the quality of code conversion. Also, the types and scope of external planners impose limitations on the generalizability. 
    \item \textit{Memory Enhanced Methods} improve the planning capabilities of LLMs with an extra memory module that stores knowledge, past experience and so on. Although effective, this type of method heavily depends on the accuracy of retrieval algorithm. How to efficiently and effectively update memory module is also a key issue.
  \end{itemize}
\end{tcolorbox}


\subsection{Finetuning-based Methods} \label{FM}
Currently, LLMs still face challenges in handling complex planning tasks. To address this limitation, researchers have proposed finetuning-based methods to update the parameters of LLMs using trajectory or feedback data, which aims to enhance LLMs’ adaptability to specific tasks and improve their ability to generate planning strategies that align more closely with human expectations. This type of methods can be further categorized into \textbf{Imitation Learning-based Methods} (\S\ref{imitation}) and \textbf{Feedback-based Methods} (\S\ref{feedback}). 

\subsubsection{Imitation Learning-based Methods} \label{imitation}
\begin{figure}[t]  
    \centering  
    \includegraphics[width=0.5\textwidth]{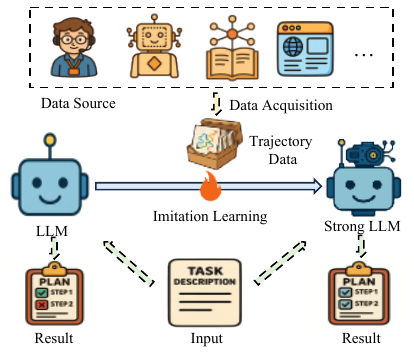}  
    \caption{The illustration of imitation learning-based methods that use trajectory data to update the parameters of LLMs for enhancing their planning capabilities.}  
    \label{fig:imitation}  
\end{figure}

Imitation learning is a machine learning paradigm that enables models to acquire task execution strategies by mimicking expert or human behavior. In contrast to reinforcement learning—which depends on environment-based reward signals—imitation learning leverages demonstration trajectories or behavioral data as direct learning targets. Its primary goal is to equip LLMs with the ability to perform complex tasks efficiently, especially in environments where reward functions are undefined or exploration is costly. This makes imitation learning particularly advantageous in such settings. Based on the source of trajectory data, these studies can be categorized into three types: \textbf{Expert Demonstration Data}, \textbf{Self-Generated Data} and \textbf{Hybrid Data}. The imitation learning-based method is shown in Figure \ref{fig:imitation}.

\textbf{Expert Demonstration Data:} 
Expert demonstration data is one of the most common and intuitive sources in imitation learning. It refers to trajectory data collected by recording the actions and decisions of human experts or high-performance systems while performing a specific task. Such data is typically of high quality, accelerates the agent's learning process, and is especially valuable in complex planning scenarios \cite{zhang2023you,furutamultimodal,zhang2024ask}. For instance, Zhang et al. \cite{zhang2023you} introduce Auto-GUI and propose an action-chain technique that leverages sequences of past actions along with future action plans to simulate behavior, thereby guiding the agent in decision-making. However, this approach may not adhere to scaling laws, meaning that increasing model size does not necessarily lead to performance gains. Although LLMs have made significant progress in various tasks, they often generate factual errors due to their limited internal knowledge. Some research has explored ways to enhance LLMs' planning capabilities from this perspective \cite{chen2024agent,cobbe2021training}. Specifically, Chen et al. \cite{chen2024agent} consider integrating agent capabilities into general-purpose LLMs to bridge the gap between specialized and general-purpose models. It aligns training corpora with chat formats and introduces an innovative method called Agent-FLAN, which effectively integrates agent capabilities into LLMs. Chen et al. \cite{chen2023fireact} propose a finetuning framework named FireAct that uses task-solving trajectories generated by high-performance systems to fine-tune language models, enabling them to imitate trajectory data for action. Ahmad et al. \cite{ahmad2025opencodereasoning} leverage DeepSeek-R1 to generate trajectory data with reasoning chains, constructing OpenCodeReasoning dataset, which includes 736,000 Python samples covering 28,904 unique competitive programming problems. 

Leveraging the concept of knowledge distillation, TAPIR \cite{yue2024distilling} uses a teacher model to select challenging tasks from a set of instructions for student model and generates high-quality task-response pairs for training  student model. From the perspective of retrieval-augmented generation (RAG), Lyu et al. \cite{lyu2024retrieve} propose Retrieve-Plan-Generation, which generates plan tokens using expert-labeled paragraphs and answer data, guiding subsequent generations. Moreover, researchers also attempt to address the issue of agent non-compliance with user-specified constraints in complex and long-horizon tasks \cite{wu2024selp,ding2024horizon}. For instance, Wu et al. \cite{wu2024selp} borrow the idea of probability distributions and introduce Linear Temporal Logic (LTL) formulas. By generating and sampling multiple LTL formulas from natural language commands, grouping equivalent formulas, and selecting a majority of the formulas to execute planning, consistency is achieved. By decomposing complex tasks into multiple steps and assigning responsibilities to different agents, Wang et al. \cite{wang2024cooperative} propose CoPlanner, which improves training efficiency through behavioral cloning initialization and difficulty-aware curriculum filtering methods. Morishita et al. \cite{morishita2024enhancing} propose four principles for synthetic sample design based on symbolic logic theory and empirical research: reasoning with unknown facts, non-logical reasoning samples, diverse inference rules, and linguistic expressivity.

\textbf{Self-Generated Data:}
The self-generated data-based methods involve allowing agent to interact with environment and autonomously generate trajectory data, which is then used to improve its own planning strategy \cite{zhu2024knowagent,qiao2024autoact,wang2024learning,wang2024adapting}.

To address issues such as planning hallucinations in complex tasks, Zhu et al. \cite{zhu2024knowagent} propose KnowAgent, which incorporates explicit action knowledge through the use of an action knowledge base and self-learning strategies, enabling trajectory synthesis. The generated trajectories are then used for iterative optimization to enhance LLM's planning capabilities. Similarly, Qiao et al. \cite{qiao2024autoact} propose AUTOACT, which synthesizes data through a toolkit and an automatic planning module and uses a division of labor strategy to generate a team of sub-agents to complete the task. Gandhi et al. \cite{gandhi2024stream} use search trajectories generated by heuristic solvers as search streams (SoS) to train the language model to learn various symbolic search strategies. In the pretraining phase, the model improves search accuracy by learning to predict the optimal search trajectories and understanding how to optimize the search process. Then, the model is fine-tuned to further optimize its search strategies and reasoning abilities. In this way, the model not only learns from the generated synthetic data but also gradually discovers and adapts to new search strategies.

Some studies seek to improve model inference speed and reduce computational costs \cite{su2024dualformer,xie2024monte}. Su et al. \cite{su2024dualformer} employ Dualformer, which integrates fast and slow reasoning modes within a Transformer framework. This model is trained using complete inference trajectories automatically generated by the A* search algorithm, providing faster reasoning while maintaining accuracy. Similarly, Lehnert et al. \cite{lehnert2024beyond} model the search steps and enables Transformer to predict the actions of A* search algorithm, focusing on longer and more complex search trajectories. To reduce data costs while still obtaining rich knowledge to complete tasks, Synatra \cite{ou2024synatra} extracts indirect knowledge from network tutorials, random webpages and generative environments, converting this knowledge into direct demonstrations to enhance data quality. Similarly, Mai et al. \cite{mai2024learning} transform unlabeled text corpora into writing action sequences through inference and use these actions to guide the language model. AGENTGEN \cite{hu2024agentgen} automatically generates diverse environments and planning tasks to synthesize high-quality trajectory data, which is then used for fine-tuning to improve LLM's planning performance, ensuring the model's ability to generalize and optimize across a wider range of tasks. Li et al. \cite{li2025self} employ a multi-objective DPO-aligned ensemble of policy models to generate diverse candidate responses using multi-preference weights and then invoke the model to produce improvement suggestions for candidate responses and rewrites them based on these suggestions, thereby synthesizing optimized trajectory data.

\textbf{Hybrid Data:} 
The hybrid data-based imitation learning method is a strategy that combines multiple data sources to compensate for the limitations of any single data source, maximizing the diversity and quality of the data to improve the effectiveness and generalization ability of the planning model \cite{wang2024mixture,fu2024camphor,qiao2024agent,gur2023real}.

A promising open direction is how to leverage the collective expertise of multiple LLMs. Wang et al. \cite{wang2024mixture} use a hierarchical mixture of experts (MoE) architecture to take advantage of the collective strengths of multiple LLMs. This approach uses supervised expert data and self-generated data from intermediate outputs of agents in the hierarchical structure to guide the behavior of next layer, thus improving LLM's planning capabilities. Similarly, CAMPHOR \cite{fu2024camphor} also use a multi-layer MoE architecture, enhancing performance through expert demonstration data and generated data. The difference is that CAMPHOR reduces model size, latency, and memory usage significantly by implementing parameter sharing among agents and using real-time compression. To address limitations in robotic planning, Li et al. \cite{li2024closed} combine model-generated planning data with environmental feedback data for end-to-end optimization, enhancing the LLM's planning capabilities. Similarly, Lai et al. \cite{lai2024autowebglm} propose AutoWebGLM, which uses expert demonstration data built by Human-AI collaboration and operation trajectories from simulated environments as guidance. Drawing on the human mental world knowledge model, which provides global prior knowledge and maintains local dynamic knowledge, Qiao et al. \cite{qiao2024agent} guides the agent to self-synthesize knowledge from expert and sampled trajectories, and then develops world knowledge model to provide prior knowledge for guidance.

\subsubsection{Feedback-based Methods} \label{feedback}
\begin{figure}[t]  
    \centering  
    \includegraphics[width=0.5\textwidth]{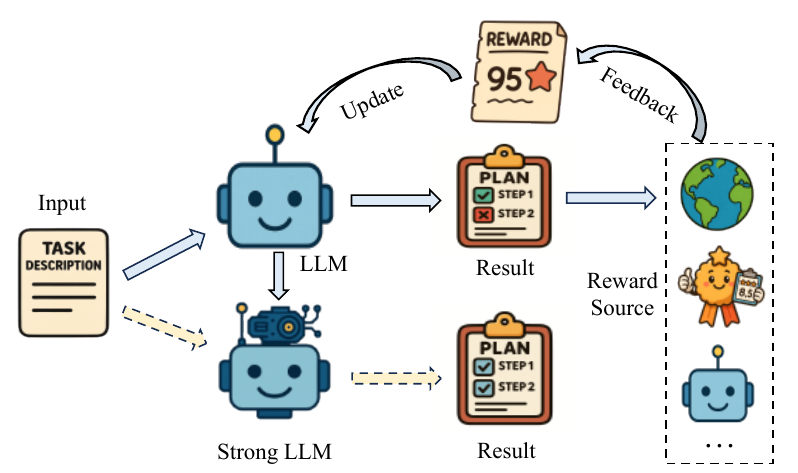}  
    \caption{The illustration of feedback-based methods that employ the feedback as the signal to guide the update of LLMs' parameters.}  
    \label{fig:feedback}  
\end{figure}

This section explores methods for optimizing the planning abilities of LLMs by introducing feedback mechanisms, as shown in Figure \ref{fig:feedback}. Specifically, we classify feedback into three categories based on its source: \textbf{Environmental Feedback}, \textbf{Reward Model Feedback}, as well as \textbf{Self-Generated and External Feedback}. These feedback mechanisms provide dynamic optimization signals for finetuning, helping LLMs gradually improve their decision-making abilities in planning tasks.

\textbf{Environmental Feedback:} 
Environmental feedback is a key factor in optimizing the planning capabilities of LLMs. By interacting with the task environment, LLMs can adjust their planning strategies based on feedback signals provided by environment (such as success/failure indicators, state changes, or rewards). In self-play gaming scenarios, such as language games\cite{cheng2024self}, chess, and Go\cite{silver2017mastering}, the rules and outcomes of these games provide direct feedback to LLM, allowing it to optimize its strategy based on game results. In the field of programming, interpreters\cite{yuan2024advancing}\cite{haluptzok2022language} can verify the correctness of programming problems and planning solutions generated by LLMs, which enables LLMs to generate problems and improve programming abilities. In the automatic annotation and optimization of agent trajectories, A3T proposed by Yang et al. \cite{yang2024react} utilizes both environmental observation and reward feedback mechanisms to automatically annotate trajectory quality, reducing manual intervention and improving decision-making abilities. Additionally, Song et al. \cite{song2024trial} optimize LLM action trajectories through trial-and-error exploration, where agents collect failure trajectories from environment and compare them with expert trajectories. 

Using contrastive learning, the model iteratively updates its strategy, enhancing performance in complex tasks. Wang et al. \cite{wang2024llms} strengthen LLMs' self-learning abilities in tool usage through simulated trial-and-error and feedback from API execution results, as well as self-reflection. S2RCQL  \cite{deng2024can} alleviates spatial hallucinations in LLMs by converting spatial cues into entity relationships. It employs the Q-learning algorithm, utilizing reward signals from environment to update the Q-table, thereby guiding LLMs to learn the optimal path. To enhance LLMs' continuous reasoning capabilities in complex tasks, some studies have designed frameworks from the perspective of long-chain-of-thought training. For example, Yu et al. \cite{yu2025think} enable models to dynamically adjust reasoning length based on problem difficulty, optimizing computational resource allocation. Du et al. \cite{du2025boost} propose BDC for code generation tasks, which systematically improves LLMs' robustness, generalization, and scenario adaptability in complex code generation through a synergistic mechanism of multi-stage reasoning enhancement, code semantic disentanglement, and dynamic parameter customization.


\textbf{Reward Model Feedback:} 
A reward model is a specific function that helps LLMs gradually optimize their strategies by providing evaluations of task completion or the quality of the planning process. This model can be rule-based, model-learned, or learned from data, with the core idea being to generate feedback signals through the reward model to guide the model’s learning and decision-making process.

In outcome reward models \cite{singh2023beyond,kumar2409training,yu2024flow}, WEBRL \cite{qi2024webrl} uses LLMs as outcome-supervised reward models to evaluate task completion and provide feedback signals to web agents, allowing them to optimize strategies in web tasks. Similarly, SCoRe \cite{kumar2409training} applies reward shaping techniques on a binary reward basis to evaluate the correctness of LLM-generated responses, guiding its self-correction behavior and significantly improving performance in mathematical and programming tasks. This result-based feedback directly correlates with task goals and can effectively improve the model's final performance. However, its limitation lies in the binary nature of feedback signals (e.g., success/failure), which lack detailed guidance on intermediate steps. To address this limitation, process reward models have emerged \cite{zhang2024rest}, whose core idea is to provide feedback or guidance on intermediate steps or states during task execution. Wang et al. \cite{wang2024q} use a process reward model to evaluate the quality of each reasoning step, providing intermediate signals for training Q-value models, allowing them to accurately estimate the expected future rewards for each state-action pair and helping LLMs select the most promising reasoning paths. Jiao et al. \cite{jiao2024learning} rank trajectories based on synthetic process rewards, using DPO to enhance LLM planning abilities. This process-based feedback can provide more detailed guidance, helping models gradually optimize their strategies in complex tasks. However, its design is complex, computationally expensive, and requires high data acquisition costs.

To enhance the performance of large language models in long-chain reasoning tasks, recent studies have developed specialized frameworks across different domains.  For GUI tasks requiring cross-platform generalization \cite{xia2025gui,lu2025ui,liu2025infigui}, Xia et al. \cite{xia2025gui} integrate interaction actions from multiple platforms and guide the model to progressively refine its reasoning process through reward signals. Actor2Reasoner \cite{liu2025infigui} leverages a teacher model to generate high-quality trajectories with explicit reasoning steps, utilizing format rewards, task-specific rewards, and GUI element localization rewards to encourage student models to learn intermediate reasoning. In the field of embodied agents, Zhao et al. \cite{zhao2025embodied} propose a perception-reasoning decoupled architecture, where large vision-language models process long videos to dynamically generate semantic representations enriched with action/spatial/task information and a small language model trained with different rewards focuses on reasoning. For scenarios requiring LLMs to interface with external tools, challenges such as inefficient tool utilization and limited generalization persist \cite{feng2025retool,qian2025toolrl,wang2025otc}. ReTool \cite{feng2025retool} addresses this by dynamically executing code and applying result-driven RL training, enabling models to autonomously discover optimal tool-calling strategies. This approach significantly improves accuracy and efficiency in complex mathematical reasoning, offering a novel paradigm for hybrid neuro-symbolic systems. For gaming environments demanding visual-spatial reasoning, Dao et al. \cite{dao2025alphamaze} design a comprehensive reward function balancing accuracy, action legality, and reasoning interpretability. This trains models to backtrack from erroneous paths and self-correct, surpassing the singular alignment objectives of traditional RLHF. In software engineering tasks, Wei et al. \cite{wei2025swe} pioneer RL application by designing a continuous reward function based on code-patch text similarity. This enables medium-scale open-source models (70B parameters) to achieve software engineering capabilities rivaling proprietary systems.


\textbf{Self-Generated and External Feedback:} 
In the process of optimizing LLM planning abilities, feedback from other models plays an important role in addition to environmental feedback and reward model feedback. This feedback typically comes from the model’s own generation, evaluation, or comparison, and can provide optimization signals to LLMs without relying on external environments or manual annotations.

One common type of feedback is self-generation and adversarial training, as seen in SPIN \cite{chen2024self}, which uses a self-adversarial mechanism to help LLMs grow from weak models to strong ones. This method uses data generated by the model itself as feedback to optimize its performance, without the need for additional manual annotations. RefAug \cite{zhang2024learn} appends a generated reflection section, which includes alternative reasoning and subsequent reasoning, to the original answer of each training instance. These reflection sections are treated as part of the training data, guiding the model's learning and optimization process. SPPO \cite{wu2024self} iterates using synthetic data generated by the model itself and feedback provided by preference models to optimize model preferences to better align with human expectations. CPO \cite{zhang2024chain} and CPL \cite{wang2024cpl} fine-tune LLMs by leveraging preference information generated through tree-structured reasoning or search-based methods, thereby aligning the model's reasoning pathways more closely with human-like reasoning processes. These methods reduce reliance on external resources by leveraging internal self-feedback mechanisms, but their effectiveness depends on the model’s own generation and evaluation capabilities. 

To address the limitations of these methods, some researchers have explored enhancing LLM planning abilities using critique models. For example, AutoMathCritique \cite{xi2024enhancing} trains critique models that provide step-level supervision and constructive feedback, ensuring the diversity and comprehensiveness of reasoning paths. Wu et al. \cite{wu2024thinking} improve answer quality by leveraging internally generated thinking processes, using evaluation models to assess generated responses, and optimizing the model's instruction-following capabilities. These methods provide multi-layered feedback mechanisms for LLMs, but their success depends on the design quality of critique models. To enhance AI agents' mathematical reasoning capabilities, Sun et al. \cite{sun2025error} developed a dynamically adaptive error classification framework that generates context-aware prompts based on problem-related error patterns to guide models in avoiding common mistakes, while constructing MWPES-300K dataset with clearly defined error categories.

\begin{tcolorbox}[colback=blue!5!white,colframe=black!75!white,
  title=\textbf{Summary of Finetuning-based Methods}, 
  fonttitle=\bfseries, coltitle=white, colbacktitle=black, rounded corners]
  \begin{itemize}[leftmargin=1.5em]
    \item \textit{Imitation Learning-based Methods} aim to quickly adapt to complex tasks by mimicking expert decisions or leveraging self-generated data. However, challenges still remain, including the cost and efficiency of data acquisition, as well as the model's generalization to adapt to new scenarios in dynamic environments. 
    \item \textit{Feedback-based Methods} introduce feedback signals, from environments, reward model and so on, to dynamically adjust model behavior and improve planning capabilities. Although popular, this type of method heavily relies on high-quality feedback signals. Moreover, it is difficult to train a universal reward model for various planning tasks.
  \end{itemize}
\end{tcolorbox}

\subsection{Searching-based Methods} \label{SM}
In addition to utilizing external modules and fine-tuning parameters, researchers have also proposed searching-based methods that fully utilize the potential of LLMs for planning. Existing studies focus on input task decomposition, target exploration, and decoding optimization. Thus, this types of methods includes \textbf{Decomposition-based Methods} (\S\ref{decomposition}), \textbf{Exploration-based Methods} (\S\ref{exploration_method}) and \textbf{Decoding-based Methods} (\S\ref{decoding}).

\begin{figure*}[t]  
    \centering  
    \includegraphics[width=0.97\textwidth]{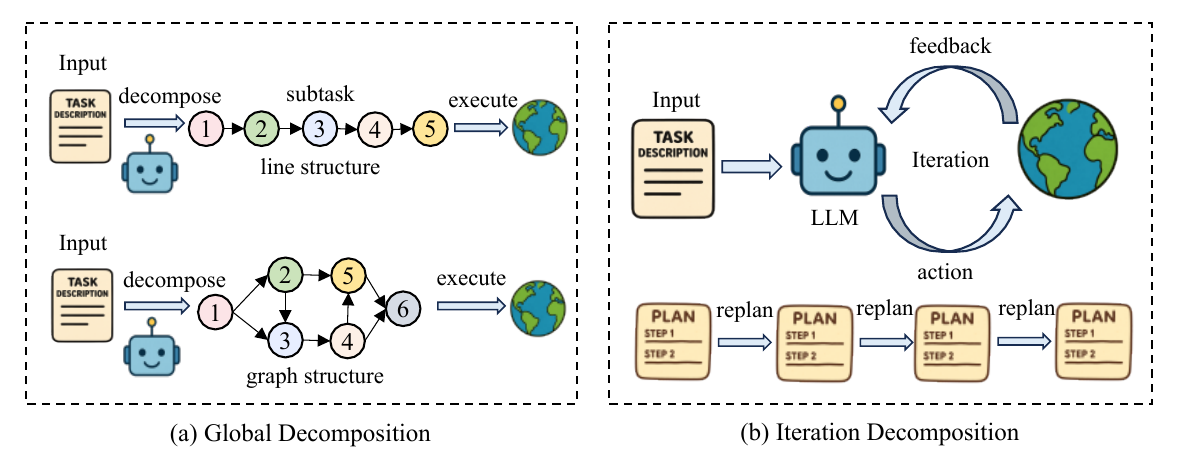}  
    \caption{An overview of decomposition-based methods that can be divided into two decomposition modes, including global decomposition and iteration decomposition.}  
    \label{fig:sbm}  
\end{figure*}

\subsubsection{Decomposition-based Methods} \label{decomposition}
In planning tasks, the decomposition strategy plays a pivotal role. Complex planning problems often require deep reasoning. To address this, many studies have explored the use of decomposition strategies, as illustrated in Figure \ref{fig:sbm}. This approach first decomposes a complex task into several simpler subtasks, and then plans for these subtasks individually \cite{kim2024rada,zhang2024meta,li2024agent,sprague2023deductive,uzunoglu2024paradise}. According to the decomposition mode, existing studies generally fall into \textbf{Global Decomposition} and \textbf{Iteration Decomposition}.  

\textbf{Global Decomposition:}
In the domain of global decomposition-based methods, these approaches are suited for fully observable environments, where complete environmental information enables thorough task analysis and planning. A common strategy is to divide the overall task into smaller subtasks, which can be addressed either concurrently or sequentially without requiring multiple rounds of interaction with the environment.

Methods in this category decompose tasks into a \textit{linear sequence} of subtasks, where each subtask depends on the output of the preceding one, forming a straightforward chain of operations. For example, Qiu et al. \cite{qiu2024optimizing} first prompt the LLM to generate an abstract arrangement plan, then employ plan-enhanced reasoning to guide subsequent steps. Similarly, HRV \cite{cornelio2025hierarchical} breaks down complex tasks into manageable subtasks and further refines them into executable atomic action sequences. Zhou et al. \cite{zhou2022least} also propose decomposing complex problems into simpler sub-problems, where the solution to each subsequent sub-problem depends on those solved earlier. Protrix \cite{wu2403protrix} first designs a reasoning path based on the given context, then allocates steps to either program-based or text-based reasoning to derive the final answer.


For task decomposition, several methods leverage the \textit{graph structure} to decompose and reason about complex tasks. This structure enables a more flexible representation of task relationships, allowing subtasks to be interconnected based on logical and data dependencies \cite{shen2023hugginggpt,jiang2024urbanllm,li2024agent}. HuggingGPT \cite{shen2023hugginggpt} proposes an inter-model cooperation protocol to fully harness the capabilities of LLMs alongside expert models. In this framework, the LLM acts as the ``brain'' for planning and decision-making, while smaller models serve as executors responsible for completing specific tasks. Similarly, UrbanLLM \cite{jiang2024urbanllm} decomposes city-related queries into manageable subtasks, constructs a graph structure, and determines appropriate spatiotemporal models for these subtasks, thus enabling the processing to generate comprehensive responses. Zhang et al. \cite{zheng2024thoughts} decompose the plan generation task based on different selection logics, offering diversified strategies for generating plans across various scenarios. Additionally, given a user query, the meta-agent \cite{li2024agent}, acting as the ``brain'' in a multi-agent system, needs to decompose the query into multiple subtasks and assign them to appropriate agents capable of solving these subtasks. Plan-RAG \cite{verma2024plan} introduces a novel directed acyclic graph-based reasoning framework that decomposes complex queries into atomic and dynamic subqueries, enabling efficient processing while preserving conditional dependencies.



\textbf{Iteration Decomposition:}
In planning tasks, iterative decomposition and step-by-step progression are highly effective strategies. The key lies in breaking down tasks into smaller components and then refining and executing them iteratively based on feedback.

A substantial body of work has adopted multistage decomposition strategies \cite{yuan2023distilling,xie2024human}. For example, Yuan et al. \cite{yuan2023distilling} decompose the constrained language planning task into two steps, in which if the generated specific goals fail to produce an executable script, the goal-generation step is modified based on feedback from the script-generation attempt. Similarly, RaDA \cite{kim2024rada} advances this concept by dividing planning into two key stages: retrieval-augmented task decomposition (RaD) and retrieval-augmented action generation (RaA). In the RaA stage, if the iteratively synthesized actions based on dynamically retrieved examples do not align with the overall task objectives, the RaD stage is revisited to adjust the high-level subtasks. Xie et al. \cite{xie2024human} apply multistage decomposition to travel planning, allowing the entire multistage process to be adjusted if new information emerges that invalidates the initial route generated during the outline-generation stage.

Task decomposition methods that leverage historical information have become increasingly sophisticated, with each new approach building on the strengths of its predecessors and addressing emerging challenges \cite{zhang2023interpretable,wang2024oscar,fu2024msi}. Zhang et al. \cite{zhang2023interpretable} use question-answer history to predict the next mathematical operation; when an incorrect prediction leads to a wrong answer, the model adjusts its prediction algorithm based on the historical data. The novel task planner based on the Task and Circuit Relation Graph (TCRG) \cite{ho2025verilogcoder} takes a more complex approach. TCRG-based planning considers historical execution data in a more structural and task-specific way, enabling more targeted and effective plan modifications in the domain of circuit-related tasks. The CR-Planner \cite{li2024can} represents a further advancement in this line of work: it iteratively selects and executes sub-goals, with a sub-goal critic providing rewards based on performance. If a sub-goal receives an unsatisfactory reward, the sub-goal selection process can be adapted accordingly.  OSCAR \cite{wang2024oscar} adopts a two-level planning approach inspired by plan-solving prompts: it first decomposes user instructions into subtasks, and then incrementally generates actions for each subtask. If the state machine detects an unexpected transition during execution, the system can revisit and revise both the task decomposition and the action-generation processes.

Task decomposition methods based on dynamic planning are also widely used.  TravelPlanner+ \cite{singh2024personal} decomposes the planning task into user model generation and strategy development, with inefficiencies during strategy execution prompting a re-evaluation of the user model. TRIP \cite{zhang2024strength} decomposes the task to enhance strategic planning capabilities; if strategic planning aimed at improving user perception fails to resolve non-cooperation, the entire decomposition framework can be redesigned. ADAPT \cite{prasad2023adapt} determines the decomposition strategy based on the execution capabilities of the LLM, allowing recursive adjustment of subtasks if difficulties arise during execution. Similarly, in budget planning, as demonstrated by the Budget Planner \cite{chen2024travelagent}, user feedback on a preliminary budget estimate (the abstract plan) can trigger a re-evaluation and refinement of the budget allocation process.

\subsubsection{Exploration-based Methods} \label{exploration_method}
\begin{figure*}[t]
\centering
  \includegraphics[width=0.98\textwidth,height=0.5\textheight,keepaspectratio]{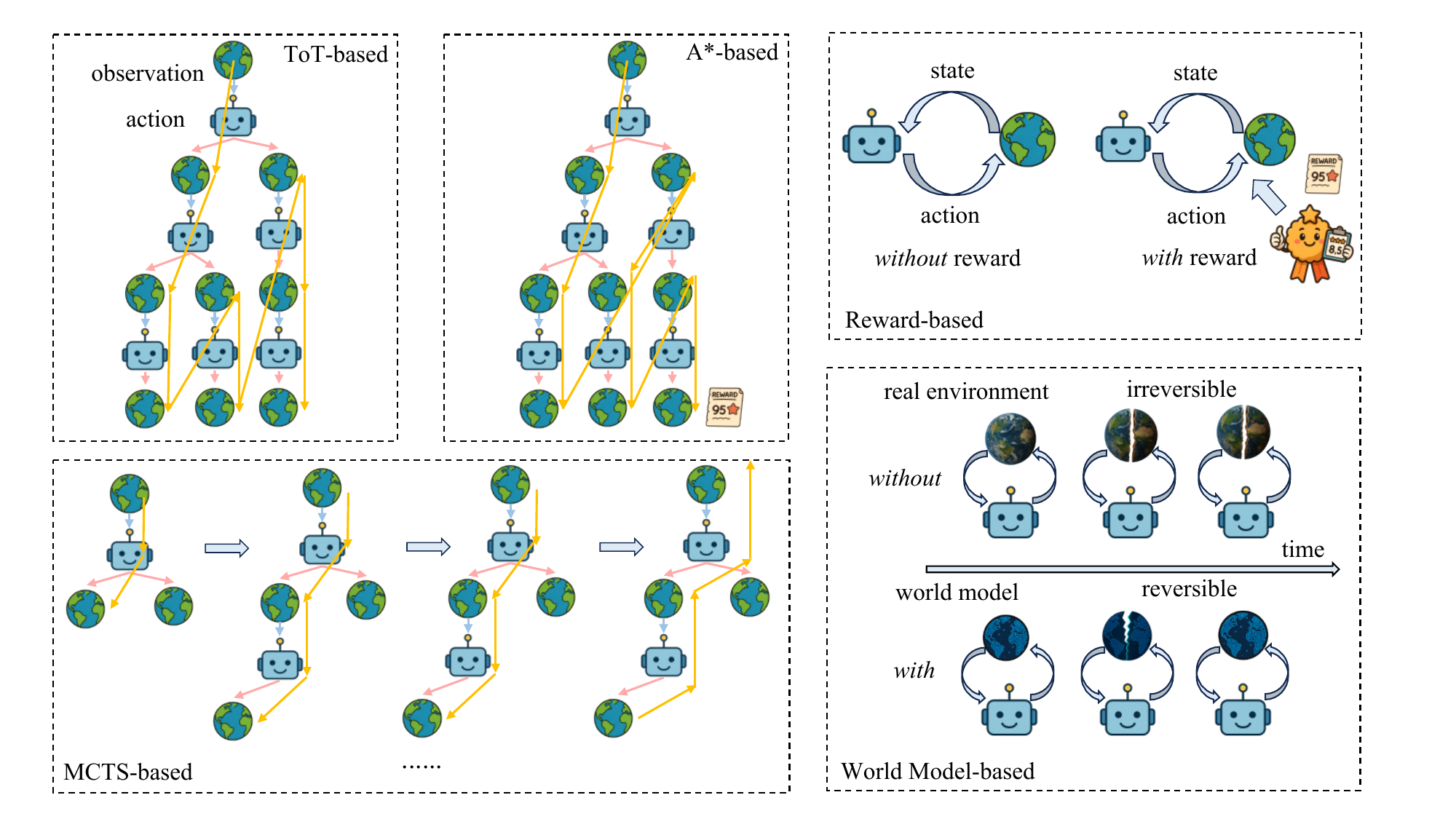}
  \caption{The illustration of exploration methods, including ToT-based, MCTS-based, A*-based, world model-based, and reward model-based exploration.}
  \label{exploration}
\end{figure*}

An effective exploration strategy is essential to the planning process, as it guides LLMs through trial and error, enabling them to backtrack within the solution space to identify optimal outcomes—akin to the workings of ``System 2'' thinking \cite{kahneman2011thinking, sloman1996empirical, stanovich1999rational}, as shown in Figure \ref{exploration}. Within this paradigm, the design of an accurate and efficient exploration strategy emerges as a key bottleneck in planning. Consequently, researchers have proposed a range of strategies to enhance both the performance and efficiency of planning systems \cite{nair2025flow}.

\textbf{ToT-based Exploration:} 
In Tree of Thoughts (ToT)-based exploration methods, the reasoning process is modeled as a tree, where each node represents a discrete step articulated in natural language, and edges denote the transitional relationships between these intermediate steps. To enhance both accuracy and efficiency, these methods employ dynamic prioritization mechanisms that adjust the order of node expansion, guiding the exploration toward a globally optimal solution \cite{yao2023tree,xu2024search,hu2023tree,besta2024graph,yu2023thought}.

Specifically, ToT \cite{yao2023tree} opens up a paradigm for modeling complex planning and reasoning processes as an exploration of a tree structure. This method models each thinking step as a node in a thought tree. Then, LLMs expand multiple child nodes from the current node, score and rank these child nodes, and finally use depth-first search (DFS) or breadth-first search (BFS) search strategies to navigate through the tree. SearChain \cite{xu2024search} and AVIS \cite{hu2023avis} also employ tree search strategies, these methods integrate feedback from retrieval results or external tools to expand nodes during the search process. Building on the tree search paradigm, subsequent research has focused on enhancing search efficiency to address its inherently high time complexity \cite{qin2023toolllm,hu2023tree,sel2023algorithm}. 
The Tree-of-Mixed-Thought \cite{hu2023tree} builds upon the ToT framework by integrating both fast and slow thinking processes. It initially attempts to generate a plan directly; if the output does not conform to the required format, it switches to the ToT mode to reduce complexity overall. Similarly, System-1.x \cite{saha2024system} is also grounded in the dual-process theory of fast and slow thinking. To further optimize the switching between these two modes, a controller mechanism is introduced to decompose the overall task: simpler tasks are handled by System 1, while more complex ones are delegated to System 2.

Beyond enhancing search efficiency, some studies seek to further integrate thinking processes \cite{besta2024graph,pandey2025adaptive,yang2025step,li2024codetree}. GoT \cite{besta2024graph} extends tree search to graph structures by introducing vertices with multiple incoming edges to aggregate the content of the incoming edge nodes. TP \cite{yu2023thought} introduces an analogy method in the planning process, proposing a ``propose-solve-aggregate'' strategy, which solves the current problem by aggregating solutions to similar problems. Tree-Planner \cite{hu2023tree} extends tree search to closed-loop robotic systems interacting with the environment, where actions are explicitly defined and belong to a finite set of types. It begins by repeatedly sampling complete trajectories. These trajectories with common prefixes are then aggregated into a trajectory action tree. Finally, the system executes the planned actions step by step within the environment. AGoT \cite{pandey2025adaptive} introduces a graph-oriented dynamic search framework that recursively decomposes problems into smaller subproblems, constructing a dynamic directed acyclic graph. By unifying chain, tree, and graph structures, it enables efficient allocation of computational resources.

\textbf{MCTS-based Exploration:}
Monte Carlo Tree Search (MCTS) \cite{coulom2006efficient,kocsis2006bandit} is an advanced tree search algorithm that balances exploration and exploitation with the goal of finding the optimal trajectory. It comprises four key steps: 1) Selection - Traverse the tree to select the most promising node for expansion, 2) Expansion - Generate one or more child nodes from the selected node, 3) Simulation - Simulate the node until the end to estimate future rewards, and 4) Back-propagation - Propagate the rewards back through the nodes. Some studies have been proposed to leverage MCTS to enhance planning capabilities of LLMs \cite{zhao2023large,zhou2023language,chen2024alphamath,liu2025doubly,fang2025synworld}.

In detail, LLM-MCTS \cite{zhao2023large} is a search strategy that uses LLMs as both the commonsense model and the heuristic policy in MCTS. LLMs serve as the commonsense model to supply prior knowledge and as the heuristic policy to guide action selection. THOUGHTSCULPT \cite{chi2024thoughtsculpt} is also an MCTS framework with reflective capabilities, which performs reflection at the step level. Unlike the previous node expansion actions, it continues by extending from the later outputs to include reflecting on and revising previous outputs. MC-DML \cite{shimonte} is a Dynamic Memory-guided MCTS search strategy, which adjusts the value scores of current action choices based on past reflective experiences through the dynamic memory components of in-trial and cross-trial. CoAT \cite{pan2025coat} combines MCTS and associative memory, which introduces an association stage between the expansion and simulation phases. This step simulates human association by either self-association or retrieving from an external knowledge base to expand the search space. WebPilot \cite{zhang2024webpilot} is a system designed for web navigation that uses a dual optimization strategy, which can narrow the search space of classical MCTS. In the global optimization phase, it decomposes complex tasks into sub-tasks, and in the local optimization phase, it employs an adapted MCTS to solve the sub-tasks.        

\textbf{A*-based Exploration:}
A* \cite{hart1968formal} is also an advanced search algorithm that estimates the value of the current node using a cumulative cost function and a future cost function. Specifically, the cumulative cost function is used to estimate the actual cost from the starting point to the current node, while the future cost function is used to estimate the expected cost from the current node to the goal node. The cumulative cost and future cost are calculated for all nodes, and the two costs are summed. Nodes that have not been explored are then expanded in order of their cost from lowest to highest.

ToolChain* \cite{zhuangtoolchain} is a tool using search method based on the A* search algorithm, which combines the reasoning capabilities of LLMs with A* search. It uses a manually defined cumulative cost function and a future cost function to guide the direction of the search, enabling efficient searching. Koh et al. \cite{koh2024tree} propose a search algorithm for multimodal web navigation based on A*, which uses a multimodal model to estimate the value of each node and performs the search based on that value. LLM-A* \cite{meng2024llm} is a path planning search framework that combines LLM global planning and A* local search. It uses LLM for high-level planning and A* for precise low-level path searching, aiming to improve search efficiency. Lehnert et al. \cite{lehnert2024beyond} propose using A* search to construct trajectory data, applied to path planning tasks.  Su et al. \cite{su2024dualformer} further apply A* search to combine fast and slow thinking.

\textbf{World Model-based Exploration:}
Agent planning relies on interactions with the environment. However, engaging with the real world poses challenges such as low efficiency, safety risks, and the potential for irreversible changes. To address these issues, a promising approach is to construct \textit{world models} \cite{ha2018world, matsuo2022deep} that predict state transitions within the environment. These models can be leveraged to simulate or generate training data during the planning process.

RAP \cite{hao2023reasoning} is a search framework that regards the LLM as a world model. It does not rely on real-world environment feedback but instead uses the LLM to predict changes in the environment state and forecast the outcomes of action executions. WEBDREAMER \cite{gu2024your} is a web navigation search framework that uses the LLM as the world model to predict the changes in web page states after the agent actions are executed. It relies on the world model rather than real web interactions, helping to prevent security risks and irreversible environmental changes that may occur during actual web operations. Feng et al. \cite{feng2025reflective} proposes a search framework for robotic manipulation based on multi-modal world models, which uses a diffusion model as the world model to predict the state changes of objects after robotic actions. WorldCoder \cite{tang2025worldcoder} uses code to build a world model, where the world knowledge is represented as code to serve as a transition function for planning based on the world model. ReasonPlanner \cite{dinh2024reasonplanner} regards a world model as a temporal knowledge graph, which records factual information according to timestamps. When an agent performs planning, it can query the temporal knowledge graph to assist the model in making forward-looking decisions.

\textbf{Reward Model-based Exploration:}
The reward model \cite{uesato2022solving,lightman2023let} plays a crucial role in reward-based exploration methods. This type of approach fixes the parameters of the policy model and constructs training data to improve the reward model’s scoring accuracy, thereby guiding the policy model toward more promising search directions. For instance, QLASS \cite{lin2025qlass} trains a Q-value network that estimates the Q-values of intermediate nodes based on environment outcome rewards. Similarly, ARMAP \cite{chen2025scaling} enhances the reward model using automatically generated positive and negative trajectory pairs, aiming to boost the effectiveness of reward-based policy search strategies.

\subsubsection{Decoding-based Methods} \label{decoding}
Decoding strategies play a crucial role in determining the output of LLMs. While traditional decoding methods, such as greedy decoding and beam search, often yield high-quality text outputs \cite{jurafskyspeech, graves2012sequence}, they fail to fully harness the reasoning and planning capabilities of LLMs. As a result, more advanced decoding techniques have been proposed to enhance the planning abilities of LLMs \cite{xie2024self, wang2024chain, roy2024flap, ma2024non}.

Specifically, Xie et al. \cite{xie2024self} propose a decoding algorithm that integrates self-evaluation guidance through stochastic beam search. This approach leverages LLMs to provide a calibrated criterion for reasoning. Additionally, it combines beam search with temperature-controlled randomness to generate more robust reasoning chains. In a similar vein, Wang et al. \cite{wang2024chain} explore whether LLMs inherently possess reasoning capabilities without relying on manually crafted prompts. Their findings suggest that Chain-of-Thought (CoT) \cite{wei2022chain} reasoning paths can be elicited from LLMs by simply using the top-k alternative tokens, rather than relying on traditional greedy decoding methods. To achieve this, they propose the CoT-decoding strategy, which uses confidence as guidance to extract CoT paths.
 
At the same time, several researchers have also explored lookahead-based decoding algorithms to improve planning accuracy. For instance, Liu et al. \cite{liu2024don} introduce a value-guided decoding algorithm that leverages Monte-Carlo Tree Search (MCTS) to build a search tree, then decodes target tokens based on sequence evaluation with a few lookahead steps. Roy et al. \cite{roy2024flap} propose a constrained decoding method with a lookahead heuristic, where the prediction probability of a token is determined by both the original prediction probability and the score of the lookahead function. Following this paradigm, Ma et al. \cite{ma2024non} introduce a predictive-decoding approach to enhance LLM planning through non-myopic generation. This method employs LLMs to sample multiple foresight trajectories, which are then used to re-weight the original generation distribution. Additionally, other decoding strategies have been developed to enhance the planning capabilities of LLMs, including contrastive decoding \cite{o2023contrastive} and grounded decoding \cite{huang2024grounded}. 

\begin{tcolorbox}[colback=blue!5!white,colframe=black!75!white,
  title=\textbf{Summary of Searching-based Methods}, 
  fonttitle=\bfseries, coltitle=white, colbacktitle=black, rounded corners]
  \begin{itemize}[leftmargin=1.5em]
    \item \textit{Decomposition-based Methods} directly or iteratively decompose the whole complex task into smaller subtasks, and then sequentially solve the subtasks. Obviously, such methods are limited by the task decomposition ability of LLMs.

    \item \textit{Exploration-based Methods} utilize some search algorithms to guide LLMs through trial and error, enabling them to backtrack within the solution space to identify optimal outcomes. However, complex search algorithms can increase computational costs, which poses practical challenges.

    \item \textit{Decoding-based Methods} harness the reasoning and planning capabilities of LLMs by improving the decoding techniques of language models. However, such methods heavily rely on the capabilities of the LLM itself and are difficult to handle very complex planning tasks.
  \end{itemize}
\end{tcolorbox}

\begin{table*}[t]
\caption{Overview of LLM Planning Evaluation Datasets. Env = Environment (Is it an interactive dynamic environment). MM = Multimodal (Is the interactive information multimodal). Obs = Observability (The environment is either fully observable or partially observable). Loop (The task requires multiple interactions with the environment to be completed, ``closed'' represents closed-loop, ``open'' represents open-loop, and ``both'' means both are present). Output (``action'' means the output is a single action, ``plan'' means the output is a sequence of actions, ``code'' means the output is code, and ``multiple'' means there are multiple types of outputs). Some dataset span multiple categories, we classify them according to one of category.}
\centering
\scalebox{0.65}{
\begin{tabular}{|l|l|l|c|c|c|c|c|l|}
\toprule
\textbf{Category} & \textbf{Subcategory} & \textbf{Dataset} & \textbf{Env} & \textbf{MM} & \textbf{Obs} & \textbf{Loop} & \textbf{Output} & \textbf{Metric} \\
\midrule
\multirow{19}{*}{Digital Scenarios} & \multirow{7}{*}{Web Navigation} & WebShop\cite{yao2022webshop} & \textcolor{green}{\checkmark} & \textcolor{green}{\checkmark} & POMDP & closed & action & Success Rate, Task Score \\
 &  & MiniWoB++\cite{liu2018reinforcement} & \textcolor{green}{\checkmark} & \textcolor{green}{\checkmark} & POMDP & closed & action & Success Rate \\
 &  & Mind2Web\cite{deng2023mind2web} & \textcolor{red}{\textbf{×}} & \textcolor{green}{\checkmark} & POMDP & closed & action & Element Accuracy, Operation F1, Success Rate \\
 &  & WebArena\cite{zhou2023webarena} & \textcolor{green}{\checkmark} & \textcolor{green}{\checkmark} & POMDP & closed & action & Success Rate \\
 &  & VisualWebArena\cite{koh2024visualwebarena} & \textcolor{green}{\checkmark} & \textcolor{green}{\checkmark} & POMDP & closed & action & Success Rate \\
 &  & WebVoyager\cite{he2024webvoyager} & \textcolor{green}{\checkmark} & \textcolor{green}{\checkmark}  & POMDP & closed & action & Task Success Rate \\
 &  & WEBLINX\cite{lu2024weblinx} & \textcolor{green}{\checkmark} & \textcolor{green}{\checkmark} & POMDP & closed & action & Intent Match, Element Similarity, etc \\
 \cmidrule{2-9}
 & \multirow{7}{*}{Mobile Navigation} & AndroidWorld\cite{rawles2024androidworld} & \textcolor{green}{\checkmark} & \textcolor{green}{\checkmark} & POMDP & closed & action & Success Rate \\
 &  & PIXELHELP\cite{li2020mapping} & \textcolor{red}{\textbf{×}} & \textcolor{green}{\checkmark} & POMDP & closed & action & Complete Matching, Partial Matching \\
 &  & META-GUI\cite{sun2022meta} & \textcolor{red}{\textbf{×}} & \textcolor{green}{\checkmark} & POMDP & closed & action & Action Completion Rate, Turn Completion Rate \\
 &  & AITW\cite{rawles2023androidinthewild} & \textcolor{red}{\textbf{×}} & \textcolor{green}{\checkmark} & POMDP & closed & action & Action Matching \\
 &  & ANDROIDCONTROL\cite{li2024effects} & \textcolor{red}{\textbf{×}} & \textcolor{green}{\checkmark} & POMDP & closed & action & Step-wise Accuracy \\
 &  & Mobile-Env\cite{zhang2023mobile} & \textcolor{green}{\checkmark} & \textcolor{green}{\checkmark} & POMDP & closed & action & Reward, Success Rate \\
 &  & A3\cite{chai2025a3} & \textcolor{green}{\checkmark} & \textcolor{green}{\checkmark} & POMDP & closed & action & Task Success Rate \\
 \cmidrule{2-9}
 & \multirow{5}{*}{Desktop Navigation} & OSWORLD\cite{xie2024osworld} & \textcolor{green}{\checkmark} & \textcolor{green}{\checkmark} & POMDP & closed & action & Success Rate \\
 &  & AgentStudio\cite{zheng2024agentstudio} & \textcolor{green}{\checkmark} & \textcolor{green}{\checkmark} & POMDP & closed & action & Success Rate, Grounding Accuracy \\
 &  & TheAgentCompany\cite{xu2024theagentcompany} & \textcolor{green}{\checkmark} & \textcolor{green}{\checkmark} & POMDP & closed & action & Full and Partial Completion Score, etc \\
 &  & WindowsAgentArena\cite{bonatti2024windows} & \textcolor{green}{\checkmark} & \textcolor{green}{\checkmark} & POMDP & closed & action & Success Rate \\
 &  & WorkArena\cite{drouin2024workarena} & \textcolor{green}{\checkmark} & \textcolor{green}{\checkmark} & POMDP & closed & action & Success Rate \\
\midrule
\multirow{26}{*}{Embodied Scenarios} &  \multirow{12}{*}{Household Robot} & ALFWorld\cite{shridhar2020alfworld} & \textcolor{green}{\checkmark} & \textcolor{red}{\textbf{×}} &  POMDP & closed & action & Success Rate \\
 &  & ALFRED\cite{shridhar2020alfred} & \textcolor{green}{\checkmark} & \textcolor{green}{\checkmark} & POMDP & closed & action & Task Success, Goal-Condition Success, etc  \\
 &  & VirtualHome\cite{puig2018virtualhome} & \textcolor{green}{\checkmark} & \textcolor{green}{\checkmark} & POMDP & closed & action & Accuracy of Action and Object Prediction \\
 &  & ScienceWorld\cite{wang2022scienceworld} & \textcolor{green}{\checkmark} & \textcolor{red}{\textbf{×}} & POMDP & closed & action & Reward \\
 &  & Watch-And-Help\cite{puig2020watch}& \textcolor{green}{\checkmark} & \textcolor{green}{\checkmark} & POMDP & closed & action & Success Rate, Speedup, Reward  \\
 &  & LangSuit·E\cite{jia2024langsuit}  & \textcolor{green}{\checkmark} & \textcolor{red}{\textbf{×}} & POMDP & closed & action & Success Rate, Fixed Strict Rate, Accuracy, etc  \\
 &  & ActPlan-1K\cite{su2024actplan}  & \textcolor{red}{\textbf{×}} & \textcolor{green}{\checkmark} & MDP & open & plan &  Longest Common Subsequence, etc \\
 &  & PARTNR\cite{chang2024partnr}  & \textcolor{green}{\checkmark} & \textcolor{green}{\checkmark} & POMDP & closed & action & Simulation Steps, Success Rate, etc \\
 &  & Embodied Agent Interface\cite{li2024embodied}  & \textcolor{green}{\checkmark} & \textcolor{red}{\textbf{×}} & POMDP & closed & multiple & Planning Success Rate, etc \\
 &  & LoTa-Bench\cite{choi2024lota} & \textcolor{green}{\checkmark} & \textcolor{red}{\textbf{×}} & POMDP & closed & action & Task Success Rate \\
 &  & GOAT-Bench\cite{khanna2024goat} & \textcolor{green}{\checkmark} & \textcolor{green}{\checkmark} & POMDP & closed & action & Success Rate, Success Weighted by Path Length \\
 &  & BEHAVIOR-1K\cite{li2024behavior} & \textcolor{green}{\checkmark} & \textcolor{green}{\checkmark} &  POMDP & closed & action & Task Success Rate, Efficiency \\
 \cmidrule{2-9}
 & \multirow{5}{*}{Manipulation Robot} & VLMbench\cite{zheng2022vlmbench} & \textcolor{green}{\checkmark} & \textcolor{green}{\checkmark} & MDP & closed & action & Success Rate \\
 &  & VLABench\cite{zhang2024vlabench} & \textcolor{green}{\checkmark} & \textcolor{green}{\checkmark} & MDP & closed & action & Progress Score, Skill Recall Rate,  etc \\
 &  & VIMA-Bench\cite{jiang2022vima}  & \textcolor{green}{\checkmark} & \textcolor{green}{\checkmark} & MDP & closed & action & Success Rate \\
 &  & EmbodiedBench\cite{yang2025embodiedbench} & \textcolor{green}{\checkmark} & \textcolor{green}{\checkmark} & BOTH & closed & action & Success Rate, Average Planner Steps, etc \\
 &  & RoCoBench\cite{mandi2024roco}  & \textcolor{green}{\checkmark} & \textcolor{green}{\checkmark} & MDP & closed & action & Success Rate, Average Environment Steps, etc \\
 \cmidrule{2-9}
 & \multirow{7}{*}{Minecraft Robot} & MineDojo\cite{fan2022minedojo} & \textcolor{green}{\checkmark} & \textcolor{green}{\checkmark} & POMDP & closed & action & Success Rate \\
 &  & Craftax\cite{matthews2024craftax} & \textcolor{green}{\checkmark} & \textcolor{green}{\checkmark} & POMDP & closed & action & Success Rate, Reward \\
 &  & MinePlanner\cite{hill2023mineplanner}  & \textcolor{green}{\checkmark} & \textcolor{red}{\textbf{×}} & POMDP & closed & action & Planning Time \\
 &  & TeamCraft\cite{long2024teamcraft} & \textcolor{green}{\checkmark} & \textcolor{green}{\checkmark} & POMDP & closed & action & Subgoal Success Rate, Task Success Rate, etc \\
 &  & Plancraft\cite{dagan2024plancraft} & \textcolor{green}{\checkmark} & \textcolor{green}{\checkmark} & MDP & closed & action & Task Success Rate, Action Efficiency \\
 &  & MINDAGENT\cite{gong2023mindagent} & \textcolor{green}{\checkmark} & \textcolor{green}{\checkmark} & POMDP & closed & action &  Collaboration Score  \\
 &  & Overcooked-AI\cite{carroll2019utility} & \textcolor{green}{\checkmark} & \textcolor{green}{\checkmark} & POMDP & closed & action & Reward \\
 \cmidrule{2-9}
 & \multirow{2}{*}{Autonomous Driving} & PCA-Bench\cite{chen2024pca}  & \textcolor{red}{\textbf{×}} & \textcolor{green}{\checkmark} & POMDP & closed & action & Perception Score, Cognition Score, Action Score \\
 &  & MetaAD\cite{jiang2025alphadrive}  & \textcolor{red}{\textbf{×}} & \textcolor{green}{\checkmark} & POMDP & closed & action & Accuracy of Meta-Action Planning \\
\bottomrule
\end{tabular}
}
\label{planning-datasets1}
\end{table*}

\begin{table*}[t]
\caption{Overview of LLM Planning Evaluation Datasets (Continuation of Table \ref{planning-datasets1}).}
\centering
\scalebox{0.68}{
\begin{tabular}{|l|l|l|c|c|c|c|c|l|}
\toprule
\textbf{Category} & \textbf{Subcategory} & \textbf{Dataset} & \textbf{Env} & \textbf{MM} & \textbf{Obs} & \textbf{Loop} & \textbf{Output} & \textbf{Metric} \\
\midrule
 \multirow{24}{*}{Everyday Scenarios}& \multirow{3}{*}{Travel Planning} & TravelPlanner\cite{xie2024travelplanner}  & \textcolor{green}{\checkmark} & \textcolor{red}{\textbf{×}} & BOTH & both & multiple & Delivery Rate, Constraint Pass Rate, etc \\
 &  & ChinaTravel\cite{shao2024chinatravel}  & \textcolor{green}{\checkmark} & \textcolor{red}{\textbf{×}} & BOTH & both & multiple & Delivery Rate, Environmental Pass Rate, etc \\
 &  & NATURAL PLAN\cite{zheng2024natural} & \textcolor{red}{\textbf{×}} & \textcolor{red}{\textbf{×}} & MDP & open & plan & Exact Match Score \\
 \cmidrule{2-9}
 & \multirow{8}{*}{Workflow} & FlowBench\cite{xiao2024flowbench} & \textcolor{green}{\checkmark} & \textcolor{red}{\textbf{×}} & POMDP & closed & plan & Tool Invocation, Success Rate, etc \\
 &  & Open Grounded Planning\cite{guo2024open}  & \textcolor{red}{\textbf{×}} & \textcolor{red}{\textbf{×}} & MDP & open & plan & Executability, Quality, Overall Pass Rate \\
 &  & HuggingGPT\cite{shen2023hugginggpt} & \textcolor{red}{\textbf{×}} & \textcolor{red}{\textbf{×}} & MDP & open & plan & Passing Rate, Rationality \\
 &  & TaskBench\cite{shen2024taskbench}  & \textcolor{red}{\textbf{×}} & \textcolor{red}{\textbf{×}} & MDP & open & plan & Tool Selection Accuracy, etc \\
 &  & TaskLAMA\cite{yuan2024tasklama} & \textcolor{red}{\textbf{×}} & \textcolor{red}{\textbf{×}} & MDP & open & plan & Task Step Generation Metrics \\
 &  & WORFBENCH\cite{qiao2024benchmarking} & \textcolor{red}{\textbf{×}} & \textcolor{red}{\textbf{×}} & MDP & open &  plan & Precision and Recall of Workflow Graph \\
 &  & WIKIPLAN\cite{lu2023multimodal} & \textcolor{red}{\textbf{×}} & \textcolor{green}{\checkmark} & MDP & open & plan & Textual Informativenss, Accuracy, etc \\
 &  & ISG-BENCH\cite{chen2024interleaved} & \textcolor{red}{\textbf{×}} & \textcolor{green}{\checkmark} & MDP & open & plan & BertScore, Human Agreement \\
 \cmidrule{2-9}
 & \multirow{6}{*}{Tool Calling} & ToolBench\cite{qin2023toolllm} & \textcolor{green}{\checkmark} & \textcolor{red}{\textbf{×}} & POMDP & closed & action & Pass Rate, Win Rate \\
 &  & AppWorld\cite{trivedi2024appworld}  & \textcolor{green}{\checkmark} & \textcolor{red}{\textbf{×}} & POMDP & closed & code & Task Goal Completion, etc \\
 &  & API-Bank\cite{li2023api}  & \textcolor{red}{\textbf{×}} & \textcolor{red}{\textbf{×}} & MDP & open & action & Accuracy, ROUGE-L \\
 &  & ToolComp\cite{nath2025toolcomp}  & \textcolor{green}{\checkmark} & \textcolor{red}{\textbf{×}} & POMDP & closed & action & LLM Grading, Exact Match \\
 &  & ToolSandbox\cite{lu2024toolsandbox}   & \textcolor{green}{\checkmark} & \textcolor{red}{\textbf{×}} & POMDP & closed & action & Tool Call Match Score  \\
 &  & ToolTalk\cite{farn2023tooltalk}& \textcolor{green}{\checkmark} & \textcolor{red}{\textbf{×}} & POMDP & closed & action & Tool Invocation Recall, Incorrect Action Rate \\
 \cmidrule{2-9}
 & \multirow{3}{*}{Code Generation} & SWE-Bench\cite{jimenez2023swe} & \textcolor{green}{\checkmark} & \textcolor{red}{\textbf{×}} & MDP & open & code & Pass Rate \\
 &  & SWE-Gym\cite{pan2024training}  & \textcolor{green}{\checkmark} & \textcolor{red}{\textbf{×}} & POMDP & closed & code & Resolve Rate, Empty Patch \\
 &  & HumanEval\cite{chen2021evaluating} & \textcolor{green}{\checkmark} & \textcolor{red}{\textbf{×}} & MDP & open & code & Functional Correctness \\
  \cmidrule{2-9}
 & \multirow{4}{*}{Game Playing} & VSP\cite{wu2024vsp}  & \textcolor{green}{\checkmark} & \textcolor{green}{\checkmark} & MDP & open & plan & Success Rate \\
 &  & TextWorld\cite{cote2019textworld}  & \textcolor{green}{\checkmark} & \textcolor{red}{\textbf{×}} & POMDP & closed & action & Reward \\
 &  & BabyAI\cite{chevalier2018babyai} & \textcolor{green}{\checkmark} & \textcolor{green}{\checkmark} & POMDP & closed & action & Success Rate, Demo Length \\
 &  & PlanBench\cite{valmeekam2023planbench}  & \textcolor{green}{\checkmark} & \textcolor{red}{\textbf{×}} & MDP & open & plan & Pass Rate \\
\midrule
\multirow{8}{*}{Vertical Scenarios} & Machine Learning & MLE-bench\cite{chan2024mle} & \textcolor{green}{\checkmark} & \textcolor{red}{\textbf{×}} & POMDP & closed & multiple & Pass Rate \\
 \cmidrule{2-9}
 & \multirow{2}{*}{AI Research} & ResearchArena\cite{kang2024researcharena}  & \textcolor{green}{\checkmark} & \textcolor{red}{\textbf{×}} & POMDP & closed & multiple & Heading Soft Recall, Heading Entity Recall  \\
 &  & CYCLERESEARCHER\cite{weng2024cycleresearcher} & \textcolor{green}{\checkmark} & \textcolor{red}{\textbf{×}} & POMDP & closed & multiple & Proxy Mean Squared Error, etc \\
  \cmidrule{2-9}
 & Biological Research & BIOPROT\cite{o2023bioplanner}  & \textcolor{red}{\textbf{×}} & \textcolor{red}{\textbf{×}} & MDP & open & multiple & Functions Accuracy, Arguments Precision, etc  \\
  \cmidrule{2-9}
 & Financial Simulation & AUCARENA \cite{chen2023put}  & \textcolor{green}{\checkmark} & \textcolor{red}{\textbf{×}} & POMDP & closed & action & TrueSkill Score \\
  \cmidrule{2-9}
 & Interior Design & DStruct2Design\cite{luo2024dstruct2design} & \textcolor{green}{\checkmark} & \textcolor{red}{\textbf{×}} & MDP & open & plan & Prompt Consistency, Self Consistency, etc \\
  \cmidrule{2-9}
 & \multirow{2}{*}{Composite} & AgentBench\cite{liu2023agentbench}  & \textcolor{green}{\checkmark} & \textcolor{red}{\textbf{×}} & POMDP & closed & action & Success Rate, Answer F1 \\
 &  & VisualAgentBench\cite{liu2024visualagentbench}   & \textcolor{green}{\checkmark} & \textcolor{green}{\checkmark} & POMDP & closed & action & Success Rate \\
\bottomrule
\end{tabular}
}
\label{planning-datasets2}
\end{table*}

\section{Planning Evaluation} \label{Eva}
The development of evaluation is crucial for advancing LLM-based planning methods. This section systematically reviews planning evaluation, including \textbf{Datasets} (\S\ref{dataset}), \textbf{Evaluation Metrics} (\S\ref{metrics}), and \textbf{Performance Comparisons} (\S\ref{performance}).


\subsection{Datasets} \label{dataset}
Currently, the research on LLMs in the planning and evaluation of interaction ability is flourishing. However, a unified system of standards and evaluation methods has yet to be established. To gain a comprehensive understanding of the current research landscape, researchers have developed a variety of datasets to assess the planning capabilities of LLMs from multiple perspectives. These datasets as shown in Table \ref{planning-datasets1} and Table \ref{planning-datasets2}\footnote{Due to space limitations, we present the dataset in two tables.} primarily fall into four categories: evaluating LLMs' planning capabilities in \textbf{Digital Scenarios} (\S\ref{digital}), \textbf{Embodiment Scenarios} (\S\ref{embodiment}), \textbf{Everyday Scenarios} (\S\ref{everyday}) and \textbf{Vertical Scenarios} (\S\ref{domains}).

\subsubsection{Digital Scenarios} \label{digital}
The goal of digital scenario tasks is to enable agents to autonomously complete complex computer operations driven by high-level user instructions. Agents perceive the environment through screen captures or HTML, and translate the instructions into interactive operations such as mouse clicks and keyboard inputs. These tasks span a variety of scenarios, including web browsing, mobile applications, file management, and enterprise software usage. Based on the type of interaction within digital environments, these datasets can be further categorized into \textbf{Web}, \textbf{Mobile} and \textbf{Desktop} navigation.

\textbf{Web Navigation:} This area focuses on assessing LLM’s capabilities to navigate and plan operations in web environments. Tasks in this area typically involve multimodal interactions, operate under partially observable states, and require multiple rounds of dynamic operations to achieve the final goal. These challenges demand that the model not only decompose complex tasks effectively, but also accurately locate and understand web elements, ground its click and interaction behaviors precisely, and backtrack or correct errors when necessary. Compared to other tasks, web navigation focuses more on keeping information retrieval and action sequences coherent within complex web page structures. In addition, models must handle more open and heterogeneous web designs, adapt to diverse layouts and interaction logics, and manage high levels of noise, information density, and dynamic content changes. These factors raise the bar for a model’s environmental perception, planning ability, and adaptability. Datasets that support research and evaluation in this domain include WebShop \cite{yao2022webshop}, MiniWoB++ \cite{liu2018reinforcement}, Mind2Web \cite{deng2023mind2web}, WebArena \cite{zhou2023webarena}, VisualWebArena \cite{koh2024visualwebarena}, WebVoyager \cite{he2024webvoyager}, and WEBLINX \cite{lu2024weblinx}.

\textbf{Mobile Navigation:} This area centers on evaluating the capabilities of LLMs to navigate and plan operations within mobile environments.  These tasks require the model to reason over screenshots or structured UI representations to accurately perceive interface elements. Given that the environment is typically static and only partially observable, achieving a goal often necessitates multiple interaction steps. In terms of evaluation priorities, mobile navigation places particular emphasis on fine-grained screen understanding, gesture-level action generation, and robust visual localization of touch targets across diverse interface layouts. High-precision gesture planning based on subtle visual details, along with maintaining contextual consistency within dynamic interfaces, represents a set of critical challenges. Datasets such as AndroidWorld \cite{rawles2024androidworld}, PIXELHELP \cite{li2020mapping}, META-GUI \cite{sun2022meta}, AITW \cite{rawles2023androidinthewild}, ANDROIDCONTROL \cite{li2024effects}, Mobile-Env \cite{zhang2023mobile}, and A3 \cite{chai2025a3} are commonly used to support research in this domain.

\textbf{Desktop Navigation:} This area focuses on evaluating the capabilities of LLMs to navigate and plan operations within operating system (OS) environments.  Relevant datasets often require models to integrate and interpret multimodal inputs, such as visual observations and textual information, to make informed decisions. To successfully accomplish these tasks, it requires the model not only have a deep understanding of the desktop ecosystem, but also the ability to process multi-source heterogeneous information, and the capacity to address complex factors such as permission management, error handling, and unexpected behaviors at the operating system level. Examples of datasets supporting research in this domain include OSWORLD \cite{xie2024osworld}, AgentStudio \cite{zheng2024agentstudio}, TheAgentCompany \cite{xu2024theagentcompany}, WindowsAgentArena \cite{bonatti2024windows}, and WorkArena \cite{drouin2024workarena}.

\subsubsection{Embodied Scenarios} \label{embodiment}
Embodied scenarios aims to enable agents, in the form of robots, to perform a series of tasks in the physical world, such as tidying up a room, cooking, resource collection, and autonomous driving, by mapping high-level instructions into low-level action sequences. These tasks are often irreversible and constrained by physical environments. Current research focuses on building realistic simulation environments and real-world tasks to evaluate agents’ capabilities in world knowledge, spatial perception, understanding of physical laws, task decomposition, and long-term planning.

\textbf{Household Robot:} A household robot is a mobile service robot capable of navigating within a home environment to perform tasks like cleaning and finding objects. Evaluation platforms for household robots characterized by dynamic home settings with partially observable conditions, requiring processing of information from multiple sensors and execution of complex action sequences in highly variable scenes that demand flexible responses to changing environmental states. This domain evaluates models across several critical capabilities, which includes spatial reasoning for navigation through home layouts with furniture and obstacles, as well as object recognition and localization in cluttered domestic environments. Additional capabilities involve temporal planning for multi-step household chores, adaptive decision-making when confronting unexpected changes, and contextual understanding of household-specific instructions. To support research and evaluation in this domain, a range of benchmark datasets has been developed, including ALFWorld \cite{shridhar2020alfworld}, ALFRED \cite{shridhar2020alfred}, VirtualHome \cite{puig2018virtualhome}, ScienceWorld \cite{wang2022scienceworld}, Watch\&Help \cite{puig2020watch}, LangSuit·E \cite{jia2024langsuit}, ActPlan-1K \cite{su2024actplan}, PARTNR \cite{chang2024partnr}, Embodied Agent Interface \cite{li2024embodied}, LoTa-Bench \cite{choi2024lota}, GOAT-Bench \cite{khanna2024goat}, and BEHAVIOR-1K \cite{li2024behavior}.

\textbf{Manipulation Robot:} A manipulation robot typically features a robotic arm and is specialized in grasping, moving, and interacting with physical objects. Benchmarks in this domain provide extensive expert-annotated trajectory data for imitation learning and model evaluation. The notable feature of this type of tasks is their continuous and interactive nature, requiring models to execute a sequence of coherent actions within a dynamically evolving environment across multiple interaction cycles to accomplish specified goals.  These benchmarks primarily assess model's capabilities in fine-grained  motion control, three-dimensional spatial reasoning, physical constraint inference, and precise operational planning based on visual input. Notable datasets supporting research in this area include VLMbench \cite{zheng2022vlmbench}, VLABench \cite{zhang2024vlabench}, VIMA-Bench \cite{jiang2022vima}, EmbodiedBench \cite{yang2025embodiedbench}, and RoCoBench \cite{mandi2024roco}.

\textbf{Minecraft Robot:} A Minecraft robot primarily utilizes the Minecraft environment to simulate a diverse and long horizon high-level embodied planning setting that mirrors the real world. The feature of this type of task is that it provides a procedurally generated, open-ended environment, which requires robot to make long-term plans and continuous decisions. These environment is typically only partially observable, compelling the robot to reason and act based on limited information in the tasks. Tasks require the robot to interact with the environment through exploration, resource gathering, and crafting over extended periods. This kind of tasks primarily evaluates several critical capabilities of the model, such as long-term planning, hierarchical task decomposition, tool utilization and so on. Datasets that support research in this domain include MineDojo \cite{fan2022minedojo}, Craftax \cite{matthews2024craftax}, MinePlanner \cite{hill2023mineplanner}, TeamCraft \cite{long2024teamcraft}, Plancraft \cite{dagan2024plancraft}, MINDAGENT \cite{gong2023mindagent}, and Overcooked-AI \cite{carroll2019utility}.

\textbf{Autonomous Driving:} An autonomous driving agent primarily leverages realistic driving simulators to emulate the dynamic, continuous, and safety-critical planning challenges of real-world road environments. Autonomous driving tasks present distinctive challenges characterized by partially observable environmental conditions, requiring comprehensive multimodal sensory perception through various inputs including visual data and radar telemetry. These tasks demand continuous decision-making processes under dynamic traffic scenarios, typically requiring iterative environmental interactions and adaptive responses. This setting is commonly explored using datasets such as PCA \cite{chen2024pca} and MetaAD \cite{jiang2025alphadrive}, which provide critical benchmarks for evaluating system performance.

\subsubsection{Everyday Scenarios} \label{everyday}
The purpose of everyday scenarios is to provide assistance through agents in complex daily tasks, such as travel planning, workflow arrangement, and tool invocation, in order to reduce user burden and enhance work efficiency. This mode is similar to Copilot, capable of replacing manual task arrangements, optimizing time management, and especially helping with resource integration and task scheduling in high-intensity work to improve execution outcomes. 

\textbf{Travel Planning:} It evaluates the planning capabilities of LLMs when multiple factors or constraints need to be considered at the same time. This evaluation paradigm typically provides an interactive dynamic environment that may involve purely textual interactions or incorporate tool-assisted functionalities, depending on the dataset. Within these environments, the model must develop comprehensive travel plans that satisfy user requirements, often dealing with partially observable decision processes where not all information is initially available. It measures the model's proficiency in multi-constraint decision planning, requiring a balanced optimization of temporal, financial, transportation, and accommodation factors. It also assesses the model's application of specialized travel knowledge, including understanding of geographical relationships, destination attributes, and transportation networks. Datasets such as TravelPlanner \cite{xie2024travelplanner}, ChinaTravel \cite{shao2024chinatravel}, and NATURAL PLAN \cite{zheng2024natural} have been commonly employed to support evaluations in this domain.

\textbf{Workflow:} It evaluates the task by decomposing a complex problem into a structured workflow, where individual tasks are represented as nodes. These nodes are interdependent, with each task relying on the completion of previous ones. These tasks are typically evaluated in static environments, do not involve dynamic interactive scenarios, and are primarily presented as pure text. The evaluation of workflow tasks focuses on several critical model capabilities. One essential aspect is the model's ability to systematically decompose complex problems and to identify and establish logical dependencies among subtasks. Additionally, workflow tasks assess the capacity to perform local adjustments and optimizations within the execution chain, thereby maintaining the overall consistency of the process even when dynamic changes occur. Datasets commonly used for evaluating such capabilities include FlowBench \cite{xiao2024flowbench}, Open Grounded Planning \cite{guo2024open}, HuggingGPT \cite{shen2023hugginggpt}, TaskBench \cite{shen2024taskbench}, TaskLAMA \cite{yuan2024tasklama}, WORFBENCH \cite{qiao2024benchmarking}, WIKIPLAN \cite{lu2023multimodal}, and ISG-BENCH \cite{chen2024interleaved}.

\textbf{Tool Calling:} This field evaluates LLM’s capability to strategize and plan the usage of external tools in complex problem. These tasks often involve multi-turn interactions in dynamic environments, where the model must handle incomplete information and adapt its strategy over multiple rounds to achieve the desired outcome. The evaluation of tool calling encompasses several essential capabilities. A fundamental requirement is the model’s ability to plan, retrieve, and execute appropriate external tools effectively. In addition, the model’s capacity to accurately interpret API documentation and construct precise, efficient requests that align with task objectives is equally important. Key resources for evaluating these capabilities include datasets such as ToolBench \cite{qin2023toolllm}, AppWorld \cite{trivedi2024appworld}, Api-bank \cite{li2023api}, ToolComp \cite{nath2025toolcomp}, Toolsandbox \cite{lu2024toolsandbox}, and Tooltalk \cite{farn2023tooltalk}.

\textbf{Code Generation:} The scene primarily assesses the software engineering capabilities of the model. These tasks are typically conducted within interactive environments that support iterative debugging and code modification. The inputs are primarily presented in plain text, without involving multimodal content. Furthermore, the problem setup often conforms to a MDP framework, as the generation of each code fragment depends primarily on the current state of the program. Code generation tasks place particular emphasis on an understanding of software engineering principles, the capacity to manage long contextual dependencies, and the mastery of syntax and best practices associated with specific programming languages. Success in this domain requires the model to reason about structural dependencies within the code and implement new functionalities without destroying the original ones. Datasets such as SWE-Bench \cite{jimenez2023swe}, SWE-Gym \cite{pan2024training}, and HumanEval \cite{chen2021evaluating} are commonly used to evaluate performance in this domain.

\textbf{Game Playing:} This assessment focuses on the game planning capabilities of LLMs within interactive and dynamically evolving environments. Models must not only understand the rules but also reason about spatial relationships and state transitions across multiple rounds of interaction in order to achieve specified goals. These tasks require models to exhibit a range of complex abilities, whereby they must comprehend and manipulate the positional relationships of objects in two- or three-dimensional spaces and strategic reasoning. Moreover, models must be capable of effective identifying efficient navigation strategies in environments populated with barriers, as well as robust error recovery, allowing them to reassess and adjust their plans when execution failures occur. Relevant datasets that support these capabilities and evaluations include VSP \cite{wu2024vsp}, TextWorld \cite{cote2019textworld}, BabyAI \cite{chevalier2018babyai}, and PlanBench \cite{valmeekam2023planbench}.

\subsubsection{Vertical Scenarios} \label{domains}
Vertical domain intelligent agents combine specialized tasks like machine learning, research, and simulation to efficiently handle data processing, model training, and analysis. They automate complex operations, replacing manual work and boosting productivity in fields like science and finance. Their goal is to expand knowledge, promote cross-domain innovation, and drive technological progress.

\textbf{Machine Learning:} This task provides an offline evaluation framework that closely mirrors real-world machine learning workflows. It requires the agent to accomplish objectives through multiple iterations of model development and evaluation, simulating real-world machine learning engineering tasks. The task primarily assesses the model’s abilities in data analysis and preprocessing, algorithm selection and optimization, experimental design and evaluation, code implementation and debugging, as well as decision-making under incomplete information. MLE-bench \cite{chan2024mle} serves as a representative benchmark for evaluating these capabilities in realistic machine learning engineering contexts. 

\textbf{AI Research:} These tasks are typically situated within multi-stage academic research workflows, requiring models to perform literature retrieval, selection, and organization in an offline setting. A defining feature of these tasks is their emphasis on evaluating a model's academic reasoning capabilities, its ability to plan research processes, its competence in integrating information across multiple interactions, and its proficiency in handling highly specialized domain knowledge. In contrast to other benchmarks, these tasks place particular importance on a model’s adaptability to environments with incomplete information, demanding the gradual construction of an internal understanding through sequential interactions. Relevant datasets that reflect these characteristics include ResearchArena \cite{kang2024researcharena} and CYCLERESEARCHER \cite{weng2024cycleresearcher}, both of which are designed to evaluate a model's performance.

\textbf{Biological Research:} The tasks within the Biological Research domain primarily aim to standardize and structure biological experimental procedures. These tasks are characterized by clearly defined steps and result-oriented objectives. Typically, they do not require dynamic, interactive environments, nor do they involve multimodal inputs. Task execution predominantly follows a one-off, sequential decision-making process, exhibiting typical properties of MDP. The core challenges in this domain center on the model’s ability to accurately comprehend complex, terminology-dense natural language descriptions of experimental protocols, to perform detailed scientific process analysis, and to convert unstructured procedural information into precise, structured representations. In particular, the model must demonstrate rigorous parsing and reconstruction of experimental operation sequences, conditional dependencies, and domain-specific knowledge. BIOPORT \cite{o2023bioplanner} serves as a benchmark for evaluating performance in such tasks.

\textbf{Financial Simulation:} Tasks within the this category are designed to provide an interactive and dynamic environment in which models must make sequential decisions and plan under uncertainty. Typically formulated as POMDPs, these tasks require models to achieve objectives through multiple rounds of interaction, rather than relying on a single-step input-output paradigm. They primarily evaluate a model’s capabilities in strategic resource allocation, decision-making under uncertainty, balancing long-term planning with short-term execution, and adapting strategies within competitive and evolving environments. AUCARENAAUCARENA \cite{chen2023put} is designed as a benchmark specifically for evaluating performance in these complex, dynamic decision-making tasks.

\textbf{Interior Design:} Datasets in this area focuses on transforming traditional image-based floor plan data into structured textual representations, thereby enabling language models to more effectively interpret and generate architectural design solutions. These datasets establish interactive and dynamic environments that support both partially and fully observable decision-making processes. This line of research primarily evaluates a model’s capacity to comprehend spatial relationships, articulate architectural structures, navigate design constraints, and optimize functional layouts. In this context, DStruct2Design \cite{luo2024dstruct2design} and Tell2Design \cite{leng2023tell2design} emerges as a benchmark for evaluating model performance across these demanding tasks.

\textbf{Composite:} This research area aims to rigorously assess the planning capabilities of LLMs within complex, interactive, and multi-dimensional environments. Evaluation tasks in this domain are dynamic and multimodal, typically exhibiting the characteristics of partially observable Markov decision processes. These tasks are designed to evaluate a wide range of core competencies, including but not limited to: complex task decomposition and hierarchical planning, multimodal information fusion and understanding, continuous environmental state tracking, and robust decision-making under conditions of uncertainty and incomplete information. Relevant benchmarks that support such evaluations include AgentBench \cite{liu2023agentbench} and VisualAgentBench \cite{liu2024visualagentbench}, which are specifically designed to reflect the complexity and diversity of these open-ended environments.

\subsection{Evaluation Metrics} \label{metrics}
\begin{figure*}[t]
\centering
  \includegraphics[width=0.99\textwidth,height=0.5\textheight,keepaspectratio]{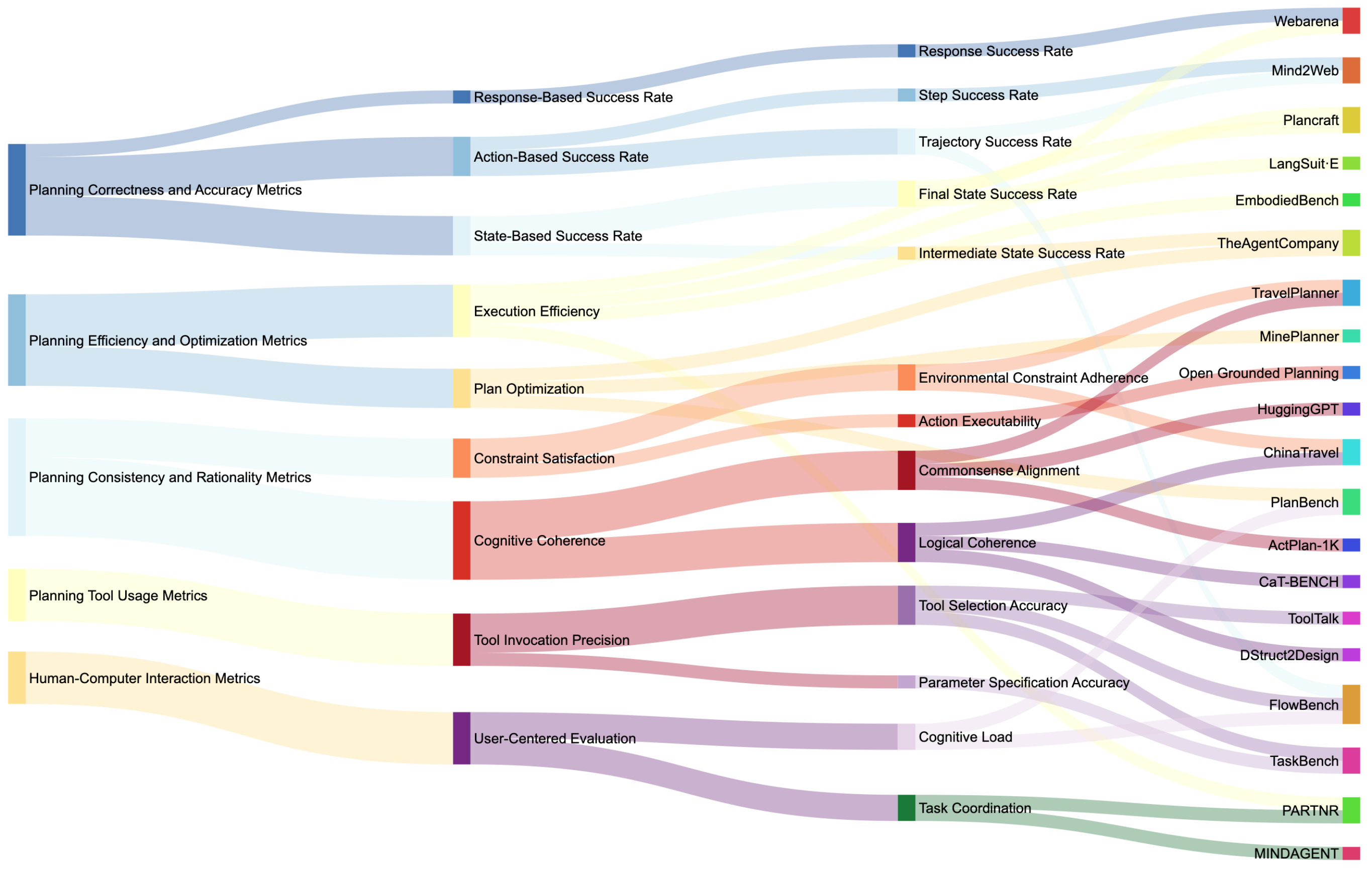}
  \caption{The corresponding relationship between planning evaluation metrics and some typical planning datasets. The left three columns represent evaluation metrics of different granularities, while the rightmost column denotes the dataset.}
  \label{sankey}
\end{figure*}

Evaluating the planning ability of LLMs is essential for assessing their practical value. This section will introduce a comprehensive set of indicators for evaluating LLM's planning capabilities, covering multiple dimensions from fundamental success rates to advanced interaction quality. The evaluation of planning ability can be categorized into five key indicators: \textbf{Planning Correctness and Accuracy} (\S\ref{accuracy}), \textbf{Planning Efficiency and Optimization} (\S\ref{quality}), \textbf{Planning Consistency and Rationality} (\S\ref{consistency}), \textbf{Planning Tool Usage} (\S\ref{tool}), and \textbf{Human-Computer Interaction} (\S\ref{interaction}). Different datasets use different evaluation metrics, and their corresponding relationships are shown in Figure \ref{sankey}.

\subsubsection{Planning Correctness and Accuracy Metrics} \label{accuracy}
The goal of correctness is to evaluate whether the agent can complete the task accurately and without errors, serving as the most fundamental evaluation dimensions. Common metrics is Success Rate (SR). SR can be divided into \textbf{Response-Based Success Rate}, \textbf{Action-Based Success Rate} and \textbf{State-Based Success Rate}. Because referred to as SR in existing papers, their meanings differ, and in the following text, they will be distinguished by different modifiers. 

For Response-Based Success Rate, the \textit{Response Success Rate} mainly measures the consistency between the agent's text response and the standard answer. In Webarena, the matching degree between the prediction and the answer is calculated by computing the three instantiation functions of r\_info \cite{zhou2023webarena}: exact\_match, must\_include, and fuzzy\_match.

For Action-Based Success Rate, the \textit{Step Success Rate} measures the consistency between the predicted single-step action and expert annotations. In Mind2Web \cite{deng2023mind2web}, Step Success Rate is used to evaluate the consistency between predicted single-step actions and manually annotated actions.
While the \textit{Trajectory Success Rate} directly evaluates the entire trajectory. In Mind2Web \cite{deng2023mind2web}, the paper uses the Success Rate to compare the consistency between the entire trajectory and the expert annotations. In FlowBench \cite{xiao2024flowbench}, the Success Rate aims to directly score the correctness of the trajectory using GPT-4 and a scoring template.

For State-Based Success Rate, since the trajectory to achieve the goal is not unique, some metrics evaluate correctness based on the online environment to address the issue of non-uniqueness of the path. 
The \textit{Final State Success Rate} checks whether the environment state reaches the expected state by constructing final state checkpoints during the program execution. In Webarena, the programmatic code function r\_prog \cite{zhou2023webarena} is designed to detect changes in the website database and web pages. The \textit{Intermediate State Success Rate} refers to the construction of intermediate state checkpoints through a program to detect whether the subtasks of the intermediate state are completed. In TheAgentCompany, the Partial Completion Score \cite{xu2024theagentcompany} is used to measure the score of intermediate checkpoints.

\subsubsection{Planning Efficiency and Optimization Metrics} \label{quality}
In addition to the basic success rate, planning quality also emphasize both the efficiency of plan execution and the optimality of the generated plan itself. These evaluation metrics can be divided into \textbf{Execution Efficiency Metrics} and \textbf{Plan Optimization Metrics}.

Execution efficiency metrics assess the operational economy during plan execution. \textit{Execution Efficiency} measure the number of actions required to complete a task, as reflected in the Average Steps used in LangSuit·E \cite{jia2024langsuit} or Simulation Steps used in PARTNR \cite{chang2024partnr}, where fewer steps generally indicate higher planning efficiency. \textit{Plan Optimization} further evaluates whether, among feasible plans, the chosen solution minimizes cumulative resource consumption relative to task-specific criteria, as exemplified by frameworks such as PlanBench \cite{valmeekam2023planbench} and the research on OpenAI’s o1 models \cite{wang2024planning}.

\subsubsection{Planning Consistency and Rationality Metrics} \label{consistency}
Consistency and rationality evaluation in the planning process aims to ensure that generated plans are not only compliant with user instructions but also aligned with environmental constraints, ensuring practical feasibility in real-world scenarios. This evaluation can be systematically divided into \textbf{Constraint Satisfaction} and \textbf{Cognitive Coherence}.  Together, these metrics provide a comprehensive framework for assessing whether generated plans are not only technically sound but also practically implementable and logically coherent.

For Constraint Satisfaction, it focuses on verifying the technical feasibility of a plan under objective conditions. \textit{Environmental Constraint Adherence} evaluates whether the plan respects external limitations. ChinaTravel \cite{shao2024chinatravel} introduces the Environmental Pass Rate (EPR), measuring satisfaction of predefined environmental constraints, while TravelPlanner \cite{xie2024travelplanner} defines the Hard Constraint Pass Rate with a focus on critical query-specified requirements. The \textit{Action Executability} assesses whether each planned action is realizable within the given action set. Open Grounded Planning \cite{guo2024open} proposes executableness indicators to verify that plan steps exist within the provided action library.

For Cognitive Coherence, it assesses the plan’s alignment with intuitive human reasoning and its internal logical structure. The \textit{Commonsense Alignment} measures whether the plan conforms to everyday human expectations and intuitive plausibility. ActPlan-1K \cite{su2024actplan} proposes Commonsense Satisfaction, and TravelPlanner \cite{xie2024travelplanner} introduces the Commonsense Pass Rate to evaluate this aspect. The \textit{Logical Coherence} evaluates whether plan steps are logically interconnected and free from contradictions. ChinaTravel \cite{shao2024chinatravel} defines the Logical Pass Rate (LPR) for this purpose, and CaT-BENCH’s \cite{lal2024cat} Temporal Consistency (TC) gauges a model’s ability to maintain proper time sequences throughout execution.

\subsubsection{Planning Tool Usage Metrics} \label{tool}
Modern LLM planning frequently requires interaction with external tool. To systematically evaluate the tool usage capabilities of LLMs, the assessment can be refined into an analysis of \textbf{Tool Invocation Precision}. These multi-faceted evaluation approaches collectively provide a comprehensive framework for assessing LLM planning capabilities across diverse professional domains and interaction requirements.

Tool Invocation Precision examines whether the agent can accurately select and operate external tools, including identifing the correct interface element and filling in the required parameters. \textit{Tool Selection Accuracy}, which measures if the correct tool is chosen given a task. TaskBench \cite{shen2024taskbench} uses Node F1 and Edge F1 to measure tool selection accuracy. \textit{Parameter Specification Accuracy}, which assesses whether the agent fills in tool parameters correctly. TaskBench proposes Parameter Name F1 and Parameter Name-Value Pair F1, evaluating the accuracy of parameter generation. 



\subsubsection{Human-Computer Interaction Metrics} \label{interaction}
Human-computer interaction indicators assess the overall performance and user experience of planning systems, highlighting their practical application value. \textbf{User-Centered Evaluation Metrics} capture the quality of human-computer interaction. Together, these metrics provide a framework for evaluating planning systems beyond technical performance, incorporating critical factors like usability, cognitive burden, and overall user satisfaction that ultimately determine real-world adoption and effectiveness.


User-centered metrics address the usability and cognitive implications of human-computer interaction. These indicators capture critical human factors such as mental effort, collaboration fluidity, and explanation clarity. \textit{Cognitive Load Metrics} assess the mental workload experienced by users when interacting with planning systems. In research on large language models’ planning abilities \cite{valmeekam2023planning},  NASA TLX is employed to quantitatively evaluate user cognitive load during interaction with the system. \textit{Task Coordination Metrics} measure the effectiveness of role distribution between human and system during joint task execution. PARTNR \cite{chang2024partnr} introduces a Task offloading metric, evaluating how efficiently tasks are distributed between human and AI.


\begin{table*}[t]
\caption{The performance comparison of different models and methods in web navigation. The value of Mind2Web is the average step success rate of the three subsets. The value of Webarena is task success rate. The value of AITW is step success rate of the subsets. The value of ScreenSpot is step success rate.}
\centering
\scalebox{0.77}{
\begin{tabular}{|l|l|c|c|c|c|c|c|c|c|c|}
\toprule
\textbf{Method} & \textbf{Base Model} & \textbf{Mind2Web} & \textbf{Webarena} & \textbf{AITW} & \textbf{ScreenSpot} & \textbf{Date} & \textbf{Finetune} & \textbf{Search}  & \textbf{Multimodal} \\
\midrule
SE-GUI \cite{yuan2025enhancing} & Qwen2.5-VL-7B \cite{bai2025qwen2} & - & - & - & 88.2 & 2025/5 & \textcolor{green}{\checkmark} & \textcolor{red}{\textbf{×}} & vision\&text \\
UIShift \cite{gao2025uishift} & Qwen2.5-VL-7B \cite{bai2025qwen2} & - & - & - & 87.8 & 2025/5 & \textcolor{green}{\checkmark} & \textcolor{red}{\textbf{×}} & vision\&text \\
ELAM \cite{ernhofer2025leveraging} & Molmo-7B \cite{deitke2024molmo} & - & - & - & 80.8 & 2025/5 & \textcolor{green}{\checkmark} & \textcolor{red}{\textbf{×}} & vision\&text \\
ScaleTrack \cite{huang2025scaletrack} & Qwen2-VL-7B \cite{wang2024qwen2} & - & - & - & 86.8 & 2025/5 & \textcolor{green}{\checkmark} & \textcolor{red}{\textbf{×}} & vision\&text \\
TongUI \cite{zhang2025tongui} & Qwen2.5-VL-7B \cite{bai2025qwen2} & 46.6 & - & 73.3 & 83.4 & 2025/4 & \textcolor{green}{\checkmark} & \textcolor{red}{\textbf{×}} & vision\&text \\
SpiritSight \cite{huang2025spiritsight} &  InternVL2-8B \cite{chen2024far} & 47.0 & - & - & 66.5 & 2025/4 & \textcolor{green}{\checkmark} & \textcolor{red}{\textbf{×}} & vision\&text\\
UI-I2E-VLM \cite{liu2025ui} & Qwen2-VL-7B \cite{wang2024qwen2} & - & - & - & 82.5 & 2025/4 & \textcolor{green}{\checkmark} & \textcolor{red}{\textbf{×}} & vision\&text \\
MP-GUI \cite{wang2025mp} &  InternVL2-8B \cite{chen2024far} & 34.9 & - & 69.2 & 64.1 & 2025/3 & \textcolor{green}{\checkmark} & \textcolor{red}{\textbf{×}} & vision\&text\\
Magma \cite{yang2025magma} &  LLaMA-3-8B \cite{grattafiori2024llama} & 45.4 & - & 67.3 & 54.5 & 2025/2 & \textcolor{green}{\checkmark} & \textcolor{red}{\textbf{×}} & vision\&text\\
Digi-Q \cite{baidigi} & LLaVA-1.5 \cite{Liu_2024} & - & - & 71.2 & - & 2025/2 & \textcolor{green}{\checkmark} & \textcolor{red}{\textbf{×}} & vision\&text \\
UI-TARS \cite{qin2025ui} & Qwen-2-VL-72B \cite{wang2024qwen2} & 64.7 & - & - & 88.4 & 2025/1 & \textcolor{green}{\checkmark} & \textcolor{red}{\textbf{×}} & vision\&text \\
UI-TARS \cite{qin2025ui} & Qwen-2-VL-7B \cite{wang2024qwen2} & 63.1 & - & - & 89.5 & 2025/1 & \textcolor{green}{\checkmark} & \textcolor{red}{\textbf{×}} & vision\&text \\
TREE SEARCH \cite{koh2024tree} & GPT-4o \cite{hurst2024gpt} & - & 19.2 & - & - & 2024/7 & \textcolor{red}{\textbf{×}} & \textcolor{green}{\checkmark} & vision\&text\\
MiniCPM-GUI \cite{chen2024guicourse} & MiniCPM-V \cite{yao2024minicpmvgpt4vlevelmllm} & 17.5 & - & 58.4 & - & 2024/6 & \textcolor{green}{\checkmark} & \textcolor{red}{\textbf{×}} & vision\&text\\
MiniCPM-V \cite{yao2024minicpm} & MiniCPM-V \cite{yao2024minicpmvgpt4vlevelmllm} & 6.5 & - & 48.9 & - & 2024/6 & \textcolor{green}{\checkmark} & \textcolor{red}{\textbf{×}}& vision\&text\\
UGround \cite{gou2024navigating} & LLaVA-NeXT-7B \cite{li2024llava} & 46.8 & - & - & 73.3 & 2024/6 & \textcolor{green}{\checkmark} & \textcolor{red}{\textbf{×}} & vision\&text\\
SeeAct \cite{zheng2024gpt} & GPT-4 \cite{achiam2023gpt} & 36.4 & - & - & - & 2024/1 & \textcolor{red}{\textbf{×}} & \textcolor{red}{\textbf{×}} & vision\&text\\
SeeAct \cite{zheng2024gpt} & LLaVA-1.5 \cite{Liu_2024}  & 8.0 & - & - & - & 2024/1 & \textcolor{red}{\textbf{×}} & \textcolor{red}{\textbf{×}} & vision\&text\\
SeeClick \cite{cheng2024seeclick} & Qwen-VL \cite{bai2023qwen} & 20.9 & - & 59.3 & 53.4 & 2024/1 & \textcolor{green}{\checkmark} & \textcolor{red}{\textbf{×}} & vision\&text\\
CogAgent \cite{hong2024cogagent} & CogVLM-17B \cite{wang2024cogvlm} & 58.5 & - & 76.8 & 47.4 & 2023/12 & \textcolor{green}{\checkmark} & \textcolor{red}{\textbf{×}} & vision\&text\\
Agent LUMOS \cite{yin2023agent} & Llama-2-13B \cite{touvron2023llama} & 31.3 & - & - & - & 2023/11 & \textcolor{green}{\checkmark} & \textcolor{red}{\textbf{×}} & text\\
AgentTuning \cite{zeng2023agenttuning} & Llama-2-13B \cite{touvron2023llama} & 8.4 & 1.6 & - & - & 2023/10 & \textcolor{green}{\checkmark} & \textcolor{red}{\textbf{×}} & text\\
MINDACT \cite{deng2023mind2web} & Flan-T5\textsubscript{XL} \cite{chung2024scaling} & 43.5 & - & - & - & 2023/6 & \textcolor{green}{\checkmark} & \textcolor{red}{\textbf{×}}& text\\
Synapse \cite{zheng2023synapse} & GPT-3.5 \cite{brown2020language} & 27.0 & - & - & - & 2023/6 & \textcolor{red}{\textbf{×}} & \textcolor{red}{\textbf{×}} & text\\
Synapse \cite{zheng2023synapse} & CodeLlama-7B \cite{roziere2023code} & 21.6 & - & - & - & 2023/6 & \textcolor{red}{\textbf{×}} & \textcolor{red}{\textbf{×}} & text\\
\bottomrule
\end{tabular}}
\label{performance1}
\end{table*}

\subsection{Performance Comparisons} \label{performance}
In order to better demonstrate the progress of LLMs in planning tasks, we present the performance of some representative methods on commonly used datasets, as shown in Table \ref{performance1} and Table \ref{performance2}. From the tables, we have the following important observations: 

(1) Although LLMs have demonstrated promising potential in planning tasks, their overall performance remains suboptimal. For example, some of the latest methods perform well on the ALFWorld datasets (i.e., the success rate of planning exceeds 80\%), but perform poorly on more complex datasets such as Mind2Web, with step success rates typically less than 65\%. This indicates that the planning task is very promising and more effective planning methods need to be designed for more complex tasks.

(2)  Among various planning tasks, fine-tuning based methods have become the mainstream approach. From the time when the method was proposed, researchers are more inclined to solve planning tasks through fine-tuning. The reason is that fine-tuning the model (i.e., imitation learning-based methods and feedback-based methods) can significantly enhance the planning performance of models compared to other types of methods. Although some researchers have attempted to use training-free methods (e.g., searching-based methods), these methods are limited by the capabilities of the base model.


(3) Multimodal agents are becoming increasingly popular. Agents are gradually evolving from understanding only textual modalities to processing purely visual inputs or a combination of visual and textual inputs. With the integration of multimodal information, the behavior of agents is becoming more and more human-like. For example, in web navigation tasks, the incorporation of multimodal inputs allows agents to perform actions such as clicking on specific icons on the screen or typing in input fields, making their operations increasingly similar to those of humans.

\begin{table*}[t]
\caption{The performance comparison of different models and methods in embodied. This refers to the average of Seen and Unseen in the original paper, or the value reported in the original paper.}
\centering
\scalebox{0.9}{
\begin{tabular}{|l|l|c|c|c|c|c|}
\toprule
\textbf{Method} & \textbf{Base Model} & \textbf{ALFWorld} & \textbf{ScienceWorld} & \textbf{Date} & \textbf{Finetune} \\
\midrule
GiGPO \cite{feng2025group} & Qwen2.5-7B \cite{qwen2024qwen25} & 90.8 & - & 2025/5 & \textcolor{green}{\checkmark} \\
CQL-Prior \cite{shihab2025cache} & Qwen-7B \cite{bai2023qwen} & 62.0 & - & 2025/5 & \textcolor{green}{\checkmark} \\
FGO \cite{liu2025divide} & GPT-4o-mini \cite{hurst2024gpt} & 83.6 & - & 2025/5 & \textcolor{red}{\textbf{×}} \\
LTC \cite{wang2023adapting} & Llama-7B \cite{touvron2023llama} & 91.0 & - & 2025/4 & \textcolor{green}{\checkmark} \\
CDMem \cite{gao2025efficient} & GPT-4o \cite{hurst2024gpt} & 90.0 & 56.0 & 2025/4 & \textcolor{red}{\textbf{×}} \\

ATLAS \cite{chen2025atlas} & Llama-3.1-8B \cite{grattafiori2024llama} & 84.5 & 42.0 & 2025/3 & \textcolor{green}{\checkmark} \\
DebFlow \cite{su2025debflow} & GPT-4o-mini \cite{hurst2024gpt} & 62.3 & - & 2025/3 & \textcolor{red}{\textbf{×}} \\
Agent-R \cite{yuan2025agent} & Llama-3.1-8B \cite{grattafiori2024llama} & - & 70.2 & 2025/1 & \textcolor{green}{\checkmark} \\
AgentRefine \cite{fu2025agentrefine} & LLama-3-8B \cite{grattafiori2024llama} & 44.8 & 14.4 & 2025/1 & \textcolor{green}{\checkmark} \\
AgentEvol \cite{xi2024agentgym} & Llama-2-7B \cite{touvron2023llama} & 88.0 & 38.0 & 2024/6 & \textcolor{green}{\checkmark} \\
WKM \cite{qiao2024agent} & Mistral-7B \cite{jiang2023mistral} & 75.2 & 57.8 & 2024/5 & \textcolor{green}{\checkmark} \\
WKM \cite{qiao2024agent} & Gemma-7B \cite{team2024gemma}  & 70.5 & 51.4 & 2024/5 & \textcolor{green}{\checkmark} \\
ETO \cite{song2024trial} & Llama-2-7B \cite{touvron2023llama} & 70.5 & 69.4 & 2024/3 & \textcolor{green}{\checkmark} \\
KNOWAGENT \cite{zhu2024knowagent} & Llama-2-7B \cite{touvron2023llama}  & 29.3 & - & 2024/3 & \textcolor{green}{\checkmark} \\
KNOWAGENT \cite{zhu2024knowagent} & Llama-2-13B \cite{touvron2023llama}  & 56.5 & - & 2024/3 & \textcolor{green}{\checkmark} \\
AgentTuning \cite{zeng2023agenttuning} & Llama-2-7B \cite{touvron2023llama} & 84.0 & 13.7 & 2023/10 & \textcolor{green}{\checkmark} \\
\bottomrule
\end{tabular}}
\label{performance2}
\end{table*}

\section{Analysis and Interpretation} \label{AI}
\begin{figure}[t]
\centering
  \includegraphics[width=0.45\textwidth,height=0.5\textheight,keepaspectratio]{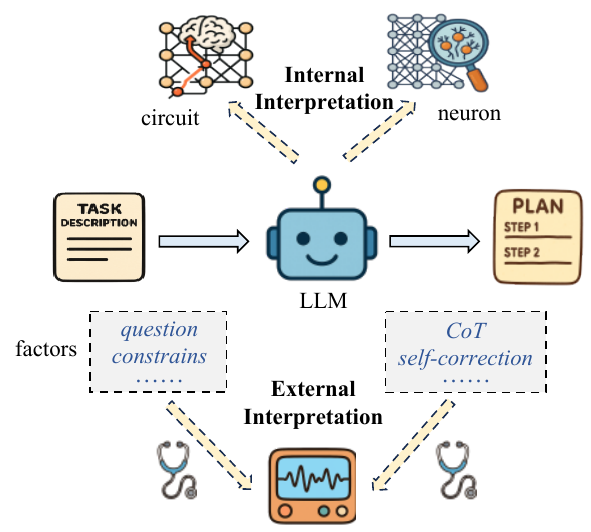}
  \caption{The illustration of external interpretation and internal interpretation, respectively.}
  \label{interpret}
\end{figure}

In addition to employing advanced techniques to enhance the planning capabilities of LLMs, several studies have also focused on analyzing and interpreting their performance in planning tasks from both external and internal perspectives \cite{stechly2024chain,bachmann2024pitfalls,yang2024large,men2024unlocking}, as shown in Figure \ref{interpret}.

\textbf{External Interpretation:} Several researchers have investigated the factors influencing LLM-based planning \cite{bachmann2024pitfalls,xie2024revealing}. For instance, Bachmann et al. \cite{bachmann2024pitfalls} argue that next-token prediction during training (i.e., teacher forcing) may introduce problematic learning dynamics in lookahead tasks. Moreover, autoregressive inference can lead to cascading errors, complicating the execution of plans during inference. In addition, Xie et al. \cite{xie2024revealing} explore the factors influencing LLM planning performance, identifying two key aspects: the limited role of constraints and the diminishing impact of questions. 

Some inherent capabilities of some LLMs are closely related to planning. For example, the excellent Chain-of-Thought (CoT) capability of LLMs supports them in completing complex planning tasks, while the mechanism behind CoT remains unclear. To this end, some researchers attempt to analyze the impact of CoT on planning. Sprague et al. \cite{sprague2024cot} attempt to evaluate where prompt-based CoT helps and why. They find that CoT is predominantly helpful tasks for tasks involving math or logic, as it can trace the intermediate steps of a problem for these tasks. Stechly et al. \cite{stechly2024chain} investigate the limitations of Chain-of-Thought (CoT) in solving classical planning problems, such as Blocksworld. Their findings indicate that CoT-like approaches rely heavily on pattern matching to enhance planning performance. In other words, the effectiveness of these methods depends on the similarity between in-context examples and the current query, meaning they cannot be easily generalized.  In addition, self-correction capabilities is crucial for the planning of LLMs, and increasingly being explored in recent works \cite{li2024confidence,madaan2024self,zhang2024small}. However, the self-correction mechanism is also unclear. Wang et al. \cite{wang2024theoretical} analyze self-correction from an in-context learning perspective, finding that when the self-assessment of LLMs is relatively correct, they can improve the responses in an in-context way.

\textbf{Internal Interpretation: } Some researchers aim to uncover the underlying mechanisms of LLM-based planning. Addressing whether LLMs perform forward thinking during inference, Wu et al. \cite{wu2024language} propose two hypotheses: 1) the pre-caching hypothesis, where the transformer model learns features that are not immediately relevant to the current inference step but may become useful in future steps, and 2) the breadcrumbs hypothesis, in which the features most pertinent to the current step are also the ones that would be most advantageous for future inference steps. They demonstrate that the transformer model can learn pre-caching using a synthetic dataset. Moreover, their findings suggest that smaller language models tend to favor the breadcrumbs hypothesis. Similarly, Men et al. \cite{men2024unlocking} examine the look-ahead planning mechanism in LLMs from the perspectives of information flow and internal representations. Their analysis indicates that the multi-head self-attention module plays a crucial role in planning, primarily extracting decision-relevant information from spans of goal states and recent steps. They also introduce the look-ahead planning decision existence hypothesis, suggesting that the middle and upper layers encode a few short-term future decisions. Wang et al. \cite{wang2024alpine} approach planning as a network path-finding task and explore the emergence of planning capabilities in transformer-based LLMs. Both their theoretical analysis and empirical experiments reveal that LLMs can learn both adjacency matrices and a limited form of reachability matrices. In addition, Cabannes et al. \cite{cabannes2024iteration} explore how CoT reasoning emerges in LLMs. They demonstrate that transformers can develop specialized attention mechanisms, or ``iteration heads'', which are dedicated to iterative reasoning. Also, the transformers may develop ``inner circuits'' for multi-hop reasoning, which can apply similar reasoning patterns to other tasks with shared logical structures.

\begin{tcolorbox}[colback=blue!5!white,colframe=black!75!white,
  title=\textbf{Summary of Analysis and Interpretation}, 
  fonttitle=\bfseries, coltitle=white, colbacktitle=black, rounded corners]
  \begin{itemize}[leftmargin=1.5em]
    \item Some studies about \textit{external interpretation} investigate the factors influencing planning, and also analyze the role of the inherent capabilities of LLMs in planning. Although these analyses can provide guidance for improving the planning ability of LLMs, they cannot reveal the internal mechanisms.

    \item Research on \textit{internal interpretation} attempts to explain planning from within LLMs, such as the emergence of planning and look-ahead planning mechanisms. While preliminary explorations have been made, a systematic understanding of the underlying processes remains lacking. Several critical issues still require attention, including the origins and transferability of planning abilities.
  \end{itemize}
\end{tcolorbox}

\section{Future Directions} \label{direction}
The following directions present promising avenues for advancing future research in LLM-based planning systems:

\subsection{Reinforcement Learning in LLM-based Planning}
Recent reports such as Deepseek-R1 \cite{guo2025deepseek} indicate that reinforcement learning (RL) is effective to enhance reasoning ability of LLMs. Thus, a promising future direction involves integrating RL technique into LLM-based planning systems. This line of research entails several key challenges and opportunities. First, designing realistic and scalable environment simulations is critical to enabling LLM agents to interact, learn, and generalize effectively across diverse planning scenarios. Second, developing reward models that reflect complex, long-horizon objectives, and goal alignment—is essential for guiding the learning process. Moreover, standard RL algorithms must be adapted to accommodate the unique properties of language models, including their high-dimensional trajectory spaces, context sensitivity and balancing exploration and exploitation during planning processes.

\subsection{Planning with Environmental Constraints}
It is also an important direction for future research lies in developing efficient planning algorithms that operate effectively under environmental constraints, particularly in settings characterized by partial observability. In such environments, agents must plan without full access to the underlying state space, making the exploration process significantly more complex. To address these challenges, it is essential to design intelligent state estimation methods that can infer hidden or uncertain aspects of the environment based on limited observations. Moreover, integrating environmental priors, such as known physical constraints, domain rules, or structural regularities, could further improve planning robustness and reduce unnecessary exploration. Balancing accuracy, efficiency, and scalability in the presence of incomplete information remains a key challenge and a fertile ground for innovation.

\subsection{Personalized Planning Systems}
The existing methods are mainly aimed at general planning tasks. For better application, future planning systems should aim to incorporate mechanisms for personalized decision-making by continuously learning user-specific goals, preferences, and behavioral patterns through ongoing interaction. Achieving effective personalization requires a combination of adaptive model tuning—either through fine-tuning of parameters or integration with modular external memory—and secure, efficient long-term memory storage. Crucially, these systems must address challenges related to privacy preservation, user data sparsity, and lifelong learning without catastrophic forgetting. The development of privacy-aware personalization frameworks and few-shot adaptation techniques will be key enablers for this direction.

\subsection{Edge Deployment for the Planning of Agents}
Although notable progress has been made in LLM-based planning, current approaches remain heavily dependent on the capabilities of large-scale base models and require substantial computational resources. A promising research direction involves developing lightweight and efficient planning models that can be deployed on resource-constrained edge devices such as mobile phones, drones, and robotic platforms. However, downsizing models often leads to a significant decline in planning performance. Key challenges include reducing model size and computational overhead while preserving robust planning capabilities and ensuring adaptability to dynamic environments. Techniques such as model compression \cite{zhu2024survey} and transfer learning \cite{zhuang2020comprehensive} are expected to play a pivotal role in addressing these challenges.

\subsection{Pretraining of Agent Base Models}
Recent efforts have focused on advancing techniques for pretraining foundational agent models using synthetically generated trajectory data. These pretrained agents serve as general-purpose backbones for downstream planning and decision-making tasks. A major challenge lies in designing scalable and efficient data synthesis pipelines that can generate diverse, high-quality trajectories at scale. Additionally, ensuring sufficient coverage of long-tail scenarios—those that are rare but critical for real-world robustness—remains an open problem. Future directions include leveraging simulation environments and data synthesis to better expose agent models to complex planning contexts during pretraining.

\subsection{Enhancing Generalization Ability of Planning Systems}
Most planning methods are typically designed for task-specific settings and often struggle to generalize across diverse tasks or dynamic environments. Their performance tends to degrade significantly when faced with unseen scenarios, primarily due to limited adaptability and reliance on task-specific training data. In contrast, LLMs trained with reinforcement learning exhibit more robust generalization capabilities, attributed to their strong reasoning and abstraction abilities. This raises a compelling question: can generalization in planning be driven more by enhanced reasoning, rather than by mere memorization of training data? Consequently, developing a universal LLM-based planner that can effectively generalize across tasks, domains, and environments represents a highly promising direction for future research.



\section{Conclusion} \label{conclusion}
In this paper, we present a comprehensive and up-to-date survey of planning with LLMs, aiming to provide a clear understanding of current progress, challenges, and future directions in this emerging area. We begin by introducing the fundamental definitions and categories of automated planning. We then provide a detailed overview of existing LLM-based planning approaches, which we categorize into three main paradigms: external module augmented methods, finetuning-based methods, and searching-based methods. Following this, we systematically review evaluation frameworks used in the field, including widely adopted datasets, evaluation metrics, and comparative performance analyses. We also delve into the underlying mechanisms and interpretability about LLM-based planning. Finally, we highlight key challenges and outline promising research directions to guide future developments in this rapidly evolving field. We hope that this survey will serve as a valuable resource for researchers and practitioners, facilitating deeper exploration and inspiring innovative contributions to LLM-based planning.


\bibliographystyle{ACM-Reference-Format}
\bibliography{sample-base}


\begin{thebibliography}{350}


\ifx \showCODEN    \undefined \def \showCODEN     #1{\unskip}     \fi
\ifx \showISBNx    \undefined \def \showISBNx     #1{\unskip}     \fi
\ifx \showISBNxiii \undefined \def \showISBNxiii  #1{\unskip}     \fi
\ifx \showISSN     \undefined \def \showISSN      #1{\unskip}     \fi
\ifx \showLCCN     \undefined \def \showLCCN      #1{\unskip}     \fi
\ifx \shownote     \undefined \def \shownote      #1{#1}          \fi
\ifx \showarticletitle \undefined \def \showarticletitle #1{#1}   \fi
\ifx \showURL      \undefined \def \showURL       {\relax}        \fi
\providecommand\bibfield[2]{#2}
\providecommand\bibinfo[2]{#2}
\providecommand\natexlab[1]{#1}
\providecommand\showeprint[2][]{arXiv:#2}

\bibitem[Achiam et~al\mbox{.}(2023)]%
        {achiam2023gpt}
\bibfield{author}{\bibinfo{person}{Josh Achiam}, \bibinfo{person}{Steven Adler}, \bibinfo{person}{Sandhini Agarwal}, \bibinfo{person}{Lama Ahmad}, \bibinfo{person}{Ilge Akkaya}, \bibinfo{person}{Florencia~Leoni Aleman}, \bibinfo{person}{Diogo Almeida}, \bibinfo{person}{Janko Altenschmidt}, \bibinfo{person}{Sam Altman}, \bibinfo{person}{Shyamal Anadkat}, {et~al\mbox{.}}} \bibinfo{year}{2023}\natexlab{}.
\newblock \showarticletitle{Gpt-4 technical report}.
\newblock \bibinfo{journal}{\emph{arXiv preprint arXiv:2303.08774}} (\bibinfo{year}{2023}).
\newblock


\bibitem[Aeronautiques et~al\mbox{.}(1998)]%
        {aeronautiques1998pddl}
\bibfield{author}{\bibinfo{person}{Constructions Aeronautiques}, \bibinfo{person}{Adele Howe}, \bibinfo{person}{Craig Knoblock}, \bibinfo{person}{ISI~Drew McDermott}, \bibinfo{person}{Ashwin Ram}, \bibinfo{person}{Manuela Veloso}, \bibinfo{person}{Daniel Weld}, \bibinfo{person}{David~Wilkins Sri}, \bibinfo{person}{Anthony Barrett}, \bibinfo{person}{Dave Christianson}, {et~al\mbox{.}}} \bibinfo{year}{1998}\natexlab{}.
\newblock \showarticletitle{Pddl| the planning domain definition language}.
\newblock \bibinfo{journal}{\emph{Technical Report, Tech. Rep.}} (\bibinfo{year}{1998}).
\newblock


\bibitem[Aghzal et~al\mbox{.}(2025)]%
        {aghzal2025survey}
\bibfield{author}{\bibinfo{person}{Mohamed Aghzal}, \bibinfo{person}{Erion Plaku}, \bibinfo{person}{Gregory~J Stein}, {and} \bibinfo{person}{Ziyu Yao}.} \bibinfo{year}{2025}\natexlab{}.
\newblock \showarticletitle{A survey on large language models for automated planning}.
\newblock \bibinfo{journal}{\emph{arXiv preprint arXiv:2502.12435}} (\bibinfo{year}{2025}).
\newblock


\bibitem[Ahmad et~al\mbox{.}(2025)]%
        {ahmad2025opencodereasoning}
\bibfield{author}{\bibinfo{person}{Wasi~Uddin Ahmad}, \bibinfo{person}{Sean Narenthiran}, \bibinfo{person}{Somshubra Majumdar}, \bibinfo{person}{Aleksander Ficek}, \bibinfo{person}{Siddhartha Jain}, \bibinfo{person}{Jocelyn Huang}, \bibinfo{person}{Vahid Noroozi}, {and} \bibinfo{person}{Boris Ginsburg}.} \bibinfo{year}{2025}\natexlab{}.
\newblock \showarticletitle{OpenCodeReasoning: Advancing Data Distillation for Competitive Coding}.
\newblock \bibinfo{journal}{\emph{arXiv preprint arXiv:2504.01943}} (\bibinfo{year}{2025}).
\newblock


\bibitem[{\AA}str{\"o}m(1965)]%
        {aastrom1965optimal}
\bibfield{author}{\bibinfo{person}{Karl~Johan {\AA}str{\"o}m}.} \bibinfo{year}{1965}\natexlab{}.
\newblock \showarticletitle{Optimal control of Markov processes with incomplete state information I}.
\newblock \bibinfo{journal}{\emph{Journal of mathematical analysis and applications}}  \bibinfo{volume}{10} (\bibinfo{year}{1965}), \bibinfo{pages}{174--205}.
\newblock


\bibitem[Bachmann and Nagarajan(2024)]%
        {bachmann2024pitfalls}
\bibfield{author}{\bibinfo{person}{Gregor Bachmann} {and} \bibinfo{person}{Vaishnavh Nagarajan}.} \bibinfo{year}{2024}\natexlab{}.
\newblock \showarticletitle{The pitfalls of next-token prediction}.
\newblock \bibinfo{journal}{\emph{arXiv preprint arXiv:2403.06963}} (\bibinfo{year}{2024}).
\newblock


\bibitem[Bai et~al\mbox{.}({[n.\,d.]})]%
        {baidigi}
\bibfield{author}{\bibinfo{person}{Hao Bai}, \bibinfo{person}{Yifei Zhou}, \bibinfo{person}{Li~Erran Li}, \bibinfo{person}{Sergey Levine}, {and} \bibinfo{person}{Aviral Kumar}.} \bibinfo{year}{[n.\,d.]}\natexlab{}.
\newblock \showarticletitle{Digi-Q: Learning VLM Q-Value Functions for Training Device-Control Agents}. In \bibinfo{booktitle}{\emph{The Thirteenth International Conference on Learning Representations}}.
\newblock


\bibitem[Bai et~al\mbox{.}(2023)]%
        {bai2023qwen}
\bibfield{author}{\bibinfo{person}{Jinze Bai}, \bibinfo{person}{Shuai Bai}, \bibinfo{person}{Shusheng Yang}, \bibinfo{person}{Shijie Wang}, \bibinfo{person}{Sinan Tan}, \bibinfo{person}{Peng Wang}, \bibinfo{person}{Junyang Lin}, \bibinfo{person}{Chang Zhou}, {and} \bibinfo{person}{Jingren Zhou}.} \bibinfo{year}{2023}\natexlab{}.
\newblock \showarticletitle{Qwen-vl: A versatile vision-language model for understanding, localization, text reading, and beyond. arXiv 2023}.
\newblock \bibinfo{journal}{\emph{arXiv preprint arXiv:2308.12966}} \bibinfo{volume}{1}, \bibinfo{number}{8} (\bibinfo{year}{2023}).
\newblock


\bibitem[Bai et~al\mbox{.}(2025)]%
        {bai2025qwen2}
\bibfield{author}{\bibinfo{person}{Shuai Bai}, \bibinfo{person}{Keqin Chen}, \bibinfo{person}{Xuejing Liu}, \bibinfo{person}{Jialin Wang}, \bibinfo{person}{Wenbin Ge}, \bibinfo{person}{Sibo Song}, \bibinfo{person}{Kai Dang}, \bibinfo{person}{Peng Wang}, \bibinfo{person}{Shijie Wang}, \bibinfo{person}{Jun Tang}, {et~al\mbox{.}}} \bibinfo{year}{2025}\natexlab{}.
\newblock \showarticletitle{Qwen2. 5-vl technical report}.
\newblock \bibinfo{journal}{\emph{arXiv preprint arXiv:2502.13923}} (\bibinfo{year}{2025}).
\newblock


\bibitem[Barrett et~al\mbox{.}(2011)]%
        {barrett2011cvc4}
\bibfield{author}{\bibinfo{person}{Clark Barrett}, \bibinfo{person}{Christopher~L Conway}, \bibinfo{person}{Morgan Deters}, \bibinfo{person}{Liana Hadarean}, \bibinfo{person}{Dejan Jovanovi{\'c}}, \bibinfo{person}{Tim King}, \bibinfo{person}{Andrew Reynolds}, {and} \bibinfo{person}{Cesare Tinelli}.} \bibinfo{year}{2011}\natexlab{}.
\newblock \showarticletitle{cvc4}. In \bibinfo{booktitle}{\emph{Computer Aided Verification: 23rd International Conference, CAV 2011, Snowbird, UT, USA, July 14-20, 2011. Proceedings 23}}. Springer, \bibinfo{pages}{171--177}.
\newblock


\bibitem[Barto and Mahadevan(2003)]%
        {barto2003recent}
\bibfield{author}{\bibinfo{person}{Andrew~G Barto} {and} \bibinfo{person}{Sridhar Mahadevan}.} \bibinfo{year}{2003}\natexlab{}.
\newblock \showarticletitle{Recent advances in hierarchical reinforcement learning}.
\newblock \bibinfo{journal}{\emph{Discrete event dynamic systems}}  \bibinfo{volume}{13} (\bibinfo{year}{2003}), \bibinfo{pages}{341--379}.
\newblock


\bibitem[Bellman(1957)]%
        {bellman1957markovian}
\bibfield{author}{\bibinfo{person}{Richard Bellman}.} \bibinfo{year}{1957}\natexlab{}.
\newblock \showarticletitle{A Markovian decision process}.
\newblock \bibinfo{journal}{\emph{Journal of mathematics and mechanics}} (\bibinfo{year}{1957}), \bibinfo{pages}{679--684}.
\newblock


\bibitem[Besta et~al\mbox{.}(2024)]%
        {besta2024graph}
\bibfield{author}{\bibinfo{person}{Maciej Besta}, \bibinfo{person}{Nils Blach}, \bibinfo{person}{Ales Kubicek}, \bibinfo{person}{Robert Gerstenberger}, \bibinfo{person}{Michal Podstawski}, \bibinfo{person}{Lukas Gianinazzi}, \bibinfo{person}{Joanna Gajda}, \bibinfo{person}{Tomasz Lehmann}, \bibinfo{person}{Hubert Niewiadomski}, \bibinfo{person}{Piotr Nyczyk}, {et~al\mbox{.}}} \bibinfo{year}{2024}\natexlab{}.
\newblock \showarticletitle{Graph of thoughts: Solving elaborate problems with large language models}. In \bibinfo{booktitle}{\emph{Proceedings of the AAAI Conference on Artificial Intelligence}}, Vol.~\bibinfo{volume}{38}. \bibinfo{pages}{17682--17690}.
\newblock


\bibitem[Bonatti et~al\mbox{.}(2024)]%
        {bonatti2024windows}
\bibfield{author}{\bibinfo{person}{Rogerio Bonatti}, \bibinfo{person}{Dan Zhao}, \bibinfo{person}{Francesco Bonacci}, \bibinfo{person}{Dillon Dupont}, \bibinfo{person}{Sara Abdali}, \bibinfo{person}{Yinheng Li}, \bibinfo{person}{Yadong Lu}, \bibinfo{person}{Justin Wagle}, \bibinfo{person}{Kazuhito Koishida}, \bibinfo{person}{Arthur Bucker}, {et~al\mbox{.}}} \bibinfo{year}{2024}\natexlab{}.
\newblock \showarticletitle{Windows agent arena: Evaluating multi-modal os agents at scale}.
\newblock \bibinfo{journal}{\emph{arXiv preprint arXiv:2409.08264}} (\bibinfo{year}{2024}).
\newblock


\bibitem[Brown et~al\mbox{.}(2020)]%
        {brown2020language}
\bibfield{author}{\bibinfo{person}{Tom Brown}, \bibinfo{person}{Benjamin Mann}, \bibinfo{person}{Nick Ryder}, \bibinfo{person}{Melanie Subbiah}, \bibinfo{person}{Jared~D Kaplan}, \bibinfo{person}{Prafulla Dhariwal}, \bibinfo{person}{Arvind Neelakantan}, \bibinfo{person}{Pranav Shyam}, \bibinfo{person}{Girish Sastry}, \bibinfo{person}{Amanda Askell}, {et~al\mbox{.}}} \bibinfo{year}{2020}\natexlab{}.
\newblock \showarticletitle{Language models are few-shot learners}.
\newblock \bibinfo{journal}{\emph{Advances in neural information processing systems}}  \bibinfo{volume}{33} (\bibinfo{year}{2020}), \bibinfo{pages}{1877--1901}.
\newblock


\bibitem[Cabannes et~al\mbox{.}(2024)]%
        {cabannes2024iteration}
\bibfield{author}{\bibinfo{person}{Vivien Cabannes}, \bibinfo{person}{Charles Arnal}, \bibinfo{person}{Wassim Bouaziz}, \bibinfo{person}{Alice Yang}, \bibinfo{person}{Francois Charton}, {and} \bibinfo{person}{Julia Kempe}.} \bibinfo{year}{2024}\natexlab{}.
\newblock \showarticletitle{Iteration Head: A Mechanistic Study of Chain-of-Thought}.
\newblock \bibinfo{journal}{\emph{arXiv preprint arXiv:2406.02128}} (\bibinfo{year}{2024}).
\newblock


\bibitem[Carroll et~al\mbox{.}(2019)]%
        {carroll2019utility}
\bibfield{author}{\bibinfo{person}{Micah Carroll}, \bibinfo{person}{Rohin Shah}, \bibinfo{person}{Mark~K Ho}, \bibinfo{person}{Tom Griffiths}, \bibinfo{person}{Sanjit Seshia}, \bibinfo{person}{Pieter Abbeel}, {and} \bibinfo{person}{Anca Dragan}.} \bibinfo{year}{2019}\natexlab{}.
\newblock \showarticletitle{On the utility of learning about humans for human-ai coordination}.
\newblock \bibinfo{journal}{\emph{Advances in neural information processing systems}}  \bibinfo{volume}{32} (\bibinfo{year}{2019}).
\newblock


\bibitem[Chai et~al\mbox{.}(2025)]%
        {chai2025a3}
\bibfield{author}{\bibinfo{person}{Yuxiang Chai}, \bibinfo{person}{Hanhao Li}, \bibinfo{person}{Jiayu Zhang}, \bibinfo{person}{Liang Liu}, \bibinfo{person}{Guangyi Liu}, \bibinfo{person}{Guozhi Wang}, \bibinfo{person}{Shuai Ren}, \bibinfo{person}{Siyuan Huang}, {and} \bibinfo{person}{Hongsheng Li}.} \bibinfo{year}{2025}\natexlab{}.
\newblock \showarticletitle{A3: Android Agent Arena for Mobile GUI Agents}.
\newblock \bibinfo{journal}{\emph{arXiv preprint arXiv:2501.01149}} (\bibinfo{year}{2025}).
\newblock


\bibitem[Chan et~al\mbox{.}(2024)]%
        {chan2024mle}
\bibfield{author}{\bibinfo{person}{Jun~Shern Chan}, \bibinfo{person}{Neil Chowdhury}, \bibinfo{person}{Oliver Jaffe}, \bibinfo{person}{James Aung}, \bibinfo{person}{Dane Sherburn}, \bibinfo{person}{Evan Mays}, \bibinfo{person}{Giulio Starace}, \bibinfo{person}{Kevin Liu}, \bibinfo{person}{Leon Maksin}, \bibinfo{person}{Tejal Patwardhan}, {et~al\mbox{.}}} \bibinfo{year}{2024}\natexlab{}.
\newblock \showarticletitle{Mle-bench: Evaluating machine learning agents on machine learning engineering}.
\newblock \bibinfo{journal}{\emph{arXiv preprint arXiv:2410.07095}} (\bibinfo{year}{2024}).
\newblock


\bibitem[Chang et~al\mbox{.}(2024)]%
        {chang2024partnr}
\bibfield{author}{\bibinfo{person}{Matthew Chang}, \bibinfo{person}{Gunjan Chhablani}, \bibinfo{person}{Alexander Clegg}, \bibinfo{person}{Mikael~Dallaire Cote}, \bibinfo{person}{Ruta Desai}, \bibinfo{person}{Michal Hlavac}, \bibinfo{person}{Vladimir Karashchuk}, \bibinfo{person}{Jacob Krantz}, \bibinfo{person}{Roozbeh Mottaghi}, \bibinfo{person}{Priyam Parashar}, {et~al\mbox{.}}} \bibinfo{year}{2024}\natexlab{}.
\newblock \showarticletitle{PARTNR: A Benchmark for Planning and Reasoning in Embodied Multi-agent Tasks}.
\newblock \bibinfo{journal}{\emph{arXiv preprint arXiv:2411.00081}} (\bibinfo{year}{2024}).
\newblock


\bibitem[Chen et~al\mbox{.}(2024d)]%
        {chen2024travelagent}
\bibfield{author}{\bibinfo{person}{Aili Chen}, \bibinfo{person}{Xuyang Ge}, \bibinfo{person}{Ziquan Fu}, \bibinfo{person}{Yanghua Xiao}, {and} \bibinfo{person}{Jiangjie Chen}.} \bibinfo{year}{2024}\natexlab{d}.
\newblock \showarticletitle{TravelAgent: An AI assistant for personalized travel planning}.
\newblock \bibinfo{journal}{\emph{arXiv preprint arXiv:2409.08069}} (\bibinfo{year}{2024}).
\newblock


\bibitem[Chen et~al\mbox{.}(2023a)]%
        {chen2023fireact}
\bibfield{author}{\bibinfo{person}{Baian Chen}, \bibinfo{person}{Chang Shu}, \bibinfo{person}{Ehsan Shareghi}, \bibinfo{person}{Nigel Collier}, \bibinfo{person}{Karthik Narasimhan}, {and} \bibinfo{person}{Shunyu Yao}.} \bibinfo{year}{2023}\natexlab{a}.
\newblock \showarticletitle{Fireact: Toward language agent fine-tuning}.
\newblock \bibinfo{journal}{\emph{arXiv preprint arXiv:2310.05915}} (\bibinfo{year}{2023}).
\newblock


\bibitem[Chen et~al\mbox{.}(2024a)]%
        {chen2024interleaved}
\bibfield{author}{\bibinfo{person}{Dongping Chen}, \bibinfo{person}{Ruoxi Chen}, \bibinfo{person}{Shu Pu}, \bibinfo{person}{Zhaoyi Liu}, \bibinfo{person}{Yanru Wu}, \bibinfo{person}{Caixi Chen}, \bibinfo{person}{Benlin Liu}, \bibinfo{person}{Yue Huang}, \bibinfo{person}{Yao Wan}, \bibinfo{person}{Pan Zhou}, {et~al\mbox{.}}} \bibinfo{year}{2024}\natexlab{a}.
\newblock \showarticletitle{Interleaved Scene Graph for Interleaved Text-and-Image Generation Assessment}.
\newblock \bibinfo{journal}{\emph{arXiv preprint arXiv:2411.17188}} (\bibinfo{year}{2024}).
\newblock


\bibitem[Chen et~al\mbox{.}(2024e)]%
        {chen2024alphamath}
\bibfield{author}{\bibinfo{person}{Guoxin Chen}, \bibinfo{person}{Minpeng Liao}, \bibinfo{person}{Chengxi Li}, {and} \bibinfo{person}{Kai Fan}.} \bibinfo{year}{2024}\natexlab{e}.
\newblock \showarticletitle{Alphamath almost zero: process supervision without process}.
\newblock \bibinfo{journal}{\emph{arXiv preprint arXiv:2405.03553}} (\bibinfo{year}{2024}).
\newblock


\bibitem[Chen et~al\mbox{.}(2023b)]%
        {chen2023put}
\bibfield{author}{\bibinfo{person}{Jiangjie Chen}, \bibinfo{person}{Siyu Yuan}, \bibinfo{person}{Rong Ye}, \bibinfo{person}{Bodhisattwa~Prasad Majumder}, {and} \bibinfo{person}{Kyle Richardson}.} \bibinfo{year}{2023}\natexlab{b}.
\newblock \showarticletitle{Put your money where your mouth is: Evaluating strategic planning and execution of llm agents in an auction arena}.
\newblock \bibinfo{journal}{\emph{arXiv preprint arXiv:2310.05746}} (\bibinfo{year}{2023}).
\newblock


\bibitem[Chen et~al\mbox{.}(2024h)]%
        {chen2024pca}
\bibfield{author}{\bibinfo{person}{Liang Chen}, \bibinfo{person}{Yichi Zhang}, \bibinfo{person}{Shuhuai Ren}, \bibinfo{person}{Haozhe Zhao}, \bibinfo{person}{Zefan Cai}, \bibinfo{person}{Yuchi Wang}, \bibinfo{person}{Peiyi Wang}, \bibinfo{person}{Xiangdi Meng}, \bibinfo{person}{Tianyu Liu}, {and} \bibinfo{person}{Baobao Chang}.} \bibinfo{year}{2024}\natexlab{h}.
\newblock \showarticletitle{Pca-bench: Evaluating multimodal large language models in perception-cognition-action chain}.
\newblock \bibinfo{journal}{\emph{arXiv preprint arXiv:2402.15527}} (\bibinfo{year}{2024}).
\newblock


\bibitem[Chen et~al\mbox{.}(2021)]%
        {chen2021evaluating}
\bibfield{author}{\bibinfo{person}{Mark Chen}, \bibinfo{person}{Jerry Tworek}, \bibinfo{person}{Heewoo Jun}, \bibinfo{person}{Qiming Yuan}, \bibinfo{person}{Henrique Ponde De~Oliveira Pinto}, \bibinfo{person}{Jared Kaplan}, \bibinfo{person}{Harri Edwards}, \bibinfo{person}{Yuri Burda}, \bibinfo{person}{Nicholas Joseph}, \bibinfo{person}{Greg Brockman}, {et~al\mbox{.}}} \bibinfo{year}{2021}\natexlab{}.
\newblock \showarticletitle{Evaluating large language models trained on code}.
\newblock \bibinfo{journal}{\emph{arXiv preprint arXiv:2107.03374}} (\bibinfo{year}{2021}).
\newblock


\bibitem[Chen et~al\mbox{.}(2024b)]%
        {chen2024guicourse}
\bibfield{author}{\bibinfo{person}{Wentong Chen}, \bibinfo{person}{Junbo Cui}, \bibinfo{person}{Jinyi Hu}, \bibinfo{person}{Yujia Qin}, \bibinfo{person}{Junjie Fang}, \bibinfo{person}{Yue Zhao}, \bibinfo{person}{Chongyi Wang}, \bibinfo{person}{Jun Liu}, \bibinfo{person}{Guirong Chen}, \bibinfo{person}{Yupeng Huo}, {et~al\mbox{.}}} \bibinfo{year}{2024}\natexlab{b}.
\newblock \showarticletitle{Guicourse: From general vision language models to versatile gui agents}.
\newblock \bibinfo{journal}{\emph{arXiv preprint arXiv:2406.11317}} (\bibinfo{year}{2024}).
\newblock


\bibitem[Chen et~al\mbox{.}(2025a)]%
        {chen2025scaling}
\bibfield{author}{\bibinfo{person}{Zhenfang Chen}, \bibinfo{person}{Delin Chen}, \bibinfo{person}{Rui Sun}, \bibinfo{person}{Wenjun Liu}, {and} \bibinfo{person}{Chuang Gan}.} \bibinfo{year}{2025}\natexlab{a}.
\newblock \showarticletitle{Scaling Autonomous Agents via Automatic Reward Modeling And Planning}.
\newblock \bibinfo{journal}{\emph{arXiv preprint arXiv:2502.12130}} (\bibinfo{year}{2025}).
\newblock


\bibitem[Chen et~al\mbox{.}(2024c)]%
        {chen2024self}
\bibfield{author}{\bibinfo{person}{Zixiang Chen}, \bibinfo{person}{Yihe Deng}, \bibinfo{person}{Huizhuo Yuan}, \bibinfo{person}{Kaixuan Ji}, {and} \bibinfo{person}{Quanquan Gu}.} \bibinfo{year}{2024}\natexlab{c}.
\newblock \showarticletitle{Self-play fine-tuning converts weak language models to strong language models}.
\newblock \bibinfo{journal}{\emph{arXiv preprint arXiv:2401.01335}} (\bibinfo{year}{2024}).
\newblock


\bibitem[Chen et~al\mbox{.}(2025b)]%
        {chen2025atlas}
\bibfield{author}{\bibinfo{person}{Zhixun Chen}, \bibinfo{person}{Ming Li}, \bibinfo{person}{Yuxuan Huang}, \bibinfo{person}{Yali Du}, \bibinfo{person}{Meng Fang}, {and} \bibinfo{person}{Tianyi Zhou}.} \bibinfo{year}{2025}\natexlab{b}.
\newblock \showarticletitle{Atlas: Agent tuning via learning critical steps}.
\newblock \bibinfo{journal}{\emph{arXiv preprint arXiv:2503.02197}} (\bibinfo{year}{2025}).
\newblock


\bibitem[Chen et~al\mbox{.}(2024f)]%
        {chen2024agent}
\bibfield{author}{\bibinfo{person}{Zehui Chen}, \bibinfo{person}{Kuikun Liu}, \bibinfo{person}{Qiuchen Wang}, \bibinfo{person}{Wenwei Zhang}, \bibinfo{person}{Jiangning Liu}, \bibinfo{person}{Dahua Lin}, \bibinfo{person}{Kai Chen}, {and} \bibinfo{person}{Feng Zhao}.} \bibinfo{year}{2024}\natexlab{f}.
\newblock \showarticletitle{Agent-FLAN: Designing Data and Methods of Effective Agent Tuning for Large Language Models}.
\newblock \bibinfo{journal}{\emph{arXiv preprint arXiv:2403.12881}} (\bibinfo{year}{2024}).
\newblock


\bibitem[Chen et~al\mbox{.}(2024g)]%
        {chen2024far}
\bibfield{author}{\bibinfo{person}{Zhe Chen}, \bibinfo{person}{Weiyun Wang}, \bibinfo{person}{Hao Tian}, \bibinfo{person}{Shenglong Ye}, \bibinfo{person}{Zhangwei Gao}, \bibinfo{person}{Erfei Cui}, \bibinfo{person}{Wenwen Tong}, \bibinfo{person}{Kongzhi Hu}, \bibinfo{person}{Jiapeng Luo}, \bibinfo{person}{Zheng Ma}, {et~al\mbox{.}}} \bibinfo{year}{2024}\natexlab{g}.
\newblock \showarticletitle{How Far Are We to GPT-4V? Closing the Gap to Commercial Multimodal Models with Open-Source Suites}.
\newblock \bibinfo{journal}{\emph{arXiv preprint arXiv:2404.16821}} (\bibinfo{year}{2024}).
\newblock


\bibitem[Cheng et~al\mbox{.}(2024b)]%
        {cheng2024seeclick}
\bibfield{author}{\bibinfo{person}{Kanzhi Cheng}, \bibinfo{person}{Qiushi Sun}, \bibinfo{person}{Yougang Chu}, \bibinfo{person}{Fangzhi Xu}, \bibinfo{person}{Yantao Li}, \bibinfo{person}{Jianbing Zhang}, {and} \bibinfo{person}{Zhiyong Wu}.} \bibinfo{year}{2024}\natexlab{b}.
\newblock \showarticletitle{Seeclick: Harnessing gui grounding for advanced visual gui agents}.
\newblock \bibinfo{journal}{\emph{arXiv preprint arXiv:2401.10935}} (\bibinfo{year}{2024}).
\newblock


\bibitem[Cheng et~al\mbox{.}(2024a)]%
        {cheng2024self}
\bibfield{author}{\bibinfo{person}{Pengyu Cheng}, \bibinfo{person}{Tianhao Hu}, \bibinfo{person}{Han Xu}, \bibinfo{person}{Zhisong Zhang}, \bibinfo{person}{Yong Dai}, \bibinfo{person}{Lei Han}, {and} \bibinfo{person}{Nan Du}.} \bibinfo{year}{2024}\natexlab{a}.
\newblock \showarticletitle{Self-playing Adversarial Language Game Enhances LLM Reasoning}.
\newblock \bibinfo{journal}{\emph{arXiv preprint arXiv:2404.10642}} (\bibinfo{year}{2024}).
\newblock


\bibitem[Chevalier-Boisvert et~al\mbox{.}(2018)]%
        {chevalier2018babyai}
\bibfield{author}{\bibinfo{person}{Maxime Chevalier-Boisvert}, \bibinfo{person}{Dzmitry Bahdanau}, \bibinfo{person}{Salem Lahlou}, \bibinfo{person}{Lucas Willems}, \bibinfo{person}{Chitwan Saharia}, \bibinfo{person}{Thien~Huu Nguyen}, {and} \bibinfo{person}{Yoshua Bengio}.} \bibinfo{year}{2018}\natexlab{}.
\newblock \showarticletitle{Babyai: A platform to study the sample efficiency of grounded language learning}.
\newblock \bibinfo{journal}{\emph{arXiv preprint arXiv:1810.08272}} (\bibinfo{year}{2018}).
\newblock


\bibitem[Chi et~al\mbox{.}(2024)]%
        {chi2024thoughtsculpt}
\bibfield{author}{\bibinfo{person}{Yizhou Chi}, \bibinfo{person}{Kevin Yang}, {and} \bibinfo{person}{Dan Klein}.} \bibinfo{year}{2024}\natexlab{}.
\newblock \showarticletitle{Thoughtsculpt: Reasoning with intermediate revision and search}.
\newblock \bibinfo{journal}{\emph{arXiv preprint arXiv:2404.05966}} (\bibinfo{year}{2024}).
\newblock


\bibitem[Choi et~al\mbox{.}(2024)]%
        {choi2024lota}
\bibfield{author}{\bibinfo{person}{Jae-Woo Choi}, \bibinfo{person}{Youngwoo Yoon}, \bibinfo{person}{Hyobin Ong}, \bibinfo{person}{Jaehong Kim}, {and} \bibinfo{person}{Minsu Jang}.} \bibinfo{year}{2024}\natexlab{}.
\newblock \showarticletitle{Lota-bench: Benchmarking language-oriented task planners for embodied agents}.
\newblock \bibinfo{journal}{\emph{arXiv preprint arXiv:2402.08178}} (\bibinfo{year}{2024}).
\newblock


\bibitem[Chu et~al\mbox{.}(2025)]%
        {chu2025llm+}
\bibfield{author}{\bibinfo{person}{Kun Chu}, \bibinfo{person}{Xufeng Zhao}, \bibinfo{person}{Cornelius Weber}, {and} \bibinfo{person}{Stefan Wermter}.} \bibinfo{year}{2025}\natexlab{}.
\newblock \showarticletitle{LLM+ MAP: Bimanual Robot Task Planning using Large Language Models and Planning Domain Definition Language}.
\newblock \bibinfo{journal}{\emph{arXiv preprint arXiv:2503.17309}} (\bibinfo{year}{2025}).
\newblock


\bibitem[Chung et~al\mbox{.}(2024)]%
        {chung2024scaling}
\bibfield{author}{\bibinfo{person}{Hyung~Won Chung}, \bibinfo{person}{Le Hou}, \bibinfo{person}{Shayne Longpre}, \bibinfo{person}{Barret Zoph}, \bibinfo{person}{Yi Tay}, \bibinfo{person}{William Fedus}, \bibinfo{person}{Yunxuan Li}, \bibinfo{person}{Xuezhi Wang}, \bibinfo{person}{Mostafa Dehghani}, \bibinfo{person}{Siddhartha Brahma}, {et~al\mbox{.}}} \bibinfo{year}{2024}\natexlab{}.
\newblock \showarticletitle{Scaling instruction-finetuned language models}.
\newblock \bibinfo{journal}{\emph{Journal of Machine Learning Research}} \bibinfo{volume}{25}, \bibinfo{number}{70} (\bibinfo{year}{2024}), \bibinfo{pages}{1--53}.
\newblock


\bibitem[Cobbe et~al\mbox{.}(2021)]%
        {cobbe2021training}
\bibfield{author}{\bibinfo{person}{Karl Cobbe}, \bibinfo{person}{Vineet Kosaraju}, \bibinfo{person}{Mohammad Bavarian}, \bibinfo{person}{Mark Chen}, \bibinfo{person}{Heewoo Jun}, \bibinfo{person}{Lukasz Kaiser}, \bibinfo{person}{Matthias Plappert}, \bibinfo{person}{Jerry Tworek}, \bibinfo{person}{Jacob Hilton}, \bibinfo{person}{Reiichiro Nakano}, {et~al\mbox{.}}} \bibinfo{year}{2021}\natexlab{}.
\newblock \showarticletitle{Training verifiers to solve math word problems}.
\newblock \bibinfo{journal}{\emph{arXiv preprint arXiv:2110.14168}} (\bibinfo{year}{2021}).
\newblock


\bibitem[Cornelio et~al\mbox{.}(2025)]%
        {cornelio2025hierarchical}
\bibfield{author}{\bibinfo{person}{Cristina Cornelio}, \bibinfo{person}{Flavio Petruzzellis}, {and} \bibinfo{person}{Pietro Lio}.} \bibinfo{year}{2025}\natexlab{}.
\newblock \showarticletitle{Hierarchical Planning for Complex Tasks with Knowledge Graph-RAG and Symbolic Verification}.
\newblock \bibinfo{journal}{\emph{arXiv preprint arXiv:2504.04578}} (\bibinfo{year}{2025}).
\newblock


\bibitem[C{\^o}t{\'e} et~al\mbox{.}(2019)]%
        {cote2019textworld}
\bibfield{author}{\bibinfo{person}{Marc-Alexandre C{\^o}t{\'e}}, \bibinfo{person}{Akos K{\'a}d{\'a}r}, \bibinfo{person}{Xingdi Yuan}, \bibinfo{person}{Ben Kybartas}, \bibinfo{person}{Tavian Barnes}, \bibinfo{person}{Emery Fine}, \bibinfo{person}{James Moore}, \bibinfo{person}{Matthew Hausknecht}, \bibinfo{person}{Layla El~Asri}, \bibinfo{person}{Mahmoud Adada}, {et~al\mbox{.}}} \bibinfo{year}{2019}\natexlab{}.
\newblock \showarticletitle{Textworld: A learning environment for text-based games}. In \bibinfo{booktitle}{\emph{Computer Games: 7th Workshop, CGW 2018, Held in Conjunction with the 27th International Conference on Artificial Intelligence, IJCAI 2018, Stockholm, Sweden, July 13, 2018, Revised Selected Papers 7}}. Springer, \bibinfo{pages}{41--75}.
\newblock


\bibitem[Coulom(2006)]%
        {coulom2006efficient}
\bibfield{author}{\bibinfo{person}{R{\'e}mi Coulom}.} \bibinfo{year}{2006}\natexlab{}.
\newblock \showarticletitle{Efficient selectivity and backup operators in Monte-Carlo tree search}. In \bibinfo{booktitle}{\emph{International conference on computers and games}}. Springer, \bibinfo{pages}{72--83}.
\newblock


\bibitem[Dagan et~al\mbox{.}(2023)]%
        {dagan2023dynamic}
\bibfield{author}{\bibinfo{person}{Gautier Dagan}, \bibinfo{person}{Frank Keller}, {and} \bibinfo{person}{Alex Lascarides}.} \bibinfo{year}{2023}\natexlab{}.
\newblock \showarticletitle{Dynamic planning with a llm}.
\newblock \bibinfo{journal}{\emph{arXiv preprint arXiv:2308.06391}} (\bibinfo{year}{2023}).
\newblock


\bibitem[Dagan et~al\mbox{.}(2024)]%
        {dagan2024plancraft}
\bibfield{author}{\bibinfo{person}{Gautier Dagan}, \bibinfo{person}{Frank Keller}, {and} \bibinfo{person}{Alex Lascarides}.} \bibinfo{year}{2024}\natexlab{}.
\newblock \showarticletitle{Plancraft: an evaluation dataset for planning with LLM agents}.
\newblock \bibinfo{journal}{\emph{arXiv preprint arXiv:2412.21033}} (\bibinfo{year}{2024}).
\newblock


\bibitem[Dao and Vu(2025)]%
        {dao2025alphamaze}
\bibfield{author}{\bibinfo{person}{Alan Dao} {and} \bibinfo{person}{Dinh~Bach Vu}.} \bibinfo{year}{2025}\natexlab{}.
\newblock \showarticletitle{AlphaMaze: Enhancing Large Language Models' Spatial Intelligence via GRPO}.
\newblock \bibinfo{journal}{\emph{arXiv preprint arXiv:2502.14669}} (\bibinfo{year}{2025}).
\newblock


\bibitem[de~la Rosa et~al\mbox{.}(2024)]%
        {de2024trip}
\bibfield{author}{\bibinfo{person}{Tomas de~la Rosa}, \bibinfo{person}{Sriram Gopalakrishnan}, \bibinfo{person}{Alberto Pozanco}, \bibinfo{person}{Zhen Zeng}, {and} \bibinfo{person}{Daniel Borrajo}.} \bibinfo{year}{2024}\natexlab{}.
\newblock \showarticletitle{TRIP-PAL: Travel Planning with Guarantees by Combining Large Language Models and Automated Planners}.
\newblock \bibinfo{journal}{\emph{arXiv preprint arXiv:2406.10196}} (\bibinfo{year}{2024}).
\newblock


\bibitem[De~Moura and Bj{\o}rner(2008)]%
        {de2008z3}
\bibfield{author}{\bibinfo{person}{Leonardo De~Moura} {and} \bibinfo{person}{Nikolaj Bj{\o}rner}.} \bibinfo{year}{2008}\natexlab{}.
\newblock \showarticletitle{Z3: An efficient SMT solver}. In \bibinfo{booktitle}{\emph{International conference on Tools and Algorithms for the Construction and Analysis of Systems}}. Springer, \bibinfo{pages}{337--340}.
\newblock


\bibitem[Deitke et~al\mbox{.}(2024)]%
        {deitke2024molmo}
\bibfield{author}{\bibinfo{person}{Matt Deitke}, \bibinfo{person}{Christopher Clark}, \bibinfo{person}{Sangho Lee}, \bibinfo{person}{Rohun Tripathi}, \bibinfo{person}{Yue Yang}, \bibinfo{person}{Jae~Sung Park}, \bibinfo{person}{Mohammadreza Salehi}, \bibinfo{person}{Niklas Muennighoff}, \bibinfo{person}{Kyle Lo}, \bibinfo{person}{Luca Soldaini}, {et~al\mbox{.}}} \bibinfo{year}{2024}\natexlab{}.
\newblock \showarticletitle{Molmo and pixmo: Open weights and open data for state-of-the-art multimodal models}.
\newblock \bibinfo{journal}{\emph{arXiv preprint arXiv:2409.17146}} (\bibinfo{year}{2024}).
\newblock


\bibitem[Deng et~al\mbox{.}(2024)]%
        {deng2024can}
\bibfield{author}{\bibinfo{person}{Hourui Deng}, \bibinfo{person}{Hongjie Zhang}, \bibinfo{person}{Jie Ou}, {and} \bibinfo{person}{Chaosheng Feng}.} \bibinfo{year}{2024}\natexlab{}.
\newblock \showarticletitle{Can llm be a good path planner based on prompt engineering? mitigating the hallucination for path planning}.
\newblock \bibinfo{journal}{\emph{arXiv preprint arXiv:2408.13184}} (\bibinfo{year}{2024}).
\newblock


\bibitem[Deng et~al\mbox{.}(2023)]%
        {deng2023mind2web}
\bibfield{author}{\bibinfo{person}{Xiang Deng}, \bibinfo{person}{Yu Gu}, \bibinfo{person}{Boyuan Zheng}, \bibinfo{person}{Shijie Chen}, \bibinfo{person}{Sam Stevens}, \bibinfo{person}{Boshi Wang}, \bibinfo{person}{Huan Sun}, {and} \bibinfo{person}{Yu Su}.} \bibinfo{year}{2023}\natexlab{}.
\newblock \showarticletitle{Mind2web: Towards a generalist agent for the web}.
\newblock \bibinfo{journal}{\emph{Advances in Neural Information Processing Systems}}  \bibinfo{volume}{36} (\bibinfo{year}{2023}), \bibinfo{pages}{28091--28114}.
\newblock


\bibitem[Ding et~al\mbox{.}(2024)]%
        {ding2024horizon}
\bibfield{author}{\bibinfo{person}{Yifeng Ding}, \bibinfo{person}{Hantian Ding}, \bibinfo{person}{Shiqi Wang}, \bibinfo{person}{Qing Sun}, \bibinfo{person}{Varun Kumar}, {and} \bibinfo{person}{Zijian Wang}.} \bibinfo{year}{2024}\natexlab{}.
\newblock \showarticletitle{Horizon-length prediction: Advancing fill-in-the-middle capabilities for code generation with lookahead planning}.
\newblock \bibinfo{journal}{\emph{arXiv preprint arXiv:2410.03103}} (\bibinfo{year}{2024}).
\newblock


\bibitem[Dinh et~al\mbox{.}(2024)]%
        {dinh2024reasonplanner}
\bibfield{author}{\bibinfo{person}{Minh~Pham Dinh}, \bibinfo{person}{Munira Syed}, \bibinfo{person}{Michael~G Yankoski}, {and} \bibinfo{person}{Trenton~W Ford}.} \bibinfo{year}{2024}\natexlab{}.
\newblock \showarticletitle{ReasonPlanner: Enhancing Autonomous Planning in Dynamic Environments with Temporal Knowledge Graphs and LLMs}.
\newblock \bibinfo{journal}{\emph{arXiv preprint arXiv:2410.09252}} (\bibinfo{year}{2024}).
\newblock


\bibitem[Drouin et~al\mbox{.}(2024)]%
        {drouin2024workarena}
\bibfield{author}{\bibinfo{person}{Alexandre Drouin}, \bibinfo{person}{Maxime Gasse}, \bibinfo{person}{Massimo Caccia}, \bibinfo{person}{Issam~H Laradji}, \bibinfo{person}{Manuel Del~Verme}, \bibinfo{person}{Tom Marty}, \bibinfo{person}{L{\'e}o Boisvert}, \bibinfo{person}{Megh Thakkar}, \bibinfo{person}{Quentin Cappart}, \bibinfo{person}{David Vazquez}, {et~al\mbox{.}}} \bibinfo{year}{2024}\natexlab{}.
\newblock \showarticletitle{Workarena: How capable are web agents at solving common knowledge work tasks?}
\newblock \bibinfo{journal}{\emph{arXiv preprint arXiv:2403.07718}} (\bibinfo{year}{2024}).
\newblock


\bibitem[Du et~al\mbox{.}(2025)]%
        {du2025boost}
\bibfield{author}{\bibinfo{person}{Kounianhua Du}, \bibinfo{person}{Hanjing Wang}, \bibinfo{person}{Jianxing Liu}, \bibinfo{person}{Jizheng Chen}, \bibinfo{person}{Xinyi Dai}, \bibinfo{person}{Yasheng Wang}, \bibinfo{person}{Ruiming Tang}, \bibinfo{person}{Yong Yu}, \bibinfo{person}{Jun Wang}, {and} \bibinfo{person}{Weinan Zhang}.} \bibinfo{year}{2025}\natexlab{}.
\newblock \showarticletitle{Boost, disentangle, and customize: A robust system2-to-system1 pipeline for code generation}.
\newblock \bibinfo{journal}{\emph{arXiv preprint arXiv:2502.12492}} (\bibinfo{year}{2025}).
\newblock


\bibitem[Ernhofer et~al\mbox{.}(2025)]%
        {ernhofer2025leveraging}
\bibfield{author}{\bibinfo{person}{Benjamin~Raphael Ernhofer}, \bibinfo{person}{Daniil Prokhorov}, \bibinfo{person}{Jannica Langner}, {and} \bibinfo{person}{Dominik Bollmann}.} \bibinfo{year}{2025}\natexlab{}.
\newblock \showarticletitle{Leveraging Vision-Language Models for Visual Grounding and Analysis of Automotive UI}.
\newblock \bibinfo{journal}{\emph{arXiv preprint arXiv:2505.05895}} (\bibinfo{year}{2025}).
\newblock


\bibitem[Fan et~al\mbox{.}(2022)]%
        {fan2022minedojo}
\bibfield{author}{\bibinfo{person}{Linxi Fan}, \bibinfo{person}{Guanzhi Wang}, \bibinfo{person}{Yunfan Jiang}, \bibinfo{person}{Ajay Mandlekar}, \bibinfo{person}{Yuncong Yang}, \bibinfo{person}{Haoyi Zhu}, \bibinfo{person}{Andrew Tang}, \bibinfo{person}{De-An Huang}, \bibinfo{person}{Yuke Zhu}, {and} \bibinfo{person}{Anima Anandkumar}.} \bibinfo{year}{2022}\natexlab{}.
\newblock \showarticletitle{Minedojo: Building open-ended embodied agents with internet-scale knowledge}.
\newblock \bibinfo{journal}{\emph{Advances in Neural Information Processing Systems}}  \bibinfo{volume}{35} (\bibinfo{year}{2022}), \bibinfo{pages}{18343--18362}.
\newblock


\bibitem[Fang et~al\mbox{.}(2025)]%
        {fang2025synworld}
\bibfield{author}{\bibinfo{person}{Runnan Fang}, \bibinfo{person}{Xiaobin Wang}, \bibinfo{person}{Yuan Liang}, \bibinfo{person}{Shuofei Qiao}, \bibinfo{person}{Jialong Wu}, \bibinfo{person}{Zekun Xi}, \bibinfo{person}{Ningyu Zhang}, \bibinfo{person}{Yong Jiang}, \bibinfo{person}{Pengjun Xie}, \bibinfo{person}{Fei Huang}, {et~al\mbox{.}}} \bibinfo{year}{2025}\natexlab{}.
\newblock \showarticletitle{SynWorld: Virtual Scenario Synthesis for Agentic Action Knowledge Refinement}.
\newblock \bibinfo{journal}{\emph{arXiv preprint arXiv:2504.03561}} (\bibinfo{year}{2025}).
\newblock


\bibitem[Farn and Shin(2023)]%
        {farn2023tooltalk}
\bibfield{author}{\bibinfo{person}{Nicholas Farn} {and} \bibinfo{person}{Richard Shin}.} \bibinfo{year}{2023}\natexlab{}.
\newblock \showarticletitle{Tooltalk: Evaluating tool-usage in a conversational setting}.
\newblock \bibinfo{journal}{\emph{arXiv preprint arXiv:2311.10775}} (\bibinfo{year}{2023}).
\newblock


\bibitem[Feng et~al\mbox{.}(2025b)]%
        {feng2025retool}
\bibfield{author}{\bibinfo{person}{Jiazhan Feng}, \bibinfo{person}{Shijue Huang}, \bibinfo{person}{Xingwei Qu}, \bibinfo{person}{Ge Zhang}, \bibinfo{person}{Yujia Qin}, \bibinfo{person}{Baoquan Zhong}, \bibinfo{person}{Chengquan Jiang}, \bibinfo{person}{Jinxin Chi}, {and} \bibinfo{person}{Wanjun Zhong}.} \bibinfo{year}{2025}\natexlab{b}.
\newblock \showarticletitle{ReTool: Reinforcement Learning for Strategic Tool Use in LLMs}.
\newblock \bibinfo{journal}{\emph{arXiv preprint arXiv:2504.11536}} (\bibinfo{year}{2025}).
\newblock


\bibitem[Feng et~al\mbox{.}(2025c)]%
        {feng2025group}
\bibfield{author}{\bibinfo{person}{Lang Feng}, \bibinfo{person}{Zhenghai Xue}, \bibinfo{person}{Tingcong Liu}, {and} \bibinfo{person}{Bo An}.} \bibinfo{year}{2025}\natexlab{c}.
\newblock \showarticletitle{Group-in-Group Policy Optimization for LLM Agent Training}.
\newblock \bibinfo{journal}{\emph{arXiv preprint arXiv:2505.10978}} (\bibinfo{year}{2025}).
\newblock


\bibitem[Feng et~al\mbox{.}(2025a)]%
        {feng2025reflective}
\bibfield{author}{\bibinfo{person}{Yunhai Feng}, \bibinfo{person}{Jiaming Han}, \bibinfo{person}{Zhuoran Yang}, \bibinfo{person}{Xiangyu Yue}, \bibinfo{person}{Sergey Levine}, {and} \bibinfo{person}{Jianlan Luo}.} \bibinfo{year}{2025}\natexlab{a}.
\newblock \showarticletitle{Reflective Planning: Vision-Language Models for Multi-Stage Long-Horizon Robotic Manipulation}.
\newblock \bibinfo{journal}{\emph{arXiv preprint arXiv:2502.16707}} (\bibinfo{year}{2025}).
\newblock


\bibitem[Fu et~al\mbox{.}(2025)]%
        {fu2025agentrefine}
\bibfield{author}{\bibinfo{person}{Dayuan Fu}, \bibinfo{person}{Keqing He}, \bibinfo{person}{Yejie Wang}, \bibinfo{person}{Wentao Hong}, \bibinfo{person}{Zhuoma Gongque}, \bibinfo{person}{Weihao Zeng}, \bibinfo{person}{Wei Wang}, \bibinfo{person}{Jingang Wang}, \bibinfo{person}{Xunliang Cai}, {and} \bibinfo{person}{Weiran Xu}.} \bibinfo{year}{2025}\natexlab{}.
\newblock \showarticletitle{AgentRefine: Enhancing Agent Generalization through Refinement Tuning}.
\newblock \bibinfo{journal}{\emph{arXiv preprint arXiv:2501.01702}} (\bibinfo{year}{2025}).
\newblock


\bibitem[Fu et~al\mbox{.}(2024b)]%
        {fu2024msi}
\bibfield{author}{\bibinfo{person}{Dayuan Fu}, \bibinfo{person}{Biqing Qi}, \bibinfo{person}{Yihuai Gao}, \bibinfo{person}{Che Jiang}, \bibinfo{person}{Guanting Dong}, {and} \bibinfo{person}{Bowen Zhou}.} \bibinfo{year}{2024}\natexlab{b}.
\newblock \showarticletitle{MSI-Agent: Incorporating Multi-Scale Insight into Embodied Agents for Superior Planning and Decision-Making}.
\newblock \bibinfo{journal}{\emph{arXiv preprint arXiv:2409.16686}} (\bibinfo{year}{2024}).
\newblock


\bibitem[Fu et~al\mbox{.}(2024a)]%
        {fu2024camphor}
\bibfield{author}{\bibinfo{person}{Yicheng Fu}, \bibinfo{person}{Raviteja Anantha}, {and} \bibinfo{person}{Jianpeng Cheng}.} \bibinfo{year}{2024}\natexlab{a}.
\newblock \showarticletitle{CAMPHOR: Collaborative Agents for Multi-input Planning and High-Order Reasoning On Device}.
\newblock \bibinfo{journal}{\emph{arXiv preprint arXiv:2410.09407}} (\bibinfo{year}{2024}).
\newblock


\bibitem[Furuta et~al\mbox{.}({[n.\,d.]})]%
        {furutamultimodal}
\bibfield{author}{\bibinfo{person}{Hiroki Furuta}, \bibinfo{person}{Kuang-Huei Lee}, \bibinfo{person}{Ofir Nachum}, \bibinfo{person}{Yutaka Matsuo}, \bibinfo{person}{Aleksandra Faust}, \bibinfo{person}{Shixiang~Shane Gu}, {and} \bibinfo{person}{Izzeddin Gur}.} \bibinfo{year}{[n.\,d.]}\natexlab{}.
\newblock \showarticletitle{Multimodal Web Navigation with Instruction-Finetuned Foundation Models}. In \bibinfo{booktitle}{\emph{The Twelfth International Conference on Learning Representations}}.
\newblock


\bibitem[Gandhi et~al\mbox{.}(2024)]%
        {gandhi2024stream}
\bibfield{author}{\bibinfo{person}{Kanishk Gandhi}, \bibinfo{person}{Denise Lee}, \bibinfo{person}{Gabriel Grand}, \bibinfo{person}{Muxin Liu}, \bibinfo{person}{Winson Cheng}, \bibinfo{person}{Archit Sharma}, {and} \bibinfo{person}{Noah~D Goodman}.} \bibinfo{year}{2024}\natexlab{}.
\newblock \showarticletitle{Stream of Search (SoS): Learning to Search in Language}.
\newblock \bibinfo{journal}{\emph{arXiv preprint arXiv:2404.03683}} (\bibinfo{year}{2024}).
\newblock


\bibitem[Gao et~al\mbox{.}(2025a)]%
        {gao2025uishift}
\bibfield{author}{\bibinfo{person}{Longxi Gao}, \bibinfo{person}{Li Zhang}, {and} \bibinfo{person}{Mengwei Xu}.} \bibinfo{year}{2025}\natexlab{a}.
\newblock \showarticletitle{UIShift: Enhancing VLM-based GUI Agents through Self-supervised Reinforcement Learning}.
\newblock \bibinfo{journal}{\emph{arXiv preprint arXiv:2505.12493}} (\bibinfo{year}{2025}).
\newblock


\bibitem[Gao et~al\mbox{.}(2025b)]%
        {gao2025efficient}
\bibfield{author}{\bibinfo{person}{Pengyu Gao}, \bibinfo{person}{Jinming Zhao}, \bibinfo{person}{Xinyue Chen}, {and} \bibinfo{person}{Long Yilin}.} \bibinfo{year}{2025}\natexlab{b}.
\newblock \showarticletitle{An Efficient Context-Dependent Memory Framework for LLM-Centric Agents}. In \bibinfo{booktitle}{\emph{Proceedings of the 2025 Conference of the Nations of the Americas Chapter of the Association for Computational Linguistics: Human Language Technologies (Volume 3: Industry Track)}}. \bibinfo{pages}{1055--1069}.
\newblock


\bibitem[Georgievski and Aiello(2015)]%
        {georgievski2015htn}
\bibfield{author}{\bibinfo{person}{Ilche Georgievski} {and} \bibinfo{person}{Marco Aiello}.} \bibinfo{year}{2015}\natexlab{}.
\newblock \showarticletitle{HTN planning: Overview, comparison, and beyond}.
\newblock \bibinfo{journal}{\emph{Artificial Intelligence}}  \bibinfo{volume}{222} (\bibinfo{year}{2015}), \bibinfo{pages}{124--156}.
\newblock


\bibitem[Ghallab et~al\mbox{.}(2004)]%
        {ghallab2004automated}
\bibfield{author}{\bibinfo{person}{Malik Ghallab}, \bibinfo{person}{Dana Nau}, {and} \bibinfo{person}{Paolo Traverso}.} \bibinfo{year}{2004}\natexlab{}.
\newblock \bibinfo{booktitle}{\emph{Automated Planning: theory and practice}}.
\newblock \bibinfo{publisher}{Elsevier}.
\newblock


\bibitem[Gong et~al\mbox{.}(2023)]%
        {gong2023mindagent}
\bibfield{author}{\bibinfo{person}{Ran Gong}, \bibinfo{person}{Qiuyuan Huang}, \bibinfo{person}{Xiaojian Ma}, \bibinfo{person}{Hoi Vo}, \bibinfo{person}{Zane Durante}, \bibinfo{person}{Yusuke Noda}, \bibinfo{person}{Zilong Zheng}, \bibinfo{person}{Song-Chun Zhu}, \bibinfo{person}{Demetri Terzopoulos}, \bibinfo{person}{Li Fei-Fei}, {et~al\mbox{.}}} \bibinfo{year}{2023}\natexlab{}.
\newblock \showarticletitle{Mindagent: Emergent gaming interaction}.
\newblock \bibinfo{journal}{\emph{arXiv preprint arXiv:2309.09971}} (\bibinfo{year}{2023}).
\newblock


\bibitem[Gou et~al\mbox{.}(2024)]%
        {gou2024navigating}
\bibfield{author}{\bibinfo{person}{Boyu Gou}, \bibinfo{person}{Ruohan Wang}, \bibinfo{person}{Boyuan Zheng}, \bibinfo{person}{Yanan Xie}, \bibinfo{person}{Cheng Chang}, \bibinfo{person}{Yiheng Shu}, \bibinfo{person}{Huan Sun}, {and} \bibinfo{person}{Yu Su}.} \bibinfo{year}{2024}\natexlab{}.
\newblock \showarticletitle{Navigating the digital world as humans do: Universal visual grounding for gui agents}.
\newblock \bibinfo{journal}{\emph{arXiv preprint arXiv:2410.05243}} (\bibinfo{year}{2024}).
\newblock


\bibitem[Grattafiori et~al\mbox{.}(2024)]%
        {grattafiori2024llama}
\bibfield{author}{\bibinfo{person}{Aaron Grattafiori}, \bibinfo{person}{Abhimanyu Dubey}, \bibinfo{person}{Abhinav Jauhri}, \bibinfo{person}{Abhinav Pandey}, \bibinfo{person}{Abhishek Kadian}, \bibinfo{person}{Ahmad Al-Dahle}, \bibinfo{person}{Aiesha Letman}, \bibinfo{person}{Akhil Mathur}, \bibinfo{person}{Alan Schelten}, \bibinfo{person}{Alex Vaughan}, {et~al\mbox{.}}} \bibinfo{year}{2024}\natexlab{}.
\newblock \showarticletitle{The llama 3 herd of models}.
\newblock \bibinfo{journal}{\emph{arXiv preprint arXiv:2407.21783}} (\bibinfo{year}{2024}).
\newblock


\bibitem[Graves(2012)]%
        {graves2012sequence}
\bibfield{author}{\bibinfo{person}{Alex Graves}.} \bibinfo{year}{2012}\natexlab{}.
\newblock \showarticletitle{Sequence transduction with recurrent neural networks}.
\newblock \bibinfo{journal}{\emph{arXiv preprint arXiv:1211.3711}} (\bibinfo{year}{2012}).
\newblock


\bibitem[Graves(2014)]%
        {graves2014neural}
\bibfield{author}{\bibinfo{person}{Alex Graves}.} \bibinfo{year}{2014}\natexlab{}.
\newblock \showarticletitle{Neural Turing Machines}.
\newblock \bibinfo{journal}{\emph{arXiv preprint arXiv:1410.5401}} (\bibinfo{year}{2014}).
\newblock


\bibitem[Gu et~al\mbox{.}(2024)]%
        {gu2024your}
\bibfield{author}{\bibinfo{person}{Yu Gu}, \bibinfo{person}{Boyuan Zheng}, \bibinfo{person}{Boyu Gou}, \bibinfo{person}{Kai Zhang}, \bibinfo{person}{Cheng Chang}, \bibinfo{person}{Sanjari Srivastava}, \bibinfo{person}{Yanan Xie}, \bibinfo{person}{Peng Qi}, \bibinfo{person}{Huan Sun}, {and} \bibinfo{person}{Yu Su}.} \bibinfo{year}{2024}\natexlab{}.
\newblock \showarticletitle{Is your llm secretly a world model of the internet? model-based planning for web agents}.
\newblock \bibinfo{journal}{\emph{arXiv preprint arXiv:2411.06559}} (\bibinfo{year}{2024}).
\newblock


\bibitem[Guan et~al\mbox{.}(2023)]%
        {guan2023leveraging}
\bibfield{author}{\bibinfo{person}{Lin Guan}, \bibinfo{person}{Karthik Valmeekam}, \bibinfo{person}{Sarath Sreedharan}, {and} \bibinfo{person}{Subbarao Kambhampati}.} \bibinfo{year}{2023}\natexlab{}.
\newblock \showarticletitle{Leveraging pre-trained large language models to construct and utilize world models for model-based task planning}.
\newblock \bibinfo{journal}{\emph{Advances in Neural Information Processing Systems}}  \bibinfo{volume}{36} (\bibinfo{year}{2023}), \bibinfo{pages}{79081--79094}.
\newblock


\bibitem[Gundawar et~al\mbox{.}(2024)]%
        {gundawar2024robust}
\bibfield{author}{\bibinfo{person}{Atharva Gundawar}, \bibinfo{person}{Karthik Valmeekam}, \bibinfo{person}{Mudit Verma}, {and} \bibinfo{person}{Subbarao Kambhampati}.} \bibinfo{year}{2024}\natexlab{}.
\newblock \showarticletitle{Robust Planning with Compound LLM Architectures: An LLM-Modulo Approach}.
\newblock \bibinfo{journal}{\emph{arXiv preprint arXiv:2411.14484}} (\bibinfo{year}{2024}).
\newblock


\bibitem[Guo et~al\mbox{.}(2025)]%
        {guo2025deepseek}
\bibfield{author}{\bibinfo{person}{Daya Guo}, \bibinfo{person}{Dejian Yang}, \bibinfo{person}{Haowei Zhang}, \bibinfo{person}{Junxiao Song}, \bibinfo{person}{Ruoyu Zhang}, \bibinfo{person}{Runxin Xu}, \bibinfo{person}{Qihao Zhu}, \bibinfo{person}{Shirong Ma}, \bibinfo{person}{Peiyi Wang}, \bibinfo{person}{Xiao Bi}, {et~al\mbox{.}}} \bibinfo{year}{2025}\natexlab{}.
\newblock \showarticletitle{Deepseek-r1: Incentivizing reasoning capability in llms via reinforcement learning}.
\newblock \bibinfo{journal}{\emph{arXiv preprint arXiv:2501.12948}} (\bibinfo{year}{2025}).
\newblock


\bibitem[Guo et~al\mbox{.}(2024)]%
        {guo2024open}
\bibfield{author}{\bibinfo{person}{Shiguang Guo}, \bibinfo{person}{Ziliang Deng}, \bibinfo{person}{Hongyu Lin}, \bibinfo{person}{Yaojie Lu}, \bibinfo{person}{Xianpei Han}, {and} \bibinfo{person}{Le Sun}.} \bibinfo{year}{2024}\natexlab{}.
\newblock \showarticletitle{Open Grounded Planning: Challenges and Benchmark Construction}. In \bibinfo{booktitle}{\emph{Proceedings of the 62nd Annual Meeting of the Association for Computational Linguistics (Volume 1: Long Papers)}}. \bibinfo{pages}{4982--5003}.
\newblock


\bibitem[Gur et~al\mbox{.}(2023)]%
        {gur2023real}
\bibfield{author}{\bibinfo{person}{Izzeddin Gur}, \bibinfo{person}{Hiroki Furuta}, \bibinfo{person}{Austin Huang}, \bibinfo{person}{Mustafa Safdari}, \bibinfo{person}{Yutaka Matsuo}, \bibinfo{person}{Douglas Eck}, {and} \bibinfo{person}{Aleksandra Faust}.} \bibinfo{year}{2023}\natexlab{}.
\newblock \showarticletitle{A real-world webagent with planning, long context understanding, and program synthesis}.
\newblock \bibinfo{journal}{\emph{arXiv preprint arXiv:2307.12856}} (\bibinfo{year}{2023}).
\newblock


\bibitem[Ha and Schmidhuber(2018)]%
        {ha2018world}
\bibfield{author}{\bibinfo{person}{David Ha} {and} \bibinfo{person}{J{\"u}rgen Schmidhuber}.} \bibinfo{year}{2018}\natexlab{}.
\newblock \showarticletitle{World models}.
\newblock \bibinfo{journal}{\emph{arXiv preprint arXiv:1803.10122}} (\bibinfo{year}{2018}).
\newblock


\bibitem[Haluptzok et~al\mbox{.}(2022)]%
        {haluptzok2022language}
\bibfield{author}{\bibinfo{person}{Patrick Haluptzok}, \bibinfo{person}{Matthew Bowers}, {and} \bibinfo{person}{Adam~Tauman Kalai}.} \bibinfo{year}{2022}\natexlab{}.
\newblock \showarticletitle{Language models can teach themselves to program better}.
\newblock \bibinfo{journal}{\emph{arXiv preprint arXiv:2207.14502}} (\bibinfo{year}{2022}).
\newblock


\bibitem[Hao et~al\mbox{.}(2023)]%
        {hao2023reasoning}
\bibfield{author}{\bibinfo{person}{Shibo Hao}, \bibinfo{person}{Yi Gu}, \bibinfo{person}{Haodi Ma}, \bibinfo{person}{Joshua~Jiahua Hong}, \bibinfo{person}{Zhen Wang}, \bibinfo{person}{Daisy~Zhe Wang}, {and} \bibinfo{person}{Zhiting Hu}.} \bibinfo{year}{2023}\natexlab{}.
\newblock \showarticletitle{Reasoning with language model is planning with world model}.
\newblock \bibinfo{journal}{\emph{arXiv preprint arXiv:2305.14992}} (\bibinfo{year}{2023}).
\newblock


\bibitem[Hao et~al\mbox{.}(2024)]%
        {hao2024large}
\bibfield{author}{\bibinfo{person}{Yilun Hao}, \bibinfo{person}{Yongchao Chen}, \bibinfo{person}{Yang Zhang}, {and} \bibinfo{person}{Chuchu Fan}.} \bibinfo{year}{2024}\natexlab{}.
\newblock \showarticletitle{Large Language Models Can Plan Your Travels Rigorously with Formal Verification Tools}.
\newblock \bibinfo{journal}{\emph{arXiv preprint arXiv:2404.11891}} (\bibinfo{year}{2024}).
\newblock


\bibitem[Hart et~al\mbox{.}(1968)]%
        {hart1968formal}
\bibfield{author}{\bibinfo{person}{Peter~E Hart}, \bibinfo{person}{Nils~J Nilsson}, {and} \bibinfo{person}{Bertram Raphael}.} \bibinfo{year}{1968}\natexlab{}.
\newblock \showarticletitle{A formal basis for the heuristic determination of minimum cost paths}.
\newblock \bibinfo{journal}{\emph{IEEE transactions on Systems Science and Cybernetics}} \bibinfo{volume}{4}, \bibinfo{number}{2} (\bibinfo{year}{1968}), \bibinfo{pages}{100--107}.
\newblock


\bibitem[Haslum et~al\mbox{.}(2019)]%
        {haslum2019introduction}
\bibfield{author}{\bibinfo{person}{Patrik Haslum}, \bibinfo{person}{Nir Lipovetzky}, \bibinfo{person}{Daniele Magazzeni}, \bibinfo{person}{Christian Muise}, \bibinfo{person}{Ronald Brachman}, \bibinfo{person}{Francesca Rossi}, {and} \bibinfo{person}{Peter Stone}.} \bibinfo{year}{2019}\natexlab{}.
\newblock \bibinfo{booktitle}{\emph{An introduction to the planning domain definition language}}. Vol.~\bibinfo{volume}{13}.
\newblock \bibinfo{publisher}{Springer}.
\newblock


\bibitem[He et~al\mbox{.}(2024)]%
        {he2024webvoyager}
\bibfield{author}{\bibinfo{person}{Hongliang He}, \bibinfo{person}{Wenlin Yao}, \bibinfo{person}{Kaixin Ma}, \bibinfo{person}{Wenhao Yu}, \bibinfo{person}{Yong Dai}, \bibinfo{person}{Hongming Zhang}, \bibinfo{person}{Zhenzhong Lan}, {and} \bibinfo{person}{Dong Yu}.} \bibinfo{year}{2024}\natexlab{}.
\newblock \showarticletitle{WebVoyager: Building an end-to-end web agent with large multimodal models}.
\newblock \bibinfo{journal}{\emph{arXiv preprint arXiv:2401.13919}} (\bibinfo{year}{2024}).
\newblock


\bibitem[Hill et~al\mbox{.}(2023)]%
        {hill2023mineplanner}
\bibfield{author}{\bibinfo{person}{William Hill}, \bibinfo{person}{Ireton Liu}, \bibinfo{person}{Anita De~Mello Koch}, \bibinfo{person}{Damion Harvey}, \bibinfo{person}{Nishanth Kumar}, \bibinfo{person}{George Konidaris}, {and} \bibinfo{person}{Steven James}.} \bibinfo{year}{2023}\natexlab{}.
\newblock \showarticletitle{MinePlanner: A Benchmark for Long-Horizon Planning in Large Minecraft Worlds}.
\newblock \bibinfo{journal}{\emph{arXiv preprint arXiv:2312.12891}} (\bibinfo{year}{2023}).
\newblock


\bibitem[Ho et~al\mbox{.}(2025)]%
        {ho2025verilogcoder}
\bibfield{author}{\bibinfo{person}{Chia-Tung Ho}, \bibinfo{person}{Haoxing Ren}, {and} \bibinfo{person}{Brucek Khailany}.} \bibinfo{year}{2025}\natexlab{}.
\newblock \showarticletitle{Verilogcoder: Autonomous verilog coding agents with graph-based planning and abstract syntax tree (ast)-based waveform tracing tool}. In \bibinfo{booktitle}{\emph{Proceedings of the AAAI Conference on Artificial Intelligence}}, Vol.~\bibinfo{volume}{39}. \bibinfo{pages}{300--307}.
\newblock


\bibitem[Hong et~al\mbox{.}(2024)]%
        {hong2024cogagent}
\bibfield{author}{\bibinfo{person}{Wenyi Hong}, \bibinfo{person}{Weihan Wang}, \bibinfo{person}{Qingsong Lv}, \bibinfo{person}{Jiazheng Xu}, \bibinfo{person}{Wenmeng Yu}, \bibinfo{person}{Junhui Ji}, \bibinfo{person}{Yan Wang}, \bibinfo{person}{Zihan Wang}, \bibinfo{person}{Yuxiao Dong}, \bibinfo{person}{Ming Ding}, {et~al\mbox{.}}} \bibinfo{year}{2024}\natexlab{}.
\newblock \showarticletitle{Cogagent: A visual language model for gui agents}. In \bibinfo{booktitle}{\emph{Proceedings of the IEEE/CVF Conference on Computer Vision and Pattern Recognition}}. \bibinfo{pages}{14281--14290}.
\newblock


\bibitem[Hu et~al\mbox{.}({[n.\,d.]})]%
        {hutree}
\bibfield{author}{\bibinfo{person}{Mengkang Hu}, \bibinfo{person}{Yao Mu}, \bibinfo{person}{Xinmiao~Chelsey Yu}, \bibinfo{person}{Mingyu Ding}, \bibinfo{person}{Shiguang Wu}, \bibinfo{person}{Wenqi Shao}, \bibinfo{person}{Qiguang Chen}, \bibinfo{person}{Bin Wang}, \bibinfo{person}{Yu Qiao}, {and} \bibinfo{person}{Ping Luo}.} \bibinfo{year}{[n.\,d.]}\natexlab{}.
\newblock \showarticletitle{Tree-Planner: Efficient Close-loop Task Planning with Large Language Models}. In \bibinfo{booktitle}{\emph{The Twelfth International Conference on Learning Representations}}.
\newblock


\bibitem[Hu et~al\mbox{.}(2024)]%
        {hu2024agentgen}
\bibfield{author}{\bibinfo{person}{Mengkang Hu}, \bibinfo{person}{Pu Zhao}, \bibinfo{person}{Can Xu}, \bibinfo{person}{Qingfeng Sun}, \bibinfo{person}{Jianguang Lou}, \bibinfo{person}{Qingwei Lin}, \bibinfo{person}{Ping Luo}, \bibinfo{person}{Saravan Rajmohan}, {and} \bibinfo{person}{Dongmei Zhang}.} \bibinfo{year}{2024}\natexlab{}.
\newblock \showarticletitle{Agentgen: Enhancing planning abilities for large language model based agent via environment and task generation}.
\newblock \bibinfo{journal}{\emph{arXiv preprint arXiv:2408.00764}} (\bibinfo{year}{2024}).
\newblock


\bibitem[Hu et~al\mbox{.}(2023b)]%
        {hu2023tree}
\bibfield{author}{\bibinfo{person}{Pengbo Hu}, \bibinfo{person}{Ji Qi}, \bibinfo{person}{Xingyu Li}, \bibinfo{person}{Hong Li}, \bibinfo{person}{Xinqi Wang}, \bibinfo{person}{Bing Quan}, \bibinfo{person}{Ruiyu Wang}, {and} \bibinfo{person}{Yi Zhou}.} \bibinfo{year}{2023}\natexlab{b}.
\newblock \showarticletitle{Tree-of-mixed-thought: Combining fast and slow thinking for multi-hop visual reasoning}.
\newblock \bibinfo{journal}{\emph{arXiv preprint arXiv:2308.09658}} (\bibinfo{year}{2023}).
\newblock


\bibitem[Hu et~al\mbox{.}(2023a)]%
        {hu2023avis}
\bibfield{author}{\bibinfo{person}{Ziniu Hu}, \bibinfo{person}{Ahmet Iscen}, \bibinfo{person}{Chen Sun}, \bibinfo{person}{Kai-Wei Chang}, \bibinfo{person}{Yizhou Sun}, \bibinfo{person}{David Ross}, \bibinfo{person}{Cordelia Schmid}, {and} \bibinfo{person}{Alireza Fathi}.} \bibinfo{year}{2023}\natexlab{a}.
\newblock \showarticletitle{Avis: Autonomous visual information seeking with large language model agent}.
\newblock \bibinfo{journal}{\emph{Advances in Neural Information Processing Systems}}  \bibinfo{volume}{36} (\bibinfo{year}{2023}), \bibinfo{pages}{867--878}.
\newblock


\bibitem[Huang et~al\mbox{.}(2025b)]%
        {huang2025scaletrack}
\bibfield{author}{\bibinfo{person}{Jing Huang}, \bibinfo{person}{Zhixiong Zeng}, \bibinfo{person}{Wenkang Han}, \bibinfo{person}{Yufeng Zhong}, \bibinfo{person}{Liming Zheng}, \bibinfo{person}{Shuai Fu}, \bibinfo{person}{Jingyuan Chen}, {and} \bibinfo{person}{Lin Ma}.} \bibinfo{year}{2025}\natexlab{b}.
\newblock \showarticletitle{ScaleTrack: Scaling and back-tracking Automated GUI Agents}.
\newblock \bibinfo{journal}{\emph{arXiv preprint arXiv:2505.00416}} (\bibinfo{year}{2025}).
\newblock


\bibitem[Huang et~al\mbox{.}(2024b)]%
        {huang2024grounded}
\bibfield{author}{\bibinfo{person}{Wenlong Huang}, \bibinfo{person}{Fei Xia}, \bibinfo{person}{Dhruv Shah}, \bibinfo{person}{Danny Driess}, \bibinfo{person}{Andy Zeng}, \bibinfo{person}{Yao Lu}, \bibinfo{person}{Pete Florence}, \bibinfo{person}{Igor Mordatch}, \bibinfo{person}{Sergey Levine}, \bibinfo{person}{Karol Hausman}, {et~al\mbox{.}}} \bibinfo{year}{2024}\natexlab{b}.
\newblock \showarticletitle{Grounded decoding: Guiding text generation with grounded models for embodied agents}.
\newblock \bibinfo{journal}{\emph{Advances in Neural Information Processing Systems}}  \bibinfo{volume}{36} (\bibinfo{year}{2024}).
\newblock


\bibitem[Huang et~al\mbox{.}(2024a)]%
        {huang2024understanding}
\bibfield{author}{\bibinfo{person}{Xu Huang}, \bibinfo{person}{Weiwen Liu}, \bibinfo{person}{Xiaolong Chen}, \bibinfo{person}{Xingmei Wang}, \bibinfo{person}{Hao Wang}, \bibinfo{person}{Defu Lian}, \bibinfo{person}{Yasheng Wang}, \bibinfo{person}{Ruiming Tang}, {and} \bibinfo{person}{Enhong Chen}.} \bibinfo{year}{2024}\natexlab{a}.
\newblock \showarticletitle{Understanding the planning of LLM agents: A survey}.
\newblock \bibinfo{journal}{\emph{arXiv preprint arXiv:2402.02716}} (\bibinfo{year}{2024}).
\newblock


\bibitem[Huang et~al\mbox{.}(2025a)]%
        {huang2025spiritsight}
\bibfield{author}{\bibinfo{person}{Zhiyuan Huang}, \bibinfo{person}{Ziming Cheng}, \bibinfo{person}{Junting Pan}, \bibinfo{person}{Zhaohui Hou}, {and} \bibinfo{person}{Mingjie Zhan}.} \bibinfo{year}{2025}\natexlab{a}.
\newblock \showarticletitle{SpiritSight Agent: Advanced GUI Agent with One Look}.
\newblock \bibinfo{journal}{\emph{arXiv preprint arXiv:2503.03196}} (\bibinfo{year}{2025}).
\newblock


\bibitem[Hurst et~al\mbox{.}(2024)]%
        {hurst2024gpt}
\bibfield{author}{\bibinfo{person}{Aaron Hurst}, \bibinfo{person}{Adam Lerer}, \bibinfo{person}{Adam~P Goucher}, \bibinfo{person}{Adam Perelman}, \bibinfo{person}{Aditya Ramesh}, \bibinfo{person}{Aidan Clark}, \bibinfo{person}{AJ Ostrow}, \bibinfo{person}{Akila Welihinda}, \bibinfo{person}{Alan Hayes}, \bibinfo{person}{Alec Radford}, {et~al\mbox{.}}} \bibinfo{year}{2024}\natexlab{}.
\newblock \showarticletitle{Gpt-4o system card}.
\newblock \bibinfo{journal}{\emph{arXiv preprint arXiv:2410.21276}} (\bibinfo{year}{2024}).
\newblock


\bibitem[Jia et~al\mbox{.}(2024)]%
        {jia2024langsuit}
\bibfield{author}{\bibinfo{person}{Zixia Jia}, \bibinfo{person}{Mengmeng Wang}, \bibinfo{person}{Baichen Tong}, \bibinfo{person}{Song-chun Zhu}, {and} \bibinfo{person}{Zilong Zheng}.} \bibinfo{year}{2024}\natexlab{}.
\newblock \showarticletitle{LangSuit{\textperiodcentered} E: Planning, Controlling and Interacting with Large Language Models in Embodied Text Environments}. In \bibinfo{booktitle}{\emph{Findings of the Association for Computational Linguistics ACL 2024}}. \bibinfo{pages}{14778--14814}.
\newblock


\bibitem[Jiang et~al\mbox{.}(2023)]%
        {jiang2023mistral}
\bibfield{author}{\bibinfo{person}{Albert~Q. Jiang}, \bibinfo{person}{Alexandre Sablayrolles}, \bibinfo{person}{Arthur Mensch}, \bibinfo{person}{Chris Bamford}, \bibinfo{person}{Devendra~Singh Chaplot}, \bibinfo{person}{Diego de~las Casas}, \bibinfo{person}{Florian Bressand}, \bibinfo{person}{Gianna Lengyel}, \bibinfo{person}{Guillaume Lample}, \bibinfo{person}{Lucile Saulnier}, \bibinfo{person}{Lélio~Renard Lavaud}, \bibinfo{person}{Marie-Anne Lachaux}, \bibinfo{person}{Pierre Stock}, \bibinfo{person}{Teven~Le Scao}, \bibinfo{person}{Thibaut Lavril}, \bibinfo{person}{Thomas Wang}, \bibinfo{person}{Timothée Lacroix}, {and} \bibinfo{person}{William~El Sayed}.} \bibinfo{year}{2023}\natexlab{}.
\newblock \bibinfo{title}{Mistral 7B}.
\newblock


\bibitem[Jiang et~al\mbox{.}(2025)]%
        {jiang2025alphadrive}
\bibfield{author}{\bibinfo{person}{Bo Jiang}, \bibinfo{person}{Shaoyu Chen}, \bibinfo{person}{Qian Zhang}, \bibinfo{person}{Wenyu Liu}, {and} \bibinfo{person}{Xinggang Wang}.} \bibinfo{year}{2025}\natexlab{}.
\newblock \showarticletitle{Alphadrive: Unleashing the power of vlms in autonomous driving via reinforcement learning and reasoning}.
\newblock \bibinfo{journal}{\emph{arXiv preprint arXiv:2503.07608}} (\bibinfo{year}{2025}).
\newblock


\bibitem[Jiang et~al\mbox{.}(2024)]%
        {jiang2024urbanllm}
\bibfield{author}{\bibinfo{person}{Yue Jiang}, \bibinfo{person}{Qin Chao}, \bibinfo{person}{Yile Chen}, \bibinfo{person}{Xiucheng Li}, \bibinfo{person}{Shuai Liu}, {and} \bibinfo{person}{Gao Cong}.} \bibinfo{year}{2024}\natexlab{}.
\newblock \showarticletitle{Urbanllm: Autonomous urban activity planning and management with large language models}.
\newblock \bibinfo{journal}{\emph{arXiv preprint arXiv:2406.12360}} (\bibinfo{year}{2024}).
\newblock


\bibitem[Jiang et~al\mbox{.}(2022)]%
        {jiang2022vima}
\bibfield{author}{\bibinfo{person}{Yunfan Jiang}, \bibinfo{person}{Agrim Gupta}, \bibinfo{person}{Zichen Zhang}, \bibinfo{person}{Guanzhi Wang}, \bibinfo{person}{Yongqiang Dou}, \bibinfo{person}{Yanjun Chen}, \bibinfo{person}{Li Fei-Fei}, \bibinfo{person}{Anima Anandkumar}, \bibinfo{person}{Yuke Zhu}, {and} \bibinfo{person}{Linxi Fan}.} \bibinfo{year}{2022}\natexlab{}.
\newblock \showarticletitle{Vima: General robot manipulation with multimodal prompts}.
\newblock \bibinfo{journal}{\emph{arXiv preprint arXiv:2210.03094}} \bibinfo{volume}{2}, \bibinfo{number}{3} (\bibinfo{year}{2022}), \bibinfo{pages}{6}.
\newblock


\bibitem[Jiao et~al\mbox{.}(2024)]%
        {jiao2024learning}
\bibfield{author}{\bibinfo{person}{Fangkai Jiao}, \bibinfo{person}{Chengwei Qin}, \bibinfo{person}{Zhengyuan Liu}, \bibinfo{person}{Nancy~F Chen}, {and} \bibinfo{person}{Shafiq Joty}.} \bibinfo{year}{2024}\natexlab{}.
\newblock \showarticletitle{Learning planning-based reasoning by trajectories collection and process reward synthesizing}.
\newblock \bibinfo{journal}{\emph{arXiv preprint arXiv:2402.00658}} (\bibinfo{year}{2024}).
\newblock


\bibitem[Jimenez et~al\mbox{.}(2023)]%
        {jimenez2023swe}
\bibfield{author}{\bibinfo{person}{Carlos~E Jimenez}, \bibinfo{person}{John Yang}, \bibinfo{person}{Alexander Wettig}, \bibinfo{person}{Shunyu Yao}, \bibinfo{person}{Kexin Pei}, \bibinfo{person}{Ofir Press}, {and} \bibinfo{person}{Karthik Narasimhan}.} \bibinfo{year}{2023}\natexlab{}.
\newblock \showarticletitle{Swe-bench: Can language models resolve real-world github issues?}
\newblock \bibinfo{journal}{\emph{arXiv preprint arXiv:2310.06770}} (\bibinfo{year}{2023}).
\newblock


\bibitem[Ju et~al\mbox{.}(2024)]%
        {ju2024globe}
\bibfield{author}{\bibinfo{person}{Da Ju}, \bibinfo{person}{Song Jiang}, \bibinfo{person}{Andrew Cohen}, \bibinfo{person}{Aaron Foss}, \bibinfo{person}{Sasha Mitts}, \bibinfo{person}{Arman Zharmagambetov}, \bibinfo{person}{Brandon Amos}, \bibinfo{person}{Xian Li}, \bibinfo{person}{Justine Kao}, \bibinfo{person}{Maryam Fazel-Zarandi}, {et~al\mbox{.}}} \bibinfo{year}{2024}\natexlab{}.
\newblock \showarticletitle{To the Globe (TTG): Towards Language-Driven Guaranteed Travel Planning}. In \bibinfo{booktitle}{\emph{Proceedings of the 2024 Conference on Empirical Methods in Natural Language Processing: System Demonstrations}}. \bibinfo{pages}{240--249}.
\newblock


\bibitem[Jurafsky and Martin({[n.\,d.]})]%
        {jurafskyspeech}
\bibfield{author}{\bibinfo{person}{Daniel Jurafsky} {and} \bibinfo{person}{James~H Martin}.} \bibinfo{year}{[n.\,d.]}\natexlab{}.
\newblock \bibinfo{title}{Speech and Language Processing: An Introduction to Natural Language Processing, Computational Linguistics, and Speech Recognition}.
\newblock


\bibitem[Kaelbling et~al\mbox{.}(1998)]%
        {kaelbling1998planning}
\bibfield{author}{\bibinfo{person}{Leslie~Pack Kaelbling}, \bibinfo{person}{Michael~L Littman}, {and} \bibinfo{person}{Anthony~R Cassandra}.} \bibinfo{year}{1998}\natexlab{}.
\newblock \showarticletitle{Planning and acting in partially observable stochastic domains}.
\newblock \bibinfo{journal}{\emph{Artificial intelligence}} \bibinfo{volume}{101}, \bibinfo{number}{1-2} (\bibinfo{year}{1998}), \bibinfo{pages}{99--134}.
\newblock


\bibitem[Kahneman(2011)]%
        {kahneman2011thinking}
\bibfield{author}{\bibinfo{person}{Daniel Kahneman}.} \bibinfo{year}{2011}\natexlab{}.
\newblock \bibinfo{booktitle}{\emph{Thinking, fast and slow}}.
\newblock \bibinfo{publisher}{macmillan}.
\newblock


\bibitem[Kambhampati et~al\mbox{.}({[n.\,d.]})]%
        {kambhampatiposition}
\bibfield{author}{\bibinfo{person}{Subbarao Kambhampati}, \bibinfo{person}{Karthik Valmeekam}, \bibinfo{person}{Lin Guan}, \bibinfo{person}{Mudit Verma}, \bibinfo{person}{Kaya Stechly}, \bibinfo{person}{Siddhant Bhambri}, \bibinfo{person}{Lucas~Paul Saldyt}, {and} \bibinfo{person}{Anil~B Murthy}.} \bibinfo{year}{[n.\,d.]}\natexlab{}.
\newblock \showarticletitle{Position: LLMs Can’t Plan, But Can Help Planning in LLM-Modulo Frameworks}. In \bibinfo{booktitle}{\emph{Forty-first International Conference on Machine Learning}}.
\newblock


\bibitem[Kang and Xiong(2024)]%
        {kang2024researcharena}
\bibfield{author}{\bibinfo{person}{Hao Kang} {and} \bibinfo{person}{Chenyan Xiong}.} \bibinfo{year}{2024}\natexlab{}.
\newblock \showarticletitle{ResearchArena: Benchmarking LLMs' Ability to Collect and Organize Information as Research Agents}.
\newblock \bibinfo{journal}{\emph{arXiv preprint arXiv:2406.10291}} (\bibinfo{year}{2024}).
\newblock


\bibitem[Kang et~al\mbox{.}(2023)]%
        {kang2023think}
\bibfield{author}{\bibinfo{person}{Jikun Kang}, \bibinfo{person}{Romain Laroche}, \bibinfo{person}{Xingdi Yuan}, \bibinfo{person}{Adam Trischler}, \bibinfo{person}{Xue Liu}, {and} \bibinfo{person}{Jie Fu}.} \bibinfo{year}{2023}\natexlab{}.
\newblock \showarticletitle{Think Before You Act: Decision Transformers with Working Memory}. In \bibinfo{booktitle}{\emph{Forty-first International Conference on Machine Learning}}.
\newblock


\bibitem[Khanna et~al\mbox{.}(2024)]%
        {khanna2024goat}
\bibfield{author}{\bibinfo{person}{Mukul Khanna}, \bibinfo{person}{Ram Ramrakhya}, \bibinfo{person}{Gunjan Chhablani}, \bibinfo{person}{Sriram Yenamandra}, \bibinfo{person}{Theophile Gervet}, \bibinfo{person}{Matthew Chang}, \bibinfo{person}{Zsolt Kira}, \bibinfo{person}{Devendra~Singh Chaplot}, \bibinfo{person}{Dhruv Batra}, {and} \bibinfo{person}{Roozbeh Mottaghi}.} \bibinfo{year}{2024}\natexlab{}.
\newblock \showarticletitle{Goat-bench: A benchmark for multi-modal lifelong navigation}. In \bibinfo{booktitle}{\emph{Proceedings of the IEEE/CVF Conference on Computer Vision and Pattern Recognition}}. \bibinfo{pages}{16373--16383}.
\newblock


\bibitem[Kim et~al\mbox{.}(2024)]%
        {kim2024rada}
\bibfield{author}{\bibinfo{person}{Minsoo Kim}, \bibinfo{person}{Victor Bursztyn}, \bibinfo{person}{Eunyee Koh}, \bibinfo{person}{Shunan Guo}, {and} \bibinfo{person}{Seung-won Hwang}.} \bibinfo{year}{2024}\natexlab{}.
\newblock \showarticletitle{Rada: Retrieval-augmented web agent planning with llms}. In \bibinfo{booktitle}{\emph{Findings of the Association for Computational Linguistics ACL 2024}}. \bibinfo{pages}{13511--13525}.
\newblock


\bibitem[Kocsis and Szepesv{\'a}ri(2006)]%
        {kocsis2006bandit}
\bibfield{author}{\bibinfo{person}{Levente Kocsis} {and} \bibinfo{person}{Csaba Szepesv{\'a}ri}.} \bibinfo{year}{2006}\natexlab{}.
\newblock \showarticletitle{Bandit based monte-carlo planning}. In \bibinfo{booktitle}{\emph{European conference on machine learning}}. Springer, \bibinfo{pages}{282--293}.
\newblock


\bibitem[Koh et~al\mbox{.}(2024a)]%
        {koh2024visualwebarena}
\bibfield{author}{\bibinfo{person}{Jing~Yu Koh}, \bibinfo{person}{Robert Lo}, \bibinfo{person}{Lawrence Jang}, \bibinfo{person}{Vikram Duvvur}, \bibinfo{person}{Ming~Chong Lim}, \bibinfo{person}{Po-Yu Huang}, \bibinfo{person}{Graham Neubig}, \bibinfo{person}{Shuyan Zhou}, \bibinfo{person}{Ruslan Salakhutdinov}, {and} \bibinfo{person}{Daniel Fried}.} \bibinfo{year}{2024}\natexlab{a}.
\newblock \showarticletitle{Visualwebarena: Evaluating multimodal agents on realistic visual web tasks}.
\newblock \bibinfo{journal}{\emph{arXiv preprint arXiv:2401.13649}} (\bibinfo{year}{2024}).
\newblock


\bibitem[Koh et~al\mbox{.}(2024b)]%
        {koh2024tree}
\bibfield{author}{\bibinfo{person}{Jing~Yu Koh}, \bibinfo{person}{Stephen McAleer}, \bibinfo{person}{Daniel Fried}, {and} \bibinfo{person}{Ruslan Salakhutdinov}.} \bibinfo{year}{2024}\natexlab{b}.
\newblock \showarticletitle{Tree search for language model agents}.
\newblock \bibinfo{journal}{\emph{arXiv preprint arXiv:2407.01476}} (\bibinfo{year}{2024}).
\newblock


\bibitem[Kumar et~al\mbox{.}({[n.\,d.]})]%
        {kumar2409training}
\bibfield{author}{\bibinfo{person}{Aviral Kumar}, \bibinfo{person}{Vincent Zhuang}, \bibinfo{person}{Rishabh Agarwal}, \bibinfo{person}{Yi Su}, \bibinfo{person}{John~D Co-Reyes}, \bibinfo{person}{Avi Singh}, \bibinfo{person}{Kate Baumli}, \bibinfo{person}{Shariq Iqbal}, \bibinfo{person}{Colton Bishop}, \bibinfo{person}{Rebecca Roelofs}, {et~al\mbox{.}}} \bibinfo{year}{[n.\,d.]}\natexlab{}.
\newblock \showarticletitle{Training language models to self-correct via reinforcement learning, 2024}.
\newblock \bibinfo{journal}{\emph{URL https://arxiv. org/abs/2409.12917}} (\bibinfo{year}{[n.\,d.]}).
\newblock


\bibitem[Lai et~al\mbox{.}(2024a)]%
        {lai2024autowebglm}
\bibfield{author}{\bibinfo{person}{Hanyu Lai}, \bibinfo{person}{Xiao Liu}, \bibinfo{person}{Iat~Long Iong}, \bibinfo{person}{Shuntian Yao}, \bibinfo{person}{Yuxuan Chen}, \bibinfo{person}{Pengbo Shen}, \bibinfo{person}{Hao Yu}, \bibinfo{person}{Hanchen Zhang}, \bibinfo{person}{Xiaohan Zhang}, \bibinfo{person}{Yuxiao Dong}, {et~al\mbox{.}}} \bibinfo{year}{2024}\natexlab{a}.
\newblock \showarticletitle{AutoWebGLM: A Large Language Model-based Web Navigating Agent}. In \bibinfo{booktitle}{\emph{Proceedings of the 30th ACM SIGKDD Conference on Knowledge Discovery and Data Mining}}. \bibinfo{pages}{5295--5306}.
\newblock


\bibitem[Lai et~al\mbox{.}(2024b)]%
        {lai2024step}
\bibfield{author}{\bibinfo{person}{Xin Lai}, \bibinfo{person}{Zhuotao Tian}, \bibinfo{person}{Yukang Chen}, \bibinfo{person}{Senqiao Yang}, \bibinfo{person}{Xiangru Peng}, {and} \bibinfo{person}{Jiaya Jia}.} \bibinfo{year}{2024}\natexlab{b}.
\newblock \showarticletitle{Step-dpo: Step-wise preference optimization for long-chain reasoning of llms}.
\newblock \bibinfo{journal}{\emph{arXiv preprint arXiv:2406.18629}} (\bibinfo{year}{2024}).
\newblock


\bibitem[Lal et~al\mbox{.}(2024)]%
        {lal2024cat}
\bibfield{author}{\bibinfo{person}{Yash~Kumar Lal}, \bibinfo{person}{Vanya Cohen}, \bibinfo{person}{Nathanael Chambers}, \bibinfo{person}{Niranjan Balasubramanian}, {and} \bibinfo{person}{Ray Mooney}.} \bibinfo{year}{2024}\natexlab{}.
\newblock \showarticletitle{CaT-Bench: Benchmarking Language Model Understanding of Causal and Temporal Dependencies in Plans}. In \bibinfo{booktitle}{\emph{Proceedings of the 2024 Conference on Empirical Methods in Natural Language Processing}}. \bibinfo{pages}{19336--19354}.
\newblock


\bibitem[Lazaro-Gredilla et~al\mbox{.}({[n.\,d.]})]%
        {lazarotype}
\bibfield{author}{\bibinfo{person}{Miguel Lazaro-Gredilla}, \bibinfo{person}{Li~Yang Ku}, \bibinfo{person}{Kevin~Patrick Murphy}, {and} \bibinfo{person}{Dileep George}.} \bibinfo{year}{[n.\,d.]}\natexlab{}.
\newblock \showarticletitle{What type of inference is planning?}. In \bibinfo{booktitle}{\emph{The Thirty-eighth Annual Conference on Neural Information Processing Systems}}.
\newblock


\bibitem[Lehnert et~al\mbox{.}({[n.\,d.]})]%
        {lehnert2024beyond}
\bibfield{author}{\bibinfo{person}{Lucas Lehnert}, \bibinfo{person}{Sainbayar Sukhbaatar}, \bibinfo{person}{Paul McVay}, \bibinfo{person}{Michael Rabbat}, {and} \bibinfo{person}{Yuandong Tian}.} \bibinfo{year}{[n.\,d.]}\natexlab{}.
\newblock \showarticletitle{Beyond A*: Better Planning with Transformers via Search Dynamics Bootstrapping}. In \bibinfo{booktitle}{\emph{ICLR 2024 Workshop on Large Language Model (LLM) Agents}}.
\newblock


\bibitem[Leng et~al\mbox{.}(2023)]%
        {leng2023tell2design}
\bibfield{author}{\bibinfo{person}{Sicong Leng}, \bibinfo{person}{Yang Zhou}, \bibinfo{person}{Mohammed~Haroon Dupty}, \bibinfo{person}{Wee~Sun Lee}, \bibinfo{person}{Sam Joyce}, {and} \bibinfo{person}{Wei Lu}.} \bibinfo{year}{2023}\natexlab{}.
\newblock \showarticletitle{Tell2Design: A Dataset for Language-Guided Floor Plan Generation}. In \bibinfo{booktitle}{\emph{Proceedings of the 61st Annual Meeting of the Association for Computational Linguistics (Volume 1: Long Papers)}}. \bibinfo{pages}{14680--14697}.
\newblock


\bibitem[Li et~al\mbox{.}(2024f)]%
        {li2024agent}
\bibfield{author}{\bibinfo{person}{Ao Li}, \bibinfo{person}{Yuexiang Xie}, \bibinfo{person}{Songze Li}, \bibinfo{person}{Fugee Tsung}, \bibinfo{person}{Bolin Ding}, {and} \bibinfo{person}{Yaliang Li}.} \bibinfo{year}{2024}\natexlab{f}.
\newblock \showarticletitle{Agent-Oriented Planning in Multi-Agent Systems}.
\newblock \bibinfo{journal}{\emph{arXiv preprint arXiv:2410.02189}} (\bibinfo{year}{2024}).
\newblock


\bibitem[Li et~al\mbox{.}(2024h)]%
        {li2024behavior}
\bibfield{author}{\bibinfo{person}{Chengshu Li}, \bibinfo{person}{Ruohan Zhang}, \bibinfo{person}{Josiah Wong}, \bibinfo{person}{Cem Gokmen}, \bibinfo{person}{Sanjana Srivastava}, \bibinfo{person}{Roberto Mart{\'\i}n-Mart{\'\i}n}, \bibinfo{person}{Chen Wang}, \bibinfo{person}{Gabrael Levine}, \bibinfo{person}{Wensi Ai}, \bibinfo{person}{Benjamin Martinez}, {et~al\mbox{.}}} \bibinfo{year}{2024}\natexlab{h}.
\newblock \showarticletitle{Behavior-1k: A human-centered, embodied ai benchmark with 1,000 everyday activities and realistic simulation}.
\newblock \bibinfo{journal}{\emph{arXiv preprint arXiv:2403.09227}} (\bibinfo{year}{2024}).
\newblock


\bibitem[Li et~al\mbox{.}(2024i)]%
        {li2024llava}
\bibfield{author}{\bibinfo{person}{Feng Li}, \bibinfo{person}{Renrui Zhang}, \bibinfo{person}{Hao Zhang}, \bibinfo{person}{Yuanhan Zhang}, \bibinfo{person}{Bo Li}, \bibinfo{person}{Wei Li}, \bibinfo{person}{Zejun Ma}, {and} \bibinfo{person}{Chunyuan Li}.} \bibinfo{year}{2024}\natexlab{i}.
\newblock \showarticletitle{Llava-next-interleave: Tackling multi-image, video, and 3d in large multimodal models}.
\newblock \bibinfo{journal}{\emph{arXiv preprint arXiv:2407.07895}} (\bibinfo{year}{2024}).
\newblock


\bibitem[Li et~al\mbox{.}(2024c)]%
        {li2024codetree}
\bibfield{author}{\bibinfo{person}{Jierui Li}, \bibinfo{person}{Hung Le}, \bibinfo{person}{Yingbo Zhou}, \bibinfo{person}{Caiming Xiong}, \bibinfo{person}{Silvio Savarese}, {and} \bibinfo{person}{Doyen Sahoo}.} \bibinfo{year}{2024}\natexlab{c}.
\newblock \showarticletitle{Codetree: Agent-guided tree search for code generation with large language models}.
\newblock \bibinfo{journal}{\emph{arXiv preprint arXiv:2411.04329}} (\bibinfo{year}{2024}).
\newblock


\bibitem[Li et~al\mbox{.}(2024d)]%
        {li2024closed}
\bibfield{author}{\bibinfo{person}{Jinghan Li}, \bibinfo{person}{Zhicheng Sun}, \bibinfo{person}{Fei Li}, \bibinfo{person}{Cao Sheng}, \bibinfo{person}{Jiazhong Yu}, {and} \bibinfo{person}{Yadong Mu}.} \bibinfo{year}{2024}\natexlab{d}.
\newblock \showarticletitle{Closed-loop long-horizon robotic planning via equilibrium sequence modeling}.
\newblock \bibinfo{journal}{\emph{arXiv preprint arXiv:2410.01440}} (\bibinfo{year}{2024}).
\newblock


\bibitem[Li et~al\mbox{.}(2024b)]%
        {li2024confidence}
\bibfield{author}{\bibinfo{person}{Loka Li}, \bibinfo{person}{Zhenhao Chen}, \bibinfo{person}{Guangyi Chen}, \bibinfo{person}{Yixuan Zhang}, \bibinfo{person}{Yusheng Su}, \bibinfo{person}{Eric Xing}, {and} \bibinfo{person}{Kun Zhang}.} \bibinfo{year}{2024}\natexlab{b}.
\newblock \showarticletitle{Confidence matters: Revisiting intrinsic self-correction capabilities of large language models}.
\newblock \bibinfo{journal}{\emph{arXiv preprint arXiv:2402.12563}} (\bibinfo{year}{2024}).
\newblock


\bibitem[Li et~al\mbox{.}(2025)]%
        {li2025self}
\bibfield{author}{\bibinfo{person}{Moxin Li}, \bibinfo{person}{Yuantao Zhang}, \bibinfo{person}{Wenjie Wang}, \bibinfo{person}{Wentao Shi}, \bibinfo{person}{Zhuo Liu}, \bibinfo{person}{Fuli Feng}, {and} \bibinfo{person}{Tat-Seng Chua}.} \bibinfo{year}{2025}\natexlab{}.
\newblock \showarticletitle{Self-improvement towards pareto optimality: Mitigating preference conflicts in multi-objective alignment}.
\newblock \bibinfo{journal}{\emph{arXiv preprint arXiv:2502.14354}} (\bibinfo{year}{2025}).
\newblock


\bibitem[Li et~al\mbox{.}(2024j)]%
        {li2024embodied}
\bibfield{author}{\bibinfo{person}{Manling Li}, \bibinfo{person}{Shiyu Zhao}, \bibinfo{person}{Qineng Wang}, \bibinfo{person}{Kangrui Wang}, \bibinfo{person}{Yu Zhou}, \bibinfo{person}{Sanjana Srivastava}, \bibinfo{person}{Cem Gokmen}, \bibinfo{person}{Tony Lee}, \bibinfo{person}{Erran~Li Li}, \bibinfo{person}{Ruohan Zhang}, {et~al\mbox{.}}} \bibinfo{year}{2024}\natexlab{j}.
\newblock \showarticletitle{Embodied agent interface: Benchmarking llms for embodied decision making}.
\newblock \bibinfo{journal}{\emph{Advances in Neural Information Processing Systems}}  \bibinfo{volume}{37} (\bibinfo{year}{2024}), \bibinfo{pages}{100428--100534}.
\newblock


\bibitem[Li et~al\mbox{.}(2023)]%
        {li2023api}
\bibfield{author}{\bibinfo{person}{Minghao Li}, \bibinfo{person}{Yingxiu Zhao}, \bibinfo{person}{Bowen Yu}, \bibinfo{person}{Feifan Song}, \bibinfo{person}{Hangyu Li}, \bibinfo{person}{Haiyang Yu}, \bibinfo{person}{Zhoujun Li}, \bibinfo{person}{Fei Huang}, {and} \bibinfo{person}{Yongbin Li}.} \bibinfo{year}{2023}\natexlab{}.
\newblock \showarticletitle{Api-bank: A comprehensive benchmark for tool-augmented llms}.
\newblock \bibinfo{journal}{\emph{arXiv preprint arXiv:2304.08244}} (\bibinfo{year}{2023}).
\newblock


\bibitem[Li et~al\mbox{.}(2024a)]%
        {li2024effects}
\bibfield{author}{\bibinfo{person}{Wei Li}, \bibinfo{person}{William~E Bishop}, \bibinfo{person}{Alice Li}, \bibinfo{person}{Christopher Rawles}, \bibinfo{person}{Folawiyo Campbell-Ajala}, \bibinfo{person}{Divya Tyamagundlu}, {and} \bibinfo{person}{Oriana Riva}.} \bibinfo{year}{2024}\natexlab{a}.
\newblock \showarticletitle{On the effects of data scale on ui control agents}.
\newblock \bibinfo{journal}{\emph{Advances in Neural Information Processing Systems}}  \bibinfo{volume}{37} (\bibinfo{year}{2024}), \bibinfo{pages}{92130--92154}.
\newblock


\bibitem[Li and Qiu(2023)]%
        {li2023mot}
\bibfield{author}{\bibinfo{person}{Xiaonan Li} {and} \bibinfo{person}{Xipeng Qiu}.} \bibinfo{year}{2023}\natexlab{}.
\newblock \showarticletitle{MoT: Memory-of-Thought Enables ChatGPT to Self-Improve}. In \bibinfo{booktitle}{\emph{Proceedings of the 2023 Conference on Empirical Methods in Natural Language Processing}}. \bibinfo{pages}{6354--6374}.
\newblock


\bibitem[Li et~al\mbox{.}(2024e)]%
        {li2024survey}
\bibfield{author}{\bibinfo{person}{Xinyi Li}, \bibinfo{person}{Sai Wang}, \bibinfo{person}{Siqi Zeng}, \bibinfo{person}{Yu Wu}, {and} \bibinfo{person}{Yi Yang}.} \bibinfo{year}{2024}\natexlab{e}.
\newblock \showarticletitle{A survey on LLM-based multi-agent systems: workflow, infrastructure, and challenges}.
\newblock \bibinfo{journal}{\emph{Vicinagearth}} \bibinfo{volume}{1}, \bibinfo{number}{1} (\bibinfo{year}{2024}), \bibinfo{pages}{9}.
\newblock


\bibitem[Li et~al\mbox{.}(2024g)]%
        {li2024can}
\bibfield{author}{\bibinfo{person}{Xingxuan Li}, \bibinfo{person}{Weiwen Xu}, \bibinfo{person}{Ruochen Zhao}, \bibinfo{person}{Fangkai Jiao}, \bibinfo{person}{Shafiq Joty}, {and} \bibinfo{person}{Lidong Bing}.} \bibinfo{year}{2024}\natexlab{g}.
\newblock \showarticletitle{Can We Further Elicit Reasoning in LLMs? Critic-Guided Planning with Retrieval-Augmentation for Solving Challenging Tasks}.
\newblock \bibinfo{journal}{\emph{arXiv preprint arXiv:2410.01428}} (\bibinfo{year}{2024}).
\newblock


\bibitem[Li et~al\mbox{.}(2020)]%
        {li2020mapping}
\bibfield{author}{\bibinfo{person}{Yang Li}, \bibinfo{person}{Jiacong He}, \bibinfo{person}{Xin Zhou}, \bibinfo{person}{Yuan Zhang}, {and} \bibinfo{person}{Jason Baldridge}.} \bibinfo{year}{2020}\natexlab{}.
\newblock \showarticletitle{Mapping natural language instructions to mobile UI action sequences}.
\newblock \bibinfo{journal}{\emph{arXiv preprint arXiv:2005.03776}} (\bibinfo{year}{2020}).
\newblock


\bibitem[Lightman et~al\mbox{.}(2023)]%
        {lightman2023let}
\bibfield{author}{\bibinfo{person}{Hunter Lightman}, \bibinfo{person}{Vineet Kosaraju}, \bibinfo{person}{Yuri Burda}, \bibinfo{person}{Harrison Edwards}, \bibinfo{person}{Bowen Baker}, \bibinfo{person}{Teddy Lee}, \bibinfo{person}{Jan Leike}, \bibinfo{person}{John Schulman}, \bibinfo{person}{Ilya Sutskever}, {and} \bibinfo{person}{Karl Cobbe}.} \bibinfo{year}{2023}\natexlab{}.
\newblock \showarticletitle{Let's verify step by step}. In \bibinfo{booktitle}{\emph{The Twelfth International Conference on Learning Representations}}.
\newblock


\bibitem[Lin et~al\mbox{.}(2025)]%
        {lin2025qlass}
\bibfield{author}{\bibinfo{person}{Zongyu Lin}, \bibinfo{person}{Yao Tang}, \bibinfo{person}{Xingcheng Yao}, \bibinfo{person}{Da Yin}, \bibinfo{person}{Ziniu Hu}, \bibinfo{person}{Yizhou Sun}, {and} \bibinfo{person}{Kai-Wei Chang}.} \bibinfo{year}{2025}\natexlab{}.
\newblock \showarticletitle{QLASS: Boosting Language Agent Inference via Q-Guided Stepwise Search}.
\newblock \bibinfo{journal}{\emph{arXiv preprint arXiv:2502.02584}} (\bibinfo{year}{2025}).
\newblock


\bibitem[Liu et~al\mbox{.}(2023a)]%
        {liu2023llm+}
\bibfield{author}{\bibinfo{person}{Bo Liu}, \bibinfo{person}{Yuqian Jiang}, \bibinfo{person}{Xiaohan Zhang}, \bibinfo{person}{Qiang Liu}, \bibinfo{person}{Shiqi Zhang}, \bibinfo{person}{Joydeep Biswas}, {and} \bibinfo{person}{Peter Stone}.} \bibinfo{year}{2023}\natexlab{a}.
\newblock \showarticletitle{Llm+ p: Empowering large language models with optimal planning proficiency}.
\newblock \bibinfo{journal}{\emph{arXiv preprint arXiv:2304.11477}} (\bibinfo{year}{2023}).
\newblock


\bibitem[Liu et~al\mbox{.}(2018)]%
        {liu2018reinforcement}
\bibfield{author}{\bibinfo{person}{Evan~Zheran Liu}, \bibinfo{person}{Kelvin Guu}, \bibinfo{person}{Panupong Pasupat}, \bibinfo{person}{Tianlin Shi}, {and} \bibinfo{person}{Percy Liang}.} \bibinfo{year}{2018}\natexlab{}.
\newblock \showarticletitle{Reinforcement learning on web interfaces using workflow-guided exploration}.
\newblock \bibinfo{journal}{\emph{arXiv preprint arXiv:1802.08802}} (\bibinfo{year}{2018}).
\newblock


\bibitem[Liu et~al\mbox{.}(2024b)]%
        {Liu_2024}
\bibfield{author}{\bibinfo{person}{Haotian Liu}, \bibinfo{person}{Chunyuan Li}, \bibinfo{person}{Yuheng Li}, {and} \bibinfo{person}{Yong~Jae Lee}.} \bibinfo{year}{2024}\natexlab{b}.
\newblock \showarticletitle{Improved Baselines with Visual Instruction Tuning}. In \bibinfo{booktitle}{\emph{2024 IEEE/CVF Conference on Computer Vision and Pattern Recognition (CVPR)}}. \bibinfo{publisher}{IEEE}, \bibinfo{pages}{26286–26296}.
\newblock


\bibitem[Liu et~al\mbox{.}(2024a)]%
        {liu2024don}
\bibfield{author}{\bibinfo{person}{Jiacheng Liu}, \bibinfo{person}{Andrew Cohen}, \bibinfo{person}{Ramakanth Pasunuru}, \bibinfo{person}{Yejin Choi}, \bibinfo{person}{Hannaneh Hajishirzi}, {and} \bibinfo{person}{Asli Celikyilmaz}.} \bibinfo{year}{2024}\natexlab{a}.
\newblock \showarticletitle{Don't throw away your value model! Generating more preferable text with Value-Guided Monte-Carlo Tree Search decoding}. In \bibinfo{booktitle}{\emph{First Conference on Language Modeling}}.
\newblock


\bibitem[Liu et~al\mbox{.}(2025b)]%
        {liu2025divide}
\bibfield{author}{\bibinfo{person}{Jiale Liu}, \bibinfo{person}{Yifan Zeng}, \bibinfo{person}{Shaokun Zhang}, \bibinfo{person}{Chi Zhang}, \bibinfo{person}{Malte H{\o}jmark-Bertelsen}, \bibinfo{person}{Marie~Normann Gadeberg}, \bibinfo{person}{Huazheng Wang}, {and} \bibinfo{person}{Qingyun Wu}.} \bibinfo{year}{2025}\natexlab{b}.
\newblock \showarticletitle{Divide, Optimize, Merge: Fine-Grained LLM Agent Optimization at Scale}.
\newblock \bibinfo{journal}{\emph{arXiv preprint arXiv:2505.03973}} (\bibinfo{year}{2025}).
\newblock


\bibitem[Liu et~al\mbox{.}(2023b)]%
        {liu2023think}
\bibfield{author}{\bibinfo{person}{Lei Liu}, \bibinfo{person}{Xiaoyan Yang}, \bibinfo{person}{Yue Shen}, \bibinfo{person}{Binbin Hu}, \bibinfo{person}{Zhiqiang Zhang}, \bibinfo{person}{Jinjie Gu}, {and} \bibinfo{person}{Guannan Zhang}.} \bibinfo{year}{2023}\natexlab{b}.
\newblock \showarticletitle{Think-in-memory: Recalling and post-thinking enable llms with long-term memory}.
\newblock \bibinfo{journal}{\emph{arXiv preprint arXiv:2311.08719}} (\bibinfo{year}{2023}).
\newblock


\bibitem[Liu and Beam(2025)]%
        {liu2025doubly}
\bibfield{author}{\bibinfo{person}{Manqing Liu} {and} \bibinfo{person}{Andrew~L Beam}.} \bibinfo{year}{2025}\natexlab{}.
\newblock \showarticletitle{Doubly Robust Monte Carlo Tree Search}.
\newblock \bibinfo{journal}{\emph{arXiv preprint arXiv:2502.01672}} (\bibinfo{year}{2025}).
\newblock


\bibitem[Liu et~al\mbox{.}(2023c)]%
        {liu2023agentbench}
\bibfield{author}{\bibinfo{person}{Xiao Liu}, \bibinfo{person}{Hao Yu}, \bibinfo{person}{Hanchen Zhang}, \bibinfo{person}{Yifan Xu}, \bibinfo{person}{Xuanyu Lei}, \bibinfo{person}{Hanyu Lai}, \bibinfo{person}{Yu Gu}, \bibinfo{person}{Hangliang Ding}, \bibinfo{person}{Kaiwen Men}, \bibinfo{person}{Kejuan Yang}, {et~al\mbox{.}}} \bibinfo{year}{2023}\natexlab{c}.
\newblock \showarticletitle{Agentbench: Evaluating llms as agents}.
\newblock \bibinfo{journal}{\emph{arXiv preprint arXiv:2308.03688}} (\bibinfo{year}{2023}).
\newblock


\bibitem[Liu et~al\mbox{.}(2024c)]%
        {liu2024visualagentbench}
\bibfield{author}{\bibinfo{person}{Xiao Liu}, \bibinfo{person}{Tianjie Zhang}, \bibinfo{person}{Yu Gu}, \bibinfo{person}{Iat~Long Iong}, \bibinfo{person}{Yifan Xu}, \bibinfo{person}{Xixuan Song}, \bibinfo{person}{Shudan Zhang}, \bibinfo{person}{Hanyu Lai}, \bibinfo{person}{Xinyi Liu}, \bibinfo{person}{Hanlin Zhao}, {et~al\mbox{.}}} \bibinfo{year}{2024}\natexlab{c}.
\newblock \showarticletitle{Visualagentbench: Towards large multimodal models as visual foundation agents}.
\newblock \bibinfo{journal}{\emph{arXiv preprint arXiv:2408.06327}} (\bibinfo{year}{2024}).
\newblock


\bibitem[Liu et~al\mbox{.}(2025c)]%
        {liu2025ui}
\bibfield{author}{\bibinfo{person}{Xinyi Liu}, \bibinfo{person}{Xiaoyi Zhang}, \bibinfo{person}{Ziyun Zhang}, {and} \bibinfo{person}{Yan Lu}.} \bibinfo{year}{2025}\natexlab{c}.
\newblock \showarticletitle{UI-E2I-Synth: Advancing GUI Grounding with Large-Scale Instruction Synthesis}.
\newblock \bibinfo{journal}{\emph{arXiv preprint arXiv:2504.11257}} (\bibinfo{year}{2025}).
\newblock


\bibitem[Liu et~al\mbox{.}(2025a)]%
        {liu2025infigui}
\bibfield{author}{\bibinfo{person}{Yuhang Liu}, \bibinfo{person}{Pengxiang Li}, \bibinfo{person}{Congkai Xie}, \bibinfo{person}{Xavier Hu}, \bibinfo{person}{Xiaotian Han}, \bibinfo{person}{Shengyu Zhang}, \bibinfo{person}{Hongxia Yang}, {and} \bibinfo{person}{Fei Wu}.} \bibinfo{year}{2025}\natexlab{a}.
\newblock \showarticletitle{InfiGUI-R1: Advancing Multimodal GUI Agents from Reactive Actors to Deliberative Reasoners}.
\newblock \bibinfo{journal}{\emph{arXiv preprint arXiv:2504.14239}} (\bibinfo{year}{2025}).
\newblock


\bibitem[Long et~al\mbox{.}(2024)]%
        {long2024teamcraft}
\bibfield{author}{\bibinfo{person}{Qian Long}, \bibinfo{person}{Zhi Li}, \bibinfo{person}{Ran Gong}, \bibinfo{person}{Ying~Nian Wu}, \bibinfo{person}{Demetri Terzopoulos}, {and} \bibinfo{person}{Xiaofeng Gao}.} \bibinfo{year}{2024}\natexlab{}.
\newblock \showarticletitle{TeamCraft: A Benchmark for Multi-Modal Multi-Agent Systems in Minecraft}.
\newblock \bibinfo{journal}{\emph{arXiv preprint arXiv:2412.05255}} (\bibinfo{year}{2024}).
\newblock


\bibitem[Lu et~al\mbox{.}(2024)]%
        {lu2024toolsandbox}
\bibfield{author}{\bibinfo{person}{Jiarui Lu}, \bibinfo{person}{Thomas Holleis}, \bibinfo{person}{Yizhe Zhang}, \bibinfo{person}{Bernhard Aumayer}, \bibinfo{person}{Feng Nan}, \bibinfo{person}{Felix Bai}, \bibinfo{person}{Shuang Ma}, \bibinfo{person}{Shen Ma}, \bibinfo{person}{Mengyu Li}, \bibinfo{person}{Guoli Yin}, {et~al\mbox{.}}} \bibinfo{year}{2024}\natexlab{}.
\newblock \showarticletitle{Toolsandbox: A stateful, conversational, interactive evaluation benchmark for llm tool use capabilities}.
\newblock \bibinfo{journal}{\emph{arXiv preprint arXiv:2408.04682}} (\bibinfo{year}{2024}).
\newblock


\bibitem[L{\`u} et~al\mbox{.}(2024)]%
        {lu2024weblinx}
\bibfield{author}{\bibinfo{person}{Xing~Han L{\`u}}, \bibinfo{person}{Zden{\v{e}}k Kasner}, {and} \bibinfo{person}{Siva Reddy}.} \bibinfo{year}{2024}\natexlab{}.
\newblock \showarticletitle{Weblinx: Real-world website navigation with multi-turn dialogue}.
\newblock \bibinfo{journal}{\emph{arXiv preprint arXiv:2402.05930}} (\bibinfo{year}{2024}).
\newblock


\bibitem[Lu et~al\mbox{.}(2023)]%
        {lu2023multimodal}
\bibfield{author}{\bibinfo{person}{Yujie Lu}, \bibinfo{person}{Pan Lu}, \bibinfo{person}{Zhiyu Chen}, \bibinfo{person}{Wanrong Zhu}, \bibinfo{person}{Xin~Eric Wang}, {and} \bibinfo{person}{William~Yang Wang}.} \bibinfo{year}{2023}\natexlab{}.
\newblock \showarticletitle{Multimodal procedural planning via dual text-image prompting}.
\newblock \bibinfo{journal}{\emph{arXiv preprint arXiv:2305.01795}} (\bibinfo{year}{2023}).
\newblock


\bibitem[Lu et~al\mbox{.}(2025)]%
        {lu2025ui}
\bibfield{author}{\bibinfo{person}{Zhengxi Lu}, \bibinfo{person}{Yuxiang Chai}, \bibinfo{person}{Yaxuan Guo}, \bibinfo{person}{Xi Yin}, \bibinfo{person}{Liang Liu}, \bibinfo{person}{Hao Wang}, \bibinfo{person}{Guanjing Xiong}, {and} \bibinfo{person}{Hongsheng Li}.} \bibinfo{year}{2025}\natexlab{}.
\newblock \showarticletitle{UI-R1: Enhancing Action Prediction of GUI Agents by Reinforcement Learning}.
\newblock \bibinfo{journal}{\emph{arXiv preprint arXiv:2503.21620}} (\bibinfo{year}{2025}).
\newblock


\bibitem[Luo et~al\mbox{.}(2025)]%
        {luo2025large}
\bibfield{author}{\bibinfo{person}{Junyu Luo}, \bibinfo{person}{Weizhi Zhang}, \bibinfo{person}{Ye Yuan}, \bibinfo{person}{Yusheng Zhao}, \bibinfo{person}{Junwei Yang}, \bibinfo{person}{Yiyang Gu}, \bibinfo{person}{Bohan Wu}, \bibinfo{person}{Binqi Chen}, \bibinfo{person}{Ziyue Qiao}, \bibinfo{person}{Qingqing Long}, {et~al\mbox{.}}} \bibinfo{year}{2025}\natexlab{}.
\newblock \showarticletitle{Large Language Model Agent: A Survey on Methodology, Applications and Challenges}.
\newblock \bibinfo{journal}{\emph{arXiv preprint arXiv:2503.21460}} (\bibinfo{year}{2025}).
\newblock


\bibitem[Luo et~al\mbox{.}(2024)]%
        {luo2024dstruct2design}
\bibfield{author}{\bibinfo{person}{Zhi~Hao Luo}, \bibinfo{person}{Luis Lara}, \bibinfo{person}{Ge~Ya Luo}, \bibinfo{person}{Florian Golemo}, \bibinfo{person}{Christopher Beckham}, {and} \bibinfo{person}{Christopher Pal}.} \bibinfo{year}{2024}\natexlab{}.
\newblock \showarticletitle{DStruct2Design: Data and Benchmarks for Data Structure Driven Generative Floor Plan Design}.
\newblock \bibinfo{journal}{\emph{arXiv preprint arXiv:2407.15723}} (\bibinfo{year}{2024}).
\newblock


\bibitem[Lyu et~al\mbox{.}(2024)]%
        {lyu2024retrieve}
\bibfield{author}{\bibinfo{person}{Yuanjie Lyu}, \bibinfo{person}{Zihan Niu}, \bibinfo{person}{Zheyong Xie}, \bibinfo{person}{Chao Zhang}, \bibinfo{person}{Tong Xu}, \bibinfo{person}{Yang Wang}, {and} \bibinfo{person}{Enhong Chen}.} \bibinfo{year}{2024}\natexlab{}.
\newblock \showarticletitle{Retrieve-plan-generation: an iterative planning and answering framework for knowledge-intensive LLM generation}.
\newblock \bibinfo{journal}{\emph{arXiv preprint arXiv:2406.14979}} (\bibinfo{year}{2024}).
\newblock


\bibitem[Ma et~al\mbox{.}(2024)]%
        {ma2024non}
\bibfield{author}{\bibinfo{person}{Chang Ma}, \bibinfo{person}{Haiteng Zhao}, \bibinfo{person}{Junlei Zhang}, \bibinfo{person}{Junxian He}, {and} \bibinfo{person}{Lingpeng Kong}.} \bibinfo{year}{2024}\natexlab{}.
\newblock \showarticletitle{Non-myopic Generation of Language Model for Reasoning and Planning}.
\newblock \bibinfo{journal}{\emph{arXiv e-prints}} (\bibinfo{year}{2024}), \bibinfo{pages}{arXiv--2410}.
\newblock


\bibitem[Madaan et~al\mbox{.}(2024)]%
        {madaan2024self}
\bibfield{author}{\bibinfo{person}{Aman Madaan}, \bibinfo{person}{Niket Tandon}, \bibinfo{person}{Prakhar Gupta}, \bibinfo{person}{Skyler Hallinan}, \bibinfo{person}{Luyu Gao}, \bibinfo{person}{Sarah Wiegreffe}, \bibinfo{person}{Uri Alon}, \bibinfo{person}{Nouha Dziri}, \bibinfo{person}{Shrimai Prabhumoye}, \bibinfo{person}{Yiming Yang}, {et~al\mbox{.}}} \bibinfo{year}{2024}\natexlab{}.
\newblock \showarticletitle{Self-refine: Iterative refinement with self-feedback}.
\newblock \bibinfo{journal}{\emph{Advances in Neural Information Processing Systems}}  \bibinfo{volume}{36} (\bibinfo{year}{2024}).
\newblock


\bibitem[Mahdavi et~al\mbox{.}({[n.\,d.]})]%
        {mahdavileveraging}
\bibfield{author}{\bibinfo{person}{Sadegh Mahdavi}, \bibinfo{person}{Raquel Aoki}, \bibinfo{person}{Keyi Tang}, {and} \bibinfo{person}{Yanshuai Cao}.} \bibinfo{year}{[n.\,d.]}\natexlab{}.
\newblock \showarticletitle{Leveraging Environment Interaction for Automated PDDL Translation and Planning with Large Language Models}. In \bibinfo{booktitle}{\emph{The Thirty-eighth Annual Conference on Neural Information Processing Systems}}.
\newblock


\bibitem[Mai et~al\mbox{.}(2024)]%
        {mai2024learning}
\bibfield{author}{\bibinfo{person}{Florian Mai}, \bibinfo{person}{Nathan Cornille}, {and} \bibinfo{person}{Marie-Francine Moens}.} \bibinfo{year}{2024}\natexlab{}.
\newblock \showarticletitle{Learning to Plan Long-Term for Language Modeling}.
\newblock \bibinfo{journal}{\emph{arXiv preprint arXiv:2409.00070}} (\bibinfo{year}{2024}).
\newblock


\bibitem[Majumder et~al\mbox{.}(2023)]%
        {majumder2023clin}
\bibfield{author}{\bibinfo{person}{Bodhisattwa~Prasad Majumder}, \bibinfo{person}{Bhavana~Dalvi Mishra}, \bibinfo{person}{Peter Jansen}, \bibinfo{person}{Oyvind Tafjord}, \bibinfo{person}{Niket Tandon}, \bibinfo{person}{Li Zhang}, \bibinfo{person}{Chris Callison-Burch}, {and} \bibinfo{person}{Peter Clark}.} \bibinfo{year}{2023}\natexlab{}.
\newblock \showarticletitle{Clin: A continually learning language agent for rapid task adaptation and generalization}.
\newblock \bibinfo{journal}{\emph{arXiv preprint arXiv:2310.10134}} (\bibinfo{year}{2023}).
\newblock


\bibitem[Mandi et~al\mbox{.}(2024)]%
        {mandi2024roco}
\bibfield{author}{\bibinfo{person}{Zhao Mandi}, \bibinfo{person}{Shreeya Jain}, {and} \bibinfo{person}{Shuran Song}.} \bibinfo{year}{2024}\natexlab{}.
\newblock \showarticletitle{Roco: Dialectic multi-robot collaboration with large language models}. In \bibinfo{booktitle}{\emph{2024 IEEE International Conference on Robotics and Automation (ICRA)}}. IEEE, \bibinfo{pages}{286--299}.
\newblock


\bibitem[Matsuo et~al\mbox{.}(2022)]%
        {matsuo2022deep}
\bibfield{author}{\bibinfo{person}{Yutaka Matsuo}, \bibinfo{person}{Yann LeCun}, \bibinfo{person}{Maneesh Sahani}, \bibinfo{person}{Doina Precup}, \bibinfo{person}{David Silver}, \bibinfo{person}{Masashi Sugiyama}, \bibinfo{person}{Eiji Uchibe}, {and} \bibinfo{person}{Jun Morimoto}.} \bibinfo{year}{2022}\natexlab{}.
\newblock \showarticletitle{Deep learning, reinforcement learning, and world models}.
\newblock \bibinfo{journal}{\emph{Neural Networks}}  \bibinfo{volume}{152} (\bibinfo{year}{2022}), \bibinfo{pages}{267--275}.
\newblock


\bibitem[Matthews et~al\mbox{.}(2024)]%
        {matthews2024craftax}
\bibfield{author}{\bibinfo{person}{Michael Matthews}, \bibinfo{person}{Michael Beukman}, \bibinfo{person}{Benjamin Ellis}, \bibinfo{person}{Mikayel Samvelyan}, \bibinfo{person}{Matthew Jackson}, \bibinfo{person}{Samuel Coward}, {and} \bibinfo{person}{Jakob Foerster}.} \bibinfo{year}{2024}\natexlab{}.
\newblock \showarticletitle{Craftax: A lightning-fast benchmark for open-ended reinforcement learning}.
\newblock \bibinfo{journal}{\emph{arXiv preprint arXiv:2402.16801}} (\bibinfo{year}{2024}).
\newblock


\bibitem[Men et~al\mbox{.}(2024)]%
        {men2024unlocking}
\bibfield{author}{\bibinfo{person}{Tianyi Men}, \bibinfo{person}{Pengfei Cao}, \bibinfo{person}{Zhuoran Jin}, \bibinfo{person}{Yubo Chen}, \bibinfo{person}{Kang Liu}, {and} \bibinfo{person}{Jun Zhao}.} \bibinfo{year}{2024}\natexlab{}.
\newblock \showarticletitle{Unlocking the Future: Exploring Look-Ahead Planning Mechanistic Interpretability in Large Language Models}. In \bibinfo{booktitle}{\emph{Proceedings of the 2024 Conference on Empirical Methods in Natural Language Processing}}. \bibinfo{pages}{7713--7724}.
\newblock


\bibitem[Meng et~al\mbox{.}(2024)]%
        {meng2024llm}
\bibfield{author}{\bibinfo{person}{Silin Meng}, \bibinfo{person}{Yiwei Wang}, \bibinfo{person}{Cheng-Fu Yang}, \bibinfo{person}{Nanyun Peng}, {and} \bibinfo{person}{Kai-Wei Chang}.} \bibinfo{year}{2024}\natexlab{}.
\newblock \showarticletitle{Llm-a*: Large language model enhanced incremental heuristic search on path planning}.
\newblock \bibinfo{journal}{\emph{arXiv preprint arXiv:2407.02511}} (\bibinfo{year}{2024}).
\newblock


\bibitem[Morishita et~al\mbox{.}(2024)]%
        {morishita2024enhancing}
\bibfield{author}{\bibinfo{person}{Terufumi Morishita}, \bibinfo{person}{Gaku Morio}, \bibinfo{person}{Atsuki Yamaguchi}, {and} \bibinfo{person}{Yasuhiro Sogawa}.} \bibinfo{year}{2024}\natexlab{}.
\newblock \showarticletitle{Enhancing reasoning capabilities of llms via principled synthetic logic corpus}.
\newblock \bibinfo{journal}{\emph{Advances in Neural Information Processing Systems}}  \bibinfo{volume}{37} (\bibinfo{year}{2024}), \bibinfo{pages}{73572--73604}.
\newblock


\bibitem[Nair et~al\mbox{.}(2025)]%
        {nair2025flow}
\bibfield{author}{\bibinfo{person}{Lakshmi Nair}, \bibinfo{person}{Ian Trase}, {and} \bibinfo{person}{Mark Kim}.} \bibinfo{year}{2025}\natexlab{}.
\newblock \showarticletitle{Flow-of-Options: Diversified and Improved LLM Reasoning by Thinking Through Options}.
\newblock \bibinfo{journal}{\emph{arXiv preprint arXiv:2502.12929}} (\bibinfo{year}{2025}).
\newblock


\bibitem[Nath et~al\mbox{.}(2025)]%
        {nath2025toolcomp}
\bibfield{author}{\bibinfo{person}{Vaskar Nath}, \bibinfo{person}{Pranav Raja}, \bibinfo{person}{Claire Yoon}, {and} \bibinfo{person}{Sean Hendryx}.} \bibinfo{year}{2025}\natexlab{}.
\newblock \showarticletitle{ToolComp: A Multi-Tool Reasoning \& Process Supervision Benchmark}.
\newblock \bibinfo{journal}{\emph{arXiv preprint arXiv:2501.01290}} (\bibinfo{year}{2025}).
\newblock


\bibitem[O'Brien and Lewis(2023)]%
        {o2023contrastive}
\bibfield{author}{\bibinfo{person}{Sean O'Brien} {and} \bibinfo{person}{Mike Lewis}.} \bibinfo{year}{2023}\natexlab{}.
\newblock \showarticletitle{Contrastive decoding improves reasoning in large language models}.
\newblock \bibinfo{journal}{\emph{arXiv preprint arXiv:2309.09117}} (\bibinfo{year}{2023}).
\newblock


\bibitem[Ou et~al\mbox{.}(2024)]%
        {ou2024synatra}
\bibfield{author}{\bibinfo{person}{Tianyue Ou}, \bibinfo{person}{Frank~F Xu}, \bibinfo{person}{Aman Madaan}, \bibinfo{person}{Jiarui Liu}, \bibinfo{person}{Robert Lo}, \bibinfo{person}{Abishek Sridhar}, \bibinfo{person}{Sudipta Sengupta}, \bibinfo{person}{Dan Roth}, \bibinfo{person}{Graham Neubig}, {and} \bibinfo{person}{Shuyan Zhou}.} \bibinfo{year}{2024}\natexlab{}.
\newblock \showarticletitle{Synatra: Turning indirect knowledge into direct demonstrations for digital agents at scale}.
\newblock \bibinfo{journal}{\emph{arXiv preprint arXiv:2409.15637}} (\bibinfo{year}{2024}).
\newblock


\bibitem[O’Donoghue et~al\mbox{.}(2023)]%
        {o2023bioplanner}
\bibfield{author}{\bibinfo{person}{Odhran O’Donoghue}, \bibinfo{person}{Aleksandar Shtedritski}, \bibinfo{person}{John Ginger}, \bibinfo{person}{Ralph Abboud}, \bibinfo{person}{Ali Ghareeb}, {and} \bibinfo{person}{Samuel Rodriques}.} \bibinfo{year}{2023}\natexlab{}.
\newblock \showarticletitle{BioPlanner: Automatic Evaluation of LLMs on Protocol Planning in Biology}. In \bibinfo{booktitle}{\emph{Proceedings of the 2023 Conference on Empirical Methods in Natural Language Processing}}. \bibinfo{pages}{2676--2694}.
\newblock


\bibitem[Packer et~al\mbox{.}(2023)]%
        {packer2023memgpt}
\bibfield{author}{\bibinfo{person}{Charles Packer}, \bibinfo{person}{Sarah Wooders}, \bibinfo{person}{Kevin Lin}, \bibinfo{person}{Vivian Fang}, \bibinfo{person}{Shishir~G Patil}, \bibinfo{person}{Ion Stoica}, {and} \bibinfo{person}{Joseph~E Gonzalez}.} \bibinfo{year}{2023}\natexlab{}.
\newblock \showarticletitle{Memgpt: Towards llms as operating systems}.
\newblock \bibinfo{journal}{\emph{arXiv preprint arXiv:2310.08560}} (\bibinfo{year}{2023}).
\newblock


\bibitem[Pan et~al\mbox{.}(2025)]%
        {pan2025coat}
\bibfield{author}{\bibinfo{person}{Jianfeng Pan}, \bibinfo{person}{Senyou Deng}, {and} \bibinfo{person}{Shaomang Huang}.} \bibinfo{year}{2025}\natexlab{}.
\newblock \showarticletitle{CoAT: Chain-of-Associated-Thoughts Framework for Enhancing Large Language Models Reasoning}.
\newblock \bibinfo{journal}{\emph{arXiv preprint arXiv:2502.02390}} (\bibinfo{year}{2025}).
\newblock


\bibitem[Pan et~al\mbox{.}(2024)]%
        {pan2024training}
\bibfield{author}{\bibinfo{person}{Jiayi Pan}, \bibinfo{person}{Xingyao Wang}, \bibinfo{person}{Graham Neubig}, \bibinfo{person}{Navdeep Jaitly}, \bibinfo{person}{Heng Ji}, \bibinfo{person}{Alane Suhr}, {and} \bibinfo{person}{Yizhe Zhang}.} \bibinfo{year}{2024}\natexlab{}.
\newblock \showarticletitle{Training Software Engineering Agents and Verifiers with SWE-Gym}.
\newblock \bibinfo{journal}{\emph{arXiv preprint arXiv:2412.21139}} (\bibinfo{year}{2024}).
\newblock


\bibitem[Pandey et~al\mbox{.}(2025)]%
        {pandey2025adaptive}
\bibfield{author}{\bibinfo{person}{Tushar Pandey}, \bibinfo{person}{Ara Ghukasyan}, \bibinfo{person}{Oktay Goktas}, {and} \bibinfo{person}{Santosh~Kumar Radha}.} \bibinfo{year}{2025}\natexlab{}.
\newblock \showarticletitle{Adaptive Graph of Thoughts: Test-Time Adaptive Reasoning Unifying Chain, Tree, and Graph Structures}.
\newblock \bibinfo{journal}{\emph{arXiv preprint arXiv:2502.05078}} (\bibinfo{year}{2025}).
\newblock


\bibitem[Park et~al\mbox{.}(2023)]%
        {park2023generative}
\bibfield{author}{\bibinfo{person}{Joon~Sung Park}, \bibinfo{person}{Joseph O'Brien}, \bibinfo{person}{Carrie~Jun Cai}, \bibinfo{person}{Meredith~Ringel Morris}, \bibinfo{person}{Percy Liang}, {and} \bibinfo{person}{Michael~S Bernstein}.} \bibinfo{year}{2023}\natexlab{}.
\newblock \showarticletitle{Generative agents: Interactive simulacra of human behavior}. In \bibinfo{booktitle}{\emph{Proceedings of the 36th annual acm symposium on user interface software and technology}}. \bibinfo{pages}{1--22}.
\newblock


\bibitem[Pateria et~al\mbox{.}(2021)]%
        {pateria2021hierarchical}
\bibfield{author}{\bibinfo{person}{Shubham Pateria}, \bibinfo{person}{Budhitama Subagdja}, \bibinfo{person}{Ah-hwee Tan}, {and} \bibinfo{person}{Chai Quek}.} \bibinfo{year}{2021}\natexlab{}.
\newblock \showarticletitle{Hierarchical reinforcement learning: A comprehensive survey}.
\newblock \bibinfo{journal}{\emph{ACM Computing Surveys (CSUR)}} \bibinfo{volume}{54}, \bibinfo{number}{5} (\bibinfo{year}{2021}), \bibinfo{pages}{1--35}.
\newblock


\bibitem[Prasad et~al\mbox{.}(2023)]%
        {prasad2023adapt}
\bibfield{author}{\bibinfo{person}{Archiki Prasad}, \bibinfo{person}{Alexander Koller}, \bibinfo{person}{Mareike Hartmann}, \bibinfo{person}{Peter Clark}, \bibinfo{person}{Ashish Sabharwal}, \bibinfo{person}{Mohit Bansal}, {and} \bibinfo{person}{Tushar Khot}.} \bibinfo{year}{2023}\natexlab{}.
\newblock \showarticletitle{Adapt: As-needed decomposition and planning with language models}.
\newblock \bibinfo{journal}{\emph{arXiv preprint arXiv:2311.05772}} (\bibinfo{year}{2023}).
\newblock


\bibitem[Puig et~al\mbox{.}(2018)]%
        {puig2018virtualhome}
\bibfield{author}{\bibinfo{person}{Xavier Puig}, \bibinfo{person}{Kevin Ra}, \bibinfo{person}{Marko Boben}, \bibinfo{person}{Jiaman Li}, \bibinfo{person}{Tingwu Wang}, \bibinfo{person}{Sanja Fidler}, {and} \bibinfo{person}{Antonio Torralba}.} \bibinfo{year}{2018}\natexlab{}.
\newblock \showarticletitle{Virtualhome: Simulating household activities via programs}. In \bibinfo{booktitle}{\emph{Proceedings of the IEEE conference on computer vision and pattern recognition}}. \bibinfo{pages}{8494--8502}.
\newblock


\bibitem[Puig et~al\mbox{.}(2020)]%
        {puig2020watch}
\bibfield{author}{\bibinfo{person}{Xavier Puig}, \bibinfo{person}{Tianmin Shu}, \bibinfo{person}{Shuang Li}, \bibinfo{person}{Zilin Wang}, \bibinfo{person}{Yuan-Hong Liao}, \bibinfo{person}{Joshua~B Tenenbaum}, \bibinfo{person}{Sanja Fidler}, {and} \bibinfo{person}{Antonio Torralba}.} \bibinfo{year}{2020}\natexlab{}.
\newblock \showarticletitle{Watch-and-help: A challenge for social perception and human-ai collaboration}.
\newblock \bibinfo{journal}{\emph{arXiv preprint arXiv:2010.09890}} (\bibinfo{year}{2020}).
\newblock


\bibitem[Qi et~al\mbox{.}(2024)]%
        {qi2024webrl}
\bibfield{author}{\bibinfo{person}{Zehan Qi}, \bibinfo{person}{Xiao Liu}, \bibinfo{person}{Iat~Long Iong}, \bibinfo{person}{Hanyu Lai}, \bibinfo{person}{Xueqiao Sun}, \bibinfo{person}{Xinyue Yang}, \bibinfo{person}{Jiadai Sun}, \bibinfo{person}{Yu Yang}, \bibinfo{person}{Shuntian Yao}, \bibinfo{person}{Tianjie Zhang}, {et~al\mbox{.}}} \bibinfo{year}{2024}\natexlab{}.
\newblock \showarticletitle{WebRL: Training LLM Web Agents via Self-Evolving Online Curriculum Reinforcement Learning}.
\newblock \bibinfo{journal}{\emph{arXiv preprint arXiv:2411.02337}} (\bibinfo{year}{2024}).
\newblock


\bibitem[Qian et~al\mbox{.}(2025)]%
        {qian2025toolrl}
\bibfield{author}{\bibinfo{person}{Cheng Qian}, \bibinfo{person}{Emre~Can Acikgoz}, \bibinfo{person}{Qi He}, \bibinfo{person}{Hongru Wang}, \bibinfo{person}{Xiusi Chen}, \bibinfo{person}{Dilek Hakkani-T{\"u}r}, \bibinfo{person}{Gokhan Tur}, {and} \bibinfo{person}{Heng Ji}.} \bibinfo{year}{2025}\natexlab{}.
\newblock \showarticletitle{ToolRL: Reward is All Tool Learning Needs}.
\newblock \bibinfo{journal}{\emph{arXiv preprint arXiv:2504.13958}} (\bibinfo{year}{2025}).
\newblock


\bibitem[Qiao et~al\mbox{.}(2024a)]%
        {qiao2024benchmarking}
\bibfield{author}{\bibinfo{person}{Shuofei Qiao}, \bibinfo{person}{Runnan Fang}, \bibinfo{person}{Zhisong Qiu}, \bibinfo{person}{Xiaobin Wang}, \bibinfo{person}{Ningyu Zhang}, \bibinfo{person}{Yong Jiang}, \bibinfo{person}{Pengjun Xie}, \bibinfo{person}{Fei Huang}, {and} \bibinfo{person}{Huajun Chen}.} \bibinfo{year}{2024}\natexlab{a}.
\newblock \showarticletitle{Benchmarking Agentic Workflow Generation}.
\newblock \bibinfo{journal}{\emph{arXiv preprint arXiv:2410.07869}} (\bibinfo{year}{2024}).
\newblock


\bibitem[Qiao et~al\mbox{.}(2024b)]%
        {qiao2024agent}
\bibfield{author}{\bibinfo{person}{Shuofei Qiao}, \bibinfo{person}{Runnan Fang}, \bibinfo{person}{Ningyu Zhang}, \bibinfo{person}{Yuqi Zhu}, \bibinfo{person}{Xiang Chen}, \bibinfo{person}{Shumin Deng}, \bibinfo{person}{Yong Jiang}, \bibinfo{person}{Pengjun Xie}, \bibinfo{person}{Fei Huang}, {and} \bibinfo{person}{Huajun Chen}.} \bibinfo{year}{2024}\natexlab{b}.
\newblock \showarticletitle{Agent planning with world knowledge model}.
\newblock \bibinfo{journal}{\emph{Advances in Neural Information Processing Systems}}  \bibinfo{volume}{37} (\bibinfo{year}{2024}), \bibinfo{pages}{114843--114871}.
\newblock


\bibitem[Qiao et~al\mbox{.}(2025)]%
        {qiao2025agentic}
\bibfield{author}{\bibinfo{person}{Shuofei Qiao}, \bibinfo{person}{Zhisong Qiu}, \bibinfo{person}{Baochang Ren}, \bibinfo{person}{Xiaobin Wang}, \bibinfo{person}{Xiangyuan Ru}, \bibinfo{person}{Ningyu Zhang}, \bibinfo{person}{Xiang Chen}, \bibinfo{person}{Yong Jiang}, \bibinfo{person}{Pengjun Xie}, \bibinfo{person}{Fei Huang}, {et~al\mbox{.}}} \bibinfo{year}{2025}\natexlab{}.
\newblock \showarticletitle{Agentic Knowledgeable Self-awareness}.
\newblock \bibinfo{journal}{\emph{arXiv preprint arXiv:2504.03553}} (\bibinfo{year}{2025}).
\newblock


\bibitem[Qiao et~al\mbox{.}(2024c)]%
        {qiao2024autoact}
\bibfield{author}{\bibinfo{person}{Shuofei Qiao}, \bibinfo{person}{Ningyu Zhang}, \bibinfo{person}{Runnan Fang}, \bibinfo{person}{Yujie Luo}, \bibinfo{person}{Wangchunshu Zhou}, \bibinfo{person}{Yuchen~Eleanor Jiang}, \bibinfo{person}{Chengfei Lv}, {and} \bibinfo{person}{Huajun Chen}.} \bibinfo{year}{2024}\natexlab{c}.
\newblock \showarticletitle{Autoact: Automatic agent learning from scratch via self-planning}.
\newblock \bibinfo{journal}{\emph{arXiv preprint arXiv:2401.05268}} (\bibinfo{year}{2024}).
\newblock


\bibitem[Qin et~al\mbox{.}(2023)]%
        {qin2023toolllm}
\bibfield{author}{\bibinfo{person}{Yujia Qin}, \bibinfo{person}{Shihao Liang}, \bibinfo{person}{Yining Ye}, \bibinfo{person}{Kunlun Zhu}, \bibinfo{person}{Lan Yan}, \bibinfo{person}{Yaxi Lu}, \bibinfo{person}{Yankai Lin}, \bibinfo{person}{Xin Cong}, \bibinfo{person}{Xiangru Tang}, \bibinfo{person}{Bill Qian}, {et~al\mbox{.}}} \bibinfo{year}{2023}\natexlab{}.
\newblock \showarticletitle{Toolllm: Facilitating large language models to master 16000+ real-world apis}.
\newblock \bibinfo{journal}{\emph{arXiv preprint arXiv:2307.16789}} (\bibinfo{year}{2023}).
\newblock


\bibitem[Qin et~al\mbox{.}(2025)]%
        {qin2025ui}
\bibfield{author}{\bibinfo{person}{Yujia Qin}, \bibinfo{person}{Yining Ye}, \bibinfo{person}{Junjie Fang}, \bibinfo{person}{Haoming Wang}, \bibinfo{person}{Shihao Liang}, \bibinfo{person}{Shizuo Tian}, \bibinfo{person}{Junda Zhang}, \bibinfo{person}{Jiahao Li}, \bibinfo{person}{Yunxin Li}, \bibinfo{person}{Shijue Huang}, {et~al\mbox{.}}} \bibinfo{year}{2025}\natexlab{}.
\newblock \showarticletitle{UI-TARS: Pioneering Automated GUI Interaction with Native Agents}.
\newblock \bibinfo{journal}{\emph{arXiv preprint arXiv:2501.12326}} (\bibinfo{year}{2025}).
\newblock


\bibitem[Qiu et~al\mbox{.}(2024)]%
        {qiu2024optimizing}
\bibfield{author}{\bibinfo{person}{Yuli Qiu}, \bibinfo{person}{Jiashu Yao}, \bibinfo{person}{Heyan Huang}, {and} \bibinfo{person}{Yuhang Guo}.} \bibinfo{year}{2024}\natexlab{}.
\newblock \showarticletitle{Optimizing Chain-of-Thought Reasoning: Tackling Arranging Bottleneck via Plan Augmentation}.
\newblock \bibinfo{journal}{\emph{arXiv preprint arXiv:2410.16812}} (\bibinfo{year}{2024}).
\newblock


\bibitem[Qwen et~al\mbox{.}(2024)]%
        {qwen2024qwen25}
\bibfield{author}{\bibinfo{person}{Qwen}, \bibinfo{person}{:}, \bibinfo{person}{An Yang}, \bibinfo{person}{Baosong Yang}, \bibinfo{person}{Beichen Zhang}, \bibinfo{person}{Binyuan Hui}, \bibinfo{person}{Bo Zheng}, \bibinfo{person}{Bowen Yu}, \bibinfo{person}{Chengyuan Li}, \bibinfo{person}{Dayiheng Liu}, \bibinfo{person}{Fei Huang}, \bibinfo{person}{Haoran Wei}, \bibinfo{person}{Huan Lin}, \bibinfo{person}{Jian Yang}, \bibinfo{person}{Jianhong Tu}, \bibinfo{person}{Jianwei Zhang}, \bibinfo{person}{Jianxin Yang}, \bibinfo{person}{Jiaxi Yang}, \bibinfo{person}{Jingren Zhou}, \bibinfo{person}{Junyang Lin}, \bibinfo{person}{Kai Dang}, \bibinfo{person}{Keming Lu}, \bibinfo{person}{Keqin Bao}, \bibinfo{person}{Kexin Yang}, \bibinfo{person}{Le Yu}, \bibinfo{person}{Mei Li}, \bibinfo{person}{Mingfeng Xue}, \bibinfo{person}{Pei Zhang}, \bibinfo{person}{Qin Zhu}, \bibinfo{person}{Rui Men}, \bibinfo{person}{Runji Lin}, \bibinfo{person}{Tianhao Li}, \bibinfo{person}{Tianyi Tang}, \bibinfo{person}{Tingyu Xia},
  \bibinfo{person}{Xingzhang Ren}, \bibinfo{person}{Xuancheng Ren}, \bibinfo{person}{Yang Fan}, \bibinfo{person}{Yang Su}, \bibinfo{person}{Yichang Zhang}, \bibinfo{person}{Yu Wan}, \bibinfo{person}{Yuqiong Liu}, \bibinfo{person}{Zeyu Cui}, \bibinfo{person}{Zhenru Zhang}, {and} \bibinfo{person}{Zihan Qiu}.} \bibinfo{year}{2024}\natexlab{}.
\newblock \bibinfo{title}{Qwen2.5 Technical Report}.
\newblock


\bibitem[Rawles et~al\mbox{.}(2024)]%
        {rawles2024androidworld}
\bibfield{author}{\bibinfo{person}{Christopher Rawles}, \bibinfo{person}{Sarah Clinckemaillie}, \bibinfo{person}{Yifan Chang}, \bibinfo{person}{Jonathan Waltz}, \bibinfo{person}{Gabrielle Lau}, \bibinfo{person}{Marybeth Fair}, \bibinfo{person}{Alice Li}, \bibinfo{person}{William Bishop}, \bibinfo{person}{Wei Li}, \bibinfo{person}{Folawiyo Campbell-Ajala}, {et~al\mbox{.}}} \bibinfo{year}{2024}\natexlab{}.
\newblock \showarticletitle{Androidworld: A dynamic benchmarking environment for autonomous agents}.
\newblock \bibinfo{journal}{\emph{arXiv preprint arXiv:2405.14573}} (\bibinfo{year}{2024}).
\newblock


\bibitem[Rawles et~al\mbox{.}(2023)]%
        {rawles2023androidinthewild}
\bibfield{author}{\bibinfo{person}{Christopher Rawles}, \bibinfo{person}{Alice Li}, \bibinfo{person}{Daniel Rodriguez}, \bibinfo{person}{Oriana Riva}, {and} \bibinfo{person}{Timothy Lillicrap}.} \bibinfo{year}{2023}\natexlab{}.
\newblock \showarticletitle{Androidinthewild: A large-scale dataset for android device control}.
\newblock \bibinfo{journal}{\emph{Advances in Neural Information Processing Systems}}  \bibinfo{volume}{36} (\bibinfo{year}{2023}), \bibinfo{pages}{59708--59728}.
\newblock


\bibitem[Roy et~al\mbox{.}(2024)]%
        {roy2024flap}
\bibfield{author}{\bibinfo{person}{Shamik Roy}, \bibinfo{person}{Sailik Sengupta}, \bibinfo{person}{Daniele Bonadiman}, \bibinfo{person}{Saab Mansour}, {and} \bibinfo{person}{Arshit Gupta}.} \bibinfo{year}{2024}\natexlab{}.
\newblock \showarticletitle{FLAP: Flow-Adhering Planning with Constrained Decoding in LLMs}. In \bibinfo{booktitle}{\emph{Proceedings of the 2024 Conference of the North American Chapter of the Association for Computational Linguistics: Human Language Technologies (Volume 1: Long Papers)}}. \bibinfo{pages}{517--539}.
\newblock


\bibitem[Roziere et~al\mbox{.}(2023)]%
        {roziere2023code}
\bibfield{author}{\bibinfo{person}{Baptiste Roziere}, \bibinfo{person}{Jonas Gehring}, \bibinfo{person}{Fabian Gloeckle}, \bibinfo{person}{Sten Sootla}, \bibinfo{person}{Itai Gat}, \bibinfo{person}{Xiaoqing~Ellen Tan}, \bibinfo{person}{Yossi Adi}, \bibinfo{person}{Jingyu Liu}, \bibinfo{person}{Romain Sauvestre}, \bibinfo{person}{Tal Remez}, {et~al\mbox{.}}} \bibinfo{year}{2023}\natexlab{}.
\newblock \showarticletitle{Code llama: Open foundation models for code}.
\newblock \bibinfo{journal}{\emph{arXiv preprint arXiv:2308.12950}} (\bibinfo{year}{2023}).
\newblock


\bibitem[Russell and Norvig(2016)]%
        {russell2016artificial}
\bibfield{author}{\bibinfo{person}{Stuart~J Russell} {and} \bibinfo{person}{Peter Norvig}.} \bibinfo{year}{2016}\natexlab{}.
\newblock \bibinfo{booktitle}{\emph{Artificial intelligence: a modern approach}}.
\newblock \bibinfo{publisher}{pearson}.
\newblock


\bibitem[Saha et~al\mbox{.}(2024)]%
        {saha2024system}
\bibfield{author}{\bibinfo{person}{Swarnadeep Saha}, \bibinfo{person}{Archiki Prasad}, \bibinfo{person}{Justin Chih-Yao Chen}, \bibinfo{person}{Peter Hase}, \bibinfo{person}{Elias Stengel-Eskin}, {and} \bibinfo{person}{Mohit Bansal}.} \bibinfo{year}{2024}\natexlab{}.
\newblock \showarticletitle{System-1. x: Learning to balance fast and slow planning with language models}.
\newblock \bibinfo{journal}{\emph{arXiv preprint arXiv:2407.14414}} (\bibinfo{year}{2024}).
\newblock


\bibitem[Sarch et~al\mbox{.}(2023)]%
        {sarch2023open}
\bibfield{author}{\bibinfo{person}{Gabriel Sarch}, \bibinfo{person}{Yue Wu}, \bibinfo{person}{Michael Tarr}, {and} \bibinfo{person}{Katerina Fragkiadaki}.} \bibinfo{year}{2023}\natexlab{}.
\newblock \showarticletitle{Open-Ended Instructable Embodied Agents with Memory-Augmented Large Language Models}. In \bibinfo{booktitle}{\emph{Findings of the Association for Computational Linguistics: EMNLP 2023}}. \bibinfo{pages}{3468--3500}.
\newblock


\bibitem[Sel et~al\mbox{.}(2023)]%
        {sel2023algorithm}
\bibfield{author}{\bibinfo{person}{Bilgehan Sel}, \bibinfo{person}{Ahmad Al-Tawaha}, \bibinfo{person}{Vanshaj Khattar}, \bibinfo{person}{Ruoxi Jia}, {and} \bibinfo{person}{Ming Jin}.} \bibinfo{year}{2023}\natexlab{}.
\newblock \showarticletitle{Algorithm of thoughts: Enhancing exploration of ideas in large language models}.
\newblock \bibinfo{journal}{\emph{arXiv preprint arXiv:2308.10379}} (\bibinfo{year}{2023}).
\newblock


\bibitem[Shao et~al\mbox{.}(2024)]%
        {shao2024chinatravel}
\bibfield{author}{\bibinfo{person}{Jie-Jing Shao}, \bibinfo{person}{Xiao-Wen Yang}, \bibinfo{person}{Bo-Wen Zhang}, \bibinfo{person}{Baizhi Chen}, \bibinfo{person}{Wen-Da Wei}, \bibinfo{person}{Guohao Cai}, \bibinfo{person}{Zhenhua Dong}, \bibinfo{person}{Lan-Zhe Guo}, {and} \bibinfo{person}{Yu-feng Li}.} \bibinfo{year}{2024}\natexlab{}.
\newblock \showarticletitle{Chinatravel: A real-world benchmark for language agents in chinese travel planning}.
\newblock \bibinfo{journal}{\emph{arXiv preprint arXiv:2412.13682}} (\bibinfo{year}{2024}).
\newblock


\bibitem[Shen et~al\mbox{.}(2023)]%
        {shen2023hugginggpt}
\bibfield{author}{\bibinfo{person}{Yongliang Shen}, \bibinfo{person}{Kaitao Song}, \bibinfo{person}{Xu Tan}, \bibinfo{person}{Dongsheng Li}, \bibinfo{person}{Weiming Lu}, {and} \bibinfo{person}{Yueting Zhuang}.} \bibinfo{year}{2023}\natexlab{}.
\newblock \showarticletitle{Hugginggpt: Solving ai tasks with chatgpt and its friends in hugging face}.
\newblock \bibinfo{journal}{\emph{Advances in Neural Information Processing Systems}}  \bibinfo{volume}{36} (\bibinfo{year}{2023}), \bibinfo{pages}{38154--38180}.
\newblock


\bibitem[Shen et~al\mbox{.}(2024)]%
        {shen2024taskbench}
\bibfield{author}{\bibinfo{person}{Yongliang Shen}, \bibinfo{person}{Kaitao Song}, \bibinfo{person}{Xu Tan}, \bibinfo{person}{Wenqi Zhang}, \bibinfo{person}{Kan Ren}, \bibinfo{person}{Siyu Yuan}, \bibinfo{person}{Weiming Lu}, \bibinfo{person}{Dongsheng Li}, {and} \bibinfo{person}{Yueting Zhuang}.} \bibinfo{year}{2024}\natexlab{}.
\newblock \showarticletitle{Taskbench: Benchmarking large language models for task automation}.
\newblock \bibinfo{journal}{\emph{Advances in Neural Information Processing Systems}}  \bibinfo{volume}{37} (\bibinfo{year}{2024}), \bibinfo{pages}{4540--4574}.
\newblock


\bibitem[Shi et~al\mbox{.}({[n.\,d.]})]%
        {shimonte}
\bibfield{author}{\bibinfo{person}{Zijing Shi}, \bibinfo{person}{Meng Fang}, {and} \bibinfo{person}{Ling Chen}.} \bibinfo{year}{[n.\,d.]}\natexlab{}.
\newblock \showarticletitle{Monte Carlo Planning with Large Language Model for Text-Based Games}. In \bibinfo{booktitle}{\emph{The Thirteenth International Conference on Learning Representations}}.
\newblock


\bibitem[Shihab et~al\mbox{.}(2025)]%
        {shihab2025cache}
\bibfield{author}{\bibinfo{person}{Ibne~Farabi Shihab}, \bibinfo{person}{Sanjeda Akter}, {and} \bibinfo{person}{Anuj Sharma}.} \bibinfo{year}{2025}\natexlab{}.
\newblock \showarticletitle{Cache-Efficient Posterior Sampling for Reinforcement Learning with LLM-Derived Priors Across Discrete and Continuous Domains}.
\newblock \bibinfo{journal}{\emph{arXiv preprint arXiv:2505.07274}} (\bibinfo{year}{2025}).
\newblock


\bibitem[Shridhar et~al\mbox{.}(2020a)]%
        {shridhar2020alfred}
\bibfield{author}{\bibinfo{person}{Mohit Shridhar}, \bibinfo{person}{Jesse Thomason}, \bibinfo{person}{Daniel Gordon}, \bibinfo{person}{Yonatan Bisk}, \bibinfo{person}{Winson Han}, \bibinfo{person}{Roozbeh Mottaghi}, \bibinfo{person}{Luke Zettlemoyer}, {and} \bibinfo{person}{Dieter Fox}.} \bibinfo{year}{2020}\natexlab{a}.
\newblock \showarticletitle{Alfred: A benchmark for interpreting grounded instructions for everyday tasks}. In \bibinfo{booktitle}{\emph{Proceedings of the IEEE/CVF conference on computer vision and pattern recognition}}. \bibinfo{pages}{10740--10749}.
\newblock


\bibitem[Shridhar et~al\mbox{.}(2020b)]%
        {shridhar2020alfworld}
\bibfield{author}{\bibinfo{person}{Mohit Shridhar}, \bibinfo{person}{Xingdi Yuan}, \bibinfo{person}{Marc-Alexandre C{\^o}t{\'e}}, \bibinfo{person}{Yonatan Bisk}, \bibinfo{person}{Adam Trischler}, {and} \bibinfo{person}{Matthew Hausknecht}.} \bibinfo{year}{2020}\natexlab{b}.
\newblock \showarticletitle{Alfworld: Aligning text and embodied environments for interactive learning}.
\newblock \bibinfo{journal}{\emph{arXiv preprint arXiv:2010.03768}} (\bibinfo{year}{2020}).
\newblock


\bibitem[Silver et~al\mbox{.}(2017)]%
        {silver2017mastering}
\bibfield{author}{\bibinfo{person}{David Silver}, \bibinfo{person}{Thomas Hubert}, \bibinfo{person}{Julian Schrittwieser}, \bibinfo{person}{Ioannis Antonoglou}, \bibinfo{person}{Matthew Lai}, \bibinfo{person}{Arthur Guez}, \bibinfo{person}{Marc Lanctot}, \bibinfo{person}{Laurent Sifre}, \bibinfo{person}{Dharshan Kumaran}, \bibinfo{person}{Thore Graepel}, {et~al\mbox{.}}} \bibinfo{year}{2017}\natexlab{}.
\newblock \showarticletitle{Mastering chess and shogi by self-play with a general reinforcement learning algorithm}.
\newblock \bibinfo{journal}{\emph{arXiv preprint arXiv:1712.01815}} (\bibinfo{year}{2017}).
\newblock


\bibitem[Singh et~al\mbox{.}(2023)]%
        {singh2023beyond}
\bibfield{author}{\bibinfo{person}{Avi Singh}, \bibinfo{person}{John~D Co-Reyes}, \bibinfo{person}{Rishabh Agarwal}, \bibinfo{person}{Ankesh Anand}, \bibinfo{person}{Piyush Patil}, \bibinfo{person}{Xavier Garcia}, \bibinfo{person}{Peter~J Liu}, \bibinfo{person}{James Harrison}, \bibinfo{person}{Jaehoon Lee}, \bibinfo{person}{Kelvin Xu}, {et~al\mbox{.}}} \bibinfo{year}{2023}\natexlab{}.
\newblock \showarticletitle{Beyond human data: Scaling self-training for problem-solving with language models}.
\newblock \bibinfo{journal}{\emph{arXiv preprint arXiv:2312.06585}} (\bibinfo{year}{2023}).
\newblock


\bibitem[Singh et~al\mbox{.}(2024)]%
        {singh2024personal}
\bibfield{author}{\bibinfo{person}{Harmanpreet Singh}, \bibinfo{person}{Nikhil Verma}, \bibinfo{person}{Yixiao Wang}, \bibinfo{person}{Manasa Bharadwaj}, \bibinfo{person}{Homa Fashandi}, \bibinfo{person}{Kevin Ferreira}, {and} \bibinfo{person}{Chul Lee}.} \bibinfo{year}{2024}\natexlab{}.
\newblock \showarticletitle{Personal Large Language Model Agents: A Case Study on Tailored Travel Planning}. In \bibinfo{booktitle}{\emph{Proceedings of the 2024 Conference on Empirical Methods in Natural Language Processing: Industry Track}}. \bibinfo{pages}{486--514}.
\newblock


\bibitem[Sloman(1996)]%
        {sloman1996empirical}
\bibfield{author}{\bibinfo{person}{Steven~A Sloman}.} \bibinfo{year}{1996}\natexlab{}.
\newblock \showarticletitle{The empirical case for two systems of reasoning.}
\newblock \bibinfo{journal}{\emph{Psychological bulletin}} \bibinfo{volume}{119}, \bibinfo{number}{1} (\bibinfo{year}{1996}), \bibinfo{pages}{3}.
\newblock


\bibitem[Song et~al\mbox{.}(2024)]%
        {song2024trial}
\bibfield{author}{\bibinfo{person}{Yifan Song}, \bibinfo{person}{Da Yin}, \bibinfo{person}{Xiang Yue}, \bibinfo{person}{Jie Huang}, \bibinfo{person}{Sujian Li}, {and} \bibinfo{person}{Bill~Yuchen Lin}.} \bibinfo{year}{2024}\natexlab{}.
\newblock \showarticletitle{Trial and error: Exploration-based trajectory optimization for llm agents}.
\newblock \bibinfo{journal}{\emph{arXiv preprint arXiv:2403.02502}} (\bibinfo{year}{2024}).
\newblock


\bibitem[Sprague et~al\mbox{.}(2023)]%
        {sprague2023deductive}
\bibfield{author}{\bibinfo{person}{Zayne Sprague}, \bibinfo{person}{Kaj Bostrom}, \bibinfo{person}{Swarat Chaudhuri}, {and} \bibinfo{person}{Greg Durrett}.} \bibinfo{year}{2023}\natexlab{}.
\newblock \showarticletitle{Deductive additivity for planning of natural language proofs}.
\newblock \bibinfo{journal}{\emph{arXiv preprint arXiv:2307.02472}} (\bibinfo{year}{2023}).
\newblock


\bibitem[Sprague et~al\mbox{.}(2024)]%
        {sprague2024cot}
\bibfield{author}{\bibinfo{person}{Zayne Sprague}, \bibinfo{person}{Fangcong Yin}, \bibinfo{person}{Juan~Diego Rodriguez}, \bibinfo{person}{Dongwei Jiang}, \bibinfo{person}{Manya Wadhwa}, \bibinfo{person}{Prasann Singhal}, \bibinfo{person}{Xinyu Zhao}, \bibinfo{person}{Xi Ye}, \bibinfo{person}{Kyle Mahowald}, {and} \bibinfo{person}{Greg Durrett}.} \bibinfo{year}{2024}\natexlab{}.
\newblock \showarticletitle{To cot or not to cot? chain-of-thought helps mainly on math and symbolic reasoning}.
\newblock \bibinfo{journal}{\emph{arXiv preprint arXiv:2409.12183}} (\bibinfo{year}{2024}).
\newblock


\bibitem[Stanovich(1999)]%
        {stanovich1999rational}
\bibfield{author}{\bibinfo{person}{Keith~E Stanovich}.} \bibinfo{year}{1999}\natexlab{}.
\newblock \bibinfo{booktitle}{\emph{Who is rational?: Studies of individual differences in reasoning}}.
\newblock \bibinfo{publisher}{Psychology Press}.
\newblock


\bibitem[Stechly et~al\mbox{.}(2024)]%
        {stechly2024chain}
\bibfield{author}{\bibinfo{person}{Kaya Stechly}, \bibinfo{person}{Karthik Valmeekam}, {and} \bibinfo{person}{Subbarao Kambhampati}.} \bibinfo{year}{2024}\natexlab{}.
\newblock \showarticletitle{Chain of thoughtlessness: An analysis of cot in planning}.
\newblock \bibinfo{journal}{\emph{arXiv preprint arXiv:2405.04776}} (\bibinfo{year}{2024}).
\newblock


\bibitem[Stein et~al\mbox{.}(2023)]%
        {stein2023autoplanbench}
\bibfield{author}{\bibinfo{person}{Katharina Stein}, \bibinfo{person}{Daniel Fi{\v{s}}er}, \bibinfo{person}{J{\"o}rg Hoffmann}, {and} \bibinfo{person}{Alexander Koller}.} \bibinfo{year}{2023}\natexlab{}.
\newblock \showarticletitle{Autoplanbench: Automatically generating benchmarks for llm planners from pddl}.
\newblock \bibinfo{journal}{\emph{arXiv preprint arXiv:2311.09830}} (\bibinfo{year}{2023}).
\newblock


\bibitem[Su et~al\mbox{.}(2024b)]%
        {su2024dualformer}
\bibfield{author}{\bibinfo{person}{DiJia Su}, \bibinfo{person}{Sainbayar Sukhbaatar}, \bibinfo{person}{Michael Rabbat}, \bibinfo{person}{Yuandong Tian}, {and} \bibinfo{person}{Qinqing Zheng}.} \bibinfo{year}{2024}\natexlab{b}.
\newblock \showarticletitle{Dualformer: Controllable fast and slow thinking by learning with randomized reasoning traces}.
\newblock \bibinfo{journal}{\emph{arXiv preprint arXiv:2410.09918}} (\bibinfo{year}{2024}).
\newblock


\bibitem[Su et~al\mbox{.}(2025)]%
        {su2025debflow}
\bibfield{author}{\bibinfo{person}{Jinwei Su}, \bibinfo{person}{Yinghui Xia}, \bibinfo{person}{Ronghua Shi}, \bibinfo{person}{Jianhui Wang}, \bibinfo{person}{Jianuo Huang}, \bibinfo{person}{Yijin Wang}, \bibinfo{person}{Tianyu Shi}, \bibinfo{person}{Yang Jingsong}, {and} \bibinfo{person}{Lewei He}.} \bibinfo{year}{2025}\natexlab{}.
\newblock \showarticletitle{DebFlow: Automating Agent Creation via Agent Debate}.
\newblock \bibinfo{journal}{\emph{arXiv preprint arXiv:2503.23781}} (\bibinfo{year}{2025}).
\newblock


\bibitem[Su et~al\mbox{.}(2024a)]%
        {su2024actplan}
\bibfield{author}{\bibinfo{person}{Ying Su}, \bibinfo{person}{Zhan Ling}, \bibinfo{person}{Haochen Shi}, \bibinfo{person}{Jiayang Cheng}, \bibinfo{person}{Yauwai Yim}, {and} \bibinfo{person}{Yangqiu Song}.} \bibinfo{year}{2024}\natexlab{a}.
\newblock \showarticletitle{Actplan-1k: Benchmarking the procedural planning ability of visual language models in household activities}.
\newblock \bibinfo{journal}{\emph{arXiv preprint arXiv:2410.03907}} (\bibinfo{year}{2024}).
\newblock


\bibitem[Sukhbaatar et~al\mbox{.}(2015)]%
        {sukhbaatar2015end}
\bibfield{author}{\bibinfo{person}{Sainbayar Sukhbaatar}, \bibinfo{person}{Jason Weston}, \bibinfo{person}{Rob Fergus}, {et~al\mbox{.}}} \bibinfo{year}{2015}\natexlab{}.
\newblock \showarticletitle{End-to-end memory networks}.
\newblock \bibinfo{journal}{\emph{Advances in neural information processing systems}}  \bibinfo{volume}{28} (\bibinfo{year}{2015}).
\newblock


\bibitem[Sun et~al\mbox{.}(2024)]%
        {sun2024retrieval}
\bibfield{author}{\bibinfo{person}{Chuanneng Sun}, \bibinfo{person}{Songjun Huang}, {and} \bibinfo{person}{Dario Pompili}.} \bibinfo{year}{2024}\natexlab{}.
\newblock \showarticletitle{Retrieval-Augmented Hierarchical in-Context Reinforcement Learning and Hindsight Modular Reflections for Task Planning with LLMs}.
\newblock \bibinfo{journal}{\emph{arXiv preprint arXiv:2408.06520}} (\bibinfo{year}{2024}).
\newblock


\bibitem[Sun et~al\mbox{.}(2023)]%
        {sun2023adaplanner}
\bibfield{author}{\bibinfo{person}{Haotian Sun}, \bibinfo{person}{Yuchen Zhuang}, \bibinfo{person}{Lingkai Kong}, \bibinfo{person}{Bo Dai}, {and} \bibinfo{person}{Chao Zhang}.} \bibinfo{year}{2023}\natexlab{}.
\newblock \showarticletitle{Adaplanner: Adaptive planning from feedback with language models}.
\newblock \bibinfo{journal}{\emph{Advances in neural information processing systems}}  \bibinfo{volume}{36} (\bibinfo{year}{2023}), \bibinfo{pages}{58202--58245}.
\newblock


\bibitem[Sun et~al\mbox{.}(2022)]%
        {sun2022meta}
\bibfield{author}{\bibinfo{person}{Liangtai Sun}, \bibinfo{person}{Xingyu Chen}, \bibinfo{person}{Lu Chen}, \bibinfo{person}{Tianle Dai}, \bibinfo{person}{Zichen Zhu}, {and} \bibinfo{person}{Kai Yu}.} \bibinfo{year}{2022}\natexlab{}.
\newblock \showarticletitle{Meta-gui: Towards multi-modal conversational agents on mobile gui}.
\newblock \bibinfo{journal}{\emph{arXiv preprint arXiv:2205.11029}} (\bibinfo{year}{2022}).
\newblock


\bibitem[Sun et~al\mbox{.}(2025)]%
        {sun2025error}
\bibfield{author}{\bibinfo{person}{Yuhong Sun}, \bibinfo{person}{Zhangyue Yin}, \bibinfo{person}{Xuanjing Huang}, \bibinfo{person}{Xipeng Qiu}, {and} \bibinfo{person}{Hui Zhao}.} \bibinfo{year}{2025}\natexlab{}.
\newblock \showarticletitle{Error Classification of Large Language Models on Math Word Problems: A Dynamically Adaptive Framework}.
\newblock \bibinfo{journal}{\emph{arXiv preprint arXiv:2501.15581}} (\bibinfo{year}{2025}).
\newblock


\bibitem[Tang et~al\mbox{.}(2025)]%
        {tang2025worldcoder}
\bibfield{author}{\bibinfo{person}{Hao Tang}, \bibinfo{person}{Darren Key}, {and} \bibinfo{person}{Kevin Ellis}.} \bibinfo{year}{2025}\natexlab{}.
\newblock \showarticletitle{Worldcoder, a model-based llm agent: Building world models by writing code and interacting with the environment}.
\newblock \bibinfo{journal}{\emph{Advances in Neural Information Processing Systems}}  \bibinfo{volume}{37} (\bibinfo{year}{2025}), \bibinfo{pages}{70148--70212}.
\newblock


\bibitem[Tang et~al\mbox{.}(2024)]%
        {tang2024itinera}
\bibfield{author}{\bibinfo{person}{Yihong Tang}, \bibinfo{person}{Zhaokai Wang}, \bibinfo{person}{Ao Qu}, \bibinfo{person}{Yihao Yan}, \bibinfo{person}{Zhaofeng Wu}, \bibinfo{person}{Dingyi Zhuang}, \bibinfo{person}{Jushi Kai}, \bibinfo{person}{Kebing Hou}, \bibinfo{person}{Xiaotong Guo}, \bibinfo{person}{Jinhua Zhao}, {et~al\mbox{.}}} \bibinfo{year}{2024}\natexlab{}.
\newblock \showarticletitle{ITINERA: Integrating Spatial Optimization with Large Language Models for Open-domain Urban Itinerary Planning}. In \bibinfo{booktitle}{\emph{Proceedings of the 2024 Conference on Empirical Methods in Natural Language Processing: Industry Track}}. \bibinfo{pages}{1413--1432}.
\newblock


\bibitem[Tantakoun et~al\mbox{.}(2025)]%
        {tantakoun2025llms}
\bibfield{author}{\bibinfo{person}{Marcus Tantakoun}, \bibinfo{person}{Xiaodan Zhu}, {and} \bibinfo{person}{Christian Muise}.} \bibinfo{year}{2025}\natexlab{}.
\newblock \showarticletitle{Llms as planning modelers: A survey for leveraging large language models to construct automated planning models}.
\newblock \bibinfo{journal}{\emph{arXiv preprint arXiv:2503.18971}} (\bibinfo{year}{2025}).
\newblock


\bibitem[Team et~al\mbox{.}(2024)]%
        {team2024gemma}
\bibfield{author}{\bibinfo{person}{Gemma Team}, \bibinfo{person}{Thomas Mesnard}, \bibinfo{person}{Cassidy Hardin}, \bibinfo{person}{Robert Dadashi}, \bibinfo{person}{Surya Bhupatiraju}, \bibinfo{person}{Shreya Pathak}, \bibinfo{person}{Laurent Sifre}, \bibinfo{person}{Morgane Rivi{\`e}re}, \bibinfo{person}{Mihir~Sanjay Kale}, \bibinfo{person}{Juliette Love}, {et~al\mbox{.}}} \bibinfo{year}{2024}\natexlab{}.
\newblock \showarticletitle{Gemma: Open models based on gemini research and technology}.
\newblock \bibinfo{journal}{\emph{arXiv preprint arXiv:2403.08295}} (\bibinfo{year}{2024}).
\newblock


\bibitem[Touvron et~al\mbox{.}(2023)]%
        {touvron2023llama}
\bibfield{author}{\bibinfo{person}{Hugo Touvron}, \bibinfo{person}{Louis Martin}, \bibinfo{person}{Kevin Stone}, \bibinfo{person}{Peter Albert}, \bibinfo{person}{Amjad Almahairi}, \bibinfo{person}{Yasmine Babaei}, \bibinfo{person}{Nikolay Bashlykov}, \bibinfo{person}{Soumya Batra}, \bibinfo{person}{Prajjwal Bhargava}, \bibinfo{person}{Shruti Bhosale}, {et~al\mbox{.}}} \bibinfo{year}{2023}\natexlab{}.
\newblock \showarticletitle{Llama 2: Open foundation and fine-tuned chat models}.
\newblock \bibinfo{journal}{\emph{arXiv preprint arXiv:2307.09288}} (\bibinfo{year}{2023}).
\newblock


\bibitem[Trivedi et~al\mbox{.}(2024)]%
        {trivedi2024appworld}
\bibfield{author}{\bibinfo{person}{Harsh Trivedi}, \bibinfo{person}{Tushar Khot}, \bibinfo{person}{Mareike Hartmann}, \bibinfo{person}{Ruskin Manku}, \bibinfo{person}{Vinty Dong}, \bibinfo{person}{Edward Li}, \bibinfo{person}{Shashank Gupta}, \bibinfo{person}{Ashish Sabharwal}, {and} \bibinfo{person}{Niranjan Balasubramanian}.} \bibinfo{year}{2024}\natexlab{}.
\newblock \showarticletitle{AppWorld: A Controllable World of Apps and People for Benchmarking Interactive Coding Agents}. In \bibinfo{booktitle}{\emph{Proceedings of the 62nd Annual Meeting of the Association for Computational Linguistics (Volume 1: Long Papers)}}. \bibinfo{pages}{16022--16076}.
\newblock


\bibitem[Uesato et~al\mbox{.}(2022)]%
        {uesato2022solving}
\bibfield{author}{\bibinfo{person}{Jonathan Uesato}, \bibinfo{person}{Nate Kushman}, \bibinfo{person}{Ramana Kumar}, \bibinfo{person}{Francis Song}, \bibinfo{person}{Noah Siegel}, \bibinfo{person}{Lisa Wang}, \bibinfo{person}{Antonia Creswell}, \bibinfo{person}{Geoffrey Irving}, {and} \bibinfo{person}{Irina Higgins}.} \bibinfo{year}{2022}\natexlab{}.
\newblock \showarticletitle{Solving math word problems with process-and outcome-based feedback}.
\newblock \bibinfo{journal}{\emph{arXiv preprint arXiv:2211.14275}} (\bibinfo{year}{2022}).
\newblock


\bibitem[Uzunoglu et~al\mbox{.}(2024)]%
        {uzunoglu2024paradise}
\bibfield{author}{\bibinfo{person}{Arda Uzunoglu}, \bibinfo{person}{Abdalfatah~Rashid Safa}, {and} \bibinfo{person}{G{\"o}zde~G{\"u}l {\c{S}}ahin}.} \bibinfo{year}{2024}\natexlab{}.
\newblock \showarticletitle{Paradise: Evaluating implicit planning skills of language models with procedural warnings and tips dataset}.
\newblock \bibinfo{journal}{\emph{arXiv preprint arXiv:2403.03167}} (\bibinfo{year}{2024}).
\newblock


\bibitem[Valmeekam et~al\mbox{.}(2023a)]%
        {valmeekam2023planbench}
\bibfield{author}{\bibinfo{person}{Karthik Valmeekam}, \bibinfo{person}{Matthew Marquez}, \bibinfo{person}{Alberto Olmo}, \bibinfo{person}{Sarath Sreedharan}, {and} \bibinfo{person}{Subbarao Kambhampati}.} \bibinfo{year}{2023}\natexlab{a}.
\newblock \showarticletitle{Planbench: An extensible benchmark for evaluating large language models on planning and reasoning about change}.
\newblock \bibinfo{journal}{\emph{Advances in Neural Information Processing Systems}}  \bibinfo{volume}{36} (\bibinfo{year}{2023}), \bibinfo{pages}{38975--38987}.
\newblock


\bibitem[Valmeekam et~al\mbox{.}(2023b)]%
        {valmeekam2023planning}
\bibfield{author}{\bibinfo{person}{Karthik Valmeekam}, \bibinfo{person}{Sarath Sreedharan}, \bibinfo{person}{Matthew Marquez}, \bibinfo{person}{Alberto Olmo}, {and} \bibinfo{person}{Subbarao Kambhampati}.} \bibinfo{year}{2023}\natexlab{b}.
\newblock \showarticletitle{On the planning abilities of large language models (a critical investigation with a proposed benchmark)}.
\newblock \bibinfo{journal}{\emph{arXiv preprint arXiv:2302.06706}} (\bibinfo{year}{2023}).
\newblock


\bibitem[Verma et~al\mbox{.}(2024)]%
        {verma2024plan}
\bibfield{author}{\bibinfo{person}{Prakhar Verma}, \bibinfo{person}{Sukruta~Prakash Midigeshi}, \bibinfo{person}{Gaurav Sinha}, \bibinfo{person}{Arno Solin}, \bibinfo{person}{Nagarajan Natarajan}, {and} \bibinfo{person}{Amit Sharma}.} \bibinfo{year}{2024}\natexlab{}.
\newblock \showarticletitle{Plan-RAG: Planning-guided Retrieval Augmented Generation}.
\newblock  (\bibinfo{year}{2024}).
\newblock


\bibitem[Wang et~al\mbox{.}(2024f)]%
        {wang2024llms}
\bibfield{author}{\bibinfo{person}{Boshi Wang}, \bibinfo{person}{Hao Fang}, \bibinfo{person}{Jason Eisner}, \bibinfo{person}{Benjamin Van~Durme}, {and} \bibinfo{person}{Yu Su}.} \bibinfo{year}{2024}\natexlab{f}.
\newblock \showarticletitle{LLMs in the Imaginarium: tool learning through simulated trial and error}.
\newblock \bibinfo{journal}{\emph{arXiv preprint arXiv:2403.04746}} (\bibinfo{year}{2024}).
\newblock


\bibitem[Wang et~al\mbox{.}(2024e)]%
        {wang2024q}
\bibfield{author}{\bibinfo{person}{Chaojie Wang}, \bibinfo{person}{Yanchen Deng}, \bibinfo{person}{Zhiyi Lyu}, \bibinfo{person}{Liang Zeng}, \bibinfo{person}{Jujie He}, \bibinfo{person}{Shuicheng Yan}, {and} \bibinfo{person}{Bo An}.} \bibinfo{year}{2024}\natexlab{e}.
\newblock \showarticletitle{Q*: Improving multi-step reasoning for llms with deliberative planning}.
\newblock \bibinfo{journal}{\emph{arXiv preprint arXiv:2406.14283}} (\bibinfo{year}{2024}).
\newblock


\bibitem[Wang et~al\mbox{.}(2024o)]%
        {wang2024cooperative}
\bibfield{author}{\bibinfo{person}{Danqing Wang}, \bibinfo{person}{Zhuorui Ye}, \bibinfo{person}{Fei Fang}, {and} \bibinfo{person}{Lei Li}.} \bibinfo{year}{2024}\natexlab{o}.
\newblock \showarticletitle{Cooperative Strategic Planning Enhances Reasoning Capabilities in Large Language Models}.
\newblock \bibinfo{journal}{\emph{arXiv preprint arXiv:2410.20007}} (\bibinfo{year}{2024}).
\newblock


\bibitem[Wang et~al\mbox{.}(2025b)]%
        {wang2025otc}
\bibfield{author}{\bibinfo{person}{Hongru Wang}, \bibinfo{person}{Cheng Qian}, \bibinfo{person}{Wanjun Zhong}, \bibinfo{person}{Xiusi Chen}, \bibinfo{person}{Jiahao Qiu}, \bibinfo{person}{Shijue Huang}, \bibinfo{person}{Bowen Jin}, \bibinfo{person}{Mengdi Wang}, \bibinfo{person}{Kam-Fai Wong}, {and} \bibinfo{person}{Heng Ji}.} \bibinfo{year}{2025}\natexlab{b}.
\newblock \showarticletitle{OTC: Optimal Tool Calls via Reinforcement Learning}.
\newblock \bibinfo{journal}{\emph{arXiv preprint arXiv:2504.14870}} (\bibinfo{year}{2025}).
\newblock


\bibitem[Wang et~al\mbox{.}(2024d)]%
        {wang2024learning}
\bibfield{author}{\bibinfo{person}{Junjie Wang}, \bibinfo{person}{Mingyang Chen}, \bibinfo{person}{Binbin Hu}, \bibinfo{person}{Dan Yang}, \bibinfo{person}{Ziqi Liu}, \bibinfo{person}{Yue Shen}, \bibinfo{person}{Peng Wei}, \bibinfo{person}{Zhiqiang Zhang}, \bibinfo{person}{Jinjie Gu}, \bibinfo{person}{Jun Zhou}, {et~al\mbox{.}}} \bibinfo{year}{2024}\natexlab{d}.
\newblock \showarticletitle{Learning to Plan for Retrieval-Augmented Large Language Models from Knowledge Graphs}.
\newblock \bibinfo{journal}{\emph{arXiv preprint arXiv:2406.14282}} (\bibinfo{year}{2024}).
\newblock


\bibitem[Wang et~al\mbox{.}(2024m)]%
        {wang2024mixture}
\bibfield{author}{\bibinfo{person}{Junlin Wang}, \bibinfo{person}{Jue Wang}, \bibinfo{person}{Ben Athiwaratkun}, \bibinfo{person}{Ce Zhang}, {and} \bibinfo{person}{James Zou}.} \bibinfo{year}{2024}\natexlab{m}.
\newblock \showarticletitle{Mixture-of-Agents Enhances Large Language Model Capabilities}.
\newblock \bibinfo{journal}{\emph{arXiv preprint arXiv:2406.04692}} (\bibinfo{year}{2024}).
\newblock


\bibitem[Wang et~al\mbox{.}(2024g)]%
        {wang2024planning}
\bibfield{author}{\bibinfo{person}{Kevin Wang}, \bibinfo{person}{Junbo Li}, \bibinfo{person}{Neel~P Bhatt}, \bibinfo{person}{Yihan Xi}, \bibinfo{person}{Qiang Liu}, \bibinfo{person}{Ufuk Topcu}, {and} \bibinfo{person}{Zhangyang Wang}.} \bibinfo{year}{2024}\natexlab{g}.
\newblock \showarticletitle{On The Planning Abilities of OpenAI's o1 Models: Feasibility, Optimality, and Generalizability}.
\newblock \bibinfo{journal}{\emph{arXiv preprint arXiv:2409.19924}} (\bibinfo{year}{2024}).
\newblock


\bibitem[Wang et~al\mbox{.}(2024i)]%
        {wang2024adapting}
\bibfield{author}{\bibinfo{person}{Kuan Wang}, \bibinfo{person}{Yadong Lu}, \bibinfo{person}{Michael Santacroce}, \bibinfo{person}{Yeyun Gong}, \bibinfo{person}{Chao Zhang}, {et~al\mbox{.}}} \bibinfo{year}{2024}\natexlab{i}.
\newblock \showarticletitle{Adapting llm agents with universal feedback in communication}. In \bibinfo{booktitle}{\emph{ICML 2024 Workshop on Foundation Models in the Wild}}.
\newblock


\bibitem[Wang et~al\mbox{.}(2023b)]%
        {wang2023adapting}
\bibfield{author}{\bibinfo{person}{Kuan Wang}, \bibinfo{person}{Yadong Lu}, \bibinfo{person}{Michael Santacroce}, \bibinfo{person}{Yeyun Gong}, \bibinfo{person}{Chao Zhang}, {and} \bibinfo{person}{Yelong Shen}.} \bibinfo{year}{2023}\natexlab{b}.
\newblock \bibinfo{title}{Adapting LLM Agents with Universal Feedback in Communication}.
\newblock


\bibitem[Wang et~al\mbox{.}(2024k)]%
        {wang2024survey}
\bibfield{author}{\bibinfo{person}{Lei Wang}, \bibinfo{person}{Chen Ma}, \bibinfo{person}{Xueyang Feng}, \bibinfo{person}{Zeyu Zhang}, \bibinfo{person}{Hao Yang}, \bibinfo{person}{Jingsen Zhang}, \bibinfo{person}{Zhiyuan Chen}, \bibinfo{person}{Jiakai Tang}, \bibinfo{person}{Xu Chen}, \bibinfo{person}{Yankai Lin}, {et~al\mbox{.}}} \bibinfo{year}{2024}\natexlab{k}.
\newblock \showarticletitle{A survey on large language model based autonomous agents}.
\newblock \bibinfo{journal}{\emph{Frontiers of Computer Science}} \bibinfo{volume}{18}, \bibinfo{number}{6} (\bibinfo{year}{2024}), \bibinfo{pages}{186345}.
\newblock


\bibitem[Wang et~al\mbox{.}(2024a)]%
        {wang2024qwen2}
\bibfield{author}{\bibinfo{person}{Peng Wang}, \bibinfo{person}{Shuai Bai}, \bibinfo{person}{Sinan Tan}, \bibinfo{person}{Shijie Wang}, \bibinfo{person}{Zhihao Fan}, \bibinfo{person}{Jinze Bai}, \bibinfo{person}{Keqin Chen}, \bibinfo{person}{Xuejing Liu}, \bibinfo{person}{Jialin Wang}, \bibinfo{person}{Wenbin Ge}, {et~al\mbox{.}}} \bibinfo{year}{2024}\natexlab{a}.
\newblock \showarticletitle{Qwen2-vl: Enhancing vision-language model's perception of the world at any resolution}.
\newblock \bibinfo{journal}{\emph{arXiv preprint arXiv:2409.12191}} (\bibinfo{year}{2024}).
\newblock


\bibitem[Wang et~al\mbox{.}(2022)]%
        {wang2022scienceworld}
\bibfield{author}{\bibinfo{person}{Ruoyao Wang}, \bibinfo{person}{Peter Jansen}, \bibinfo{person}{Marc-Alexandre C{\^o}t{\'e}}, {and} \bibinfo{person}{Prithviraj Ammanabrolu}.} \bibinfo{year}{2022}\natexlab{}.
\newblock \showarticletitle{Scienceworld: Is your agent smarter than a 5th grader?}
\newblock \bibinfo{journal}{\emph{arXiv preprint arXiv:2203.07540}} (\bibinfo{year}{2022}).
\newblock


\bibitem[Wang et~al\mbox{.}(2024l)]%
        {wang2024alpine}
\bibfield{author}{\bibinfo{person}{Siwei Wang}, \bibinfo{person}{Yifei Shen}, \bibinfo{person}{Shi Feng}, \bibinfo{person}{Haoran Sun}, \bibinfo{person}{Shang-Hua Teng}, {and} \bibinfo{person}{Wei Chen}.} \bibinfo{year}{2024}\natexlab{l}.
\newblock \showarticletitle{ALPINE: Unveiling the Planning Capability of Autoregressive Learning in Language Models}.
\newblock \bibinfo{journal}{\emph{arXiv preprint arXiv:2405.09220}} (\bibinfo{year}{2024}).
\newblock


\bibitem[Wang et~al\mbox{.}(2024c)]%
        {wang2024cpl}
\bibfield{author}{\bibinfo{person}{Tianlong Wang}, \bibinfo{person}{Junzhe Chen}, \bibinfo{person}{Xueting Han}, {and} \bibinfo{person}{Jing Bai}.} \bibinfo{year}{2024}\natexlab{c}.
\newblock \showarticletitle{CPL: Critical Plan Step Learning Boosts LLM Generalization in Reasoning Tasks}.
\newblock \bibinfo{journal}{\emph{arXiv preprint arXiv:2409.08642}} (\bibinfo{year}{2024}).
\newblock


\bibitem[Wang et~al\mbox{.}(2024j)]%
        {wang2024cogvlm}
\bibfield{author}{\bibinfo{person}{Weihan Wang}, \bibinfo{person}{Qingsong Lv}, \bibinfo{person}{Wenmeng Yu}, \bibinfo{person}{Wenyi Hong}, \bibinfo{person}{Ji Qi}, \bibinfo{person}{Yan Wang}, \bibinfo{person}{Junhui Ji}, \bibinfo{person}{Zhuoyi Yang}, \bibinfo{person}{Lei Zhao}, \bibinfo{person}{Song XiXuan}, {et~al\mbox{.}}} \bibinfo{year}{2024}\natexlab{j}.
\newblock \showarticletitle{Cogvlm: Visual expert for pretrained language models}.
\newblock \bibinfo{journal}{\emph{Advances in Neural Information Processing Systems}}  \bibinfo{volume}{37} (\bibinfo{year}{2024}), \bibinfo{pages}{121475--121499}.
\newblock


\bibitem[Wang et~al\mbox{.}(2023a)]%
        {wang2023ldm2}
\bibfield{author}{\bibinfo{person}{Xingjin Wang}, \bibinfo{person}{Linjing Li}, {and} \bibinfo{person}{Daniel Zeng}.} \bibinfo{year}{2023}\natexlab{a}.
\newblock \showarticletitle{LDM2: A Large Decision Model Imitating Human Cognition with Dynamic Memory Enhancement}. In \bibinfo{booktitle}{\emph{Findings of the Association for Computational Linguistics: EMNLP 2023}}. \bibinfo{pages}{4660--4681}.
\newblock


\bibitem[Wang and Liu(2024)]%
        {wang2024oscar}
\bibfield{author}{\bibinfo{person}{Xiaoqiang Wang} {and} \bibinfo{person}{Bang Liu}.} \bibinfo{year}{2024}\natexlab{}.
\newblock \showarticletitle{Oscar: Operating system control via state-aware reasoning and re-planning}.
\newblock \bibinfo{journal}{\emph{arXiv preprint arXiv:2410.18963}} (\bibinfo{year}{2024}).
\newblock


\bibitem[Wang and Zhou(2024)]%
        {wang2024chain}
\bibfield{author}{\bibinfo{person}{Xuezhi Wang} {and} \bibinfo{person}{Denny Zhou}.} \bibinfo{year}{2024}\natexlab{}.
\newblock \showarticletitle{Chain-of-thought reasoning without prompting}.
\newblock \bibinfo{journal}{\emph{arXiv preprint arXiv:2402.10200}} (\bibinfo{year}{2024}).
\newblock


\bibitem[Wang et~al\mbox{.}(2024n)]%
        {wang2024theoretical}
\bibfield{author}{\bibinfo{person}{Yifei Wang}, \bibinfo{person}{Yuyang Wu}, \bibinfo{person}{Zeming Wei}, \bibinfo{person}{Stefanie Jegelka}, {and} \bibinfo{person}{Yisen Wang}.} \bibinfo{year}{2024}\natexlab{n}.
\newblock \showarticletitle{A Theoretical Understanding of Self-Correction through In-context Alignment}.
\newblock \bibinfo{journal}{\emph{arXiv preprint arXiv:2405.18634}} (\bibinfo{year}{2024}).
\newblock


\bibitem[Wang et~al\mbox{.}(2024b)]%
        {wang2024jarvis}
\bibfield{author}{\bibinfo{person}{Zihao Wang}, \bibinfo{person}{Shaofei Cai}, \bibinfo{person}{Anji Liu}, \bibinfo{person}{Yonggang Jin}, \bibinfo{person}{Jinbing Hou}, \bibinfo{person}{Bowei Zhang}, \bibinfo{person}{Haowei Lin}, \bibinfo{person}{Zhaofeng He}, \bibinfo{person}{Zilong Zheng}, \bibinfo{person}{Yaodong Yang}, {et~al\mbox{.}}} \bibinfo{year}{2024}\natexlab{b}.
\newblock \showarticletitle{Jarvis-1: Open-world multi-task agents with memory-augmented multimodal language models}.
\newblock \bibinfo{journal}{\emph{IEEE Transactions on Pattern Analysis and Machine Intelligence}} (\bibinfo{year}{2024}).
\newblock


\bibitem[Wang et~al\mbox{.}(2025a)]%
        {wang2025mp}
\bibfield{author}{\bibinfo{person}{Ziwei Wang}, \bibinfo{person}{Weizhi Chen}, \bibinfo{person}{Leyang Yang}, \bibinfo{person}{Sheng Zhou}, \bibinfo{person}{Shengchu Zhao}, \bibinfo{person}{Hanbei Zhan}, \bibinfo{person}{Jiongchao Jin}, \bibinfo{person}{Liangcheng Li}, \bibinfo{person}{Zirui Shao}, {and} \bibinfo{person}{Jiajun Bu}.} \bibinfo{year}{2025}\natexlab{a}.
\newblock \showarticletitle{MP-GUI: Modality Perception with MLLMs for GUI Understanding}.
\newblock \bibinfo{journal}{\emph{arXiv preprint arXiv:2503.14021}} (\bibinfo{year}{2025}).
\newblock


\bibitem[Wang et~al\mbox{.}(2024h)]%
        {wang2024crafting}
\bibfield{author}{\bibinfo{person}{Zheng Wang}, \bibinfo{person}{Zhongyang Li}, \bibinfo{person}{Zeren Jiang}, \bibinfo{person}{Dandan Tu}, {and} \bibinfo{person}{Wei Shi}.} \bibinfo{year}{2024}\natexlab{h}.
\newblock \showarticletitle{Crafting Personalized Agents through Retrieval-Augmented Generation on Editable Memory Graphs}. In \bibinfo{booktitle}{\emph{Proceedings of the 2024 Conference on Empirical Methods in Natural Language Processing}}. \bibinfo{pages}{4891--4906}.
\newblock


\bibitem[Wei et~al\mbox{.}(2025b)]%
        {wei2025plangenllms}
\bibfield{author}{\bibinfo{person}{Hui Wei}, \bibinfo{person}{Zihao Zhang}, \bibinfo{person}{Shenghua He}, \bibinfo{person}{Tian Xia}, \bibinfo{person}{Shijia Pan}, {and} \bibinfo{person}{Fei Liu}.} \bibinfo{year}{2025}\natexlab{b}.
\newblock \showarticletitle{PlanGenLLMs: A Modern Survey of LLM Planning Capabilities}.
\newblock \bibinfo{journal}{\emph{arXiv preprint arXiv:2502.11221}} (\bibinfo{year}{2025}).
\newblock


\bibitem[Wei et~al\mbox{.}(2022)]%
        {wei2022chain}
\bibfield{author}{\bibinfo{person}{Jason Wei}, \bibinfo{person}{Xuezhi Wang}, \bibinfo{person}{Dale Schuurmans}, \bibinfo{person}{Maarten Bosma}, \bibinfo{person}{Fei Xia}, \bibinfo{person}{Ed Chi}, \bibinfo{person}{Quoc~V Le}, \bibinfo{person}{Denny Zhou}, {et~al\mbox{.}}} \bibinfo{year}{2022}\natexlab{}.
\newblock \showarticletitle{Chain-of-thought prompting elicits reasoning in large language models}.
\newblock \bibinfo{journal}{\emph{Advances in neural information processing systems}}  \bibinfo{volume}{35} (\bibinfo{year}{2022}), \bibinfo{pages}{24824--24837}.
\newblock


\bibitem[Wei et~al\mbox{.}(2025a)]%
        {wei2025swe}
\bibfield{author}{\bibinfo{person}{Yuxiang Wei}, \bibinfo{person}{Olivier Duchenne}, \bibinfo{person}{Jade Copet}, \bibinfo{person}{Quentin Carbonneaux}, \bibinfo{person}{Lingming Zhang}, \bibinfo{person}{Daniel Fried}, \bibinfo{person}{Gabriel Synnaeve}, \bibinfo{person}{Rishabh Singh}, {and} \bibinfo{person}{Sida~I Wang}.} \bibinfo{year}{2025}\natexlab{a}.
\newblock \showarticletitle{SWE-RL: Advancing LLM Reasoning via Reinforcement Learning on Open Software Evolution}.
\newblock \bibinfo{journal}{\emph{arXiv preprint arXiv:2502.18449}} (\bibinfo{year}{2025}).
\newblock


\bibitem[Weng et~al\mbox{.}(2024)]%
        {weng2024cycleresearcher}
\bibfield{author}{\bibinfo{person}{Yixuan Weng}, \bibinfo{person}{Minjun Zhu}, \bibinfo{person}{Guangsheng Bao}, \bibinfo{person}{Hongbo Zhang}, \bibinfo{person}{Jindong Wang}, \bibinfo{person}{Yue Zhang}, {and} \bibinfo{person}{Linyi Yang}.} \bibinfo{year}{2024}\natexlab{}.
\newblock \showarticletitle{Cycleresearcher: Improving automated research via automated review}.
\newblock \bibinfo{journal}{\emph{arXiv preprint arXiv:2411.00816}} (\bibinfo{year}{2024}).
\newblock


\bibitem[Wong et~al\mbox{.}(2023)]%
        {wong2023learning}
\bibfield{author}{\bibinfo{person}{Lionel Wong}, \bibinfo{person}{Jiayuan Mao}, \bibinfo{person}{Pratyusha Sharma}, \bibinfo{person}{Zachary~S Siegel}, \bibinfo{person}{Jiahai Feng}, \bibinfo{person}{Noa Korneev}, \bibinfo{person}{Joshua~B Tenenbaum}, {and} \bibinfo{person}{Jacob Andreas}.} \bibinfo{year}{2023}\natexlab{}.
\newblock \showarticletitle{Learning adaptive planning representations with natural language guidance}.
\newblock \bibinfo{journal}{\emph{arXiv preprint arXiv:2312.08566}} (\bibinfo{year}{2023}).
\newblock


\bibitem[Wu and Mitra(2024)]%
        {wu2024can}
\bibfield{author}{\bibinfo{person}{Erik Wu} {and} \bibinfo{person}{Sayan Mitra}.} \bibinfo{year}{2024}\natexlab{}.
\newblock \showarticletitle{Can LLMs plan paths with extra hints from solvers?}
\newblock \bibinfo{journal}{\emph{arXiv preprint arXiv:2410.05045}} (\bibinfo{year}{2024}).
\newblock


\bibitem[Wu et~al\mbox{.}(2024b)]%
        {wu2024toolplanner}
\bibfield{author}{\bibinfo{person}{Qinzhuo Wu}, \bibinfo{person}{Wei Liu}, \bibinfo{person}{Jian Luan}, {and} \bibinfo{person}{Bin Wang}.} \bibinfo{year}{2024}\natexlab{b}.
\newblock \showarticletitle{ToolPlanner: A Tool Augmented LLM for Multi Granularity Instructions with Path Planning and Feedback}. In \bibinfo{booktitle}{\emph{Proceedings of the 2024 Conference on Empirical Methods in Natural Language Processing}}. \bibinfo{pages}{18315--18339}.
\newblock


\bibitem[Wu et~al\mbox{.}(2024f)]%
        {wu2024vsp}
\bibfield{author}{\bibinfo{person}{Qiucheng Wu}, \bibinfo{person}{Handong Zhao}, \bibinfo{person}{Michael Saxon}, \bibinfo{person}{Trung Bui}, \bibinfo{person}{William~Yang Wang}, \bibinfo{person}{Yang Zhang}, {and} \bibinfo{person}{Shiyu Chang}.} \bibinfo{year}{2024}\natexlab{f}.
\newblock \showarticletitle{Vsp: Assessing the dual challenges of perception and reasoning in spatial planning tasks for vlms}.
\newblock \bibinfo{journal}{\emph{arXiv preprint arXiv:2407.01863}} (\bibinfo{year}{2024}).
\newblock


\bibitem[Wu et~al\mbox{.}(2024a)]%
        {wu2024thinking}
\bibfield{author}{\bibinfo{person}{Tianhao Wu}, \bibinfo{person}{Janice Lan}, \bibinfo{person}{Weizhe Yuan}, \bibinfo{person}{Jiantao Jiao}, \bibinfo{person}{Jason Weston}, {and} \bibinfo{person}{Sainbayar Sukhbaatar}.} \bibinfo{year}{2024}\natexlab{a}.
\newblock \showarticletitle{Thinking LLMs: General Instruction Following with Thought Generation}.
\newblock \bibinfo{journal}{\emph{arXiv preprint arXiv:2410.10630}} (\bibinfo{year}{2024}).
\newblock


\bibitem[Wu et~al\mbox{.}(2024c)]%
        {wu2024language}
\bibfield{author}{\bibinfo{person}{Wilson Wu}, \bibinfo{person}{John~X Morris}, {and} \bibinfo{person}{Lionel Levine}.} \bibinfo{year}{2024}\natexlab{c}.
\newblock \showarticletitle{Do language models plan ahead for future tokens?}
\newblock \bibinfo{journal}{\emph{arXiv preprint arXiv:2404.00859}} (\bibinfo{year}{2024}).
\newblock


\bibitem[Wu et~al\mbox{.}({[n.\,d.]})]%
        {wucan}
\bibfield{author}{\bibinfo{person}{Xixi Wu}, \bibinfo{person}{Yifei Shen}, \bibinfo{person}{Caihua Shan}, \bibinfo{person}{Kaitao Song}, \bibinfo{person}{Siwei Wang}, \bibinfo{person}{Bohang Zhang}, \bibinfo{person}{Jiarui Feng}, \bibinfo{person}{Hong Cheng}, \bibinfo{person}{Wei Chen}, \bibinfo{person}{Yun Xiong}, {et~al\mbox{.}}} \bibinfo{year}{[n.\,d.]}\natexlab{}.
\newblock \showarticletitle{Can Graph Learning Improve Planning in LLM-based Agents?}. In \bibinfo{booktitle}{\emph{The Thirty-eighth Annual Conference on Neural Information Processing Systems}}.
\newblock


\bibitem[Wu et~al\mbox{.}(2024d)]%
        {wu2024self}
\bibfield{author}{\bibinfo{person}{Yue Wu}, \bibinfo{person}{Zhiqing Sun}, \bibinfo{person}{Huizhuo Yuan}, \bibinfo{person}{Kaixuan Ji}, \bibinfo{person}{Yiming Yang}, {and} \bibinfo{person}{Quanquan Gu}.} \bibinfo{year}{2024}\natexlab{d}.
\newblock \showarticletitle{Self-play preference optimization for language model alignment}.
\newblock \bibinfo{journal}{\emph{arXiv preprint arXiv:2405.00675}} (\bibinfo{year}{2024}).
\newblock


\bibitem[Wu et~al\mbox{.}(2024e)]%
        {wu2024selp}
\bibfield{author}{\bibinfo{person}{Yi Wu}, \bibinfo{person}{Zikang Xiong}, \bibinfo{person}{Yiran Hu}, \bibinfo{person}{Shreyash~S Iyengar}, \bibinfo{person}{Nan Jiang}, \bibinfo{person}{Aniket Bera}, \bibinfo{person}{Lin Tan}, {and} \bibinfo{person}{Suresh Jagannathan}.} \bibinfo{year}{2024}\natexlab{e}.
\newblock \showarticletitle{SELP: Generating Safe and Efficient Task Plans for Robot Agents with Large Language Models}.
\newblock \bibinfo{journal}{\emph{arXiv preprint arXiv:2409.19471}} (\bibinfo{year}{2024}).
\newblock


\bibitem[Wu and Feng(2024)]%
        {wu2403protrix}
\bibfield{author}{\bibinfo{person}{Zirui Wu} {and} \bibinfo{person}{Yansong Feng}.} \bibinfo{year}{2024}\natexlab{}.
\newblock \showarticletitle{Protrix: Building models for planning and reasoning over tables with sentence context}.
\newblock \bibinfo{journal}{\emph{URL https://arxiv. org/abs/2403.02177}} (\bibinfo{year}{2024}).
\newblock


\bibitem[Xi et~al\mbox{.}(2023)]%
        {xi2023rise}
\bibfield{author}{\bibinfo{person}{Zhiheng Xi}, \bibinfo{person}{Wenxiang Chen}, \bibinfo{person}{Xin Guo}, \bibinfo{person}{Wei He}, \bibinfo{person}{Yiwen Ding}, \bibinfo{person}{Boyang Hong}, \bibinfo{person}{Ming Zhang}, \bibinfo{person}{Junzhe Wang}, \bibinfo{person}{Senjie Jin}, \bibinfo{person}{Enyu Zhou}, {et~al\mbox{.}}} \bibinfo{year}{2023}\natexlab{}.
\newblock \showarticletitle{The rise and potential of large language model based agents: A survey}.
\newblock \bibinfo{journal}{\emph{arXiv preprint arXiv:2309.07864}} (\bibinfo{year}{2023}).
\newblock


\bibitem[Xi et~al\mbox{.}(2024a)]%
        {xi2024agentgym}
\bibfield{author}{\bibinfo{person}{Zhiheng Xi}, \bibinfo{person}{Yiwen Ding}, \bibinfo{person}{Wenxiang Chen}, \bibinfo{person}{Boyang Hong}, \bibinfo{person}{Honglin Guo}, \bibinfo{person}{Junzhe Wang}, \bibinfo{person}{Dingwen Yang}, \bibinfo{person}{Chenyang Liao}, \bibinfo{person}{Xin Guo}, \bibinfo{person}{Wei He}, {et~al\mbox{.}}} \bibinfo{year}{2024}\natexlab{a}.
\newblock \showarticletitle{Agentgym: Evolving large language model-based agents across diverse environments}.
\newblock \bibinfo{journal}{\emph{arXiv preprint arXiv:2406.04151}} (\bibinfo{year}{2024}).
\newblock


\bibitem[Xi et~al\mbox{.}(2024b)]%
        {xi2024enhancing}
\bibfield{author}{\bibinfo{person}{Zhiheng Xi}, \bibinfo{person}{Dingwen Yang}, \bibinfo{person}{Jixuan Huang}, \bibinfo{person}{Jiafu Tang}, \bibinfo{person}{Guanyu Li}, \bibinfo{person}{Yiwen Ding}, \bibinfo{person}{Wei He}, \bibinfo{person}{Boyang Hong}, \bibinfo{person}{Shihan Do}, \bibinfo{person}{Wenyu Zhan}, {et~al\mbox{.}}} \bibinfo{year}{2024}\natexlab{b}.
\newblock \showarticletitle{Enhancing LLM Reasoning via Critique Models with Test-Time and Training-Time Supervision}.
\newblock \bibinfo{journal}{\emph{arXiv preprint arXiv:2411.16579}} (\bibinfo{year}{2024}).
\newblock


\bibitem[Xia and Luo(2025)]%
        {xia2025gui}
\bibfield{author}{\bibinfo{person}{Xiaobo Xia} {and} \bibinfo{person}{Run Luo}.} \bibinfo{year}{2025}\natexlab{}.
\newblock \showarticletitle{GUI-R1: A Generalist R1-Style Vision-Language Action Model For GUI Agents}.
\newblock \bibinfo{journal}{\emph{arXiv preprint arXiv:2504.10458}} (\bibinfo{year}{2025}).
\newblock


\bibitem[Xiao et~al\mbox{.}(2024)]%
        {xiao2024flowbench}
\bibfield{author}{\bibinfo{person}{Ruixuan Xiao}, \bibinfo{person}{Wentao Ma}, \bibinfo{person}{Ke Wang}, \bibinfo{person}{Yuchuan Wu}, \bibinfo{person}{Junbo Zhao}, \bibinfo{person}{Haobo Wang}, \bibinfo{person}{Fei Huang}, {and} \bibinfo{person}{Yongbin Li}.} \bibinfo{year}{2024}\natexlab{}.
\newblock \showarticletitle{Flowbench: Revisiting and benchmarking workflow-guided planning for llm-based agents}.
\newblock \bibinfo{journal}{\emph{arXiv preprint arXiv:2406.14884}} (\bibinfo{year}{2024}).
\newblock


\bibitem[Xie and Zou(2024)]%
        {xie2024human}
\bibfield{author}{\bibinfo{person}{Chengxing Xie} {and} \bibinfo{person}{Difan Zou}.} \bibinfo{year}{2024}\natexlab{}.
\newblock \showarticletitle{A human-like reasoning framework for multi-phases planning task with large language models}.
\newblock \bibinfo{journal}{\emph{arXiv preprint arXiv:2405.18208}} (\bibinfo{year}{2024}).
\newblock


\bibitem[Xie et~al\mbox{.}(2024d)]%
        {xie2024revealing}
\bibfield{author}{\bibinfo{person}{Jian Xie}, \bibinfo{person}{Kexun Zhang}, \bibinfo{person}{Jiangjie Chen}, \bibinfo{person}{Siyu Yuan}, \bibinfo{person}{Kai Zhang}, \bibinfo{person}{Yikai Zhang}, \bibinfo{person}{Lei Li}, {and} \bibinfo{person}{Yanghua Xiao}.} \bibinfo{year}{2024}\natexlab{d}.
\newblock \showarticletitle{Revealing the Barriers of Language Agents in Planning}.
\newblock \bibinfo{journal}{\emph{arXiv preprint arXiv:2410.12409}} (\bibinfo{year}{2024}).
\newblock


\bibitem[Xie et~al\mbox{.}(2024e)]%
        {xie2024travelplanner}
\bibfield{author}{\bibinfo{person}{Jian Xie}, \bibinfo{person}{Kai Zhang}, \bibinfo{person}{Jiangjie Chen}, \bibinfo{person}{Tinghui Zhu}, \bibinfo{person}{Renze Lou}, \bibinfo{person}{Yuandong Tian}, \bibinfo{person}{Yanghua Xiao}, {and} \bibinfo{person}{Yu Su}.} \bibinfo{year}{2024}\natexlab{e}.
\newblock \showarticletitle{TravelPlanner: a benchmark for real-world planning with language agents}. In \bibinfo{booktitle}{\emph{Proceedings of the 41st International Conference on Machine Learning}}. \bibinfo{pages}{54590--54613}.
\newblock


\bibitem[Xie et~al\mbox{.}(2024c)]%
        {xie2024osworld}
\bibfield{author}{\bibinfo{person}{Tianbao Xie}, \bibinfo{person}{Danyang Zhang}, \bibinfo{person}{Jixuan Chen}, \bibinfo{person}{Xiaochuan Li}, \bibinfo{person}{Siheng Zhao}, \bibinfo{person}{Ruisheng Cao}, \bibinfo{person}{Toh~J Hua}, \bibinfo{person}{Zhoujun Cheng}, \bibinfo{person}{Dongchan Shin}, \bibinfo{person}{Fangyu Lei}, {et~al\mbox{.}}} \bibinfo{year}{2024}\natexlab{c}.
\newblock \showarticletitle{Osworld: Benchmarking multimodal agents for open-ended tasks in real computer environments}.
\newblock \bibinfo{journal}{\emph{Advances in Neural Information Processing Systems}}  \bibinfo{volume}{37} (\bibinfo{year}{2024}), \bibinfo{pages}{52040--52094}.
\newblock


\bibitem[Xie et~al\mbox{.}(2024a)]%
        {xie2024monte}
\bibfield{author}{\bibinfo{person}{Yuxi Xie}, \bibinfo{person}{Anirudh Goyal}, \bibinfo{person}{Wenyue Zheng}, \bibinfo{person}{Min-Yen Kan}, \bibinfo{person}{Timothy~P Lillicrap}, \bibinfo{person}{Kenji Kawaguchi}, {and} \bibinfo{person}{Michael Shieh}.} \bibinfo{year}{2024}\natexlab{a}.
\newblock \showarticletitle{Monte Carlo Tree Search Boosts Reasoning via Iterative Preference Learning}.
\newblock \bibinfo{journal}{\emph{arXiv preprint arXiv:2405.00451}} (\bibinfo{year}{2024}).
\newblock


\bibitem[Xie et~al\mbox{.}(2024b)]%
        {xie2024self}
\bibfield{author}{\bibinfo{person}{Yuxi Xie}, \bibinfo{person}{Kenji Kawaguchi}, \bibinfo{person}{Yiran Zhao}, \bibinfo{person}{James~Xu Zhao}, \bibinfo{person}{Min-Yen Kan}, \bibinfo{person}{Junxian He}, {and} \bibinfo{person}{Michael Xie}.} \bibinfo{year}{2024}\natexlab{b}.
\newblock \showarticletitle{Self-evaluation guided beam search for reasoning}.
\newblock \bibinfo{journal}{\emph{Advances in Neural Information Processing Systems}}  \bibinfo{volume}{36} (\bibinfo{year}{2024}).
\newblock


\bibitem[Xie et~al\mbox{.}(2023)]%
        {xie2023translating}
\bibfield{author}{\bibinfo{person}{Yaqi Xie}, \bibinfo{person}{Chen Yu}, \bibinfo{person}{Tongyao Zhu}, \bibinfo{person}{Jinbin Bai}, \bibinfo{person}{Ze Gong}, {and} \bibinfo{person}{Harold Soh}.} \bibinfo{year}{2023}\natexlab{}.
\newblock \showarticletitle{Translating natural language to planning goals with large-language models}.
\newblock \bibinfo{journal}{\emph{arXiv preprint arXiv:2302.05128}} (\bibinfo{year}{2023}).
\newblock


\bibitem[Xu et~al\mbox{.}(2024b)]%
        {xu2024theagentcompany}
\bibfield{author}{\bibinfo{person}{Frank~F Xu}, \bibinfo{person}{Yufan Song}, \bibinfo{person}{Boxuan Li}, \bibinfo{person}{Yuxuan Tang}, \bibinfo{person}{Kritanjali Jain}, \bibinfo{person}{Mengxue Bao}, \bibinfo{person}{Zora~Z Wang}, \bibinfo{person}{Xuhui Zhou}, \bibinfo{person}{Zhitong Guo}, \bibinfo{person}{Murong Cao}, {et~al\mbox{.}}} \bibinfo{year}{2024}\natexlab{b}.
\newblock \showarticletitle{Theagentcompany: benchmarking llm agents on consequential real world tasks}.
\newblock \bibinfo{journal}{\emph{arXiv preprint arXiv:2412.14161}} (\bibinfo{year}{2024}).
\newblock


\bibitem[Xu et~al\mbox{.}(2024a)]%
        {xu2024search}
\bibfield{author}{\bibinfo{person}{Shicheng Xu}, \bibinfo{person}{Liang Pang}, \bibinfo{person}{Huawei Shen}, \bibinfo{person}{Xueqi Cheng}, {and} \bibinfo{person}{Tat-Seng Chua}.} \bibinfo{year}{2024}\natexlab{a}.
\newblock \showarticletitle{Search-in-the-chain: Interactively enhancing large language models with search for knowledge-intensive tasks}. In \bibinfo{booktitle}{\emph{Proceedings of the ACM Web Conference 2024}}. \bibinfo{pages}{1362--1373}.
\newblock


\bibitem[Yang et~al\mbox{.}(2025b)]%
        {yang2025magma}
\bibfield{author}{\bibinfo{person}{Jianwei Yang}, \bibinfo{person}{Reuben Tan}, \bibinfo{person}{Qianhui Wu}, \bibinfo{person}{Ruijie Zheng}, \bibinfo{person}{Baolin Peng}, \bibinfo{person}{Yongyuan Liang}, \bibinfo{person}{Yu Gu}, \bibinfo{person}{Mu Cai}, \bibinfo{person}{Seonghyeon Ye}, \bibinfo{person}{Joel Jang}, {et~al\mbox{.}}} \bibinfo{year}{2025}\natexlab{b}.
\newblock \showarticletitle{Magma: A foundation model for multimodal ai agents}.
\newblock \bibinfo{journal}{\emph{arXiv preprint arXiv:2502.13130}} (\bibinfo{year}{2025}).
\newblock


\bibitem[Yang et~al\mbox{.}(2025a)]%
        {yang2025embodiedbench}
\bibfield{author}{\bibinfo{person}{Rui Yang}, \bibinfo{person}{Hanyang Chen}, \bibinfo{person}{Junyu Zhang}, \bibinfo{person}{Mark Zhao}, \bibinfo{person}{Cheng Qian}, \bibinfo{person}{Kangrui Wang}, \bibinfo{person}{Qineng Wang}, \bibinfo{person}{Teja~Venkat Koripella}, \bibinfo{person}{Marziyeh Movahedi}, \bibinfo{person}{Manling Li}, {et~al\mbox{.}}} \bibinfo{year}{2025}\natexlab{a}.
\newblock \showarticletitle{EmbodiedBench: Comprehensive Benchmarking Multi-modal Large Language Models for Vision-Driven Embodied Agents}.
\newblock \bibinfo{journal}{\emph{arXiv preprint arXiv:2502.09560}} (\bibinfo{year}{2025}).
\newblock


\bibitem[Yang et~al\mbox{.}(2024a)]%
        {yang2024large}
\bibfield{author}{\bibinfo{person}{Sohee Yang}, \bibinfo{person}{Elena Gribovskaya}, \bibinfo{person}{Nora Kassner}, \bibinfo{person}{Mor Geva}, {and} \bibinfo{person}{Sebastian Riedel}.} \bibinfo{year}{2024}\natexlab{a}.
\newblock \showarticletitle{Do Large Language Models Latently Perform Multi-Hop Reasoning?}
\newblock \bibinfo{journal}{\emph{arXiv preprint arXiv:2402.16837}} (\bibinfo{year}{2024}).
\newblock


\bibitem[Yang et~al\mbox{.}(2025c)]%
        {yang2025step}
\bibfield{author}{\bibinfo{person}{Xiao-Wen Yang}, \bibinfo{person}{Xuan-Yi Zhu}, \bibinfo{person}{Wen-Da Wei}, \bibinfo{person}{Ding-Chu Zhang}, \bibinfo{person}{Jie-Jing Shao}, \bibinfo{person}{Zhi Zhou}, \bibinfo{person}{Lan-Zhe Guo}, {and} \bibinfo{person}{Yu-Feng Li}.} \bibinfo{year}{2025}\natexlab{c}.
\newblock \showarticletitle{Step Back to Leap Forward: Self-Backtracking for Boosting Reasoning of Language Models}.
\newblock \bibinfo{journal}{\emph{arXiv preprint arXiv:2502.04404}} (\bibinfo{year}{2025}).
\newblock


\bibitem[Yang et~al\mbox{.}(2023)]%
        {yang2023coupling}
\bibfield{author}{\bibinfo{person}{Zhun Yang}, \bibinfo{person}{Adam Ishay}, {and} \bibinfo{person}{Joohyung Lee}.} \bibinfo{year}{2023}\natexlab{}.
\newblock \showarticletitle{Coupling Large Language Models with Logic Programming for Robust and General Reasoning from Text}. In \bibinfo{booktitle}{\emph{Findings of the Association for Computational Linguistics: ACL 2023}}. \bibinfo{pages}{5186--5219}.
\newblock


\bibitem[Yang et~al\mbox{.}(2024b)]%
        {yang2024react}
\bibfield{author}{\bibinfo{person}{Zonghan Yang}, \bibinfo{person}{Peng Li}, \bibinfo{person}{Ming Yan}, \bibinfo{person}{Ji Zhang}, \bibinfo{person}{Fei Huang}, {and} \bibinfo{person}{Yang Liu}.} \bibinfo{year}{2024}\natexlab{b}.
\newblock \showarticletitle{ReAct Meets ActRe: When Language Agents Enjoy Training Data Autonomy.}
\newblock \bibinfo{journal}{\emph{CoRR, abs/2403.14589}} (\bibinfo{year}{2024}).
\newblock


\bibitem[Yao et~al\mbox{.}(2022)]%
        {yao2022webshop}
\bibfield{author}{\bibinfo{person}{Shunyu Yao}, \bibinfo{person}{Howard Chen}, \bibinfo{person}{John Yang}, {and} \bibinfo{person}{Karthik Narasimhan}.} \bibinfo{year}{2022}\natexlab{}.
\newblock \showarticletitle{Webshop: Towards scalable real-world web interaction with grounded language agents}.
\newblock \bibinfo{journal}{\emph{Advances in Neural Information Processing Systems}}  \bibinfo{volume}{35} (\bibinfo{year}{2022}), \bibinfo{pages}{20744--20757}.
\newblock


\bibitem[Yao et~al\mbox{.}(2023)]%
        {yao2023tree}
\bibfield{author}{\bibinfo{person}{Shunyu Yao}, \bibinfo{person}{Dian Yu}, \bibinfo{person}{Jeffrey Zhao}, \bibinfo{person}{Izhak Shafran}, \bibinfo{person}{Tom Griffiths}, \bibinfo{person}{Yuan Cao}, {and} \bibinfo{person}{Karthik Narasimhan}.} \bibinfo{year}{2023}\natexlab{}.
\newblock \showarticletitle{Tree of thoughts: Deliberate problem solving with large language models}.
\newblock \bibinfo{journal}{\emph{Advances in neural information processing systems}}  \bibinfo{volume}{36} (\bibinfo{year}{2023}), \bibinfo{pages}{11809--11822}.
\newblock


\bibitem[Yao et~al\mbox{.}(2024b)]%
        {yao2024minicpm}
\bibfield{author}{\bibinfo{person}{Yuan Yao}, \bibinfo{person}{Tianyu Yu}, \bibinfo{person}{Ao Zhang}, \bibinfo{person}{Chongyi Wang}, \bibinfo{person}{Junbo Cui}, \bibinfo{person}{Hongji Zhu}, \bibinfo{person}{Tianchi Cai}, \bibinfo{person}{Haoyu Li}, \bibinfo{person}{Weilin Zhao}, \bibinfo{person}{Zhihui He}, {et~al\mbox{.}}} \bibinfo{year}{2024}\natexlab{b}.
\newblock \showarticletitle{Minicpm-v: A gpt-4v level mllm on your phone}.
\newblock \bibinfo{journal}{\emph{arXiv preprint arXiv:2408.01800}} (\bibinfo{year}{2024}).
\newblock


\bibitem[Yao et~al\mbox{.}(2024a)]%
        {yao2024minicpmvgpt4vlevelmllm}
\bibfield{author}{\bibinfo{person}{Yuan Yao}, \bibinfo{person}{Tianyu Yu}, \bibinfo{person}{Ao Zhang}, \bibinfo{person}{Chongyi Wang}, \bibinfo{person}{Junbo Cui}, \bibinfo{person}{Hongji Zhu}, \bibinfo{person}{Tianchi Cai}, \bibinfo{person}{Haoyu Li}, \bibinfo{person}{Weilin Zhao}, \bibinfo{person}{Zhihui He}, \bibinfo{person}{Qianyu Chen}, \bibinfo{person}{Huarong Zhou}, \bibinfo{person}{Zhensheng Zou}, \bibinfo{person}{Haoye Zhang}, \bibinfo{person}{Shengding Hu}, \bibinfo{person}{Zhi Zheng}, \bibinfo{person}{Jie Zhou}, \bibinfo{person}{Jie Cai}, \bibinfo{person}{Xu Han}, \bibinfo{person}{Guoyang Zeng}, \bibinfo{person}{Dahai Li}, \bibinfo{person}{Zhiyuan Liu}, {and} \bibinfo{person}{Maosong Sun}.} \bibinfo{year}{2024}\natexlab{a}.
\newblock \bibinfo{title}{MiniCPM-V: A GPT-4V Level MLLM on Your Phone}.
\newblock


\bibitem[Yin et~al\mbox{.}(2023)]%
        {yin2023agent}
\bibfield{author}{\bibinfo{person}{Da Yin}, \bibinfo{person}{Faeze Brahman}, \bibinfo{person}{Abhilasha Ravichander}, \bibinfo{person}{Khyathi Chandu}, \bibinfo{person}{Kai-Wei Chang}, \bibinfo{person}{Yejin Choi}, {and} \bibinfo{person}{Bill~Yuchen Lin}.} \bibinfo{year}{2023}\natexlab{}.
\newblock \showarticletitle{Agent lumos: Unified and modular training for open-source language agents}.
\newblock \bibinfo{journal}{\emph{arXiv preprint arXiv:2311.05657}} (\bibinfo{year}{2023}).
\newblock


\bibitem[Yin et~al\mbox{.}(2024)]%
        {yin2024explicit}
\bibfield{author}{\bibinfo{person}{Zhangyue Yin}, \bibinfo{person}{Qiushi Sun}, \bibinfo{person}{Qipeng Guo}, \bibinfo{person}{Zhiyuan Zeng}, \bibinfo{person}{Qinyuan Cheng}, \bibinfo{person}{Xipeng Qiu}, {and} \bibinfo{person}{Xuan-Jing Huang}.} \bibinfo{year}{2024}\natexlab{}.
\newblock \showarticletitle{Explicit Memory Learning with Expectation Maximization}. In \bibinfo{booktitle}{\emph{Proceedings of the 2024 Conference on Empirical Methods in Natural Language Processing}}. \bibinfo{pages}{16618--16635}.
\newblock


\bibitem[Yoo et~al\mbox{.}(2024)]%
        {yoo2024exploratory}
\bibfield{author}{\bibinfo{person}{Minjong Yoo}, \bibinfo{person}{Jinwoo Jang}, \bibinfo{person}{Wei-Jin Park}, {and} \bibinfo{person}{Honguk Woo}.} \bibinfo{year}{2024}\natexlab{}.
\newblock \showarticletitle{Exploratory Retrieval-Augmented Planning For Continual Embodied Instruction Following}. In \bibinfo{booktitle}{\emph{The Thirty-eighth Annual Conference on Neural Information Processing Systems}}.
\newblock


\bibitem[Yu et~al\mbox{.}(2024)]%
        {yu2024flow}
\bibfield{author}{\bibinfo{person}{Fangxu Yu}, \bibinfo{person}{Lai Jiang}, \bibinfo{person}{Haoqiang Kang}, \bibinfo{person}{Shibo Hao}, {and} \bibinfo{person}{Lianhui Qin}.} \bibinfo{year}{2024}\natexlab{}.
\newblock \showarticletitle{Flow of Reasoning: Training LLMs for Divergent Problem Solving with Minimal Examples}.
\newblock \bibinfo{journal}{\emph{arXiv preprint arXiv:2406.05673}} (\bibinfo{year}{2024}).
\newblock


\bibitem[Yu et~al\mbox{.}(2023)]%
        {yu2023thought}
\bibfield{author}{\bibinfo{person}{Junchi Yu}, \bibinfo{person}{Ran He}, {and} \bibinfo{person}{Rex Ying}.} \bibinfo{year}{2023}\natexlab{}.
\newblock \showarticletitle{Thought propagation: An analogical approach to complex reasoning with large language models}.
\newblock \bibinfo{journal}{\emph{arXiv preprint arXiv:2310.03965}} (\bibinfo{year}{2023}).
\newblock


\bibitem[Yu et~al\mbox{.}(2025)]%
        {yu2025think}
\bibfield{author}{\bibinfo{person}{Zishun Yu}, \bibinfo{person}{Tengyu Xu}, \bibinfo{person}{Di Jin}, \bibinfo{person}{Karthik~Abinav Sankararaman}, \bibinfo{person}{Yun He}, \bibinfo{person}{Wenxuan Zhou}, \bibinfo{person}{Zhouhao Zeng}, \bibinfo{person}{Eryk Helenowski}, \bibinfo{person}{Chen Zhu}, \bibinfo{person}{Sinong Wang}, {et~al\mbox{.}}} \bibinfo{year}{2025}\natexlab{}.
\newblock \showarticletitle{Think Smarter not Harder: Adaptive Reasoning with Inference Aware Optimization}.
\newblock \bibinfo{journal}{\emph{arXiv preprint arXiv:2501.17974}} (\bibinfo{year}{2025}).
\newblock


\bibitem[Yuan et~al\mbox{.}(2024a)]%
        {yuan2024advancing}
\bibfield{author}{\bibinfo{person}{Lifan Yuan}, \bibinfo{person}{Ganqu Cui}, \bibinfo{person}{Hanbin Wang}, \bibinfo{person}{Ning Ding}, \bibinfo{person}{Xingyao Wang}, \bibinfo{person}{Jia Deng}, \bibinfo{person}{Boji Shan}, \bibinfo{person}{Huimin Chen}, \bibinfo{person}{Ruobing Xie}, \bibinfo{person}{Yankai Lin}, {et~al\mbox{.}}} \bibinfo{year}{2024}\natexlab{a}.
\newblock \showarticletitle{Advancing llm reasoning generalists with preference trees}.
\newblock \bibinfo{journal}{\emph{arXiv preprint arXiv:2404.02078}} (\bibinfo{year}{2024}).
\newblock


\bibitem[Yuan et~al\mbox{.}(2024b)]%
        {yuan2024tasklama}
\bibfield{author}{\bibinfo{person}{Quan Yuan}, \bibinfo{person}{Mehran Kazemi}, \bibinfo{person}{Xin Xu}, \bibinfo{person}{Isaac Noble}, \bibinfo{person}{Vaiva Imbrasaite}, {and} \bibinfo{person}{Deepak Ramachandran}.} \bibinfo{year}{2024}\natexlab{b}.
\newblock \showarticletitle{Tasklama: probing the complex task understanding of language models}. In \bibinfo{booktitle}{\emph{Proceedings of the AAAI Conference on Artificial Intelligence}}, Vol.~\bibinfo{volume}{38}. \bibinfo{pages}{19468--19476}.
\newblock


\bibitem[Yuan et~al\mbox{.}(2023)]%
        {yuan2023distilling}
\bibfield{author}{\bibinfo{person}{Siyu Yuan}, \bibinfo{person}{Jiangjie Chen}, \bibinfo{person}{Ziquan Fu}, \bibinfo{person}{Xuyang Ge}, \bibinfo{person}{Soham Shah}, \bibinfo{person}{Charles~Robert Jankowski}, \bibinfo{person}{Yanghua Xiao}, {and} \bibinfo{person}{Deqing Yang}.} \bibinfo{year}{2023}\natexlab{}.
\newblock \showarticletitle{Distilling script knowledge from large language models for constrained language planning}.
\newblock \bibinfo{journal}{\emph{arXiv preprint arXiv:2305.05252}} (\bibinfo{year}{2023}).
\newblock


\bibitem[Yuan et~al\mbox{.}(2025a)]%
        {yuan2025agent}
\bibfield{author}{\bibinfo{person}{Siyu Yuan}, \bibinfo{person}{Zehui Chen}, \bibinfo{person}{Zhiheng Xi}, \bibinfo{person}{Junjie Ye}, \bibinfo{person}{Zhengyin Du}, {and} \bibinfo{person}{Jiecao Chen}.} \bibinfo{year}{2025}\natexlab{a}.
\newblock \showarticletitle{Agent-R: Training Language Model Agents to Reflect via Iterative Self-Training}.
\newblock \bibinfo{journal}{\emph{arXiv preprint arXiv:2501.11425}} (\bibinfo{year}{2025}).
\newblock


\bibitem[Yuan et~al\mbox{.}(2025b)]%
        {yuan2025enhancing}
\bibfield{author}{\bibinfo{person}{Xinbin Yuan}, \bibinfo{person}{Jian Zhang}, \bibinfo{person}{Kaixin Li}, \bibinfo{person}{Zhuoxuan Cai}, \bibinfo{person}{Lujian Yao}, \bibinfo{person}{Jie Chen}, \bibinfo{person}{Enguang Wang}, \bibinfo{person}{Qibin Hou}, \bibinfo{person}{Jinwei Chen}, \bibinfo{person}{Peng-Tao Jiang}, {et~al\mbox{.}}} \bibinfo{year}{2025}\natexlab{b}.
\newblock \showarticletitle{Enhancing Visual Grounding for GUI Agents via Self-Evolutionary Reinforcement Learning}.
\newblock \bibinfo{journal}{\emph{arXiv preprint arXiv:2505.12370}} (\bibinfo{year}{2025}).
\newblock


\bibitem[Yue et~al\mbox{.}(2024a)]%
        {yue2024learning}
\bibfield{author}{\bibinfo{person}{William Yue}, \bibinfo{person}{Bo Liu}, {and} \bibinfo{person}{Peter Stone}.} \bibinfo{year}{2024}\natexlab{a}.
\newblock \showarticletitle{Learning Memory Mechanisms for Decision Making through Demonstrations}.
\newblock \bibinfo{journal}{\emph{arXiv preprint arXiv:2411.07954}} (\bibinfo{year}{2024}).
\newblock


\bibitem[Yue et~al\mbox{.}(2024b)]%
        {yue2024distilling}
\bibfield{author}{\bibinfo{person}{Yuanhao Yue}, \bibinfo{person}{Chengyu Wang}, \bibinfo{person}{Jun Huang}, {and} \bibinfo{person}{Peng Wang}.} \bibinfo{year}{2024}\natexlab{b}.
\newblock \showarticletitle{Distilling Instruction-following Abilities of Large Language Models with Task-aware Curriculum Planning}.
\newblock \bibinfo{journal}{\emph{arXiv preprint arXiv:2405.13448}} (\bibinfo{year}{2024}).
\newblock


\bibitem[Zeng et~al\mbox{.}(2023)]%
        {zeng2023agenttuning}
\bibfield{author}{\bibinfo{person}{Aohan Zeng}, \bibinfo{person}{Mingdao Liu}, \bibinfo{person}{Rui Lu}, \bibinfo{person}{Bowen Wang}, \bibinfo{person}{Xiao Liu}, \bibinfo{person}{Yuxiao Dong}, {and} \bibinfo{person}{Jie Tang}.} \bibinfo{year}{2023}\natexlab{}.
\newblock \showarticletitle{Agenttuning: Enabling generalized agent abilities for llms}.
\newblock \bibinfo{journal}{\emph{arXiv preprint arXiv:2310.12823}} (\bibinfo{year}{2023}).
\newblock


\bibitem[Zhang et~al\mbox{.}(2025)]%
        {zhang2025tongui}
\bibfield{author}{\bibinfo{person}{Bofei Zhang}, \bibinfo{person}{Zirui Shang}, \bibinfo{person}{Zhi Gao}, \bibinfo{person}{Wang Zhang}, \bibinfo{person}{Rui Xie}, \bibinfo{person}{Xiaojian Ma}, \bibinfo{person}{Tao Yuan}, \bibinfo{person}{Xinxiao Wu}, \bibinfo{person}{Song-Chun Zhu}, {and} \bibinfo{person}{Qing Li}.} \bibinfo{year}{2025}\natexlab{}.
\newblock \showarticletitle{TongUI: Building Generalized GUI Agents by Learning from Multimodal Web Tutorials}.
\newblock \bibinfo{journal}{\emph{arXiv preprint arXiv:2504.12679}} (\bibinfo{year}{2025}).
\newblock


\bibitem[Zhang et~al\mbox{.}(2024b)]%
        {zhang2024meta}
\bibfield{author}{\bibinfo{person}{Cong Zhang}, \bibinfo{person}{Derrick Goh~Xin Deik}, \bibinfo{person}{Dexun Li}, \bibinfo{person}{Hao Zhang}, {and} \bibinfo{person}{Yong Liu}.} \bibinfo{year}{2024}\natexlab{b}.
\newblock \showarticletitle{Meta-task planning for language agents}.
\newblock \bibinfo{journal}{\emph{arXiv preprint arXiv:2405.16510}} (\bibinfo{year}{2024}).
\newblock


\bibitem[Zhang et~al\mbox{.}(2024a)]%
        {zhang2024large}
\bibfield{author}{\bibinfo{person}{Danyang Zhang}, \bibinfo{person}{Lu Chen}, \bibinfo{person}{Situo Zhang}, \bibinfo{person}{Hongshen Xu}, \bibinfo{person}{Zihan Zhao}, {and} \bibinfo{person}{Kai Yu}.} \bibinfo{year}{2024}\natexlab{a}.
\newblock \showarticletitle{Large language models are semi-parametric reinforcement learning agents}.
\newblock \bibinfo{journal}{\emph{Advances in Neural Information Processing Systems}}  \bibinfo{volume}{36} (\bibinfo{year}{2024}).
\newblock


\bibitem[Zhang et~al\mbox{.}(2023a)]%
        {zhang2023mobile}
\bibfield{author}{\bibinfo{person}{Danyang Zhang}, \bibinfo{person}{Zhennan Shen}, \bibinfo{person}{Rui Xie}, \bibinfo{person}{Situo Zhang}, \bibinfo{person}{Tianbao Xie}, \bibinfo{person}{Zihan Zhao}, \bibinfo{person}{Siyuan Chen}, \bibinfo{person}{Lu Chen}, \bibinfo{person}{Hongshen Xu}, \bibinfo{person}{Ruisheng Cao}, {et~al\mbox{.}}} \bibinfo{year}{2023}\natexlab{a}.
\newblock \showarticletitle{Mobile-Env: Building Qualified Evaluation Benchmarks for LLM-GUI Interaction}.
\newblock \bibinfo{journal}{\emph{arXiv preprint arXiv:2305.08144}} (\bibinfo{year}{2023}).
\newblock


\bibitem[Zhang et~al\mbox{.}(2024l)]%
        {zhang2024rest}
\bibfield{author}{\bibinfo{person}{Dan Zhang}, \bibinfo{person}{Sining Zhoubian}, \bibinfo{person}{Ziniu Hu}, \bibinfo{person}{Yisong Yue}, \bibinfo{person}{Yuxiao Dong}, {and} \bibinfo{person}{Jie Tang}.} \bibinfo{year}{2024}\natexlab{l}.
\newblock \showarticletitle{Rest-mcts*: Llm self-training via process reward guided tree search}.
\newblock \bibinfo{journal}{\emph{arXiv preprint arXiv:2406.03816}} (\bibinfo{year}{2024}).
\newblock


\bibitem[Zhang et~al\mbox{.}(2024g)]%
        {zhang2024pddlego}
\bibfield{author}{\bibinfo{person}{Li Zhang}, \bibinfo{person}{Peter Jansen}, \bibinfo{person}{Tianyi Zhang}, \bibinfo{person}{Peter Clark}, \bibinfo{person}{Chris Callison-Burch}, {and} \bibinfo{person}{Niket Tandon}.} \bibinfo{year}{2024}\natexlab{g}.
\newblock \showarticletitle{PDDLEGO: Iterative Planning in Textual Environments}.
\newblock \bibinfo{journal}{\emph{arXiv preprint arXiv:2405.19793}} (\bibinfo{year}{2024}).
\newblock


\bibitem[Zhang et~al\mbox{.}(2023b)]%
        {zhang2023interpretable}
\bibfield{author}{\bibinfo{person}{Mengxue Zhang}, \bibinfo{person}{Zichao Wang}, \bibinfo{person}{Zhichao Yang}, \bibinfo{person}{Weiqi Feng}, {and} \bibinfo{person}{Andrew Lan}.} \bibinfo{year}{2023}\natexlab{b}.
\newblock \showarticletitle{Interpretable math word problem solution generation via step-by-step planning}.
\newblock \bibinfo{journal}{\emph{arXiv preprint arXiv:2306.00784}} (\bibinfo{year}{2023}).
\newblock


\bibitem[Zhang et~al\mbox{.}(2024j)]%
        {zhang2024vlabench}
\bibfield{author}{\bibinfo{person}{Shiduo Zhang}, \bibinfo{person}{Zhe Xu}, \bibinfo{person}{Peiju Liu}, \bibinfo{person}{Xiaopeng Yu}, \bibinfo{person}{Yuan Li}, \bibinfo{person}{Qinghui Gao}, \bibinfo{person}{Zhaoye Fei}, \bibinfo{person}{Zhangyue Yin}, \bibinfo{person}{Zuxuan Wu}, \bibinfo{person}{Yu-Gang Jiang}, {et~al\mbox{.}}} \bibinfo{year}{2024}\natexlab{j}.
\newblock \showarticletitle{VLABench: A Large-Scale Benchmark for Language-Conditioned Robotics Manipulation with Long-Horizon Reasoning Tasks}.
\newblock \bibinfo{journal}{\emph{arXiv preprint arXiv:2412.18194}} (\bibinfo{year}{2024}).
\newblock


\bibitem[Zhang et~al\mbox{.}(2024f)]%
        {zhang2024strength}
\bibfield{author}{\bibinfo{person}{Tong Zhang}, \bibinfo{person}{Chen Huang}, \bibinfo{person}{Yang Deng}, \bibinfo{person}{Hongru Liang}, \bibinfo{person}{Jia Liu}, \bibinfo{person}{Zujie Wen}, \bibinfo{person}{Wenqiang Lei}, {and} \bibinfo{person}{Tat-Seng Chua}.} \bibinfo{year}{2024}\natexlab{f}.
\newblock \showarticletitle{Strength Lies in Differences! Improving Strategy Planning for Non-collaborative Dialogues via Diversified User Simulation}.
\newblock \bibinfo{journal}{\emph{arXiv preprint arXiv:2403.06769}} (\bibinfo{year}{2024}).
\newblock


\bibitem[Zhang et~al\mbox{.}(2024k)]%
        {zhang2024proc2pddl}
\bibfield{author}{\bibinfo{person}{Tianyi Zhang}, \bibinfo{person}{Li Zhang}, \bibinfo{person}{Zhaoyi Hou}, \bibinfo{person}{Ziyu Wang}, \bibinfo{person}{Yuling Gu}, \bibinfo{person}{Peter Clark}, \bibinfo{person}{Chris Callison-Burch}, {and} \bibinfo{person}{Niket Tandon}.} \bibinfo{year}{2024}\natexlab{k}.
\newblock \showarticletitle{Proc2pddl: Open-domain planning representations from texts}.
\newblock \bibinfo{journal}{\emph{arXiv preprint arXiv:2403.00092}} (\bibinfo{year}{2024}).
\newblock


\bibitem[Zhang et~al\mbox{.}(2024c)]%
        {zhang2024ask}
\bibfield{author}{\bibinfo{person}{Xuan Zhang}, \bibinfo{person}{Yang Deng}, \bibinfo{person}{Zifeng Ren}, \bibinfo{person}{See-Kiong Ng}, {and} \bibinfo{person}{Tat-Seng Chua}.} \bibinfo{year}{2024}\natexlab{c}.
\newblock \showarticletitle{Ask-before-plan: Proactive language agents for real-world planning}.
\newblock \bibinfo{journal}{\emph{arXiv preprint arXiv:2406.12639}} (\bibinfo{year}{2024}).
\newblock


\bibitem[Zhang et~al\mbox{.}(2024d)]%
        {zhang2024chain}
\bibfield{author}{\bibinfo{person}{Xuan Zhang}, \bibinfo{person}{Chao Du}, \bibinfo{person}{Tianyu Pang}, \bibinfo{person}{Qian Liu}, \bibinfo{person}{Wei Gao}, {and} \bibinfo{person}{Min Lin}.} \bibinfo{year}{2024}\natexlab{d}.
\newblock \showarticletitle{Chain of Preference Optimization: Improving Chain-of-Thought Reasoning in LLMs}.
\newblock \bibinfo{journal}{\emph{arXiv preprint arXiv:2406.09136}} (\bibinfo{year}{2024}).
\newblock


\bibitem[Zhang et~al\mbox{.}(2024h)]%
        {zhang2024small}
\bibfield{author}{\bibinfo{person}{Yunxiang Zhang}, \bibinfo{person}{Muhammad Khalifa}, \bibinfo{person}{Lajanugen Logeswaran}, \bibinfo{person}{Jaekyeom Kim}, \bibinfo{person}{Moontae Lee}, \bibinfo{person}{Honglak Lee}, {and} \bibinfo{person}{Lu Wang}.} \bibinfo{year}{2024}\natexlab{h}.
\newblock \showarticletitle{Small Language Models Need Strong Verifiers to Self-Correct Reasoning}.
\newblock \bibinfo{journal}{\emph{arXiv preprint arXiv:2404.17140}} (\bibinfo{year}{2024}).
\newblock


\bibitem[Zhang et~al\mbox{.}(2024i)]%
        {zhang2024webpilot}
\bibfield{author}{\bibinfo{person}{Yao Zhang}, \bibinfo{person}{Zijian Ma}, \bibinfo{person}{Yunpu Ma}, \bibinfo{person}{Zhen Han}, \bibinfo{person}{Yu Wu}, {and} \bibinfo{person}{Volker Tresp}.} \bibinfo{year}{2024}\natexlab{i}.
\newblock \showarticletitle{Webpilot: A versatile and autonomous multi-agent system for web task execution with strategic exploration}.
\newblock \bibinfo{journal}{\emph{arXiv preprint arXiv:2408.15978}} (\bibinfo{year}{2024}).
\newblock


\bibitem[Zhang et~al\mbox{.}(2024e)]%
        {zhang2024learn}
\bibfield{author}{\bibinfo{person}{Zhihan Zhang}, \bibinfo{person}{Tao Ge}, \bibinfo{person}{Zhenwen Liang}, \bibinfo{person}{Wenhao Yu}, \bibinfo{person}{Dian Yu}, \bibinfo{person}{Mengzhao Jia}, \bibinfo{person}{Dong Yu}, {and} \bibinfo{person}{Meng Jiang}.} \bibinfo{year}{2024}\natexlab{e}.
\newblock \showarticletitle{Learn beyond the answer: Training language models with reflection for mathematical reasoning}.
\newblock \bibinfo{journal}{\emph{arXiv preprint arXiv:2406.12050}} (\bibinfo{year}{2024}).
\newblock


\bibitem[Zhang and Zhang(2023)]%
        {zhang2023you}
\bibfield{author}{\bibinfo{person}{Zhuosheng Zhang} {and} \bibinfo{person}{Aston Zhang}.} \bibinfo{year}{2023}\natexlab{}.
\newblock \showarticletitle{You only look at screens: Multimodal chain-of-action agents}.
\newblock \bibinfo{journal}{\emph{arXiv preprint arXiv:2309.11436}} (\bibinfo{year}{2023}).
\newblock


\bibitem[Zhao et~al\mbox{.}(2025)]%
        {zhao2025embodied}
\bibfield{author}{\bibinfo{person}{Baining Zhao}, \bibinfo{person}{Ziyou Wang}, \bibinfo{person}{Jianjie Fang}, \bibinfo{person}{Chen Gao}, \bibinfo{person}{Fanhang Man}, \bibinfo{person}{Jinqiang Cui}, \bibinfo{person}{Xin Wang}, \bibinfo{person}{Xinlei Chen}, \bibinfo{person}{Yong Li}, {and} \bibinfo{person}{Wenwu Zhu}.} \bibinfo{year}{2025}\natexlab{}.
\newblock \showarticletitle{Embodied-R: Collaborative Framework for Activating Embodied Spatial Reasoning in Foundation Models via Reinforcement Learning}.
\newblock \bibinfo{journal}{\emph{arXiv preprint arXiv:2504.12680}} (\bibinfo{year}{2025}).
\newblock


\bibitem[Zhao et~al\mbox{.}(2023)]%
        {zhao2023large}
\bibfield{author}{\bibinfo{person}{Zirui Zhao}, \bibinfo{person}{Wee~Sun Lee}, {and} \bibinfo{person}{David Hsu}.} \bibinfo{year}{2023}\natexlab{}.
\newblock \showarticletitle{Large language models as commonsense knowledge for large-scale task planning}.
\newblock \bibinfo{journal}{\emph{Advances in Neural Information Processing Systems}}  \bibinfo{volume}{36} (\bibinfo{year}{2023}), \bibinfo{pages}{31967--31987}.
\newblock


\bibitem[Zheng et~al\mbox{.}(2024a)]%
        {zheng2024gpt}
\bibfield{author}{\bibinfo{person}{Boyuan Zheng}, \bibinfo{person}{Boyu Gou}, \bibinfo{person}{Jihyung Kil}, \bibinfo{person}{Huan Sun}, {and} \bibinfo{person}{Yu Su}.} \bibinfo{year}{2024}\natexlab{a}.
\newblock \showarticletitle{Gpt-4v (ision) is a generalist web agent, if grounded}.
\newblock \bibinfo{journal}{\emph{arXiv preprint arXiv:2401.01614}} (\bibinfo{year}{2024}).
\newblock


\bibitem[Zheng et~al\mbox{.}(2024d)]%
        {zheng2024natural}
\bibfield{author}{\bibinfo{person}{Huaixiu~Steven Zheng}, \bibinfo{person}{Swaroop Mishra}, \bibinfo{person}{Hugh Zhang}, \bibinfo{person}{Xinyun Chen}, \bibinfo{person}{Minmin Chen}, \bibinfo{person}{Azade Nova}, \bibinfo{person}{Le Hou}, \bibinfo{person}{Heng-Tze Cheng}, \bibinfo{person}{Quoc~V Le}, \bibinfo{person}{Ed~H Chi}, {et~al\mbox{.}}} \bibinfo{year}{2024}\natexlab{d}.
\newblock \showarticletitle{Natural plan: Benchmarking llms on natural language planning}.
\newblock \bibinfo{journal}{\emph{arXiv preprint arXiv:2406.04520}} (\bibinfo{year}{2024}).
\newblock


\bibitem[Zheng et~al\mbox{.}(2022)]%
        {zheng2022vlmbench}
\bibfield{author}{\bibinfo{person}{Kaizhi Zheng}, \bibinfo{person}{Xiaotong Chen}, \bibinfo{person}{Odest~Chadwicke Jenkins}, {and} \bibinfo{person}{Xin Wang}.} \bibinfo{year}{2022}\natexlab{}.
\newblock \showarticletitle{Vlmbench: A compositional benchmark for vision-and-language manipulation}.
\newblock \bibinfo{journal}{\emph{Advances in Neural Information Processing Systems}}  \bibinfo{volume}{35} (\bibinfo{year}{2022}), \bibinfo{pages}{665--678}.
\newblock


\bibitem[Zheng et~al\mbox{.}(2024b)]%
        {zheng2024agentstudio}
\bibfield{author}{\bibinfo{person}{Longtao Zheng}, \bibinfo{person}{Zhiyuan Huang}, \bibinfo{person}{Zhenghai Xue}, \bibinfo{person}{Xinrun Wang}, \bibinfo{person}{Bo An}, {and} \bibinfo{person}{Shuicheng Yan}.} \bibinfo{year}{2024}\natexlab{b}.
\newblock \showarticletitle{Agentstudio: A toolkit for building general virtual agents}.
\newblock \bibinfo{journal}{\emph{arXiv preprint arXiv:2403.17918}} (\bibinfo{year}{2024}).
\newblock


\bibitem[Zheng et~al\mbox{.}({[n.\,d.]})]%
        {zheng2023synapse}
\bibfield{author}{\bibinfo{person}{Longtao Zheng}, \bibinfo{person}{Rundong Wang}, \bibinfo{person}{Xinrun Wang}, {and} \bibinfo{person}{Bo An}.} \bibinfo{year}{[n.\,d.]}\natexlab{}.
\newblock \showarticletitle{Synapse: Trajectory-as-exemplar prompting with memory for computer control}. In \bibinfo{booktitle}{\emph{The Twelfth International Conference on Learning Representations}}.
\newblock


\bibitem[Zheng et~al\mbox{.}(2024c)]%
        {zheng2024thoughts}
\bibfield{author}{\bibinfo{person}{Zhonghua Zheng}, \bibinfo{person}{Lizi Liao}, \bibinfo{person}{Yang Deng}, \bibinfo{person}{Ee-Peng Lim}, \bibinfo{person}{Minlie Huang}, {and} \bibinfo{person}{Liqiang Nie}.} \bibinfo{year}{2024}\natexlab{c}.
\newblock \showarticletitle{Thoughts to target: Enhance planning for target-driven conversation}. In \bibinfo{booktitle}{\emph{Proceedings of the 2024 Conference on Empirical Methods in Natural Language Processing}}. \bibinfo{pages}{21108--21124}.
\newblock


\bibitem[Zhong et~al\mbox{.}(2024)]%
        {zhong2024memorybank}
\bibfield{author}{\bibinfo{person}{Wanjun Zhong}, \bibinfo{person}{Lianghong Guo}, \bibinfo{person}{Qiqi Gao}, \bibinfo{person}{He Ye}, {and} \bibinfo{person}{Yanlin Wang}.} \bibinfo{year}{2024}\natexlab{}.
\newblock \showarticletitle{Memorybank: Enhancing large language models with long-term memory}. In \bibinfo{booktitle}{\emph{Proceedings of the AAAI Conference on Artificial Intelligence}}, Vol.~\bibinfo{volume}{38}. \bibinfo{pages}{19724--19731}.
\newblock


\bibitem[Zhou et~al\mbox{.}(2023b)]%
        {zhou2023language}
\bibfield{author}{\bibinfo{person}{Andy Zhou}, \bibinfo{person}{Kai Yan}, \bibinfo{person}{Michal Shlapentokh-Rothman}, \bibinfo{person}{Haohan Wang}, {and} \bibinfo{person}{Yu-Xiong Wang}.} \bibinfo{year}{2023}\natexlab{b}.
\newblock \showarticletitle{Language agent tree search unifies reasoning acting and planning in language models}.
\newblock \bibinfo{journal}{\emph{arXiv preprint arXiv:2310.04406}} (\bibinfo{year}{2023}).
\newblock


\bibitem[Zhou et~al\mbox{.}(2022)]%
        {zhou2022least}
\bibfield{author}{\bibinfo{person}{Denny Zhou}, \bibinfo{person}{Nathanael Sch{\"a}rli}, \bibinfo{person}{Le Hou}, \bibinfo{person}{Jason Wei}, \bibinfo{person}{Nathan Scales}, \bibinfo{person}{Xuezhi Wang}, \bibinfo{person}{Dale Schuurmans}, \bibinfo{person}{Claire Cui}, \bibinfo{person}{Olivier Bousquet}, \bibinfo{person}{Quoc Le}, {et~al\mbox{.}}} \bibinfo{year}{2022}\natexlab{}.
\newblock \showarticletitle{Least-to-most prompting enables complex reasoning in large language models}.
\newblock \bibinfo{journal}{\emph{arXiv preprint arXiv:2205.10625}} (\bibinfo{year}{2022}).
\newblock


\bibitem[Zhou et~al\mbox{.}(2023a)]%
        {zhou2023webarena}
\bibfield{author}{\bibinfo{person}{Shuyan Zhou}, \bibinfo{person}{Frank~F Xu}, \bibinfo{person}{Hao Zhu}, \bibinfo{person}{Xuhui Zhou}, \bibinfo{person}{Robert Lo}, \bibinfo{person}{Abishek Sridhar}, \bibinfo{person}{Xianyi Cheng}, \bibinfo{person}{Tianyue Ou}, \bibinfo{person}{Yonatan Bisk}, \bibinfo{person}{Daniel Fried}, {et~al\mbox{.}}} \bibinfo{year}{2023}\natexlab{a}.
\newblock \showarticletitle{Webarena: A realistic web environment for building autonomous agents}.
\newblock \bibinfo{journal}{\emph{arXiv preprint arXiv:2307.13854}} (\bibinfo{year}{2023}).
\newblock


\bibitem[Zhu et~al\mbox{.}(2023)]%
        {zhu2023ghost}
\bibfield{author}{\bibinfo{person}{Xizhou Zhu}, \bibinfo{person}{Yuntao Chen}, \bibinfo{person}{Hao Tian}, \bibinfo{person}{Chenxin Tao}, \bibinfo{person}{Weijie Su}, \bibinfo{person}{Chenyu Yang}, \bibinfo{person}{Gao Huang}, \bibinfo{person}{Bin Li}, \bibinfo{person}{Lewei Lu}, \bibinfo{person}{Xiaogang Wang}, {et~al\mbox{.}}} \bibinfo{year}{2023}\natexlab{}.
\newblock \showarticletitle{Ghost in the minecraft: Generally capable agents for open-world environments via large language models with text-based knowledge and memory}.
\newblock \bibinfo{journal}{\emph{arXiv preprint arXiv:2305.17144}} (\bibinfo{year}{2023}).
\newblock


\bibitem[Zhu et~al\mbox{.}(2024a)]%
        {zhu2024survey}
\bibfield{author}{\bibinfo{person}{Xunyu Zhu}, \bibinfo{person}{Jian Li}, \bibinfo{person}{Yong Liu}, \bibinfo{person}{Can Ma}, {and} \bibinfo{person}{Weiping Wang}.} \bibinfo{year}{2024}\natexlab{a}.
\newblock \showarticletitle{A survey on model compression for large language models}.
\newblock \bibinfo{journal}{\emph{Transactions of the Association for Computational Linguistics}}  \bibinfo{volume}{12} (\bibinfo{year}{2024}), \bibinfo{pages}{1556--1577}.
\newblock


\bibitem[Zhu et~al\mbox{.}(2024b)]%
        {zhu2024knowagent}
\bibfield{author}{\bibinfo{person}{Yuqi Zhu}, \bibinfo{person}{Shuofei Qiao}, \bibinfo{person}{Yixin Ou}, \bibinfo{person}{Shumin Deng}, \bibinfo{person}{Ningyu Zhang}, \bibinfo{person}{Shiwei Lyu}, \bibinfo{person}{Yue Shen}, \bibinfo{person}{Lei Liang}, \bibinfo{person}{Jinjie Gu}, {and} \bibinfo{person}{Huajun Chen}.} \bibinfo{year}{2024}\natexlab{b}.
\newblock \showarticletitle{Knowagent: Knowledge-augmented planning for llm-based agents}.
\newblock \bibinfo{journal}{\emph{arXiv preprint arXiv:2403.03101}} (\bibinfo{year}{2024}).
\newblock


\bibitem[Zhuang et~al\mbox{.}(2020)]%
        {zhuang2020comprehensive}
\bibfield{author}{\bibinfo{person}{Fuzhen Zhuang}, \bibinfo{person}{Zhiyuan Qi}, \bibinfo{person}{Keyu Duan}, \bibinfo{person}{Dongbo Xi}, \bibinfo{person}{Yongchun Zhu}, \bibinfo{person}{Hengshu Zhu}, \bibinfo{person}{Hui Xiong}, {and} \bibinfo{person}{Qing He}.} \bibinfo{year}{2020}\natexlab{}.
\newblock \showarticletitle{A comprehensive survey on transfer learning}.
\newblock \bibinfo{journal}{\emph{Proc. IEEE}} \bibinfo{volume}{109}, \bibinfo{number}{1} (\bibinfo{year}{2020}), \bibinfo{pages}{43--76}.
\newblock


\bibitem[Zhuang et~al\mbox{.}({[n.\,d.]})]%
        {zhuangtoolchain}
\bibfield{author}{\bibinfo{person}{Yuchen Zhuang}, \bibinfo{person}{Xiang Chen}, \bibinfo{person}{Tong Yu}, \bibinfo{person}{Saayan Mitra}, \bibinfo{person}{Victor Bursztyn}, \bibinfo{person}{Ryan~A Rossi}, \bibinfo{person}{Somdeb Sarkhel}, {and} \bibinfo{person}{Chao Zhang}.} \bibinfo{year}{[n.\,d.]}\natexlab{}.
\newblock \showarticletitle{ToolChain*: Efficient Action Space Navigation in Large Language Models with A* Search}. In \bibinfo{booktitle}{\emph{The Twelfth International Conference on Learning Representations}}.
\newblock


\bibitem[Zuo et~al\mbox{.}(2024)]%
        {zuo2024planetarium}
\bibfield{author}{\bibinfo{person}{Max Zuo}, \bibinfo{person}{Francisco~Piedrahita Velez}, \bibinfo{person}{Xiaochen Li}, \bibinfo{person}{Michael~L Littman}, {and} \bibinfo{person}{Stephen~H Bach}.} \bibinfo{year}{2024}\natexlab{}.
\newblock \showarticletitle{Planetarium: A rigorous benchmark for translating text to structured planning languages}.
\newblock \bibinfo{journal}{\emph{arXiv preprint arXiv:2407.03321}} (\bibinfo{year}{2024}).
\newblock


\end{thebibliography}


\end{document}